%% file: main.tex
\documentclass[12pt]{report}
\usepackage[toc,page]{appendix} 
\usepackage{amsmath}
\usepackage{graphicx}
\usepackage{booktabs}
\usepackage{caption}
\usepackage{lineno,hyperref}
\usepackage{multirow}
\usepackage[outdir=./]{epstopdf}
\usepackage[export]{adjustbox}
\usepackage{subfig}
\usepackage{stmaryrd}
\usepackage[left=1.5in, right=1.25in, top=1.25in, bottom=1.25in]{geometry}
\usepackage{siunitx}
\usepackage{array}
\usepackage{subfig}
\usepackage{color, colortbl}
\usepackage[section]{placeins}
\usepackage{newunicodechar}
\usepackage[utf8]{inputenc}
\usepackage{textcomp}
\usepackage{xcolor}
\usepackage{pdfpages}
\usepackage{pifont}
\usepackage{xspace}
\usepackage{amssymb}
\usepackage{arydshln}
\usepackage{titletoc}
\usepackage{natbib}
\newcolumntype{L}[1]{>{\raggedright\arraybackslash}p{#1}}

\usepackage{algorithm}
\usepackage{algpseudocode}

\usepackage{hyperref}
\usepackage{url}
\usepackage{graphicx}
\usepackage{xspace}
\usepackage{booktabs}

\usepackage{multirow}
\usepackage{siunitx}
\usepackage{arydshln}
\usepackage{pifont}
\usepackage{subcaption}
\usepackage{enumitem}
\usepackage{wrapfig}

\usepackage{bm}

\usepackage[capitalize]{cleveref}

\usepackage{xspace}
\newcommand{\benchmarkname}{MMWorld\xspace}
\usepackage{multirow}

\usepackage{stmaryrd}

\definecolor{my_green}{RGB}{51,102,0}
\definecolor{my_red}{RGB}{204, 0, 0}
\newcommand{\cmark}{\textcolor{my_green}{\ding{51}}}

\usepackage{colortbl}
\definecolor{mygray}{RGB}{230,230,250}
\definecolor{mygray2}{RGB}{176,196,222}
\definecolor{dblue}{RGB}{0, 112, 192}
\definecolor{dred}{RGB}{192, 0, 0}
\definecolor{Gray}{gray}{0.9}
\definecolor{tblue}{HTML}{174992}
\definecolor{barca-blue}{RGB}{0, 76, 153}
\definecolor{barca-red}{RGB}{167, 0, 66}
\definecolor{lblue}{RGB}{13, 152, 255}
\definecolor{lred}{RGB}{255, 108, 108}

\usepackage{setspace}

\newcommand{\diff}[1]{\textcolor{magenta}{[#1]}}
\newcommand{\inc}[1]{\textcolor{teal}{[+#1]}}

\makeatletter
\def\@makechapterhead#1{%
  \vspace*{20\p@}
  {\parindent \z@ \raggedright \normalfont
    \ifnum \c@secnumdepth >\m@ne
        \huge\bfseries \@chapapp\space \thechapter
        \par\nobreak
        \vskip 20\p@
    \fi
    \interlinepenalty\@M
    \Huge \bfseries #1\par\nobreak
    \vskip 20\p@
  }}
\def\@makeschapterhead#1{%
  {\parindent \z@ \raggedright
    \normalfont
    \interlinepenalty\@M
    \Huge \bfseries  #1\par\nobreak
    \vskip 20\p@
  }}
\makeatother

\begin{document}


\begin{titlepage}
    \centering
    {\Huge\bfseries Bridging the Gap Between Multimodal Foundation Models and World Models\\[0.3in]\par}
    \vspace{1cm}
    {Submitted by: Xuehai He\par}
    {Computer Science and Engineering\par}
    {University of California, Santa Cruz\par}
    \vspace{0.5cm}
    {Supervisor: Dr. Xin Eric Wang, Assistant Professor\par}
    \vspace{4cm}
    {Committee Approval: \par}
    \vspace{1cm}
    \begin{tabular}{ll}
        Dr. Xin Eric Wang  & \raisebox{-2ex}{\underline{\hspace{2.5in}}} \\
        (Committee Chair) & \\
        \vspace{0.5cm} & \\
        Dr. Yi Zhang & \raisebox{-2ex}{\underline{\hspace{2.5in}}} \\
        (Committee Member) & \\
        \vspace{0.5cm} & \\
        Dr. Chunyuan Li & \raisebox{-2ex}{\underline{\hspace{2.5in}}} \\
        (Committee Member) & \\
    \end{tabular}
    \vfill
    {\today\par}
\end{titlepage}

\doublespacing

\clearpage

\titlecontents{part}[0em]
    {\vspace{2ex}} 
    {\bfseries\large\partname\ \thecontentslabel\quad} 
    {\bfseries\large} 
    {\hfill\contentspage} 
    [\vspace{1ex}] 

\clearpage

\pagenumbering{roman}
\setcounter{page}{3}
\tableofcontents

\clearpage

\listoftables

\clearpage

\listoffigures

\clearpage

\begin{abstract}
\thispagestyle{plain}

\setcounter{page}{32}
{\centering
Bridging the Gap Between Multimodal Foundation Models and World Models\\
by\\
Xuehai He\\
}

Humans understand the world through the integration of multiple sensory modalities, enabling them to perceive, reason about, and imagine dynamic physical processes. Inspired by this capability, multimodal foundation models (MFMs) have emerged as powerful tools for multimodal understanding and generation. However, today's MFMs fall short of serving as effective world models. They lack the essential ability such as perform counterfactual reasoning, simulate dynamics, understand the spatiotemporal information, control generated visual outcomes, and perform multifaceted reasoning.

This thesis investigates what it takes to bridge the gap between multimodal foundation models and world models. We begin by improving the reasoning capabilities of MFMs through discriminative tasks and equipping MFMs with structured reasoning skills, such as causal inference, counterfactual thinking, and spatiotemporal reasoning, enabling them to go beyond surface correlations and understand deeper relationships within visual and textual data. Next, we explore generative capabilities of multimodal foundation models across both image and video modalities, introducing new frameworks for structured and controllable generation. Our approaches incorporate scene graphs, multimodal conditioning, and multimodal alignment strategies to guide the generation process, ensuring consistency with high-level semantics and fine-grained user intent. We further extend these techniques to controllable 4D generation, enabling interactive, editable, and morphable object synthesis over time and space. To comprehensively evaluate progress in this direction and towards the eventual goal of world models, we introduce the MMWorld benchmark for evaluation multimodal foundation models on multi-discipline and multi-faceted reasoning tasks. 

Together, this thesis aims to move beyond static perception toward the creation of intelligent systems that can imagine, reason, and act within richly structured environments. By pushing multimodal foundation models closer to world models, this work takes a step toward building models that do not just see and describe the world, but can reason, simulate, and interact with the world like humans do.
\end{abstract}

\addcontentsline{toc}{chapter}{Abstract}

\chapter*{}
\begin{center}
\vspace*{\fill}

\setcounter{page}{34}
\textit{To my family}
\vspace*{\fill}
\end{center}

\chapter*{Acknowledgements}
My heart is filled with immense gratitude when I am writing this PhD thesis. There are more people to thank than I can possibly name.

First and foremost, I want to express my deepest appreciation to my advisor Xin Eric Wang. His unwavering support, guidance, and encouragement have been the cornerstone of this dissertation. Throughout my PhD journey, he has always been supportive, both professionally and personally. He brought me to the advanced area of multimodal research, guided me to meaningful research directions, and taught me critical skills—from selecting impactful topics and conducting rigorous research, to writing papers and eventually becoming an independent researcher. I still cherish those good old days when our lab were playing basketballs and bowling together. He has been more than an advisor; he is a great friend.

I am also sincerely grateful to my committee members,  Yi Zhang and Chunyuan Li. Professor Yi Zhang provided me with invaluable career advice and insightful suggestions on conducting impactful research. Her thoughtful suggestions and deep insights helped me appreciate the broader impact of academic endeavors. Beyond this, I warmly remember her habit of always bringing sweet candies to our lab, making our research environment even sweeter. Chunyuan mentored me at Microsoft Research even before I formally began my PhD journey, demonstrating what impactful research looks like and how industry researchers approach problems. He also offered invaluable advice on formalizing ideas, coding and debugging, and writing papers. Without his unparalleled mentorship and guidance, navigating the early stages of my PhD would have been much more difficult. 

During my PhD years, I have also been fortunate to work alongside many outstanding friends, colleagues, and collaborators. Their useful discussions, keen insights, encouragement, and suggestions have been essential companions on this journey: Arjun Akula, Mohit Bansal, Sugato Basu, Zonglin Di, Simon Shaolei Du, Jacob Zhiyuan Fang, Yue Fan, Weixi Feng, Tsu-jui Fu, Qiaozi Gao, Xiaofeng Gao, Reza Ghanadan, Jing Gu, Varun Jampani, Kenan Jiang, Michael Johnston, Kevin Lin, Jiachen Li, Linjie Li, Yujie Lu, Pradyumna Narayana, Vicente Ordonez, Nanyun Peng, Robinson Piramuthu, Olatunji Ruwase, Gunnar A Sigurdsson, Hangjie Shi, Yelong Shen, Jialu Wang, Jianfeng Wang, Kuan Wang, Lijuan Wang, Shuohang Wang, William Yang Wang, Yiping Wang, Xiaoxia Wu, Pengtao Xie, Ruize Xu, Qianqi Yan, Jianwei Yang, Zhengyuan Yang, Xiang Yue, Zheng Zhan, Pengchuan Zhang, Xingchen Zhao, Jian Zheng, Kaizhi Zheng, Kaiwen Zhou, Wanrong Zhu.

Above all else, my deepest gratitude goes to my family. Their continuous support, understanding, and trust have been my greatest sources of strength, especially during the most challenging times of this journey. Without their encouragement and presence, I would not have reached this milestone.

Finally, I would like to honor the memory of my beloved grandfather, Jialu He. He gave me my name and hoped that one day I would study abroad overseas and become an expert in my field. Sadly, he could not make it to this day and left me forever during the COVID pandemic, and I could not be by his side to see him one last time. Throughout these PhD years and especially during those toughest and struggling moments, every time when I was debugging alone in the middle of the night—I would look up at the brightest stars, knowing he was always there with me.

\chapter{Introduction}

\pagenumbering{arabic}
\setcounter{page}{1}

Humans perceive and understand the world through multiple sensory modalities, including vision, sound, and touch. Inspired by this multimodal nature of human cognition, multimodal foundational models have emerged as foundational tools to simulate, comprehend, and generate representations of our physical world~\cite{clip, awadalla2023openflamingo, blip2}. This dissertation mainly aims at answering the following question: \emph{What does it take to bridge the gap between multimodal foundation models (MFMs) and world models?}

While current multimodal models have shown impressive performance on a wide range of tasks~\cite{clip, multimodalqa, blip2}, they often struggle with core world modeling capabilities such as counterfactual reasoning~\cite{worldmodels}, causal and temporal inference~\cite{lecun2022path}, estimated missing states of the world based on observations, spatial and temporal awareness of the environment, and dynamic understanding of how the world evolves. These shortcomings limit their ability to simulate, plan, and interact with the world in a human-like way. To this end, this PhD dissertation first proposes novel approaches to inject various reasoning abilities into multimodal foundation models. This includes equip MFMs with counterfactual thinking abilities by enabling them to ask "what-if" questions, infer cause-effect relationships, and reason over spatial-temporal dynamics.

We then turn to the generative aspect of world modeling, where the MFMs should simulate not only what the world looks like, but how it changes. Beyond generating visually realistic content, we propose new mechanisms for structured and controllable generation. This includes multimodal controlled image generation~\cite{he2024flexecontrol}, motion-conditioned video generation~\cite{he2024mojito}, and interactivae, controllable, and editable 4D scene generation from text.

To systematically and comprehensively evaluate the progress in this direction, we introduce new benchmarks and evaluation protocols that emphasize reasoning, controllability, and multi-discipline understanding. One such contribution is the MMWorld benchmark~\cite{he2024mmworld}, designed to assess world models along diverse axes such as temporal understanding, multimodal alignment, and generalization across domains and concepts. Our evaluation framework allows for standardized comparisons and fosters progress toward the development of general-purpose, reasoning-capable multimodal systems.

Overall, in this dissertation, we \underline{first} investigate how multimodal foundation models can effectively perceive and understand the world, particularly focusing on discriminative tasks that require causal reasoning, structural reasoning, and counterfactual thinking capabilities. \underline{Second}, we explore their generative capabilities to simulate realistic scenarios and generate new content reflecting the complexities of the world. \underline{Finally}, aiming towards the ultimate goal of creating effective world models, we introduce benchmarks designed specifically for evaluating MFMs from these perspectives.

This dissertation is organized into two parts:

\section{Part I: Discriminative World Modeling (Perception \& Reasoning)}

The first part of the thesis focuses on advancing the \textit{perceptual} and \textit{reasoning} capabilities of multimodal foundation models. 

\begin{itemize}
    \item \textbf{Chapter 2} proposes techniques for adapting foundation models to perception tasks. We introduce subspace-based training strategies and analyze the inductive biases of different adaptation approaches for multimodal foundation models, especially vision-language models.
    
    \item \textbf{Chapter 3} incorporates \textit{counterfactual thinking} into multimodal models through a new prompt learning paradigm. This approach enables models to reason about what could have been, rather than only what is, improving robustness on unseen or ambiguous inputs~\cite{he2022cpl}.
    
    \item \textbf{Chapter 4} enhances \textit{compositional reasoning} of multimodal foundation models by aligning visual and linguistic structures with causal relations. We show that integrating causality helps models generalize to novel combinations of known concepts and improves performance on structured reasoning tasks.
    
    \item \textbf{Chapter 5} explores the use of \textit{generative models for perception}~\cite{he2023discffusion}, especially diffusion-based models. We demonstrate how to adapt diffusion models for discriminative tasks and propose mechanisms for accelerated, zero-shot classification using generative priors.
    \item \textbf{Chapter 6} introduces a graph-based framework for multimodal generation~\cite{multimodalgraphtransformer}. We design a Multimodal Graph Transformer that conditions generation on structured scene graphs and other symbolic representations, enhancing compositional and relational fidelity.
    \item \textbf{Chapter 7} presents MMWorld, a benchmark and evaluation suite for multimodal world models~\cite{he2024mmworld}. It covers tasks spanning perception, reasoning, and generation across diverse domains and temporal scales and most importantly, different disciplines. The benchmark includes new evaluation metrics and a broad taxonomy of tasks, enabling standardized comparisons of multimodal models.
       \item \textbf{Chapter 8} presents VLM4D~\cite{zhouvlm4d}, a benchmark and evaluation suite for evaluating spatiotemporal awareness in multimodal world models~\cite{he2024mmworld}. We also introduce future solutions for improving the spatiotemporal awareness in world models.
\end{itemize}

\section{Part II: Generative World Modeling}

The second part of the thesis moves from perception to \textit{generation}, with a focus on controllable and interactive generation. 

\begin{itemize}
    
    \item \textbf{Chapter 9} addresses \textit{controllable text-to-image generation}. We develop a flexible framework that allows users to guide generation via prompts, sketch maps, depth maps, and many other modalities, supporting multimodal control on image generation~\cite{he2024flexecontrol}.
    
    \item \textbf{Chapter 10} extends controllability to the video domain. We propose Mojito, a framework for \textit{text-to-video generation} with controllable motion dynamics, and show how to manipulate motion intensity and direction without retraining the backbone model~\cite{he2024mojito}.
    \item \textbf{Chapter 11} introduces 4D generation via Morpho4D, which synthesizes interactive, controllable, and editable 4d scenes over time. This allows for a richer simulation of the
physical world, opening up new avenues for 4D generative modeling.
\end{itemize}

\section{Conclusion and Future Work} For future reference, the final chapters of the dissertation (\textbf{Chapter 12} and \textbf{Chapter 13}) summarizes the core findings and contributions and discusses open challenges and future research directions in this dissertation. These include scaling up to real-world environments, improving long-horizon reasoning, and developing interactive, embodied multimodal agents.

\part{Discriminative World Modeling (Perception \& Reasoning)}
\chapter{Efficient Adaptation of Multimodal Foundation Models for Multimodal Perception}
\section{Introduction}
In the last few years, large-scale vision models and language models pretrained on web-scale data have seen a great surge of interest with promising performance~\cite{gpt2,bert,yang2019xlnet,liu2019roberta}. Meanwhile, aided by the rapid gains in hardware, their sizes keep growing rapidly. Currently, vision transformers~\cite{vision_transformer} (ViTs) with billions of parameters such as {\em ViT-Large}~\cite{vision_transformer} have been released. 
It is expected that pretrained vision models with even larger orders of magnitude will emerge in the foreseeable future.

\begin{figure}[t]
	\begin{center}
 	\includegraphics[width = 0.6\columnwidth]{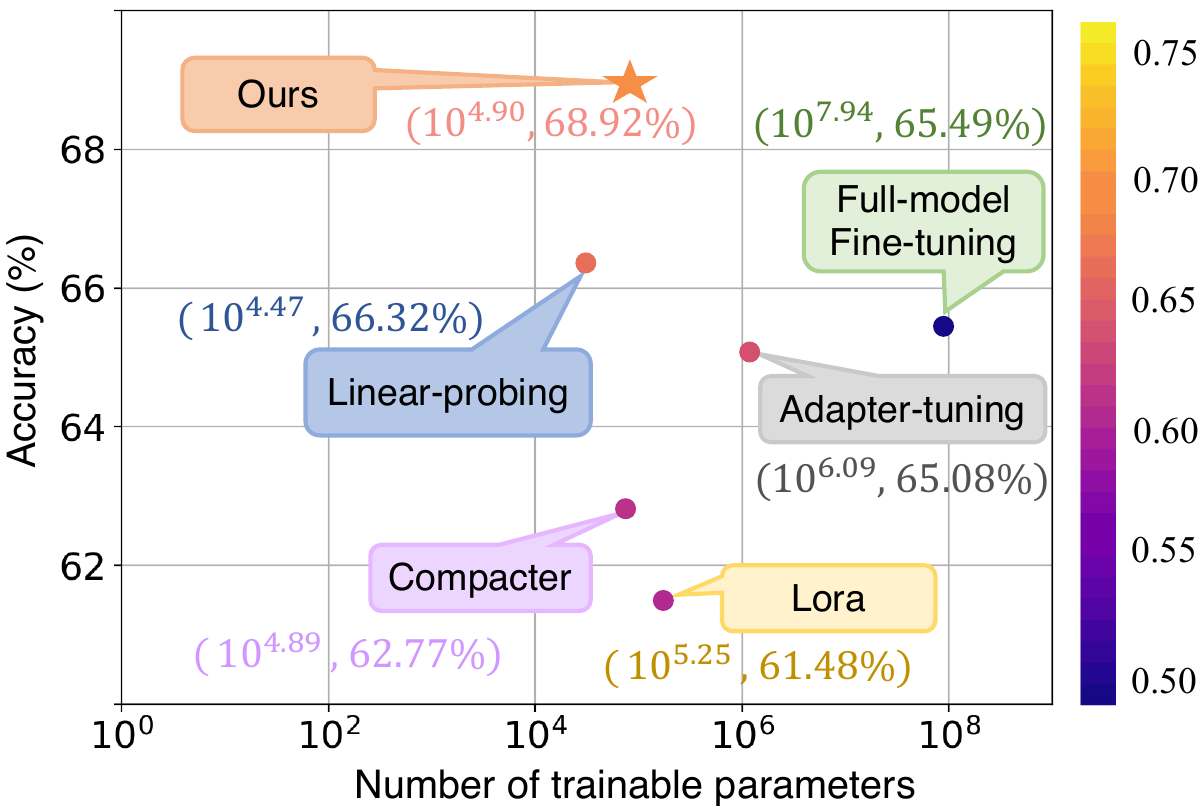}
 	\caption[The tradeoff between accuracy and parameter numbers of various model adaptation methods.]{ The results are measured using the vision transformer
(ViT-B-224/32) via CLIP pretraining across the average of 20 image classification datasets. Our method places in the topleft corner and achieves the best tradeoff between accuracy and parameter efficiency. The color of points and numbers denote the performance-efficiency (PE) metric (higher is better). }\label{fig:trend}
 	\vspace{-0.2cm}
	\end{center}
 \end{figure}

These large-scale pretrained models are powerful when transferred to downstream vision tasks. However, deploying many independent instances of fine-tuned models can also cause substantial storage and deployment costs and hinder the applicability of large-scale ViTs to real-world problems. Motivated by this and the importance of parameter-efficient learning~\cite{houlsbyParameterEfficientTransferLearning2019,huLoRALowRankAdaptation2021,zakenBitFitSimpleParameterefficient2021,mahabadiCompacterEfficientLowRank2021,he2021towards}, we aim to study the parameter-efficient model adaptation strategy for vision transformers. Conventional wisdom for transfer learning in our computer vision community is fine-tuning all model parameters or leveraging linear probes. However, performing full-model fine-tuning of pretrained ViTs may incur both financial and environmental costs~\cite{patterson2021carbon}, requires a high computational budget, and becomes increasingly infeasible as the model size continuously grows. Another go-to strategy is performing linear-probing by stacking an additional trainable multi-layer perceptron (MLP) layer in the end. It is parameter-efficient yet suboptimal in terms of performance. 
Ideally, we hope to design model adaptation strategies that can achieve the best tradeoff between efficiency and effectiveness (see Figure~\ref{fig:trend}) --- optimizing adaptation parameter-efficiency while allowing for the model to maintain the effectiveness of transfer learning on downstream vision tasks, especially the image classification task.
 
In this chapter, we answer the question: {\em what are the general guidelines one should adopt while adapting large-scale pretrained vision models to the downstream tasks?}  We first build a benchmark for adapting ViTs and proposing a more parameter-efficient model adaptation method. 
We choose ViTs as the pretrained vision models, which are representative mainstream state-of-the-art (SOTA) models on a wide range of downstream vision tasks. 
Specifically, we experiment with two off-the-shelf pretrained ViTs in the remainder of this paper: the one via Contrastive Language-Image Pretraining (also known as CLIP)~\cite{clip}, and the one via supervised pretraining (we refer to as Supervised ViT)~\cite{vision_transformer}. In addition to Full-model Fine-tuning and linear-probing, we re-implement several SOTA efficient adaptation methods~\cite{houlsbyParameterEfficientTransferLearning2019,ruckleAdapterDropEfficiencyAdapters2021,huLoRALowRankAdaptation2021,zakenBitFitSimpleParameterefficient2021,liPrefixTuningOptimizingContinuous2021} (originally proposed for pretrained language models) on vision tasks, and design various new baseline methods for comparison. 

Aghajanyan~\cite{aghajanyanIntrinsicDimensionalityExplains2020}  show that pretrained language models have a low intrinsic dimension and can still learn efficiently despite a low-dimensional reparameterization. 
Motivated by this observation, we reformulate the task of efficient model adaptation as a subspace training problem.
Within this framework, we measure the \emph{local intrinsic dimension} of each module in ViTs, which reveals that the attention module dominates the training progress. 
Moreover, we introduce a novel parameter-efficient model adaptation framework named \emph{Kronecker Adaptation (KAdaptation)}, where during adaptation, pretrained weights are frozen, and only the updates to the weights receive gradients.  And the weight updates are decomposed to a set of Kronecker products, with the {\em slow weights}~\cite{fast_weights} shared across layers and {\em fast weights}~\cite{fast_weights} further decomposed into low-rank matrices product to improve parameter efficiency. 
We apply KAdaptation to attention weights, and it achieves the best average accuracy among efficient model adaptation methods while containing much less trainable parameters,  e.g., around \textbf{45\%} parameters of LoRA~\cite{huLoRALowRankAdaptation2021} and \textbf{0.09\%} of all the model parameters in CLIP under the few-shot setting.

\section{Efficient Model Adaptation with Subspace Training}

Given a large pretrained vision transformer $\mathcal{M}$ with size $\vert \mathcal{M} \vert$. Our goal is to develop a parameter-efficient model adaptation technique with trainable parameters $\theta$ of size $d \ll \vert \mathcal{M} \vert$, that can attain comparable performance with fine-tuning the whole model. Our ultimate goal is that one could
achieve satisfactory results in both efficacy and efficiency without the hassle of fine-tuning the full model.

\subsection{Subspace Training}
A typical neural network contains numerous dense layers that perform matrix multiplication. The weight matrices in these layers can be full-rank. When adapting to a specific task, however, ~\cite{aghajanyanIntrinsicDimensionalityExplains2020} show that the pretrained language models have a low \emph{intrinsic dimension} and can still learn efficiently despite a low-dimensional reparameterization.

Drawing inspiration from their observation and study, we hypothesize that the updates to weights of ViTs during each step in model adaptation also have a low intrinsic rank and develop our method accordingly. The intuition behind our method is to perform subspace training on weight updates. In the de-facto training paradigm of neural network models, the gradient is computed first, followed by gradient steps taken by the optimizer in the entire parameter space $D$. While in subspace training, we instead build a random $d$-dimensional parameter subspace from $ \mathcal{M}$, where generally $d \ll \vert \mathcal{M} \vert$, and optimize directly in this subspace.

In fact, most current parameter-efficient NLP model adaptation strategies perform subspace training. Given a large pretrained language model $\mathcal{M}$ with size $\vert \mathcal{M} \vert$, existing methods either select a submodule from $\mathcal{M}$ or inject an additional module to $\mathcal{M}$. For the parameter vector $\Theta\in \mathbb{R}^{D}$ from this module, they learn a projection $\mathcal{P}$ mapping $\Theta$ into a random $d$-dimensional subspace and perform training in that subspace to minimize computational cost. With this observation, we motivate our study on the efficient model adaptation problem in the principle of subspace training. We approach the problem by addressing two scientific questions:  \emph{how to choose these submodules} and \emph{how to make the subspace projection}.

\subsection{The Proposed Kronecker Adaptation} 
To answer the two fundamental questions of efficient model adaptation, \emph{how to choose these submodules} and \emph{how to make the subspace projection},  we propose a novel framework that consists of two corresponding strategies. First, we define the local intrinsic dimension and we choose submodules based on their measured local intrinsic dimensions. Second, we propose a Kronecker Adaptation method to perform the subspace projection on the selected submodules by exploiting parameterized hypercomplex multiplication layers (PHM)~\cite{phm}.  

\subsubsection{Local Intrinsic Dimension} 
Measuring the intrinsic dimension of an objective function was first proposed in~\cite{liMeasuringIntrinsicDimension2018}. ~\cite{aghajanyanIntrinsicDimensionalityExplains2020} extended it to analyze the quality of pretrained language models. They point out that analyzing model adaptation through the lens of intrinsic dimension offers empirical and theoretical intuitions. Both of them study the intrinsic dimension of the entire model. 

Unlike them, we propose to measure the intrinsic dimension of each individual submodule in ViT. We define the intrinsic dimension of the submodule as \emph{local intrinsic dimension}, to distinguish it from the intrinsic dimension of the whole model. The local intrinsic dimension is indicative of the contribution of each submodule during model adaptation and measuring it will tell us how many free parameters are required to approximate the optimization problem closely. The conventional standard method of measuring the intrinsic dimensionality of an objective~\cite{liMeasuringIntrinsicDimension2018} asks for performing grid search over different subspace dimensions $d$, training using standard SGD~\cite{sgd} over the subspace reparameterization, and selecting the smallest $d$ which can produce a satisfactory solution (e.g., 90\% of the full training metric). 
Likewise, we measure the local intrinsic dimension via finding the smallest $d$ for the measured submodule that can reach 90\% of the full accuracy.

To this end, we first follow the similar definition in~\cite{liMeasuringIntrinsicDimension2018} and define $\Theta$ in a subspace in the following way: \begin{equation}
\Theta=\Theta_{0}+P \theta,
\end{equation}
where $\Theta_{0}\in \mathbb{R}^{D}$ is the initial parameter vector of $\Theta$ when the training begins, $P\in \mathbb{R}^{D \times d}$ is the projection matrix generated by the Fastfood transform~\cite{le2014fastfood}, and $\theta \in \mathbb{R}^{d}$ is the parameter vector in the subspace. Subspace training proceeds by computing gradients with respect to $\theta$ and taking steps in that subspace. By performing experiments with gradually larger values of $d$, we can find the subspace dimension $d_t$ at which the performance of the model $\mathcal{M}$ reaches 90\% of the full accuracy. We refer to $d_t$ the \emph{local intrinsic dimension} of the measured submodule.

The module with the lowest local intrinsic dimension --- attention module is selected. We project them into subspace via our proposed KAdaptation method for the sake of efficient model adaptation.
KAdaptation fine-tunes attention weight matrices indirectly by optimizing decomposition matrices of the updates to attention weight matrices. To lower the parameter cost, the decomposition is computed as the sum of Kronecker products while the original matrices remain frozen.

\subsubsection{Kronecker Product}
The Kronecker product between matrix $\boldsymbol{A} \in \mathbb{R}^{m \times n}$ and $\boldsymbol{B} \in \mathbb{R}^{p \times q}$, denoted by $\boldsymbol{A} \otimes \boldsymbol{B} \in \mathbb{R}^{m p \times n q}$, is mathematically written in the following form:
\begin{equation}
\boldsymbol{A} \otimes \boldsymbol{B}=\left(\begin{array}{ccc}
a_{11} \boldsymbol{B} & \cdots & a_{1 n} \boldsymbol{B} \\
\vdots & \ddots & \vdots \\
a_{m 1} \boldsymbol{B} & \cdots & a_{m n} \boldsymbol{B},
\end{array}\right)
\end{equation}
where $a_{i j}$ shows the element in the $i$-{th} row and $j$-{th} column of $\boldsymbol{A}$.

\begin{figure}[t]
	\begin{center}
 	\includegraphics[width = 0.8\columnwidth]{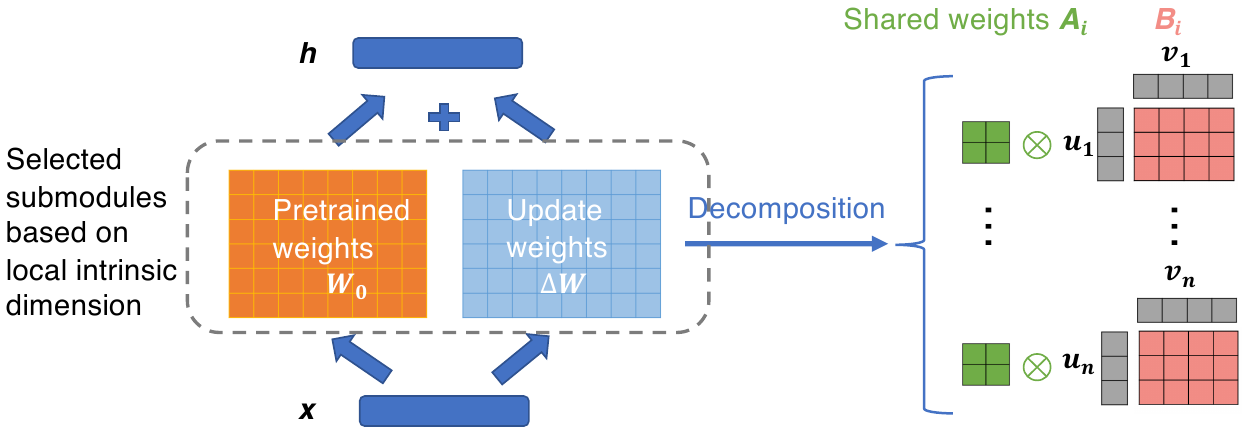}
 	\caption[An illustration of KAdaptation.]{  $\boldsymbol{A_i}$ denotes the shared weight matrix, with $i \in\{1, \ldots, n\}$. $\boldsymbol{B_i}$ is decomposed into two low-rank matrices $\boldsymbol{u}_{\boldsymbol{i}}$ and $\boldsymbol{v}_{\boldsymbol{i}}$. $\boldsymbol{h}$ is the output of the selected ViT submodule. $\boldsymbol{x}$ is the input to the submodule. During model adaptation process, only matrices $\boldsymbol{A_i}$, $\boldsymbol{u}_{\boldsymbol{i}}$, and $\boldsymbol{v}_{\boldsymbol{i}}$ receive gradients to improve parameter efficiency.
 	}\label{fig:overview_pevit}
 	\vspace{-0.2cm}
	\end{center}
 \end{figure}
\subsubsection{Kronecker Adaptation}
Leveraging the Kronecker product to perform language model compression has been shown to be beneficial in prior works~\cite{kroneckerbert, kroneckergpt}. Recently,~\cite{phm} introduces PHM layers, theoretically demonstrating that Kronecker products can help to reduce learnable parameters in language models and maintain performance. Built upon the success of PHM, for an update matrix $\Delta\boldsymbol{W} \in \mathbb{R}^{k \times d}$ in the ViT,
we propose the KAdaptation to adapt it into subspace. The illustration is shown in Fig.~\ref{fig:overview_pevit}. Mathematically, we compute $\Delta\boldsymbol{W}$ as follows: \begin{equation}
\Delta\boldsymbol{W}=\sum_{i=1}^{n} \boldsymbol{A}_{\boldsymbol{i}} \otimes \boldsymbol{B}_{\boldsymbol{i}},
\label{eq:kronecker}
\end{equation}
where $n$ is the user-defined hyperparameter  representing the number of Kronecker products, $\boldsymbol{A}_{\boldsymbol{i}} \in \mathbb{R}^{n \times n} \text {, and } \boldsymbol{B}_{\boldsymbol{i}} \in \mathbb{R}^{\frac{k}{n} \times \frac{d}{n}}$. The new representation of the update weights in Eq.~\ref{eq:kronecker} is composed of a sum of $n$ Kronecker
products between shared {\em slow weights} $\boldsymbol{A}_i$ and independent {\em fast weights} $\boldsymbol{B}_i$\text {, with } $i \in\{1, \ldots, n\}$.

Meanwhile, low-rank methods~\cite{aghajanyanIntrinsicDimensionalityExplains2020,liMeasuringIntrinsicDimension2018, low_rank} have demonstrated that strong performance can be achieved by optimizing models in a low-rank subspace. Similarly, we hypothesize that $\Delta\boldsymbol{W} $ can be effectively adapted by learning transformations in a low-rank subspace to reduce parameter cost further. Therefore, we parameterize $\boldsymbol{B}_i \in \mathbb{R}^{\frac{k}{n} \times \frac{d}{n}}$ as low rank and further decompose it into the product of two low-rank matrices $\boldsymbol{u}_{\boldsymbol{i}} \in \mathbb{R}^{\frac{k}{n} \times r}$ and $\boldsymbol{v}_{\boldsymbol{i}} \in \mathbb{R}^{r \times \frac{d}{n}}$, where $r$ is the rank of the matrix. Overall, similar to the low-rank parameterized hypercomplex multiplication layer (LPHM) proposed in~\cite{mahabadiCompacterEfficientLowRank2021}, the expression of the update matrix $\Delta\boldsymbol{W} $ is then:
\begin{equation}
    \Delta\boldsymbol{W}=\sum_{i=1}^{n} \boldsymbol{A}_{\boldsymbol{i}} \otimes \boldsymbol{B}_{\boldsymbol{i}}=\sum_{i=1}^{n} \boldsymbol{A}_{\boldsymbol{i}} \otimes\left(\boldsymbol{u}_{\boldsymbol{i}} \boldsymbol{v}_{\boldsymbol{i}}^{\top}\right).
\end{equation}
The number of trainable parameters is now substantially saved. Note that similar to $\boldsymbol{B}_i \in \mathbb{R}^{\frac{k}{n} \times \frac{d}{n}}$, the shared {\em slow weights} $\boldsymbol{A}_i$ can also be further decomposed into the product of low-rank matrices. Additional bias terms can also be applied to the update matrix. We give the analysis of parameter efficiency in the next section.

\section{Analysis of Model Adaptation Methods} 
\subsection{Discussion of State-of-the-art Methods}
In what follows, we discuss connections between our method and state-of-the-art parameter-efficient tuning methods on NLP tasks and provide additional insight into the characteristics of our method.

\paragraph{Adapter-tuning~\cite{houlsbyParameterEfficientTransferLearning2019}} is  the first efficient model adaptation work in the NLP community. It brings in an additional trainable set of modules by adding a trainable bottleneck layer after the
feedforward network in each Transformer layer of
the pretrained language models. A bottleneck layer consists of a down and up projection pair that shrinks and recovers the size of
token hidden states. 

Similar to the Adapter-tuning method where they use the bottleneck structure in the additional layer, our method implements low-rank decomposition on the {\em fast} rank-one matrices~\cite{fast_weights}. The critical functional difference is that our learned weights can be merged with the main weights during inference, thus introducing no latency.

\paragraph{LoRA~\cite{huLoRALowRankAdaptation2021}} is another line of work for parameter-efficient language model tuning: it treats the model parameters after fine-tuning as an addition of the pretrained parameters $\Theta_{\text {pretrained }}$ and task-specific differences $\theta_{\text {task }}$, where $\Theta_{\text {pretrained }}$ is fixed and a new subset of model parameters are added on top. 
Given a pretrained weight matrix $\boldsymbol{W}_{0} \in \mathbb{R}^{d \times k}$, they constrain its update by 
performing low-rank decomposition: $\boldsymbol{W_{0}}+\Delta \boldsymbol{W}=\boldsymbol{W}_{0}+\boldsymbol{B A}$, where $\boldsymbol{A} \in \mathbb{R}^{r \times k}$, $\boldsymbol{B} \in \mathbb{R}^{d \times r}$, and the rank $r \ll \min (d, k)$. By doing this, the weight matrices are split into two parts, where during training, $\boldsymbol{W}_{0}$ is frozen and receives no gradient updates, while only $\boldsymbol{A}$ and $\boldsymbol{B}$ contain trainable parameters.

Our work differs from LoRA mainly in that we decompose weight updates to a set of Kronecker product decomposition. The decomposed {\em slow weight} are shared across layers, further reducing the parameter cost.

\paragraph{Compacter~\cite{mahabadiCompacterEfficientLowRank2021}} 
inserts task-specific weight matrices into weights of pretrained models. Each Compacter weight matrix is computed as the sum of Kronecker products between shared {\em slow} weights and {\em fast} matrices defined per Compacter layer. 

In a similar vein to Compacter, we also leverage the Kronecker product in our method to reduce parameter cost further. Yet, apart from application domains, our method fundamentally differs from Adapter/Compacter based methods in that: first, our method brings in no additional layer and introduces no latency; Second,
our method first selects submodules by measuring the local intrinsic dimension and then performs the KAdaptation over the update weights to selected submodules; Third, during adaptation, only updates to the weights of selected submodules receive gradients and tuned, while pretrained weights are always fixed.

\begin{table}[t]
  \centering
   \resizebox{0.6\columnwidth}{!}{
  \begin{tabular}{lcc}
    \toprule
    Method & \#Params & Complexity\\
    \midrule
      Adapter-tuning & $4Lkd$&$\mathcal{O}\left(kd\right)$
      \\ LoRA & $2Lrd_{model}$&$\mathcal{O}\left(rd_{model}\right)$
      \\ Compacter &$4L(\frac{k}{n}+\frac{d}{n})+n^3$&$\mathcal{O}\left(\frac{k+d}{n}\right)$
      \\ KAdaptation & $2L(\frac{d_{model}}{n}+\frac{r}{n})+n^3$&$\mathcal{O}\left(\frac{r+d_{model}}{n}\right)$ \\
    \bottomrule
  \end{tabular}}
   \caption{Parameter count in Adapter-tuning, LoRA, Compacter, and KAdaptation. $L$ is the number of layers in the Transformer. $k$ is the size of the input dimension to the Adapter layer. $d$ is the bottleneck dimension in the Adapter layer. $d_{model}$ is the Transformer hidden size. $r$ denotes the rank in the low-rank decomposition step. $n$ is the number of Kronecker products usually very small.}
 \label{tab:num_par}
\end{table}

\subsection{Analysis of Parameter Efficiency} 
We analyze the parameter-efficiency of our KAdaptation and other model adaptation methods as below:

\paragraph{Adapter-tuning} In the standard setting, two Adapters are added per layer of a Transformer model~\cite{Baevski2019AdaptiveIR}. Each Adapter layer consists of $2\times k\times d$ parameters for the down and up-projection matrices, where $k$ is the size of the input dimension and $d$ is the Adapter's bottleneck dimension.
The total number of parameters for Adapters for a $L-$layer Transformer is, $|\Theta| = 2\times L \times 2\times k\times d $s.

\paragraph{LoRA} LoRA adds trainable pairs of rank decomposition matrices to existing weight matrices. The number of trainable parameters is determined by the rank $r$: $|\Theta|=2 \times L \times d_{\text {model }} \times r$, where $d_{\text {model }}$ is Transformer hidden size.
\paragraph{Compacter} Compacter shares the trained weight matrices $\left\{\boldsymbol{A}_{\boldsymbol{i}}\right\}_{i=1}^{n}$ consisting
of $n^{3}$ parameters across all layers, where $n$ is the number of Kronecker products. Compacter also has two rank-one weights for each Adapter layer consisting of $\frac{k}{n}+\frac{d}{n}$ parameters, where the Adapter layers are of size $k \times d$, resulting in a total of $2\times \left(\frac{k}{n}+\frac{d}{n}\right)$ parameters for down and up-projection weights. Therefore, the total number of parameters of Compacter is $4 \times L \times \left(\frac{k}{n}+\frac{d}{n}\right)+n^{3}$ for a Transformer with $L$ layers in the encoder and decoder.

\paragraph{Our Approach} we analyze the parameter efficiency of our approach under the scenario where we decompose the updates to weights into a sum of Kronecker products first and then further perform low-rank decomposition for the {\em fast weights}. The total number of parameters in this scenario will be:
$2 \times L \times \left(\frac{r+d_{model}}{n}\right)+n^{3}$.

The overall comparison of parameter counts is shown in Table~\ref{tab:num_par}. Our method has a complexity of $\mathcal{O}\left(\frac{r+d_{model}}{n}\right)$ with $r$ being a small integer. Our approach greatly reduces the number of parameters. The exact numbers of trainable parameters are present in Table~\ref{tab:acc}.

\section{Experiments}
\input{PEViT/fewshot}
\subsection{Datasets} 
For few-shot benchmark experiments, we conduct experiments on 20 image classification datasets from the ELEVATER benchmark~\cite{vision_benchmark} on four Quadro RTX A6000 GPUs. Detailed dataset statistics are given in the supplementary material. For full-shot experiments, we summarize the results by computing the average performance on CIFAR10~\cite{cifar}, CIFAR100~\cite{cifar}, SUN397~\cite{sun}, DTD~\cite{dtd}, STL10~\cite{stl10}, FGVCAircraft~\cite{fgvc}, and FER2013~\cite{fer}. We use the official split for each of these datasets.

\subsection{Implementation Details}
For benchmark experiments, we use the SGD~\cite{sgd} optimizer with the learning rate and weight decay being automatically searched for all methods so that these two hyperparameters have the optimum combination. We borrow the automatic hyper-parameter tuning toolkit from~\cite{vision_benchmark}. Training epochs are set via grid search. We test two pretrained $12$-layer ViTs: the one using ViT-B-224/32 via unsupervised pretraining ({\em CLIP}) and the one using ViT-B-224/16 via supervised pretraining ({\em Supervised ViT}).

For intrinsic dimension experiments, we use the AdamW~\cite{adam} as the optimizer, with the weight decay of $10^{-8}$, learning rate of $10^{-5}$, and batch size of 32 following the setting in Li~\cite{liMeasuringIntrinsicDimension2018}. The Fastfood transform~\cite{le2014fastfood} is applied to the attention and multi-layer perceptron (MLP) module in the first layer of Supervised ViT, respectively. The dimension $d$ is measured from $0-2000$ in both scenarios. Each model is fine-tuned for 300 epochs.

\subsection{Baselines}
We test the baselines below. Unless otherwise specified, the task-specific classification layer and added parameters are tuned while the pretrained ViTs are frozen.

First are commonly-used model adaptation methods for vision models.
\begin{itemize}
\item \emph{Full-model Fine-tuning}: fine-tunes all model parameters.
\item \emph{Linear-probing}: only tune the task-specific classification layer. 
\end{itemize}

The second types are SOTA methods borrowed from the NLP community.

\begin{itemize}
\item \emph{BitFit}~\cite{zakenBitFitSimpleParameterefficient2021}: freezes all ViT parameters except for the bias terms and the task-specific classification layer.
\item \emph{Adapter}-tuning~\cite{houlsbyParameterEfficientTransferLearning2019}: two Adapters are added and tuned in each Transformer layer.
\item \emph{AdapterDrop}~\cite{ruckleAdapterDropEfficiencyAdapters2021}: only keep Adapters from the last Transformer layer.
\item \emph{LoRA}~\cite{huLoRALowRankAdaptation2021}: apply LoRA to $\boldsymbol{W}_q$ and $\boldsymbol{W}_v$ matrices in the attention module and tune the low-rank decomposition matrices.
\item \emph{Compacter}~\cite{mahabadiCompacterEfficientLowRank2021}: we experiment with $n=4$.
\end{itemize}

The third types are new baseline methods we developed.

\begin{itemize}
\item \emph{Transformer-probing}: an additional trainable Transformer block is stacked before the task-specific classification layer and tuned.
\item \emph{LoRA-Fix}: the matrix $\boldsymbol{A}$ in LoRA~\cite{huLoRALowRankAdaptation2021} is fixed and only the matrix $\boldsymbol{B}$ is tuned.
\item \emph{LayerNorm Tuning}: the layer norm layers are tuned.
\item\emph{Attention Tuning}: the attention layers are tuned.
\item \emph{LePE Tuning}~\cite{dong2021cswin}: locally-enhanced positional encoding (LePE) is added to the ViT and tuned. We implement it by the depthwise convolution operator~\cite{depthwidth} on the matrix $\boldsymbol{V}$ in the attention layer:
$
\operatorname{Attention}(\boldsymbol{Q}, \boldsymbol{K}, \boldsymbol{V})= \operatorname{SoftMax}\left(\boldsymbol{Q} \boldsymbol{K}^{T} / \sqrt{d}\right) \boldsymbol{V}+\operatorname{DWConv}(\boldsymbol{V}).$
\item \emph{Relative Position Bias (RPB) Tuning}~\cite{liuSwinTransformerHierarchical2021}: an additional relative position bias term $\boldsymbol{B}$ is included in computing self-attention in the ViT and tuned:
$
\operatorname{Attention}(\boldsymbol{Q}, \boldsymbol{K}, \boldsymbol{V})= \operatorname{SoftMax}\left(\boldsymbol{Q} \boldsymbol{K}^{T} / \sqrt{d}+\boldsymbol{B}\right)\boldsymbol{V}.
$
\end{itemize}
LayerNorm Tuning, Attention Tuning, and BitFit shed light on which parameters in ViT matter more during model adaptation. Among all modules in ViT, multi-layer perceptron (MLP) tuning is not considered a baseline because it is prohibitively costly compared to others. Given that the special structure of ViT and its variants, e.g., depthwise convolution operator and relative position bias, are different from the general transformers in natural language processing,  we actually made the first step towards parameter-efficient model adaptation for the ViT
via LePE Tuning and Relative Position Bias Tuning.

\subsection{Results and Analysis}
\paragraph{Metric with performance-efficiency trade-off}
To better compare different methods with a single number that considers both prediction accuracy and parameter-efficiency, we resort to the performance-efficiency (PE) metric defined in~\cite{li2022elevater}:
$$
\text {PE}=\text {score} * \exp \left(-\log _{10}(\text {\# trainable-parameters } / M_0+1)\right)
$$ where score is the prediction accuracy, while \# trainable-parameters is the number of updated parameters in the model adaptation stage, and $M_0$ is the normalization constant. $M_0$ is set to $10^8$ because most existing vision backbone model size are in this magnitude, for example, ViT-Base (80M parameters). 

The experimental results of measured average accuracy across the 20 datasets in the low-data regime and under the 5-shot setting using random seeds of $0$, $1$, and $2$ are shown in Table~\ref{tab:fewshot}. As observed, the parameter cost of linear-probing is the lowest while that of full-model fine-tuning is the highest. Our method has the highest average accuracy and remains the ideal approach with the optimum tradeoff: our method has much less trainable parameters than other adaptation methods --- the second lowest and is only higher than Linear-probing. From the performance-efficiency trade-off metric, it can also be seen that \textbf{ours has the highest PE}.

To further compare our method with SOTA methods for NLP models and more baselines, we investigate the performance of adaptation approaches in the full-data regime and test under the full-shot setting. The results across the seven datasets are shown in Table~\ref{tab:acc}. In our analytical experiments, we first observe that Full-model Fine-tuning has the highest accuracy in both scenarios, serving as a performance upper bound. Second, different efficient model adaptation methods exhibit diverse characteristics
and perform differently on the same task. Third, the results from CLIP are mostly consistent with the results from Supervised ViT. This suggests that the pretraining strategy may not affect the selection of downstream model adaptation strategy much. Fourth, previous methods such as Adapter-tuning~\cite{houlsbyParameterEfficientTransferLearning2019} and LoRA~\cite{huLoRALowRankAdaptation2021} are still effective, and their accuracy is substantially higher than naive baselines, including BitFit and Attention-tuning regardless of the pretrained checkpoint. 
Fifth, among naive baselines where only submodules or task-specific classification heads are tuned, tuning the parameters of the attention layer turns out to be a surprisingly effective approach even compared to some SOTA methods, though its parameter cost is significantly higher. This further validates the effectiveness of our method by applying KAdaptation to attention weights. Finally, our method outperforms all the SOTA methods borrowed from the NLP community as well as their variants in both scenarios.

Furthermore, the average number of trainable parameters across seven datasets is also shown in Table~\ref{tab:acc}. As can be seen, our KAdaptation method contains the lowest parameter cost compared with other SOTA methods. This phenomenon is obviously noticeable when compared with Full-model Fine-tuning, where our method takes less than 0.14\% of trainable parameters of end-to-end Full-model Fine-tuning but it is capable of achieving comparable performance.

\begin{table*}[t]
\small
  \resizebox{\columnwidth}{!}{
  \centering
  \setlength{\tabcolsep}{2.5pt}
  \begin{tabular}{l ccccccccc  ccccccccc}
    \toprule
    \multirow{2}*{Method} & \multicolumn{9}{c}{CLIP} & \multicolumn{9}{c}{Supervised ViT}\\
    \cmidrule(lr){2-10} \cmidrule(lr){11-19}&CIFAR10&CIFAR100&SUN397&DTD&FER2013&FGVCAircraft&STL10&Average ($\uparrow$) &\#Params ($\downarrow$) &CIFAR10&CIFAR100&SUN397&DTD&FER2013&FGVCAircraft&STL10&Average ($\uparrow$)&\#Params ($\downarrow$)\\
    \midrule
     \multicolumn{19}{c}{Commonly-used model adaptation methods for vision models} \\
    \midrule
   Full-model Fine-tuning &\textbf{97.7}&\textbf{85.4}&73.8&\textbf{79.0}&\textbf{69.8}&\textbf{59.0}&\textbf{99.7}&\textbf{80.6} &87,897,654 &\textbf{99.0}&\textbf{92.4}&75.0&72.4&\textbf{68.2}&52.6&\textbf{99.6}&\textbf{79.9} & 86,630,561\\
      Linear-probing & 94.8&80.1&72.4&75.4&67.3&49.7&98.4&76.9 &49,175 &96.3&87.7&70.1&\textbf{72.7}&60.1&45.0&98.7&75.8 &49,175
      \\
      \midrule
      \multicolumn{19}{c}{SOTA methods for NLP models} \\
      \midrule
      BitFit&92.1&76.0&70.8 &75.9&68.0&54.5&98.8&76.6 & 179,049 &92.3&81.0&71.8&\textcolor{dblue}{72.6}&60.4&45.9&99.0&74.7 &358,741
      \\Adapter-tuning&94.7&81.4&\textcolor{dblue}{77.1} &78.0&68.4&55.3&99.0&79.1 &1,242,843 &98.4&90.6&74.2&71.0&63.4&52.4&99.3&78.5 &1,505,654
      \\AdapterDrop &93.3&78.3&71.4&77.1&67.1&51.3&98.0&76.6 &91,487 &96.8&88.4&72.3&70.2&46.9&35.6&99.6&72.8 &174,646
      \\ LoRA &95.1&78.1&\textbf{80.8}&78.1&67.7 &55.8&99.2&79.3 &147,236 &\textcolor{dblue}{98.7}&90.6&73.6&70.4&62.7&\textcolor{dblue}{54.9}&99.4&78.6 & 219,601\\
      \midrule
      \multicolumn{19}{c}{Baseline methods developed in this work} \\
   \midrule
      Transformer-probing &95.6&80.1&74.3&75.9&67.6&50.9&98.5&77.6 &3,198,999&96.5&86.9&\textbf{76.7}&72.0&60.7&45.5&99.0&76.8 &3,198,999
      \\ LoRA-Fix&92.5&77.1&60.0&77.7&65.5&44.4&88.6&72.3 & 98,481 &96.2&88.3&72.0&65.5&53.4&51.7&99.0&75.2 & 148,704
      \\ LayerNorm Tuning &82.5&76.6&66.7&72.4&61.0&37.6&99.1&70.8& 52,405 &92.2&71.7&72.0&69.0&52.7&51.0&98.8&72.5 &75,413
      \\  Attention Tuning &\textcolor{dblue}{96.8}&81.8&73.1&75.0&62.2&54.2&97.6&77.2&41,005,636 &93.9&85.7&73.8&69.2&55.2&51.9&99.2&75.6 &28,405,278
      \\
      LePE Tuning &95.1&78.9&68.0&75.4&65.2&54.0&98.0&76.4&112,556
      &93.7&90.8&73.2&69.8&60.0&49.3&99.1& 76.6 &167,225
      \\RPB Tuning &94.7&77.1& 68.4&75.2&65.1&54.1&97.9&76.1& 66,768
      &96.7&87.0&72.4&70.4&50.9&51.4&98.9&75.4&145,920
      \\ 
       \midrule
       KAdaptation &95.9&\textcolor{dblue}{84.8}&74.0&\textcolor{dblue}{78.1}&\textcolor{dblue}{69.0}&\textcolor{dblue}{56.0}&\textcolor{dblue}{99.2}&\textcolor{dblue}{79.6} &80,726 &97.9&\textcolor{dblue}{91.2}&\textcolor{dblue}{75.1}&71.4&\textcolor{dblue}{63.8}&\textbf{55.5}&\textcolor{dblue}{99.4}&\textcolor{dblue}{79.2} & 114,079\\
    \bottomrule
  \end{tabular}
  }
   \caption{Experimental result comparison on CIFAR10~\cite{cifar}, CIFAR100~\cite{cifar}, SUN397~\cite{sun}, DTD~\cite{dtd}, STL10~\cite{stl10}, FGVCAircraft~\cite{fgvc}, and FER2013~\cite{fer} datasets in terms of accuracy (\%) and number of trainable parameters (\#Params). 
 }
 \label{tab:acc}
\end{table*}

To further validate the efficiency of our proposed method, in addition to parameter costs, we perform additional evaluation on memory footprint and inference time. We compare the per-sample memory usage of each method in Table~\ref{tab:time}. Our method reduces memory overhead by $-86.0$\% compared to Full-model Fine-tuning and is in the same order of magnitude as other efficient model adaptation methods. We compare the inference time cost per batch in Table~\ref{tab:time} as well. On average,
our method costs $6.93$s per batch, the same as the vanilla ViT and LoRA, while Adapter-tuning costs $12.97$s and
Compacter takes $14.90$s.  Our method is the most efficient. It's within expectation as our method does not bring any additional layer to the original ViT, suffering from no inference latency. 

    \begin{table}[t]
\centering
  \resizebox{ \columnwidth}{!}{
         \begin{tabular}{lccc}
    \toprule
    Method   & Average Accuracy ($\uparrow$)  & Inference time ($\downarrow$)   & Memory ($\downarrow$)    \\
    \midrule
     Full-model Fine-tuning&\textbf{79.9} &6.93&421.5  
     \\ Linear-probing&   75.8 & 6.93 &\textbf{27.1}  
  \\  Adapter-tuning&  78.5 & \textcolor{dblue}{12.97}& 70.2 
      \\ LoRA& 78.6 &6.93&\textcolor{dblue}{56.0} 
  \\ Compacter&  78.6 & 14.90  & 70.0\\
   \hdashline
       KAdaptation&\textcolor{dblue}{79.2}&\textbf{6.93} &59.1  \\
    \bottomrule
\end{tabular}}
\caption{Average accuracy (\%), average inference time/throughput (s) per batch, and average peak memory (MB) for each method. Our method is time-efficient, and our memory footprint is in the same order of magnitude as other efficient model adaptation methods and much less than Full-model Fine-tuning.}
\label{tab:time}
\end{table}

\subsection{Local Intrinsic Dimension}
\begin{figure}[t]
	\begin{center}
 	\includegraphics[width = 0.7\columnwidth]{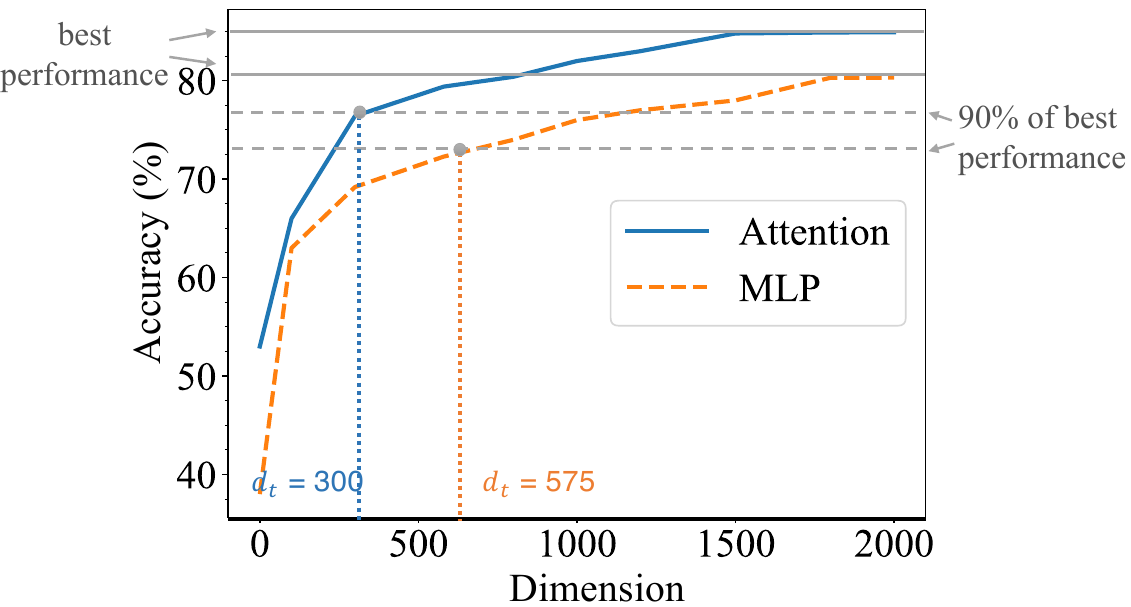}
 	\caption{Validation Accuracy \textit{vs.} Subspace Dimension $d$ of MLP and the attention module for Supervised ViT on CIFAR100. The local intrinsic dimension $d_t$ of the attention module is lower than that of the MLP.}\label{fig:dim}
	\vspace{-0.2cm}
	\end{center}
 \end{figure}

Local intrinsic dimension~\cite{liMeasuringIntrinsicDimension2018} informs us of the importance of each module in the ViT and we select submodules to perform KAdaptation based on the measurement results of the local intrinsic dimension. We measure the local intrinsic dimension of the two fundamental architectural components in the ViT --- the MLP module and the attention module. We use the remarkable Fastfood transform~\cite{le2014fastfood} to do the projection. The accuracy results averaged across $\{1, 6, 12\}$-th ViT layers are shown in Fig.~\ref{fig:dim}. As a substantiating point to performing Kronecker Adaptation on attention layers, we can see the attention module has a lower intrinsic dimension than the MLP module (300 \textit{vs.} 575).

 \begin{table}[t]
\centering
  \resizebox{0.7\columnwidth}{!}{
  \begin{tabular}{lc}
    \toprule
    Method & \ Average Accuracy \\
    \midrule
      Adapters on attention layer & 54.1 \\
      Standard Adapter-tuning & 87.7 \\
      KAdaptation to MLP & 86.6\\
    KAdaptation  & 88.1\\
    \bottomrule
  \end{tabular}}
  \caption{KAdaptation and Adapter-tuning ablation experiments with Supervised ViT on CIFAR10~\cite{cifar}, CIFAR100~\cite{cifar}, and SUN397~\cite{sun}. We report the average accuracy (\%) across the three datasets.}
  \label{tab:ablation1}
  \vspace{-0.2cm}
  \end{table}

 \subsection{Ablation Studies}
 We ablate our method and Adapter-tuning using the settings in Table~\ref{tab:acc}. As can be seen in Table~\ref{tab:ablation1}, several intriguing properties are observed.
First, applying KAdaptation to MLP modules performs worse than the original method where we apply KAdaptation to attention modules. This phenomenon is consistent with our findings from naive baseline experiments and intrinsic dimension experiments.
Second, we test another variant of Adapter-tuning. Instead of inserting two Adapters after the attention and feedforward modules respectively following Houlsby~\cite{houlsbyParameterEfficientTransferLearning2019}, we add Adapters in the attention layers. It can be observed that the standard Adapter-tuning outperforms this variance, indicating the effectiveness of the vanilla Adapter-tuning when it is adapted to vision tasks.

\section{Related Work}
\paragraph{Vision Transformer}
Fine-tuning large-scale pretrained ViTs has shown prominent performance for computer vision tasks, such as image classification~\cite{vision_transformer}, object detection~\cite{detection_transformer}, and etc. Recently, there are also other variants, including hierarchical ViTs with varying resolutions and spatial embeddings~\cite{liuSwinTransformerHierarchical2021,dong2021cswin} been proposed. Undoubtedly, the recent progress of large ViTs posts great demands for developing efficient model adaptation strategies. 

\paragraph{Efficient Model Adaptation in NLP}
In the natural language processing domain, efficient model
adaptation techniques typically involve adding to or modifying a limited number of parameters of the model — limiting the dimension of the optimization problem can prevent catastrophic forgetting~\cite{catastrophic,yang2020transfer,zhao2020learning}. Exiting methods are mainly divided into two categories depending on whether new trainable parameters are introduced. 
Specifically, one is to train a subset of the model parameters, where the common approach is to use a linear probe on top of pretrained features~\cite{clip}. The other alternatives include new parameters in between the network~\cite{liPrefixTuningOptimizingContinuous2021,ruckleAdapterDropEfficiencyAdapters2021,gupta2022self,huLoRALowRankAdaptation2021,pfeifferAdapterFusionNonDestructiveTask2021, vladapter,zheng2023minigpt}.
Nevertheless, these methodologies normally have not been investigated in the computer vision scenario~\cite{xie2023improve} and it is furthermore uncertain if findings from NLP tasks (e.g., question answering~\cite{rajpurkar2016squad,zhou2021generation,yan2024worse,zhang2025soft}, natural language understanding~\cite{wang2018glue,zheng2022jarvis,feng2023layoutgpt,yan2024med}, etc.) can transfer to downstream vision applications~\cite{xie2024simultaneous,fan2025grit}. Spurred by those facts, we establish a benchmark to compare these methods and we further advocate our method which can gain a better tradeoff under both the full-shot and few-shot settings.

\chapter{Robust Multimodal Representation Learning through Counterfactual Thinking}
\section{Introduction}
Pre-trained vision and language foundation models~\cite{clip,align} have shown encouraging results toward open-domain visual-concept matching.
Benefiting from prompt engineering~\cite{entailment,declaration}, where free-form text prompts are designed for specific task goals, those foundation models can be easily transferred to a wide array of tasks under zero-shot and few-shot scenarios, including image classification~\cite{imagenet}, visual question answering~\cite{how_much_can_clip}, image-text retrieval~\cite{align}, etc. 
But manually constructing prompts for vision and language models such as CLIP is a tedious, time-consuming process, which usually requires prior domain knowledge and leads to suboptimal solutions.

\begin{figure}[t]
\centering
\includegraphics[width=0.6\linewidth]{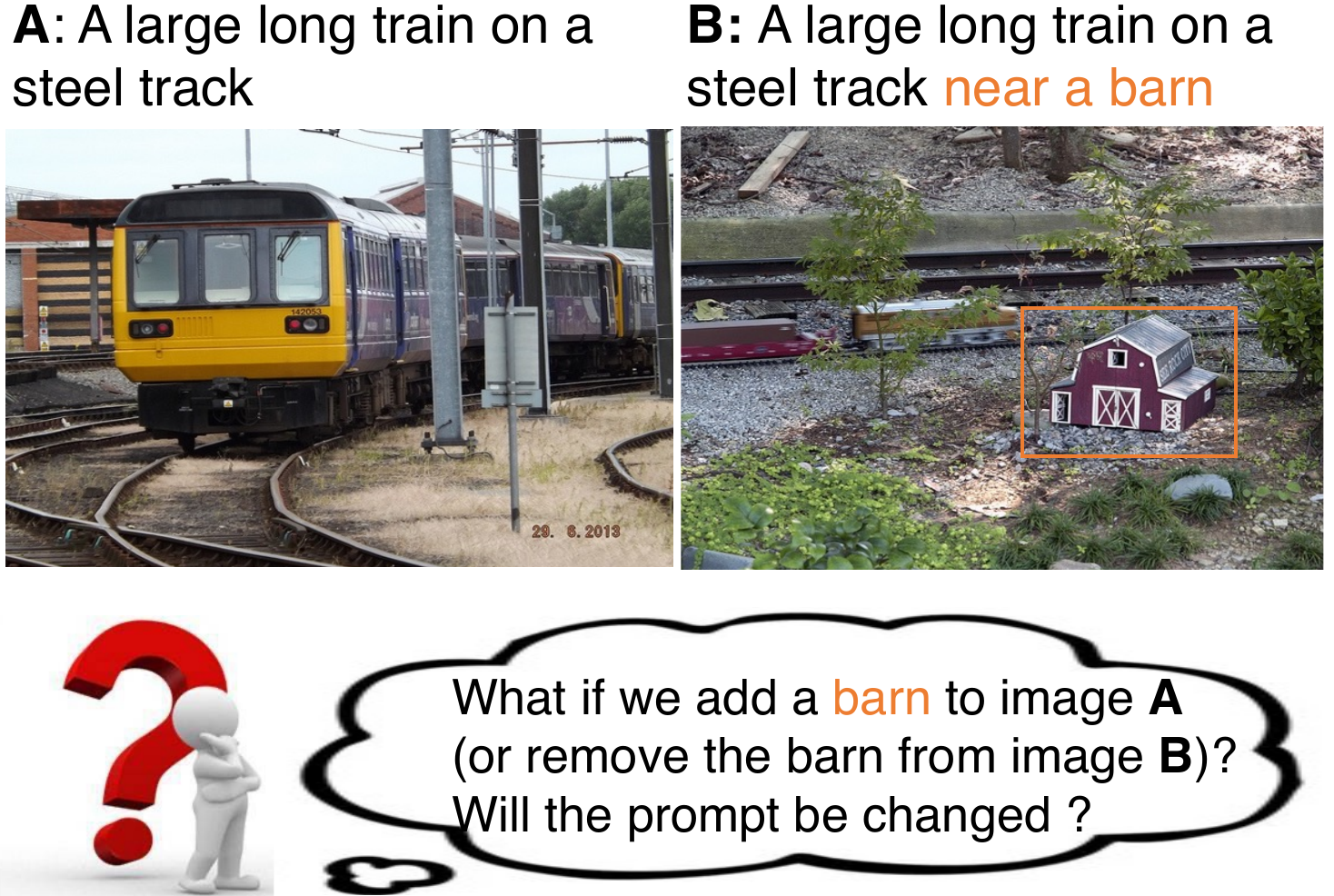}
\caption{A conceptual overview of counterfactual prompt learning. CPL constructs counterfactuals by identifying non-spurious feature change that causally causes the prompt change. In this case, the ``\textcolor{orange}{barn}'' feature is the essential cause between Prompt \textbf{A} and \textbf{B}.}
\label{fig:teaser_cpl}
\end{figure}

Prompt tuning~\cite{prompt_tuning}, on the other hand, liberates us from manual prompt engineering and automates this process. Prompt tuning methods~\cite{promptingvl,coco,cocoop} are proposed to effectively transfer CLIP to image recognition tasks after tuning a learnable prompt with a few examples of the classes.
However, those methods purely conduct empirical risk minimization (ERM) and optimize for predictive accuracy, which often produces spurious, inefficient, or entangled representations~\cite{wang2021desiderata}. 
Therefore, the generalization ability of existing prompt tuning methods for vision and language models is limited, and they often fail to transfer well to unseen classes or concepts. For example, the image classification performance of the SOTA method CoCoOp~\cite{cocoop} is similar or even degrades on unseen classes when compared with zero-shot CLIP.

Learning non-spurious representation for better generalization requires disentangling features that causally determine the prompts. One solution is counterfactual reasoning.
Counterfactual (``counter to the facts'') is a concept that describes the human capacity to learn from limited prior experiences by imagining the outcome of an alternative action that could have been taken.
So we can do counterfactual intervention by asking ``what if ...'' questions in prompt learning. For example, as shown in Figure~\ref{fig:teaser_cpl}, a change in the visual feature of the barn would cause the label to change (if we view the two prompts as two labels).

Therefore, in this chapter, we introduce a new causality-based approach, \underline{\textbf{C}}ounterfactual \underline{\textbf{P}}rompt \underline{\textbf{L}}earning (CPL), for non-spurious and efficient prompt learning, and to avoid time-consuming prompt engineering and learn more generalizable prompt representation for vision and language models.
First, we introduce a text-based negative sampling strategy to discover the most semantically-similar negative sample based on text similarity.
Then we generate a counterfactual example by identifying minimal non-spurious feature change between semantically-similar positive and negative samples that causally causes prompt change. Finally, we adopt contrastive learning in the joint optimization framework (with counterfactual construction) to tune the learnable prompts using both factual and counterfactual examples.
The causally fine-tuned prompts will eventually guide vision-and-language foundation models to distinguish images from unseen concepts, thereby improving the generalization ability of prompt learning.

We extensively evaluate CPL using seven standard datasets for image classification, two for image-text-retrieval, and one for visual question answering (VQA). We show that CPL outperforms the baseline on all three tasks: on image classification, our method achieves $3.55\%$ average relative improvement on unseen classes across the seven datasets in terms of accuracy; on image-text retrieval, our method improves the most ($4.09\%$ relative improvement in terms of Recall@1) when using $0.5\%$ of total training instances on MSCOCO~\cite{coco} and Flickr30K~\cite{flickr}; on VQA, we gain up to $25.08\%$ relative improvement on the VQAv2~\cite{vqav2} dataset.

\begin{figure*}[t]
\centering
\includegraphics[width=\linewidth]{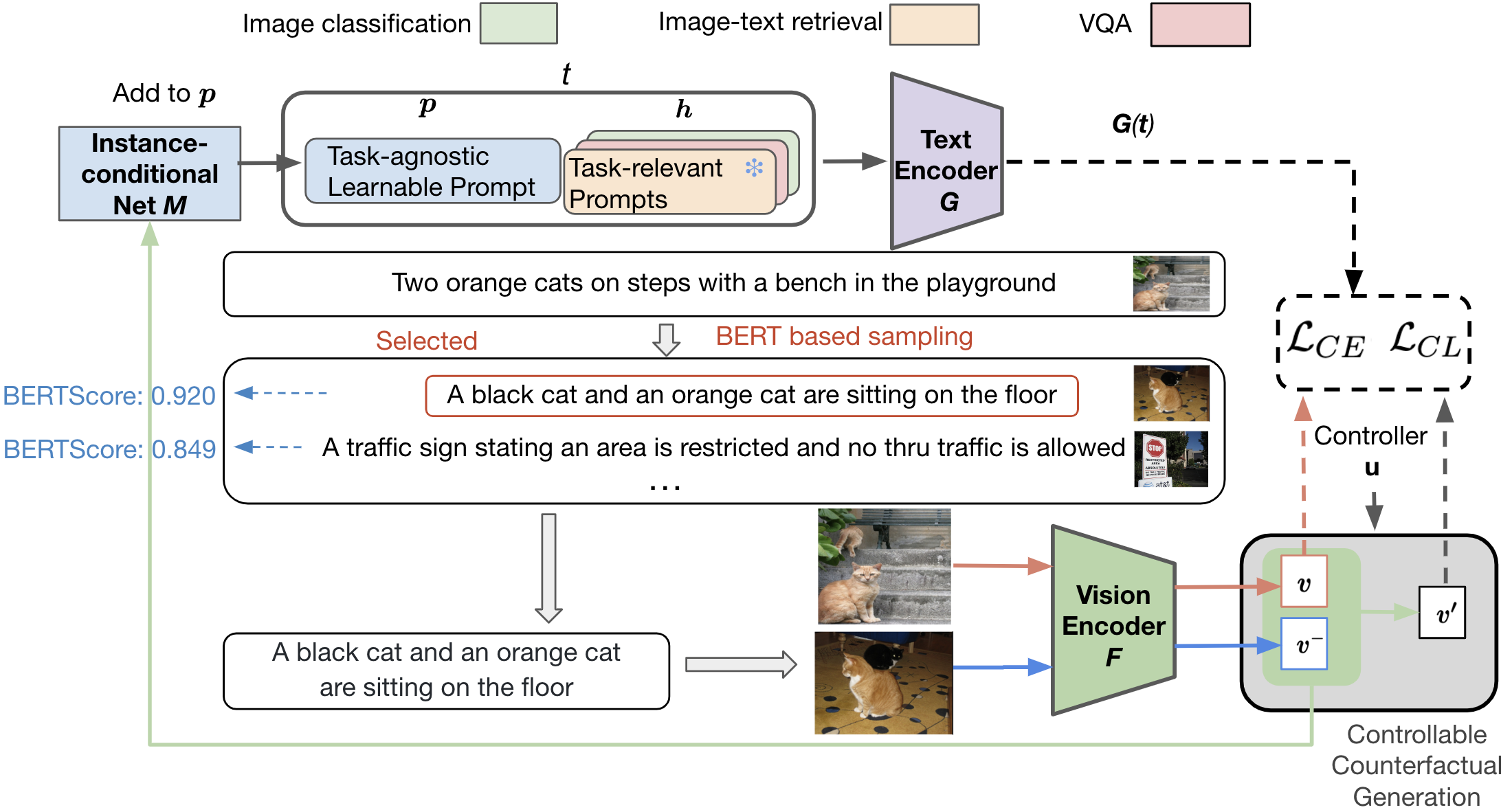}
\caption{The counterfactual prompt learning framework. We freeze the vision encoder $F$ and the text encoder $G$, and only optimize the task-agnostic prompts and the instance-conditioned net $M$ (blue blocks). Please refer to Section~\ref{sec:overview} for the explanation.
}
\label{fig:overview_cpl}
\end{figure*}

\section{From Counterfactual Prompt Learning towards Robust Multimodal Representation}
\label{sec:method}
\subsection{Problem Formulation}
Our goal is to learn generalizable prompt representation with limited data. The prompt in CLIP is divided into two parts: task-agnostic prompt $\boldsymbol{p}$
and task-relevant prompt $\boldsymbol{h}$.
Task-agnostic prompt $\boldsymbol{p}$ is learned end-to-end automatically. The set of task-relevant prompts $\mathbb{H}=\left\{\boldsymbol{h}_{0}, \boldsymbol{h}_{1}, \ldots, \boldsymbol{h}_{C}\right\}$ is mapped from the label space $\mathbb{Y}$ with some predefined rules hinging on the task type, where $C$ is the total number of classes. The final prompt $\boldsymbol{t}_c$ is the concatenation of the task-agnostic prompt and the task-relevant prompt fed into CLIP's text encoder:
$\boldsymbol{t}_{c}=[\boldsymbol{p}, \boldsymbol{h}_{c}]$.

Existing works to this problem~\cite{coop,cocoop} propose to first extract visual feature $\boldsymbol{v}$ of each input image by feeding it into CLIP’s vision encoder $F$; and text embeddings are generated by feeding $\left\{\boldsymbol{t}_{c}\right\}_{c=1}^{C}$ into the CLIP’s text encoder  $G$.
The probability of $i$-th class is computed as
\begin{equation}
p(\boldsymbol{t}_{i}\mid \boldsymbol{x})=\frac{e^ \frac{<G\left(\boldsymbol{t}_{i}\right), \boldsymbol{v}>}{ \tau}}{\sum_{c=1}^{C} e^ \frac{<G\left(\boldsymbol{t}_{c}\right), \boldsymbol{v}>}{ \tau}},
\label{eq:1}
\end{equation}
where $\tau$ is the temperature parameter, $<\cdot>$ denotes the cosine similarity. Cross-entropy loss is then minimized and the gradients can be back-propagated via the text encoder $G$ to update the learnable prompt representation $\boldsymbol{p}$. During training, the weights of CLIP always remain frozen. During inference, Eq.~\ref{eq:1} is used to compute the probability for each class.

\subsection{Method Overview}\label{sec:overview}
An  overview of the Counterfactual Prompt Learning (CPL) framework is shown in Figure~\ref{fig:overview_cpl}. For pre-processing, we construct task-relevant prompts for all training samples. 
The goal is to optimize the task-agnostic prompt $\boldsymbol{p}$.\footnote{Together with the instance-conditional net $\boldsymbol{M}$ as introduced in \cite{cocoop}. For simplicity, we will only use $\boldsymbol{p}$ hereafter as $\boldsymbol{p}$ and $\boldsymbol{M}$ are always optimized together.}
During training, given a positive image-prompt pair, we first perform \emph{text-based negative sampling} to find the most semantically-similar negative sample based on text similarity scores. 
Then we adopt a \emph{controllable counterfactual generation} strategy to construct the counterfactual from the positive and negative samples in the visual feature space. 
Finally, we perform contrastive learning using both generated counterfactual image features and factual image features in a joint optimization framework to fine-tune the task-agnostic prompt $\boldsymbol{p}$, allowing the model to understand non-spurious semantic information and learn generalized prompt representations.

\subsection{Controllable Counterfactual Generation}\label{sec:generation}
\begin{figure}[t]
\centering
\includegraphics[width=0.6\linewidth]{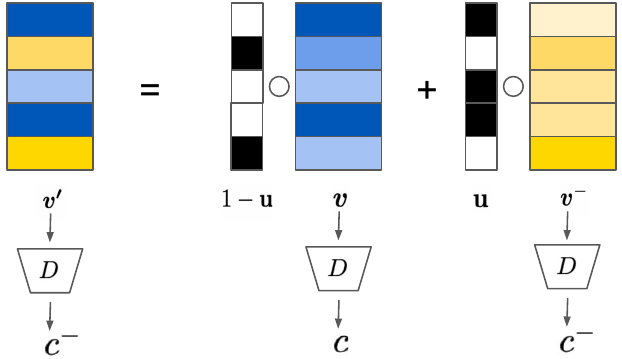}
\caption{Counterfactual generation process. $\boldsymbol{v}$ and $c$ are the positive image feature and label, while $\boldsymbol{v}^-$ and $c^-$ are the negative image feature and label. $\circ$ is element-wise multiplication. By mixing $\boldsymbol{v}$ and $\boldsymbol{v}^-$, the counterfactual image feature $\boldsymbol{v'}$ is predicted as a negative label $c^-$ by the discriminator $D$. $\mathbf{u}$ is minimized so a minimal change to the positive image feature $\mathbf{u}$ is captured here to causally change the label.}
\label{fig:generation}
\end{figure}

By viewing image feature $ \boldsymbol{v}$ as a potential cause of the label, a non-spurious feature shall be a sufficient cause of the label. So we would like to generate counterfactuals by identifying minimal non-spurious feature change that causes the label change.
The illustration of the counterfactual construction process is shown in Figure~\ref{fig:generation}.
Given positive image features $\boldsymbol{v}$ and negative image features $\boldsymbol{v^-}$, we can generate negative counterfactual image features $\boldsymbol{v'}$ as below:
 \begin{equation}
 \boldsymbol{v'} =(1-\mathbf{u}) \circ  \boldsymbol{v} + \mathbf{u}\circ \boldsymbol{v}^{-},
 \label{generation}
 \end{equation}
where $\circ$ is the element-wise multiplication and $\mathbf{u}$ is the parameter controlling the amount of negative image feature that replaces the positive image feature. 
The negative image features are extracted from those images similar to the original image at the semantic level, which we will introduce in Section~\ref{sec:nagative_sampling}.

To capture the non-spuriousness, we would like to construct counterfactuals by replacing essential non-spurious features only. This can be achieved by minimizing the amount of feature change $\mathbf{u^*}$ to the original image that can causally incur label change:
\begin{equation}
\begin{array}{cl}
\underset{ \mathbf{u}^*}{\operatorname{minimize}} &\|\mathbf{u}^*\|_{1} \\
\text { s.t. } & \mathbf{u}^*=\arg \underset{\mathbf{u}}{\max} D_{c^-}(\boldsymbol{v'}).
\end{array}
\label{eq:min-u}
\end{equation} 

Given the factual and counterfactual features $\boldsymbol{v}$ and $\boldsymbol{v'}$, we aim to learn the prompt that can help CLIP better align visual features $\boldsymbol{v}$ and textual features $G(\boldsymbol{t})$ with same semantic meanings. This can be achieved by maximizing the mutual information (MI) between $\boldsymbol{v}$ and $G(\boldsymbol{t})$. Therefore, by minimizing the InfoNCE loss~\cite{infonce}, we can maximize the lower bound on MI$(\boldsymbol{v},G(\boldsymbol{t}))$.
To this end, we define the contrastive objective function based on the InfoNCE estimator following~\cite{supervised_contrastive_learning}:
\begin{equation}
\mathcal{L}_{CL}(\boldsymbol{p}, \mathbf{u}^*) = -log(\frac{e^{\frac{S(\boldsymbol{v},G(\boldsymbol{t}))}{ \tau}}}{ e^{ \frac{S(\boldsymbol{v}, G(\boldsymbol{t}))}{\tau}}+ e^{\frac{S(\boldsymbol{v'}, G(\boldsymbol{t}))}{ \tau}}}),
\label{eq:cl}
\end{equation}
where
$S\left(\cdot, \cdot\right) $
is normally the cosine similarity function and $\tau$ is the temperature value.

\subsection{Text-based Negative Sampling}
\label{sec:nagative_sampling}
We then discuss how to perform negative sampling for constructing counterfactual features. 
As suggested in~\cite{contrastive_learning_hard_sampling}, good negative samples have different labels and are difficult to be distinguished from an anchor point, while their semantic representations are close~\cite{suresh2021not}.
Since not all negative samples can serve as useful negatives~\cite{chuang2020debiased}, indiscriminate leverage of these data may harm model robustness and algorithm efficiency.
Therefore, during training, in each batch, we only utilize the most 
semantically-similar one to generate counterfactual image features. Other image samples are filtered out.

Semantic concepts may be highly complex in the visual representations, and thus it is hard to directly measure semantic similarity in the visual space. While language is more expressive and naturally preserves semantic meanings.
Therefore, we propose a text-based negative sampling method. We first measure the text similarity between prompts with BERTScore~\cite{zhang2019bertscore}, which computes pairwise cosine similarity between reference sentences and candidate sentences using BERT contextual embedding~\cite{devlin-etal-2019-bert}. We compute a similarity matrix with the value of each element being:
\begin{equation}
\operatorname{sim}({i, j}) = \operatorname{BERTScore}(\boldsymbol{h}_i, \boldsymbol{h}_j).
\label{similarity}
\end{equation}
Denote $\mathcal{B}$ as the collection of sampled instances. During training, each prompt $ \boldsymbol{h}_c\in \mathcal{B}$ ($1 \leq c \leq C$, where $C$ is the size of sampled instances) can be treated as a query. Given a query prompt $\boldsymbol{h}_q$, its most semantically similar prompt (the one with the highest BERTScore) $\boldsymbol{h}_k$ is searched from $\mathcal{B}$.
Then we use the CLIP vision encoder to obtain the features of the corresponding positive and negative images $\boldsymbol{v}$ and $\boldsymbol{v}^{-}$.

\subsection{Joint Optimization}\label{sec:optimization}
In addition to the contrastive learning loss as introduced in Eq.~\ref{eq:cl}, we also adopt the standard cross-entropy loss for training:
\begin{equation}
\mathcal{L}_{CE}(\boldsymbol{p})=-\sum_{c} \boldsymbol{y}_{c} \log p\left(\boldsymbol{t}_{c} \mid \boldsymbol{x}\right),
\label{eq:ce}
\end{equation}
where $\boldsymbol{y}_c$ denotes the one-hot ground-truth annotation of the label. We treat all downstream tasks in this work as classification tasks, where the model predicts if the image and text prompt pair is matched or not. 

Then the task-agnostic prompt $\boldsymbol{p}$ is learned by minimizing the weighted combination of contrastive learning loss and cross-entropy loss:
\begin{equation}
   \mathcal{L}(\boldsymbol{p})= \mathcal{L}_{CE}(\boldsymbol{p})+\lambda \cdot \mathcal{L}_{CL}(\boldsymbol{p}, \mathbf{u}^*),
   \label{eq:loss}
\end{equation}
where $\lambda$ determines the weight of $\mathcal{L}_{CL}$.

\begin{algorithm}[t]
\begin{small}
	\caption{Counterfactual Prompt Learning}
\label{vqa_alg}
	\begin{algorithmic}[1]
	\State $\mathbb{X}$: image space
	\State $\mathbb{Y}$: label space
	\State $\boldsymbol{h}_{c}$: task-relevant prompt for the $c$-th class
	\State $\mathbb{H}$: the set of task-relevant prompts
	\State $\boldsymbol{p}$: the task-agnostic prompt 
	\State $\boldsymbol{v}$: image features
	\State $\boldsymbol{v}^-$: negative image features
	\State $\mathbf{u}$: parameter controls the generation of counterfactual image features
		\Function {$\mathcal{\textcolor{blue}{CPL}}$}{$\mathbb{X}, \mathbb{Y}$}
		\State $\mathbb{H}\leftarrow \mathbb{Y}$  
		\State $\boldsymbol{t}_{c}\leftarrow[\boldsymbol{p}, \boldsymbol{h}_{c}]$
		\For{each $i,j$}  
		\State $\operatorname{sim}({i, j}) = \operatorname{BERTScore}(\boldsymbol{h}_i, \boldsymbol{h}_j)$~\Comment{Eq.~\ref{similarity}} 
		\EndFor
		\For{$q$ in the batch}  
		\State $\boldsymbol{v} \leftarrow \boldsymbol{v_q}$
		\State Find the index $k$ that maximize $\operatorname{sim}({q, k})$ with the given index $q$
		\State $\boldsymbol{v}^- \leftarrow \boldsymbol{v_k}$
		\State Generate counterfactual image features 
		\label{step:generation}
		\Comment{Eq.~\ref{generation}}
		\State $\mathcal{L}_{CE} \leftarrow$ cross-entropy loss~\Comment{Eq.~\ref{eq:ce}}
		\State $\mathcal{L}_{CL} \leftarrow$ contrastive loss~\Comment{Eq.~\ref{eq:cl}} 
		\State Update $\boldsymbol{p}$ and $\mathbf{u}$ with the joint optimization loss~\Comment{Eq.~\ref{eq:loss}} 
		\EndFor
		\label{step:optimizing}
		\EndFunction
	\end{algorithmic}
	\end{small}
\end{algorithm}

In fact, we can seek to put Eq.~\ref{eq:min-u} and Eq.~\ref{eq:loss} in a single-stage optimization framework. 
The intuition is that we generate counterfactual image features with minimal feature change that can maximize the negative prediction probability, and at the same time, utilize contrastive learning to learn the prompt that can guide CLIP to explicitly distinguish between factual images and counterfactual images. 
Putting all pieces together, we have:
\begin{equation}
\begin{array}{cl}
\underset{\boldsymbol{p}, \mathbf{u}^*}{\operatorname{minimize}} & \mathcal{L}_{CE}(\boldsymbol{p}) +\lambda \cdot \mathcal{L}_{CL}(\boldsymbol{p}, \mathbf{u}^*) + \|\mathbf{u}^*\|_{1} \\
\text { s.t. } & \mathbf{u}^*=\arg \underset{\mathbf{u}}{\max} D_{c^-}(\boldsymbol{v'}) \\
\text{where ~}  \boldsymbol{v'} &= (1-\mathbf{u}) \circ  \boldsymbol{v} + \mathbf{u}\circ \boldsymbol{v}^{-}.
\end{array}
\label{eq:6}
 \end{equation}
In Eq.~\ref{eq:6}, the gradients can be back-propagated all the way through the text encoder $G$ to the task-agnostic prompt, making use of the rich knowledge encoded in the pre-trained CLIP model to optimize the prompt.

Algorithm~\ref{vqa_alg} presents the learning algorithm of CPL. In summary, given few input training samples $\left\{\left(x_{1}, y_{1}\right), \ldots,\left(x_{n}, y_{n}\right)\right\}$, CPL consists of three main steps:
(1) compute the similarity matrix between different text prompts within the sampled batch;
(2) generate counterfactual image features;
(3) optimize $\boldsymbol{p}$ and $\boldsymbol{u}$ with contrastive learning loss and cross-entropy loss.

\subsection{Task-relevant Prompt Construction}\label{sec:task-relevant}
We construct task-relevant prompts $\mathbb{H}$ for image classification, image-text retrieval, and visual question answering, respectively. For image classification, the prompts are class labels for each task; for image-text retrieval, captions for each image are adopted as prompts; for visual question answering, we first use a pre-trained generative T5 model~\cite{t5} to convert the question-answer pairs into declarative sentences referring to the VQA prompt generation method proposed in~\cite{vqa_prompt}. Then, motivated by~\cite{chain_of_thought}, we add additional category information into the prompt generated from templates based on the question type to help the model perform intermediate reasoning steps. Specifically, we add ``The question is asking about others'' for \emph{Other} questions before the generated declarative sentence. In a similar vein, ``The question is asking about yes or no'' and ``The question is asking about numbers'' are added for \emph{Yes/No} and \emph{Number} questions.

\begin{table*}[t]
 \resizebox{\linewidth}{!}{
  \centering
  \setlength{\tabcolsep}{3pt}
  \begin{tabular}{llllllllll}
    \toprule
      Classes  &Method & SUN397& Caltech101 & ImageNet & OxfordPets & StanfordCars & Flowers102 & Food101 & Average\\
    \midrule
    \multirow{3}{*}{Seen} 
        &  CLIP & 69.40& 96.51& 72.46& 91.33 & 74.85 & 72.17 & 90.12& 80.98\\
        & CoCoOp & 79.08 \inc{13.95}& 97.66 \inc{1.19}& 76.01 \inc{4.90}& 95.18 \inc{4.22}& 70.91 \diff{-5.26} & \textbf{94.65} \inc{31.15}& 90.67 \inc{0.61} & 86.31 \inc{6.58}\\
        & CPL (ours)& \textbf{81.05} \inc{16.79}& \textbf{97.70} \inc{1.23}& \textbf{78.81} \inc{8.76} &\textbf{96.69} \inc{5.87} & \textbf{75.51} \inc{0.88} & 93.91 \inc{30.12} & \textbf{93.01} \inc{3.21} & \textbf{88.10} \inc{8.79} \\
    \midrule
    \multirow{3}{*}{Unseen} 
        &  CLIP & 75.40& 94.10& 68.09& 97.04& 74.95& \textbf{77.87}& 91.30&82.68\\
        & CoCoOp & 76.83 \inc{1.90} & 93.92 \diff{-0.19} & 70.44 \inc{3.45} &97.78 \inc{0.76} & 73.09 \diff{-2.48} & 69.24 \diff{-11.08} & 91.53 \inc{0.25} &81.83 \diff{-1.02} \\
        & CPL (ours) & \textbf{80.19} \inc{6.35} &\textbf{94.94} \inc{0.89} & \textbf{73.17} \inc{7.46} & \textbf{98.81} \inc{1.82} & \textbf{78.90} \inc{5.27} & 72.30 \diff{-7.15} & \textbf{93.44} \inc{2.34} &\textbf{84.54} \inc{2.25} \\
    \bottomrule
  \end{tabular}}
  \caption{Result comparison between CPL and CoCoOp~\cite{cocoop} on seen and unseen classes across seven image classification datasets in terms of accuracy (\%) under the few-shot setting.
  The relative difference (\%) compared with CLIP is reported in color. 
  }
 \label{tab:classification}
\end{table*}

\begin{table}[t]
  \centering
 \resizebox{0.8\columnwidth}{!}{
  \begin{tabular}{lllll}
    \toprule
   Training data used & Method & {Flickr30k}& {MSCOCO}& Average \\
   \midrule
   0 & CLIP & 83.00 & 53.35&68.18\\
    \hline
   \multirow{2}{*}{0.5\%} 
    & CoCoOp & 82.40 \diff{-0.72} &55.55 \inc{4.12}&68.98 \inc{1.17}\\
    & CPL (ours) & \textbf{85.64} \inc{3.18} &\textbf{57.91} \inc{8.55}&\textbf{71.78} \inc{5.28}\\
   \midrule
   \multirow{2}{*}{1\%} 
    & CoCoOp & 84.80 \inc{2.17}&56.62 \inc{6.13}&70.71 \inc{3.71}\\
    & CPL (ours) & \textbf{86.91} \inc{4.71} &\textbf{58.43} \inc{9.52}&\textbf{72.67} \inc{6.59}\\
   \midrule
   \multirow{2}{*}{3\%} 
    & CoCoOp & 85.90 \inc{3.49}&58.08 \inc{8.87}&71.99 \inc{5.59}\\
    & CPL (ours) & \textbf{87.74} \inc{5.71} &\textbf{59.96} \inc{12.39}&\textbf{73.85} \inc{8.32}\\
    \bottomrule
  \end{tabular}
  }
  \caption{Result comparison between CPL and CoCoOp on two image-text retrieval datasets, Flickr30k~\cite{flickr} and MSCOCO~\cite{coco}, on the unseen test sets in terms of Recall@1 (\%).  The relative difference (\%) over CLIP is reported in color.}
 \label{tab:ir}
\end{table}

\begin{table}[t]
  \centering
 \resizebox{0.7\columnwidth}{!}{
  \begin{tabular}{lll}
    \toprule
    Training data used & Method & VQAv2 \\
    \midrule
    0 &{CLIP}&11.83\\
    \midrule
     \multirow{3}{*}{0.5\%} & {CoCoOp}& 27.98 \inc{136.52}\\
     & {CPL w/o. Category Information}& 31.68 \inc{167.79}\\
     & {CPL }& \textbf{33.39} \inc{182.25} \\
     \midrule
      \multirow{3}{*}{1\%} & {CoCoOp}& 28.51 \inc{141.00}\\
     & {CPL w/o. Category Information}& 34.70 \inc{193.32}\\
     & {CPL }& \textbf{35.66} \inc{201.44}\\
     \midrule
      \multirow{3}{*}{3\%} & {CoCoOp}& 30.18 \inc{155.11}\\
     & {CPL w/o. Category Information}& 35.41 \inc{199.32}\\
     & {CPL }& \textbf{36.32} \inc{207.02}\\
    \bottomrule
  \end{tabular}}
 \caption{Result comparison on the VQAv2 dataset~\cite{vqav2} in terms of accuracy (\%). The relative improvements over CLIP are reported in color. Incorporating category information into task-relevant prompts can further improve the performance.}
 \label{tab:vqa}
\end{table}

\section{Experiments}
\subsection{Tasks and Datasets}
\paragraph{Image Classification.~}
We employ seven publicly available image classification datasets used in CLIP: SUN397~\cite{sun397}, Caltech101~\cite{caltech}, ImageNet~\cite{imagenet}, OxfordPets~\cite{oxfordpet}, StandfordCars~\cite{standfordcars}, Flowers102~\cite{flower}, and Food101~\cite{food101}. These datasets constitute a comprehensive benchmark, which covers a diverse set of vision tasks
including the classification of generic objects, fine-grained image recognition, action classification, etc. To evaluate the generalization ability of methods, we split those datasets into seen and unseen classes. Only images in the seen classes will be used for training. The setting follows the few-shot evaluation protocol in CLIP, where we use 16 shots for training and full test sets for testing.

\paragraph{Image-Text Retrieval.~} 
We consider two datasets for image-text retrieval: MSCOCO~\cite{coco} and Flickr30K~\cite{flickr}. We adopt the
widely used Karpathy split~\cite{karpathy2015deep} for both the MSCOCO and Flickr30K datasets, where MSCOCO contains 113/5K/5K for train/validation/test.
Flickr30K contains 29K/1K/1K images for train/validation/test. We construct few-shot setting subsets for both CoCoOp and CPL by taking $0.5\%$, $1\%$, and $3\%$ of training instances. We train the model with the subsets and evaluate its performance on the complete
test set. We use Recall at 1 (R@1) as the default evaluation metric.

\paragraph{Visual Question Answering.~}
VQAv2~\cite{goyal} is an extended dataset from the VQA~\cite{vqa} dataset. The questions are categorized
into three types: \emph{Number}, \emph{Yes/No}, and \emph{Other}. We set up the experiments following~\cite{bottom}, which treats visual question answering as a classification problem: for each question, the model picks the corresponding answer from a given set of predefined most frequent candidate answers and matches it with the image. The questions are first converted into a masked template using the pre-trained T5 model and predefined rules. The infilled template along with the questions will be turned into prompts that naturally connect questions and answers. The model will predict whether the given prompt and image pairs are matched. We construct the few-shot setting by taking $0.5\%$, $1\%$, and $3\%$ instances for training.

\subsection{Implementation Details}
\paragraph{Baselines.~} 
We mainly compare CPL with CoCoOp~\cite{cocoop}, one of the earliest prompt tuning methods proposed for vision-and-language pre-trained models.
CoCoOp considers each input image and injects the learnable instance-aware tokens into the context vectors as the final prompt. For a fair comparison, both CPL and CoCoOp adopt CLIP~\cite{clip} as the pre-trained vision-and-language backbone and are compared with respect to their relative improvements over zero-shot CLIP.

\paragraph{Prompt Tuning.~} The task-agnostic prompt is randomly initialized from a zero-mean Gaussian distribution with the standard deviation $0.02$, where we set length $L=4$ by default.  For vision and language tasks, in contrast to image classification, where an image is labeled by a category, the task-relevant prompts comprise more fine-grained details, usually a sentence. We here similarly tokenize the whole sentence using the CLIP word embedding~\cite{clip}, and feed the tokenized results to the text encoder with task-agnostic prompt vectors, to generate the language embedding for each prompt. In both the image-text retrieval and visual question answering, all data in the test set can be treated as belonging to unseen classes.

\subsection{Main Results}
\paragraph{Image Classification.~}
The experimental results for image classification are shown in Table~\ref{tab:classification}. 
With better prompts learned from counterfactual examples, our CPL method achieves clear advantages over CoCoOp for both seen and unseen classes across almost all datasets. Particularly on unseen classes, we gain an average relative improvement of $3.55\%$.

Meanwhile, CoCoOp shows its poor generalization ability. Specifically, we found that CoCoOp performs worse than CLIP on StandfordCars on both seen and unseen classes, and on Caltech101 and Flower102 on unseen classes, indicating that it tends to learn and leverage spurious relations and could not generalize well on unseen classes in some cases. We believe all these mentioned above can be sufficient evidence that
the main idea of CPL, learning non-spurious prompt representation can aid CLIP adapting at test time, is practical. 

\paragraph{Image-Text Retrieval.~}
Table~\ref{tab:ir} reports results on image-text retrieval on the unseen test set. CPL can beat the zero-shot CLIP consistently across the three different settings, demonstrating that CPL can also learn better prompt representation and more effectively exploit the limited amount of data on image-text retrieval. Meanwhile, CoCoOp performs even worse than CLIP on Flickr30k using $0.5\%$ training data, which suggests that a tiny quantity of training data for image-text retrieval can lead to spurious prompt representation if using naïve instance-conditional prompt tuning method.

\paragraph{Visual Question Answering.~}
For visual question answering, the results are shown in Table~\ref{tab:vqa}. As can be seen, CPL surpasses the baseline CoCoOp with a relative improvement of up to $25.08\%$ when using $1\%$ instances for training. This proves the concept that CPL can be effective on more complicated vision-and-language tasks. In fact, visual question answering is more challenging for zero-shot CLIP which is pre-trained for image-text matching. During pre-training, CLIP sees most sentences similar to captions in image-text retrieval and those captions can be directly used as prompts; while for VQA, question-answer pairs have to be adapted into declarative prompts. Therefore, zero-shot CLIP has poor performance on VQA, but few-shot prompt tuning via CPL can help reduce the prompt domain gap significantly. Apart from the vanilla CPL method, we examined another variant of CPL where we do not add additional category information into the prompt (denoted as CPL w/o. Category Information), the results indicate that constructing task-relevant prompts by adding categorical information contributes to the improvement. 

\begin{figure}[t]
\centering
\includegraphics[width=0.8\linewidth]{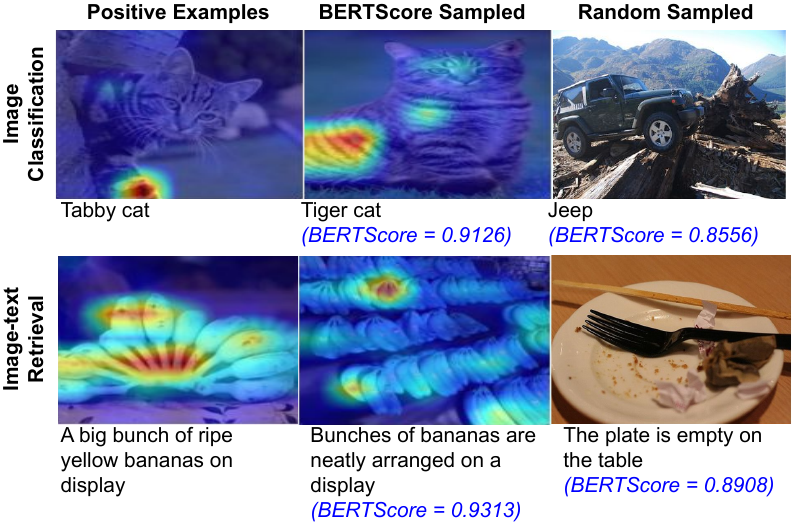}
\caption{Visualization of the weights of the controller parameter $\mathbf{u}$ on images. The first column is the original positive examples; the second column is BERT-sampled negative examples; the third column is randomly-sampled negative examples for comparison. The BERTScore between the text prompts of positive examples and sampled examples are shown at the bottom.
}
\label{fig:visualize}
\end{figure}

\subsection{Ablation Analysis}
\paragraph{Negative Sampling.~} We compare the random sampling vs. BERTScore sampling over ImageNet for image classification, MSCOCO for image-text retrieval, and VQAv2 for visual question answering in Table~\ref{tab:sample}. With more challenging negative examples, BERTScore sampling leads to more effective prompt tuning and overbeats random sampling on all three tasks. The qualitative visualizations of the two sampling strategies are shown in Figure~\ref{fig:visualize}, from which it can be seen that BERTScore-sampled images are much more semantically similar to the original images.

\paragraph{Non-spurious Feature Visualization.}
We visualize the heatmap of the learned non-spurious feature weights in the image level in Figure~\ref{fig:visualize}. The weights are mainly centralized on the semantically meaningful regions that are aligned to the text prompts. 
    
\begin{table}[t]
  \centering
 \resizebox{0.7\columnwidth}{!}{
  \begin{tabular}{llll}
    \toprule
    Method & {ImageNet}& {MSCOCO}& VQAv2 \\
   \midrule
    Random sampling & 75.28&57.78& 33.01
\\
    BERTScore sampling &\textbf{76.02}&\textbf{58.43}&\textbf{35.66}\\
   \bottomrule
  \end{tabular}
  }
  \caption{Random sampling vs. BERTScore sampling for CPL over three tasks.  On ImageNet, we measure the average accuracy across seen and unseen classes. On MSCOCO and VQAv2, we both use 1\% instances for few-shot learning.}
 \label{tab:sample}
\end{table}

\begin{figure}[t]
\centering
\includegraphics[width=0.5\linewidth]{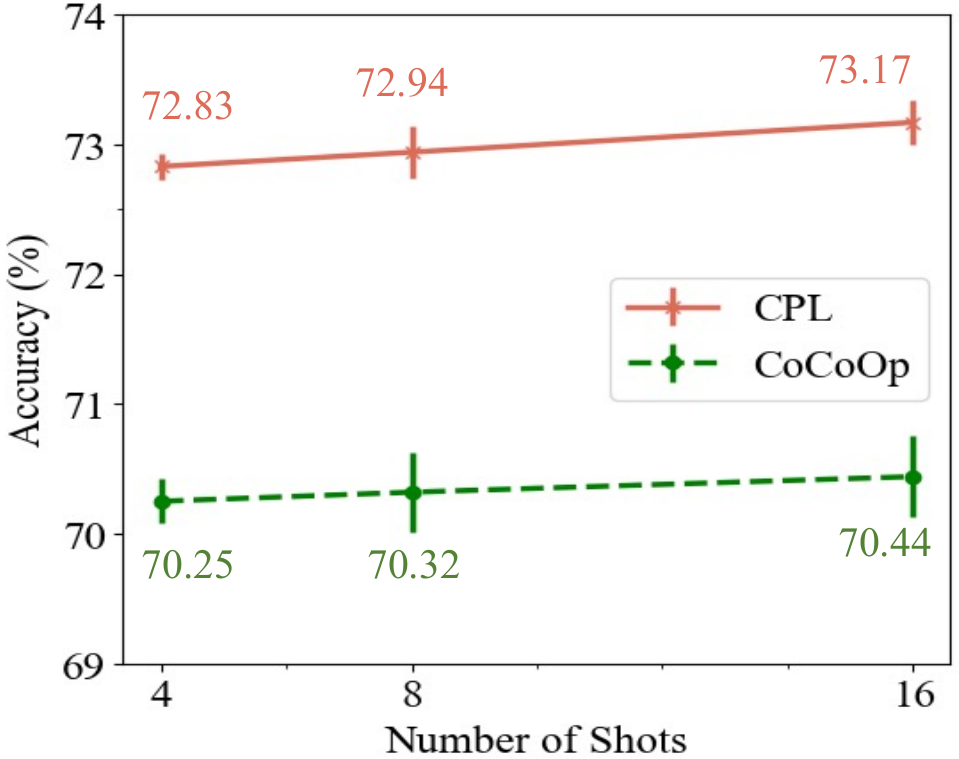}
\caption{Accuracy comparison on ImageNet~\cite{imagenet} unseen classes under three different shots. CPL performs better than CoCoOp consistently and has lower standard errors.
}
\label{fig:shots}
\end{figure}

\paragraph{Number of Shots in Image Classification.~} 
We then study the effects of the number of shots on CPL for image classification. Following the few-shot evaluation protocol adopted in CLIP, we use $4$, $8$, and $16$ shots for training on ImageNet. From Figure~\ref{fig:shots}, increasing the number of shots keeps improving the performance of both two methods on unseen classes. Meanwhile, CPL outperforms CoCoOp under the three different settings and has lower standard errors.

\begin{figure}[t]
\centering
\includegraphics[width=0.5\linewidth]{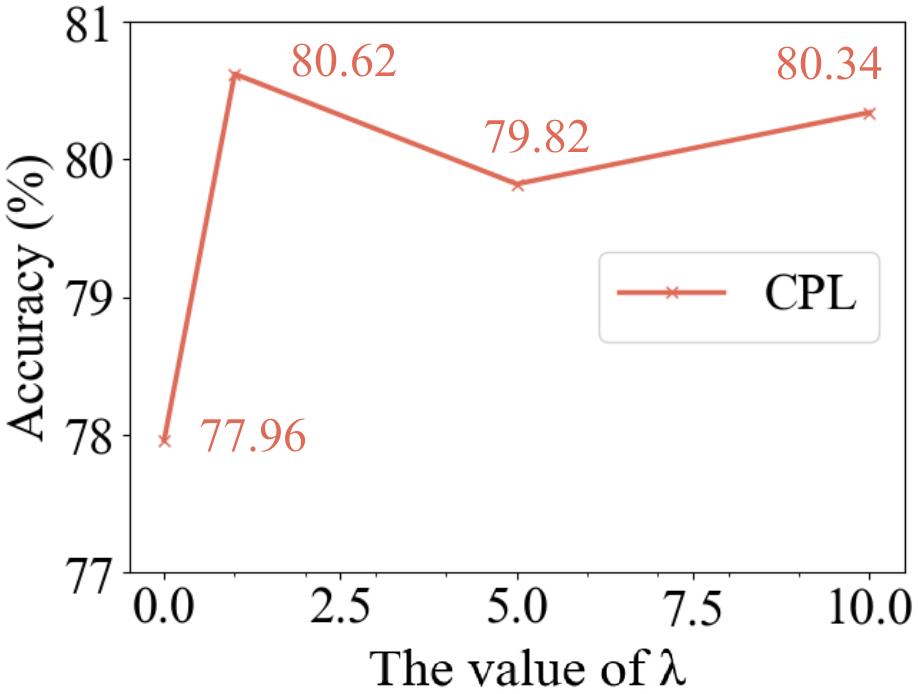}
\caption{Ablation of four different $\lambda$ values on the SUN397 dataset in terms of average accuracy (\%). The performance of CPL peaks at $\lambda =1$.}
\label{fig:lambda}
\end{figure}

\paragraph{Contribution of Contrastive Learning.} 
In Section~\ref{sec:method}, we use the coefficient $\lambda$ to weigh the contrastive learning loss and combine it with the cross-entropy loss. It is observed that the scale of contrastive learning loss is smaller, hence we try to use a larger $\lambda$ to balance the two loss terms. Figure~\ref{fig:lambda}
shows the average accuracy result across seen and unseen classes on the SUN397 dataset under four different $\lambda$ values. Note that when $\lambda$ is zero, there is no contribution from the contrastive loss and the method actually learns the prompt using standard cross-entropy loss. From experimental results obtained on the SUN397 dataset, we can observe that using $\lambda = 1$
leads to the best performance.

\section{Related Work}
\paragraph{Vision-and-Language Models.~}
Vision-and-Language models pre-trained on large-scale image-text pairs have demonstrated great potential in multimodal representation learning~\cite{align,flip,florence,pathvqa,li2023mastering}. Among them, the representative CLIP~\cite{clip} benefits from 400M curated data and defines various prompt templates to carry out zero-shot image classification. However, those prompts still require hand-crafted designs. In this work, we automatically learn task-agnostic and task-relevant prompts without human priors. In addition, by considering the counterfactual examples, we can further improve various vision-and-language tasks, including visual question answering and image-text retrieval in a few-shot scenario.

\paragraph{Prompt Tuning.~}
Many works focus on learning from discrete natural language prompts, e.g., AutoPrompt~\cite{shin2020autoprompt} elicits knowledge from language models with automatically generated discrete prompts. Lately, many other works~\cite{coop, cocoop} directly tune prompts in continuous vector forms. ~\cite{prompt_q_learning} introduces Q-Learning to optimize the soft prompt.  P-Tuning v2~\cite{ptuningv2} shows that continuous prompt tuning achieves the same performance as fine-tuning in various settings. Prompt tuning also receives great interest in the computer vision domain. For example, CoOp proposes a continuous prompt optimization strategy to avoid prompt design.  CoCoOp~\cite{cocoop} extends CoOp by further learning an instance-conditional network to generate an input-conditional token for each image. However, these methods trained with empirical risk minimization (ERM) may learn to rely on correlations between class labels and spurious attributes by minimizing average training error~\cite{correct_contrast}. They usually learn spurious, inefficient, and entangled representation, lacking generalization ability to unseen scenarios. 

\paragraph{Counterfactual Reasoning.~}
A number of recent works have investigated  generating counterfactual images \cite{besserve2020counterfactuals}, or counterfactual text in specific language domains (e.g., court view~\cite{wu2020biased}, dialogue generation~\cite{zhu2020counterfactual}, Natural Language Inference~\cite{kaushik2019learning, semantically_robust_optimization}, named entity recognition~\cite{zeng2020counterfactual}); On the vision end, ~\cite{causal_pose_estimator} proposes to add intervention over the changed domain on images during the data-generation process and steer the generative model to produce counterfactual features to augment the training process. ~\cite{causalvqa} uses automated semantic image manipulations to generate synthetic data to make models more robust against spurious correlations; On the vision and language end, ~\cite{counterfactual_vqa} proposes to generate counterfactual VQA samples by masking critical objects in images or words in questions to augment the training data and gain a huge improvement on the VQAv2 dataset.  ~\cite{mutant} proposes  template-based counterfactual image augmentation methods. ~\cite{counterfactual_vln} proposes a novel training strategy for visual language navigation that dynamically generates counterfactuals to account for unseen scenarios.
To our best knowledge, CPL is the first to apply counterfactual generation to prompt-based few-shot learning for vision and language models.

\paragraph{Few-shot Learning.}
Recently, many few-shot and efficient learning methods on vision~\cite{PEViT} and language~\cite{efficient_language_learning} tasks have been widely studied. At the same time, like CLIP, several different few-shot learners were proposed. GPT~\cite{gpt3}, as a strong few-shot learner, is capable of performing a new language task by learning from only a few training instances.  Frozen~\cite{frozen} is developed based on GPT and made into a multimodal few-shot learner by expanding the soft prompting to include a collection of images and text. Their 
method demonstrates strong few-shot capabilities on visual question answering and image classification tasks. Similarly, CoCa~\cite{coca} is pre-trained from scratch and end-to-end using both web-scale data and annotated
images by considering all labels as text, therefore unifying supervision for learning representations through natural language. It can achieve state-of-the-art performance with few-shot transfer or by minimal task-specific adaptation on a wide range of downstream vision-and-language tasks, including visual recognition, multimodal understanding, crossmodal retrieval, and image
captioning. SimVLM~\cite{simvlm} is pre-trained with prefix language modeling on datasets with weak supervision. It exhibits its efficacy on few-shot captioning tasks. Even though all these models mentioned above can already achieve improvement on some few-shot tasks, how to exploit their few-shot reasoning ability using limited training examples still deserves the effort. In this work, we study this direction via the lens of prompt learning utilizing CLIP as a starting point.

\chapter{Enhancing Compositional Reasoning in Multimodal Models}
\section{Introduction} 

\begin{figure}[t]
    \centering
    \includegraphics[width=\textwidth]{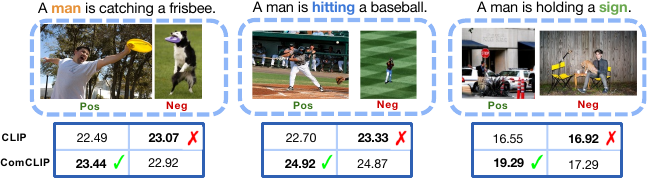}
    \caption{Examples of the compositional image-text matching problem, in which the positive and negative images have very similar semantics except for the only difference in subject, predicate/verb, or object. CLIP mistakenly connects the text prompts with the wrong images on the right (high similarity scores with negative images), while our ComCLIP model does compositional matching more effectively. 
    }
    \label{fig:teaser_comclip}
\end{figure}
In this chapter, we study the compositional visual and text alignment from the causal view, where we start from the image and text matching task.
Image and text matching~\cite{flickr30k, coco} is a fundamental task for vision-language research that involves multimodal reasoning and multi-level visual and text concept alignment.
Recently, a growing number of pretrained vision-language foundation models~\cite{clip,align,blip, GLIP} have shown encouraging results towards open-domain visual and language concept matching. 
Among these models, CLIP~\cite{clip} can be easily transferred to image and text matching under zero-shot and few-shot scenarios. 
However, CLIP treats the image and the text as a whole for alignment and ignores the compositional matching of disentangled concepts, especially for tasks that require the model's compositional understanding ability.
For instance, Figure~\ref{fig:teaser_comclip} shows some examples that CLIP fails at, which require a compositional generalization of the model to understand different subject, predicate, or object concepts.

In fact, it is widely observed that current pretrained vision-language models struggle to recognize actions from the image, distinguishing objects from subjects~\cite{SVO_dataset}, or failing to identify objects in unseen surroundings~\cite{elephant}. 
These may be ascribed to shortcut learning~\cite{shortcut} and dataset biases in pretraining, where the models learn the correspondence between entities and images implicitly and are thus vulnerable to spurious correlations, incurring biases toward particular objects/subjects/predicates and combinations. 

Therefore, there are primarily two challenges to address when adopting CLIP for compositional image and text matching.  \emph{Challenge 1}: the pretrained language model in CLIP is biased and tends to rely on spurious relationships learned in pretraining. For example, in Figure~\ref{fig:teaser_comclip} (A), CLIP associates ``frisbee'' with ``dog'' because of their more frequent co-occurrence and makes the wrong prediction.  Meanwhile, the richness of entities in text descriptions brings \emph{Challenge 2}: entity embeddings should contribute dynamically for compositional matching. In Figure~\ref{fig:teaser_comclip}, the subject/predicate/object entities ``{\color{orange}man}/{\color{blue}hitting}/{\color{teal}sign}'', as identifiers for correct matching in each scenario, should be endowed with more importance. Based on the semantics of input images, CLIP should adjust the weights for these entity embeddings. Yet existing approaches often calculate the similarities merely based on the global embedding of images and texts and overlook fine-grained concept matching~\cite{Visual_semantic_reasoning_for_image_text_matching}.

To address the above limitations, we propose a new \textit{\textbf{training-free}} framework based on CLIP-like models from the causal viewpoint, named ComCLIP. Specifically, we disentangle the visual scene into individual visual concepts and construct counterfactual subimages containing subject/object/predicate entities only. Then we utilize backdoor adjustment~\cite{backdoor_adjustment} to implement interventions over the disentangled subimages to mitigate the effect of spurious correlations. 
With this design, ComCLIP can bind the disentangled visual components with the correct word concept and avoid matching solely based on spurious correlations learned during pretraining and fine-tuning, achieving compositional generalization. 
To validate our approach, we formalize the~\underline{compositional image and text matching} task and construct a new Compositional Visual Genome (ComVG) dataset from the Visual Genome~\cite{visualgenome} dataset for this task. We evaluated on multiple datasets: Winoground, VL-checklist, SVO-Probes~\cite{SVO_dataset}, Flickr30K~\cite{flickr30k}, MSCOCO~\cite{coco}, and the ComVG dataset. Notably, ComCLIP gains an absolute accuracy improvement of 4.50\% on the image score and 2.34\% on the group score over CLIP and SLIP respectively on the challenging Winoground dataset.

\begin{figure*}[t]
    \centering
    \includegraphics[width=\textwidth]{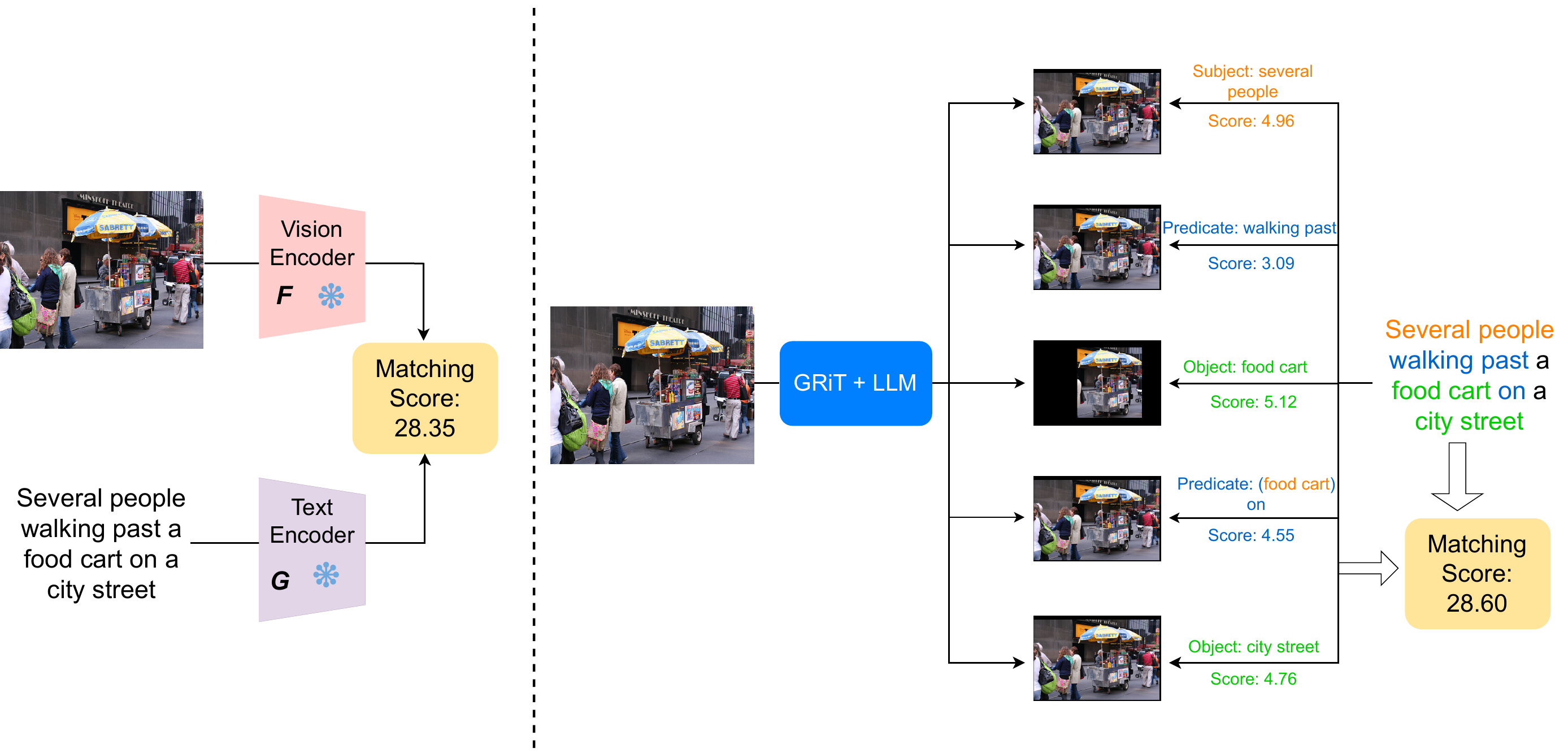}
    \caption{Overview of our ComCLIP framework using CLIP as the backbone. We disentangle the input image using GRiT~\cite{wu2022grit} and the Large Language Model (LLM) by obeying the rules of encoding object, subject, and predicate respectively. 
    The figure shows the case where multiple subjects/objects/predicates are involved (this is a positive example from Flickr30K).
    }
    \label{case_study1}
\end{figure*}

\begin{figure*}[t]
    \centering
    \includegraphics[width=\textwidth]{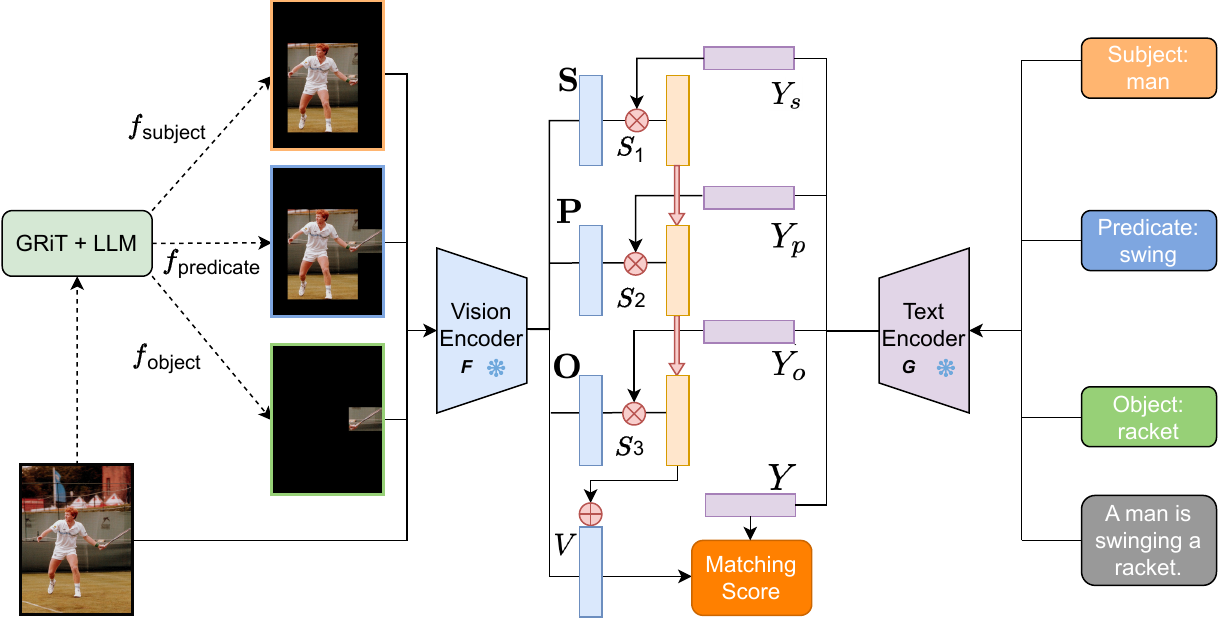}
    \caption{Overview of our ComCLIP framework using CLIP as the backbone. We disentangle the input image using three independent encoding mechanisms by obeying the rules of encoding object, subject, and predicate respectively. The entity information is introduced to the global embedding of the whole image. Module components from CLIP (vision encoder $F(\cdot)$, text encoder $G(\cdot)$) are always frozen. During implementation, the process for matching and calculating the score begins with the input image being processed into object, subject, and predicate subimages. This is followed by feeding both the original sentence and image, along with their parsed words and subimages, into the CLIP text and vision encoders. Subsequently, cosine similarity scores are computed for each pairing of subimage and word embeddings. These scores are then subjected to a Softmax layer, resulting in three positive weights. The next step involves adding the reweighted subimage embeddings to the embedding of the original image. Finally, the ultimate matching score is derived from comparing this aggregated image embedding and the global text embedding.The whole framework is \textbf{\textit{training-free}}.
    }
    \label{fig:overview_comclip}
      \vspace{-1ex}
\end{figure*}

\section{Compositional Visual and Language Alignment}
We first introduce the task of compositional image and text matching, where we are interested in improving the compositional understanding, more specifically, subject/object/predicate understanding of vision-language models. Compositional image and text matching is a task focused on enhancing the understanding of compositional elements such as subjects, objects, and predicates within CLIP-like models. This task requires an appreciation of fine distinctions between texts and their underlying compositional structure, as illustrated in Figure~\ref{fig:teaser_comclip} with phrases like ``{\color{orange}man}/{\color{blue}hitting}/{\color{teal}sign}." The model's ability to differentiate images that only vary by one conceptual element in their accompanying text highlights its comprehension of compositionality.

We formally define this task as follows: given text prompts $Y$ (e.g., 
"\texttt{A man is hitting a baseball}") and a set of entities $T^E=\{e^k\}_{k=1}^K$ such as {\color{blue}hitting}, where $K$ denotes the total number of entities and $e^k$ represents the $k$-th entity, the model's objective is to match the text prompts with the corresponding images. The challenge lies in the inclusion of negative images that contain mismatched entities $\{e^k\}_{k=1}^n$, where $n<k$. These negative images are designed to confuse the model, demanding a nuanced understanding of the entities within a sentence. Simply relying on nouns or spurious relations would not succeed at this task.
To evaluate how well the model grasps this concept of compositionality in texts and matches them with the right images, we introduce an additional ComVG dataset as an extended testing platform.

\section{Improve the Composition Reasoning via Causality} 
We propose ComCLIP to incorporate a causal view into the CLIP-like models. We briefly introduce the background of ComCLIP in view of structured causal models in Section~\ref{sec:background}. Then, we present the overview of ComCLIP pipeline in Section~\ref{subsec:m1}. We introduce its critical components in depth in Section~\ref{visual_concept} and~\ref{composition}. Our objectives are: (i) We aim at disentangling visual input into subimages containing fine-grained compositional concepts. (ii) We intend to utilize those disentangled concepts to perform entity-level matching dynamically and mitigate the effect of spurious relations in the pretrained vision-language models learned during training. 

\subsection{Background}
\label{sec:background}
Causal inference aims to understand how changing one variable can affect another, often represented using concepts such as confounders, interventions, counterfactuals, and do-operations. In the realm of computer vision and natural language processing, the causal relationships can provide insights into the underlying generative processes.

Consider a dataset comprised of (high-dimensional) observations $X$ (i.e., images) and corresponding text prompts $Y$. Assume that each $X$ can be described by lower-dimensional, semantically meaningful factors of variation $z$ (e.g., objects, subjects, or action relations between objects and subjects (i.e., predicates in the image)). These factors, which we term confounders $Z$, may affect either $X$ or $Y$. By disentangling these factors, we can achieve more granular image and text matching. This idea of disentanglement resonates with the principles of structural causal models (SCMs)~\cite{structural_causal_models} and independent mechanisms (IMs). An SCM is a mathematical formulation representing how variables influence one another, often composed of multiple IMs, the individual causal processes. Inspired by SCMs, our approach decomposes the subimage generation process into three independent mechanisms: object mechanism $f_{\text {object }}$, subject mechanism $f_{\text {subject}}$, and predicate mechanism $f_{\text {predicate}}$.

\subsection{Method Overview}
\label{subsec:m1}
We introduce the overview of our method from a conceptual view. The pipeline is shown in Figure~\ref{case_study1} and Figure~\ref{fig:overview_comclip}. 
Our goal is to refine a pretrained vision-language model for fine-grained compositional image-text matching. This involves disentangling an input image to create entity-specific subimages, calculating similarity scores between these subimages and their textual counterparts, and integrating these weighted embeddings with the global image embedding. This process enables the model to capture non-spurious semantic entity information and conduct concept matching at the granular level.

\subsection{Counterfactual Subimage Generation}
\label{visual_concept}
Our method centers on the concept of causality, particularly, the Independent Mechanism (IM) assumption. In the realm of causality, the IM assumption posits that a system's variable generation process comprises autonomous modules that operate without mutual interference~\cite{elements_of_causal_inference}. We adopt this principle and tailor it to our context by considering three independent mechanisms for generating object, subject, and predicate subimages.

While our method is inspired by causal mechanisms, we do not make strong causal claims. Instead, we utilize the intuition that in a complex system, certain variables (or mechanisms) operate autonomously. Given the aforementioned setup, our structural causal model (SCM) takes the form:
$
\mathbf{O} :=f_{\text {object }}\left(X\right), \mathbf{S} :=f_{\text {subject}}\left(X\right), \mathbf{P} :=f_{\text {predicate}}\left(X\right).
$
Where $\mathbf{O}$ is the object image, $\mathbf{S}$ is the subject image, and $\mathbf{P}$ is the predicate image.

With the structural framework above, we answer counterfactual questions, a fundamental concept in causality. Specifically, we pose questions like "What if we retain only the subject/object/predicate in the original image?". The responses to such inquiries allow us to generate what we term as \emph{counterfactual subimages}. The essence of these images is that they exclusively feature the entity in question (see Figure~\ref{fig:overview_comclip}). This procedure leads to the disentanglement of the input image into three distinct and causally independent subimages.

With these foundational blocks in place, our method is geared to connect each disentangled image entity with its corresponding textual counterpart. When each entity is independently and aptly encoded, matching becomes streamlined and efficient. The remaining challenge is to craft a mechanism that effectively governs the composition process of distinct entity regions within an image.

\subsection{Entity Composition}
\label{composition}
As mentioned, the pretrained CLIP-like model is prone to be biased toward specific subjects, objects or predicates, or even rely solely on one of them in the sentence.

From the causal perspective, to match image $X$ with text prompt $Y$ correctly, we want to infer $P(Y|X)$ while at the same time mitigating the effect of detrimental confounders $z$. The confounders may introduce spurious correlations in the model when directly inferring from $P(Y \mid X)$. 

Our goal is to infer $P(Y \mid X)$ while mitigating the effects of detrimental confounders $z$.
Leveraging Bayes Rule, 
\begin{align}
P(Y \mid X)=&\sum_z P(Y, z \mid X)
\\ =&\sum_z P(Y \mid X, z) {P(z \mid X)},
\end{align}
the confounder $z$ introduces the bias of word concept via $P(z \mid X)$. To adjust the effect of confounder $z$, we can intervene $X$ by first disentangling it and then intervening with it using $do$-operation~\footnote{$P(Y \mid do(X))$ uses the do-operator~\cite{do-operator}. Given random variables $X, Y$, we write $P(Y= y \mid d o(X=x))$ to indicate the probability that $Y=y$ when we intervene and set $X$ to be $x$. }: \begin{equation}
P(Y \mid do(X))=\sum P(Y \mid X, z){P(z)}.
\end{equation}
$do(X)$ refers to the process of mitigating the effect of harmful confounders $z$. These confounders $z$, as explained in Section 4.1, are lower-dimensional and semantically meaningful factors that include objects, subjects, and predicates within the image. By mitigating the impact of these confounders, we aim to refine our compositional matching process between the image and text.
We now seek an implicit way to compute $P(Y \mid X, z)$ and ${P(z)}$. Considering the SCMs mentioned above,
we interpret  $f_{\text {object }}(X), f_{\text {subject}}(X), f_{\text {predicate}}(X)$ as incorporating entity semantics into attended regions. 

To do concept matching over the text prompt $Y$ and the entity set $T^E=\left\{e^k\right\}_{k=1}^K$, where $K$ is the total number of entities, and $e^k$ is the $k$-th entity. $T^E$ represents a set of entities extracted from text prompts, during testing, both the image and its corresponding text, along with these parsed entities and their associated subimages, are processed through the CLIP text and vision encoders.

This interpretation motivates us to compute the similarity with different word entity embeddings to achieve concept-wise semantic fusion and guidance. The prediction $P(Y \mid X, z)$ can be regarded as a classifier: $P(Y \mid X, z)=$ Softmax $f_i(X, z)$. Similar to~\cite{wangVisualCommonsenseRCNN2020}, using the approximation of NGSM (Normalized Weighted Geometric Mean)~\cite{show_attend_and_tell}, we have:
$
P(Y \mid do(X)) \approx \operatorname{Softmax}\left[\mathbb{E}_z\left(f_i(X, z)\right)\right].
$
Specifically, to implement this on the ComVG dataset, given an input image $X$ and IMs $f_{\text {object }}(\cdot), f_{\text {subject}}(\cdot), f_{\text {predicate}}(\cdot)$, we first extract a collection of visual concepts from input images. For the language side, given a prompt $Y$ and its entity set $T^E$, we extract all (subject, object, predicate) words ($Y_s, Y_o, Y_p$) from the input text prompts. Using cosine similarity score $\mathcal{S}$ as an example, we compute the concept-level similarity separately:
\begin{equation}
\begin{aligned}
    & S_1 =   \mathcal{S}(F(f_{\text {object }}(X)), G(Y_s)), \\& S_2 = \mathcal{S}(F(f_{\text {subject }}(X)), G(Y_o)), \\& S_3=S(F(f_{\text {predicate}}(X)), G(Y_p)),\; \\& \text{where} \ F(\cdot)=\text{CLIP}_{\text{vision}}(\cdot),\; G(\cdot) = \text{CLIP}_{\text{text}}(\cdot).
\end{aligned}
\end{equation}
The final visual feature is composed by:
\begin{equation}
\begin{aligned}
   V = F(X) +& F(f_{\text {object }}(X))S_1  + F(f_{\text {subject }}(X))S_2  +  F(f_{\text {predicate}}(X))S_3.\; 
   \label{eq:visual_feature}
\end{aligned}
\end{equation}
By adding compositional features back to the global image feature (as in Eq~\ref{eq:visual_feature}) and matching them with the global text features, we balance the need for detailed matching with overall context preservation.

We can compute the image-text matching score by:
$
    O = S(G(Y), V).
 $
With this design, the language part of CLIP is aware of connections between entities from both the visual and language input when doing the concept matching. During implementation, we calculate cosine similarity scores for each pair of subimage and word embedding. These scores are then transformed into weights using a Softmax layer. Subsequently, we enhance the original image embedding by adding these reweighted subimage embeddings. The final step involves computing the overall matching score by comparing this augmented image embedding with the global text embedding, thus finalizing our image-text matching process.

\section{Experiments} \label{sec:exp}
\subsection{Datasets}
\paragraph{Winoground~\cite{thrush2022winoground}} Designed to evaluate vision-language models, this dataset contains 400 instances with two image-text pairs per instance. The challenge is the differing arrangement of identical words across the pairs. Our evaluation spanned the entire dataset.

\paragraph{VL-checklist~\cite{vlchecklist}} Distinguishing itself by combining multiple sources, VL-checklist classifies 410,000 images into three categories. We analyzed a subset of 2000 images from each category to gauge our method's effectiveness.

\paragraph{Flickr30K~\cite{flickr30k}} Each of the 1000 test images has 5 annotations; one annotation is selected randomly. CLIP is evaluated across the dataset; for ComCLIP, the top 10 similar images from CLIP are taken. We create subimages for the top 10 similar images and apply ComCLIP to them.

\paragraph{MSCOCO~\cite{coco}} Like Flickr30K, for each of the 1000 test images, one annotation is selected randomly. The top 10 images from CLIP undergo ComCLIP processing, and subimages are created based on parsed elements.

\paragraph{SVO-Probes~\cite{SVO_dataset}} Built to assess language-image models on distinctions within image elements. From its initial 30,000 data points, we utilized 13,000 due to accessibility issues. We conducted tests using three random divisions and presented the average accuracy.

\paragraph{Compositional Visual Genome (ComVG)~} 
Derived from Visual Genome's~\cite{visualgenome} 2.3 million relationships, we developed ComVG. These relationships, encompassing action and spatial aspects, are in subject-predicate-object triplets. Using these, we created image descriptions and selected 542 distinct relationship images from Visual Genome. Similar to SVO-Probes, we identified variants for each image with single discrepancies in subject, object, or predicate, resulting in 5400 curated test samples with grammatical corrections. ComVG stands out for its high-quality images and focus on text-to-image retrieval.
For comprehensive dataset statistics, kindly refer Table~\ref{tab:main1}. Our evaluation covered the entire ComVG.

More data examples are presented in Appendix.

\begin{table}[t]  
\caption{
  The number of data samples in the dataset that have one of their subjects, objects, or predicates changed between positive and negative images and the number of unique types of subjects, predicates, and objects across ComVG and SVO-Probes (SVO).
  }
 \resizebox{\linewidth}{!}{
  \centering
  \begin{tabular}{ccccccc}
    \toprule
    & Sub-Neg & Pred-Neg &Obj-Neg  &    Subjects &
   Predicates  & Objects   \\
     \midrule
ComVG & 2,584 & 1,536&1,280 &   30 &
65 & 82  \\
  SVO & 5,679 & 23,525&7,637 &    100 &
  421 & 275   \\
\bottomrule
  \end{tabular}}
 \label{tab:main1}
  \end{table}

\begin{figure}[t]
    \centering
    \includegraphics[width=\textwidth]{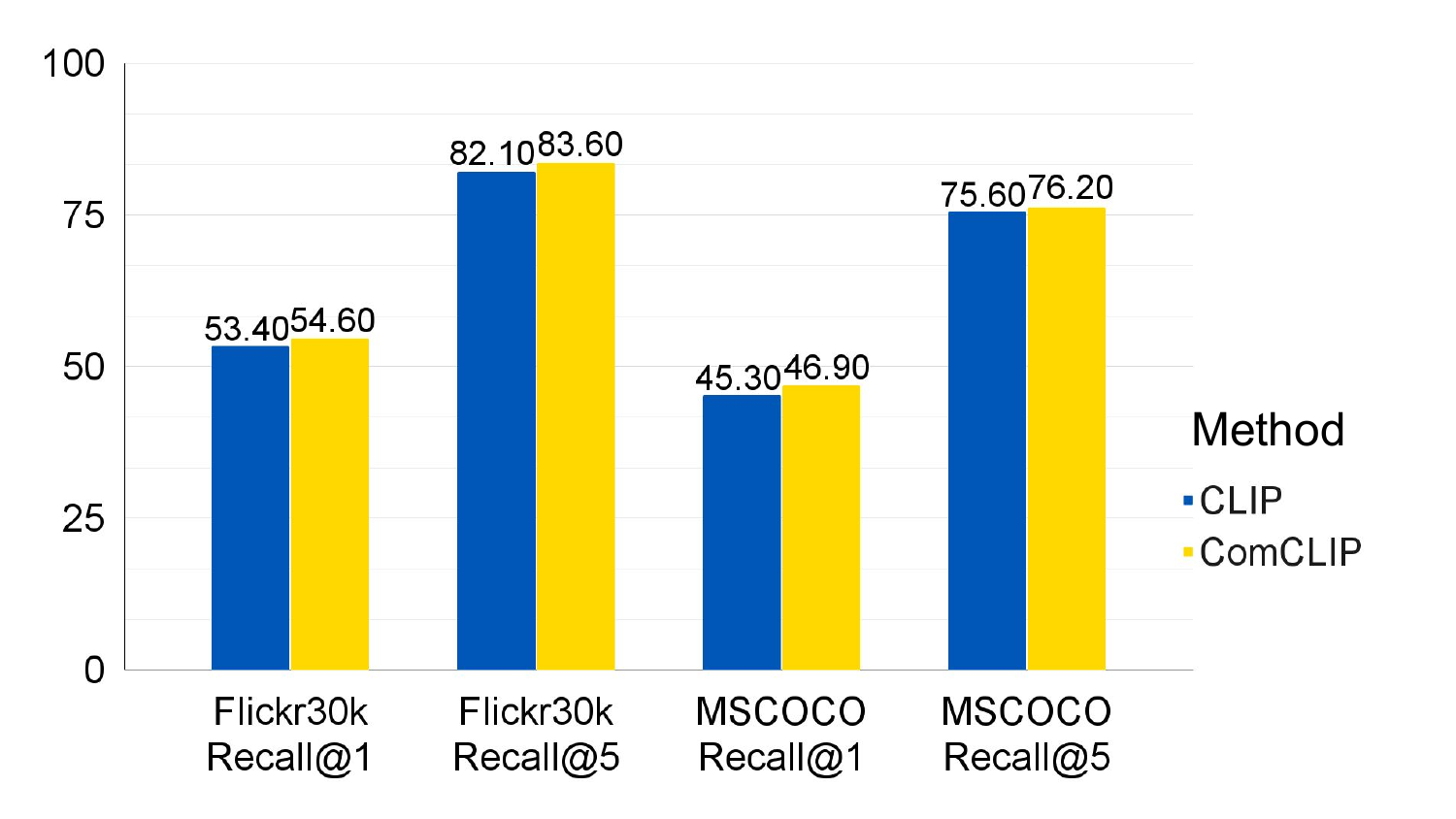}
    \caption{Comparison of Recall@1 (\%) and Recall@5 (\%) using CLIP and ComCLIP over the general image-text retrieval datasets. 
    }
    \label{fig:Flickr8K}
\end{figure}

\begin{table}[t]
  \caption{Comparison of accuracy (\%) on Winoground and VL-checklist using SLIP, and CLIP, and BLIP2. Results marked with $\spadesuit$ are our methods.}
  \centering
  \resizebox{\columnwidth}{!}{
    \begin{tabular}{llllllll}
      \toprule
      \multirow{2}*{Method} & \multicolumn{3}{c}{Winoground} & \multicolumn{4}{c}{VL-checklist} \\
      \cmidrule(lr){2-4} \cmidrule(lr){5-8}
      & {Text} & {Image} & {Group} & {Attribute} & {Object} & {Relation} & {Ave} \\
      \midrule 
      SLIP &  23.25 &  10.00 & 6.75  &  65.95 &  76.81 & 65.30 & 69.35  \\
      ComSLIP $\spadesuit$ & 26.76 \tiny $\mathcolor{red}{(+3.51)}$ &  12.12 \tiny $\mathcolor{red}{(+2.12)}$ & 9.09 \tiny $\mathcolor{red}{(+2.34)}$ & 67.64 \tiny $\mathcolor{red}{(+1.69)}$ & 77.79 \tiny $\mathcolor{red}{(+0.98)}$ & 67.02 \tiny $\mathcolor{red}{(+1.72)}$ & 70.82 \tiny $\mathcolor{red}{(+1.47)}$ \\
     \hdashline
      CLIP & 31.25 & 11.25 & 9.00 & 67.85 & 75.70 & 67.15 & 70.23 \\
      ComCLIP $\spadesuit$ & \textbf{34.00} \tiny $\mathcolor{red}{(+2.75)}$ & \textbf{15.75} \tiny $\mathcolor{red}{(+4.50)}$ & \textbf{10.50} \tiny $\mathcolor{red}{(+1.50)}$ & 69.90 \tiny $\mathcolor{red}{(+2.05)}$ & 79.00 \tiny $\mathcolor{red}{(+3.30)}$ & 69.30 \tiny $\mathcolor{red}{(+2.15)}$ & 72.73 \tiny $\mathcolor{red}{(+2.50)}$ \\
      \hdashline
      BLIP & 29.25 & 12.00 & 8.75 & 79.00 & 84.05 & 73.55 & 78.87 \\
      ComBLIP $\spadesuit$ & 28.75 \tiny $\mathcolor{red}{(-0.50)}$ & 13.00 \tiny $\mathcolor{red}{(+1.00)}$ & 10.00 \tiny $\mathcolor{red}{(+1.25)}$ & \textbf{79.15} \tiny $\mathcolor{red}{(+0.15)}$ & \textbf{84.70} \tiny $\mathcolor{red}{(+0.65)}$ & \textbf{73.95} \tiny $\mathcolor{red}{(+0.40)}$ & \textbf{79.27} \tiny $\mathcolor{red}{(+0.40)}$ \\
      \bottomrule
    \end{tabular}
  }
    \vspace{-1ex}
  \label{tab:winoground}
\end{table}

\begin{table}[t]
  \caption{Comparison of accuracy (\%) on ComVG, and average accuracy (\%) across the three splits on SVO-Probes using  CLIP, GLIP, and ComCLIP. Results marked with $\spadesuit$ are our methods. Ours could also beat GLIP, showing the superiority of our method compared with region-based vision-language pretrained models. 
  }
  \centering
 \resizebox{\columnwidth}{!}{
  \begin{tabular}{lcccccccc}
    \toprule
    \multirow{2}*{Method} & \multicolumn{4}{c}{ ComVG} & \multicolumn{4}{c}{SVO-Probes}\\
    \cmidrule(lr){2-5} \cmidrule(lr){6-9}
    & {Sub} &{Pred} &{Obj}& Ave&{Sub} &{Pred} &{Obj}&Ave \\
       \midrule 
        GLIP & 65.95 & 57.50 & 65.75 &63.85 & 68.91 & 65.14 & 74.94 &67.81\\ \hdashline
      SLIP & 86.20 & 61.33 & 85.84 &80.13 & 79.62 & 79.92 & 78.43 &79.57\\
ComSLIP $\spadesuit$ & 87.43 & 61.25 & 87.11 &81.07 & 79.73 & 80.83 & 79.63 &80.42\\ \hdashline
   CLIP &88.61 & 68.52 & 93.85&86.38 & 85.53 & 80.77 & 90.53 &85.60\\
   ComCLIP $\spadesuit$ &\textbf{90.04} & \textbf{69.06} & \textbf{94.78} &\textbf{87.40} & \textbf{86.70} & \textbf{81.87} & \textbf{90.67} &\textbf{86.41}\\
  \bottomrule
  \end{tabular}
  }
    \vspace{-1ex}
 \label{tab:main}
\end{table}

\begin{table}[t]
  \caption{Comparison of accuracy (\%) on Compositional Visual Genome and SVO-Probes using CLIP, OpenCLIP, and ComCLIP.}
  \vspace{-1ex}
 \resizebox{\columnwidth}{!}{
  \centering
  \begin{tabular}{lcccccccc}
    \toprule
    & \multicolumn{3}{c}{Compositional Visual Genome} & \multicolumn{3}{c}{SVO-Probes} \\
    \cmidrule(lr){2-4}\cmidrule(lr){5-7}
    Vision Encoder & {CLIP} & {OpenCLIP} & {ComCLIP} & {CLIP} & {OpenCLIP} & {ComCLIP} \\
  \midrule
   ResNet-50 &82.25 & 82.21 & \textbf{83.73} & 83.07 & 83.06 & \textbf{84.17} \\
   ViT-B-32 &82.45 & 82.41 & \textbf{84.75} & 84.28 & 84.27 & \textbf{85.18} \\
   ViT-L-14 &86.38 & 86.38 & \textbf{87.40} & 85.61 & 85.60 & \textbf{86.41} \\
  \bottomrule
  \end{tabular}
  }
 \label{tab:mergedResults}
\end{table}

\begin{table}[t]
  \caption{Comparison of accuracy (\%) on Compostional Visual Genome and SVO-Probes using different subimage configuration.
  }
    \vspace{-1ex}
 \setlength{\tabcolsep}{1pt}
 \resizebox{\columnwidth}{!}{
  \centering
  \begin{tabular}{lcccccc}
    \toprule
    & \multicolumn {3}{c}{Compositional Visual Genome} & \multicolumn{3}{c}{SVO-Probes} \\
    \cmidrule(lr){2-4}\cmidrule(lr){5-7}
    {Subimage Configuraion} & {ResNet-50} & {ViT-B-32} & {ViT-L-14} & {ResNet-50} & {ViT-B-32} & {ViT-L-14} \\
  \midrule
   ComCLIP & \textbf{83.73} & \textbf{84.73} & \textbf{87.40} & \textbf{84.17} & \textbf{85.18} & \textbf{86.41} \\
   All black subimages & 82.75 & 83.33 & 86.35&83.09 & 83.83 & 84.47\\
   All original images & 82.25 & 82.45 & 86.38 & 83.07 &84.27 & 85.60\\
   All subject subimages & 82.46 & 82.55 & 86.46 & 83.18 & 84.10&85.24 \\
   All object subimages & 83.28 & 83.73 & 86.48 & 83.85 & 84.53 & 85.72 \\
   All predicate subimages & 82.79 & 83.33 & 86.37 & 83.30 & 84.22 &85.34 \\
  \bottomrule
  \end{tabular}
  }
 \label{tab:subimages}
\end{table}

\subsection{Baselines}
\paragraph{CLIP~\cite{clip}} We use standard CLIP, where image embeddings are generated by CLIP’s vision encoder $F$; and text embeddings are generated by CLIP’s text encoder  $G$. The cosine similarity between them is computed to do matching.

\paragraph{SLIP~\cite{slip}} 
We use the SLIP ViT-L-16. Similar to CLIP, the cosine similarity between the image embeddings and text embeddings is computed to do matching. 

\paragraph{GLIP~\cite{GLIP}}
As GLIP has no global sentence and image embedding, we perform the following rule-based matching: 1) The image with more matched objects is predicted to be matching; 2) For images with the same set of objects, we compute the average confidence score of each object on both images. Larger score image is predicted. 

\paragraph{BLIP2~\cite{blip2}}
We employed the official pretrained BLIP2. For the cosine similarity between image and text features, we adopted BLIP2's image-text contrastive learning match head as our BLIP2 baseline. Specifically, BLIP2 computes the cosine similarity score between each image embedding from each query output and the text embedding of the [CLS] token, selecting the highest similarity score as the ultimate outcome.

\subsection{Implementation Details}
The process begins by processing the original image with the dense caption module of GRiT~\cite{wu2022grit}, producing dense image captions based on object. The input text sentence is then parsed using the large language model (LLM), \texttt{gpt-3.5-turbo}, extracting entity words and organizing them into a subject-predicate-object format. We provide the prompt for parsing sentences for entities: \texttt{Analyze the objects in this sentence, the attributes of the objects and how each object is connected.} The prompt to match objects to text entities: \texttt{Find labels of the image that refer to this object from the sentence.} The alignment between dense image captions and entity words is realized using the same LLM, mapping entity words to their image counterparts based on captions.

For creating a predicate subimage, related object and subject subimages are combined. The original sentence and image, along with their respective parsed words and subimages, are fed into the CLIP text and vision encoders. Cosine similarity scores between each image and word embedding are computed and processed through a Softmax~\cite{softmax} layer, yielding three positive weights. The weighted sum of the subimage embeddings is then added to the original image's global embedding to obtain the final image embedding. The methodology remains similar for SLIP~\cite{slip} and BLIP2~\cite{blip2}, termed as ComSLIP and ComBLIP respectively. Notably, for BLIP2, we project the final image embedding to the sentence embedding dimension for the score computation.

\paragraph{Evaluation Metrics~} 
We use Accuracy as the evaluation metric on the ComVG, SVO-Probes and VL-checklist datasets. For Winoground, we use three accuracy scores: text, image, and group score. The text score quantifies the proportion of both images correctly matched to their corresponding texts. The image score indicates the rate of both texts correctly matched to their corresponding images. Lastly, the group score signifies the accuracy of all texts and images matched correctly. We use Recall~\cite{recall} for Flickr30K and MSCOCO over the general image-text retrieval task.

\subsection{Main Results}
\paragraph{Compositional Image and Text Matching}

\emph{Results on Winoground and VL-checklist} From Table~\ref{tab:winoground}, ComCLIP and ComSLIP consistently outperforms CLIP and SLIP respectively across both datasets, emphasizing their ability to grasp complex image-text relationships.
ComBLIP shows modest improvements, because BLIP2, pretrained on the Visual Genome dataset, already performs strongly.
Overall, it shows that our method's capability to be generalized to other stronger vision-language pretrained models.

\emph{Results on ComVG and SVO-Probes} In this subsection, we show the evaluation results on ComVG and SVO-Probes datasets in Table~\ref{tab:main}. Our ComCLIP can outperform zero-shot CLIP on both ComVG and SVO-Probes datasets. Separately reviewing the results, we see improvements in all negative types. This indicates that incorporating the information of subimages at inference time is helping CLIP attend to the semantic details of images and make fine-grained alignment. Apart from CLIP, we also validate the effectiveness of our method on SLIP~\cite{slip}, denoted by ComSLIP, with the results shown in Table~\ref{tab:main}. As presented, ours can beat SLIP on both the ComVG and SVO-Probes datasets, validating the effectiveness of our method on other CLIP-like models. In addition, we realize that our methods have lower performance improvement on the SVO-Probes dataset compared to ComVG on both CLIP and SLIP. This is because SVO-Probes contains sketchy data samples that we can not fully remove. We discuss some poor examples from SVO-Probes in the Appendix.

\emph{Comparison with GLIP} We compare our methods with GLIP in Table~\ref{tab:main}. Ours outperforms GLIP by a large margin on the compositional image-text matching task, further suggesting the effectiveness of our method compared with other region-based vision-language pretrained models.

\paragraph{General Image-Text Retrieval} 
Results on two image-text retrieval datasets are shown in Figure~\ref{fig:Flickr8K}. CLIP and ComCLIP both perform well in Recall@5, particularly in general image-text retrieval tasks like those in the Flickr30K, where compositionality comprehension is not crucial. ComCLIP outperforms CLIP in Recall@1 on both Flickr30K and MSCOCO, due to its focus on entities and their relations, steering CLIP away from decisions based on single nouns or spurious associations. Overall, these results suggest that our method is also competitive for general image-text retrieval tasks.

\section{Related Work} \label{sec:related}
\paragraph{Image-Text Matching}
Most existing image-text matching datasets are evaluated in a classification setting. For example, ~\cite{hico,Visual_relationship_detection} focus on the relationship or interaction detection.
~\cite{contrastive_weakly_phrase_grounding,faghri2017vse++} explore how creating hard negatives (e.g., by substituting words in train examples) leads to better test performance. FOIL benchmark~\cite{shekhar2017foil} tests if vision-language models can differentiate between sentences that vary with respect to only one noun. SVO-Probes adds hard evaluation examples to test the model's understanding of verbs as well as subjects and objects in a controlled way. To associate local regions in an image with texts to do matching,  ~\cite{xu2015show} incorporates a soft form of attention into their recurrent model.
~\cite{ma2015multimodal} learns multiple networks that capture words, phrases, and sentence-level interactions with images and combines the scores of these networks to obtain a whole image-sentence score.  ~\cite{hu2016natural} leverages spatial information and global context to predict where objects are likely to occur. ~\cite{wang2016structured} formulates a linear program to localize all the phrases from a caption jointly. In this paper, we focus on the task of matching error-prone texts with images, requiring distinguishing words on a granular level --- compositional image and text matching.

\paragraph{Pretrained Vision-Language Models}
Vision-Language models pretrained on large-scale image-text pairs have demonstrated great potential in multimodal representation  learning~\cite{align,flip,florence,GLIP,clip}. Among them, CLIP~\cite{clip} benefits from 400M curated data and defines various prompt templates to carry out zero-shot image classification. GLIP~\cite{GLIP} has incorporated region-level alignment in its pretraining. 
However, these models can suffer from connecting verbs/subjects/objects concepts with visual components correctly~\cite{SVO_dataset} and bias towards spurious relations they have seen in the pretraining data, referred to as ``confounders" ~\cite{devlbert}.
By modeling using a structural causal model (SCM) network~\cite{structural_causal_models}, ~\cite{devlbert} executes a hard intervention to eliminate dataset bias via a backdoor intervention during pretraining. Different from them, in this work, we focus on mitigating the effect of spurious relations and improving the zero-shot inference and compositonal generalization abilities of off-the-shelf pretrained vision-language models. We develop a new training-free paradigm that gains superior performance on compositional image and text matching.

\paragraph{Disentangled Representation Learning}
It is often assumed that real-world observations like images can be disentangled~\cite{representation_learning, elements_of_causal_inference}. \cite{disentangle_image_generation} disentangles background, texture, shape, etc., and uses object bounding boxes as supervision to synthesize images. 
~\cite{besserve2020counterfactuals} leverages the idea of independent mechanisms to identify modularity in pretrained generative models. ~\cite{personReID_align} performs hierarchical alignments in three different granularities, i.e., global-global, global-local, and local-local alignments for description-based person
re-id. ~\cite{fine_video_text_retrieval} improves fine-grained video-text retrieval by decomposing video-text matching into global-to-local levels. 
~\cite{zhang2022multi} proposes a multi-granularity semantic collaborative reasoning network and employs different granularity semantic representations of the question and dialog history to collaboratively identify the relevant information from multiple inputs based on attention mechanisms. ~\cite{counterfactual_generative_networks} utilizes independent mechanisms to generate images to improve image classification. ~\cite{eiclip} disentangles word entities from the conventional meanings of special entities encoded in the pretrained language model. None of these works consider the alignment of subjects, objects, and predicate entities. Different from them~\cite{elements_of_causal_inference}, we employ independent mechanisms to disentangle images and use generated subimages to improve fine-grained visual and language concept matching, which can mitigate spurious correlations introduced by the pretrained model.

\chapter{Leveraging Generative Models for Perception and Reasoning}
\section{Introduction} \label{sec:label}
\begin{figure}[t]
\centering
\includegraphics[width=0.7\linewidth]{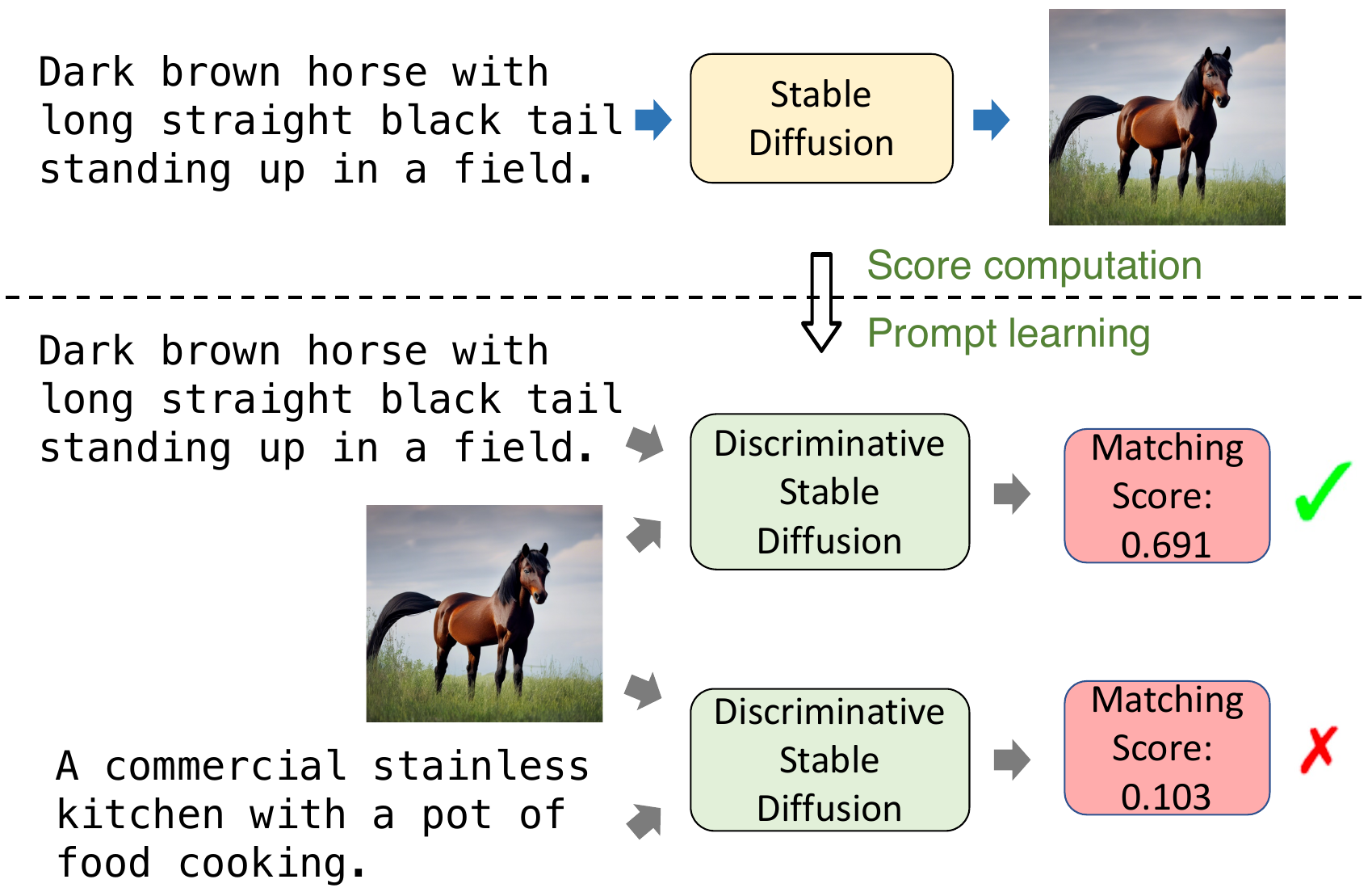}
\caption{The upper subfigure in the teaser image illustrates the ability of Stable Diffusion to generate realistic images given a text prompt. The bottom subfigure illustrates the process of our proposed method, Discriminative Stable Diffusion (Discffusion), for utilizing Stable Diffusion for the image-text matching task. Discffusion can output a matching score for a given text prompt and image, with a higher score indicating a stronger match.}
\label{fig:teaser_dsd}
\end{figure}

Despite the success of various methods in the image-text matching task~\cite{image_text_matching1, image_text_matching2}, there is still a need for more advanced models that can better capture the fine-grained details, spatial relationships, and compositionality. Meanwhile, diffusion models~\cite{diffusion_models,stable_diffusion} have been shown to produce high-quality and diverse images from text descriptions.  
Therefore, in this paper, we investigate the idea of leveraging the power of pre-trained Diffusion Models, specifically the state-of-the-art text-to-image generative model---Stable Diffusion~\cite{stable_diffusion}, for the discriminative image-text matching task, as shown in Figure~\ref{fig:teaser_dsd}. The success of Stable Diffusion in generative tasks suggests that it has a strong understanding of the relationship between visual and textual information, and we aim to harness this understanding for image-text matching tasks.

The key advantages of using Stable Diffusion for image-text matching are two folds: ~\underline{first}, Stable Diffusion uses a VQVAE~\cite{VAE,van2017neural} and cross-attention layers in its architecture, which provides strong compressed representations and shed information about the alignment of data from different modalities. ~\underline{Second}, Stable Diffusion, through proper adaptations, demonstrates a capacity to comprehend spatial relations~\cite{wu2023harnessing,derakhshani2023unlocking} and to discern fine-grained, disentangled concepts~\cite{matsunaga2022fine,rambhatla2023selfeval}. It can generate images that closely align with the specifics of textual prompts, while traditional vision and language model such as CLIP~\cite{clip}, which are pre-trained on discriminative tasks, predominantly facilitate coarse-grained, contextual image-text alignment. Such models lack the capability to perform compositional matching at finer granularity, specifically failing to achieve detailed cross-modal alignment at the region-word level~\cite{comclip}. 

However, to efficiently adapt Stable Diffusion to the image-text matching task, two key challenges need to be addressed: (1) how to disentangle the degree of alignment between the image and text from the latent space of Stable Diffusion? In text-to-image generation, the model is trained to generate an image that is semantically consistent with a given text prompt. However, in image-text matching, the task is to determine the degree of alignment between a given image and text. Therefore, it is important to disentangle the degree of alignment between the image and text in the latent space of Stable Diffusion, to effectively use it for image-text matching; (2) How to efficiently adapt the model in a few-shot setting. Adapting a text-to-image generation model such as Stable Diffusion for image-text matching involves transitioning the model from a generative to a discriminative task. This shift presents significant challenges due to the differences in task requirements and underlying model architectures.

To address these challenges, we propose the Discriminative Stable Diffusion (Discffusion) method, which includes two key ideas: (1) identifying and leveraging attention scores from the selected cross-attention maps as the matching score and (2) employing attention-based prompt learning for model fine-tuning.

\section{Preliminaries on Diffusion Models}

In this section, we provide a brief overview of the concepts and techniques in denoising diffusion models that are necessary to understand our proposed method.  Diffusion models are a class of generative models that are particularly effective at generating high-quality images~\cite{sohl2015thermodynamics,nichol2021glide,ramesh2022hierarchical,imagen2022,rombach2021highresolution}. They aim to model a distribution $p_\theta\left(x_0\right)$ that approximates the data distribution $q\left(x_0\right)$ and is easy to sample from. DPMs model a "forward process" in the space of $x_0$ from data to noise by adding noise to real data, and a reverse process that tries to reconstruct the original data from the noisy version. The forward process is described by the equation \begin{equation}
q(x_t|x_0) = \mathcal{N}(x_t; \sqrt{\bar{\alpha}_t}x_0, (1 - \bar{\alpha}_t)\mathbf{I}),
\end{equation} where $x_{1: T}$ defines a set of noisy images and $x_0$ is the initial image. $\mathcal{N}$ denotes a Gaussian distribution, and $\bar{\alpha}_t$ are hyper-parameters. The reverse process is modeled by a Gaussian distribution 
\begin{equation}
p_\theta(x_{t-1}|x_t) = \mathcal{N}(\mu_\theta(x_t), \Sigma_\theta(x_t)),
\end{equation}
where neural networks are used to predict the mean and covariance of the distribution. The parameters of the model, $\theta$, are learned by optimizing a variational lower bound on the log-likelihood of the real data. Once trained, new images can be generated by starting from a noise sample and iteratively sampling from the reverse process distribution until reaching the final time step. In latent diffusion probabilistic models such as Stable Diffusion, these two processes are similar, while they proceeds in the latent space: $x_0$ is encoded into $z_0$ in an efficient, low-dimensional latent space first and then do the diffusion process. And in the case where a DPM is conditioned on additional information, such as text information $c$, the reverse process becomes $p_\theta(z_{t-1}|z_t,y)$, where $y$ is the input text.

\section{Discriminative Diffusion Models }
\begin{figure*}[t]
\centering
\includegraphics[width=0.9\linewidth]{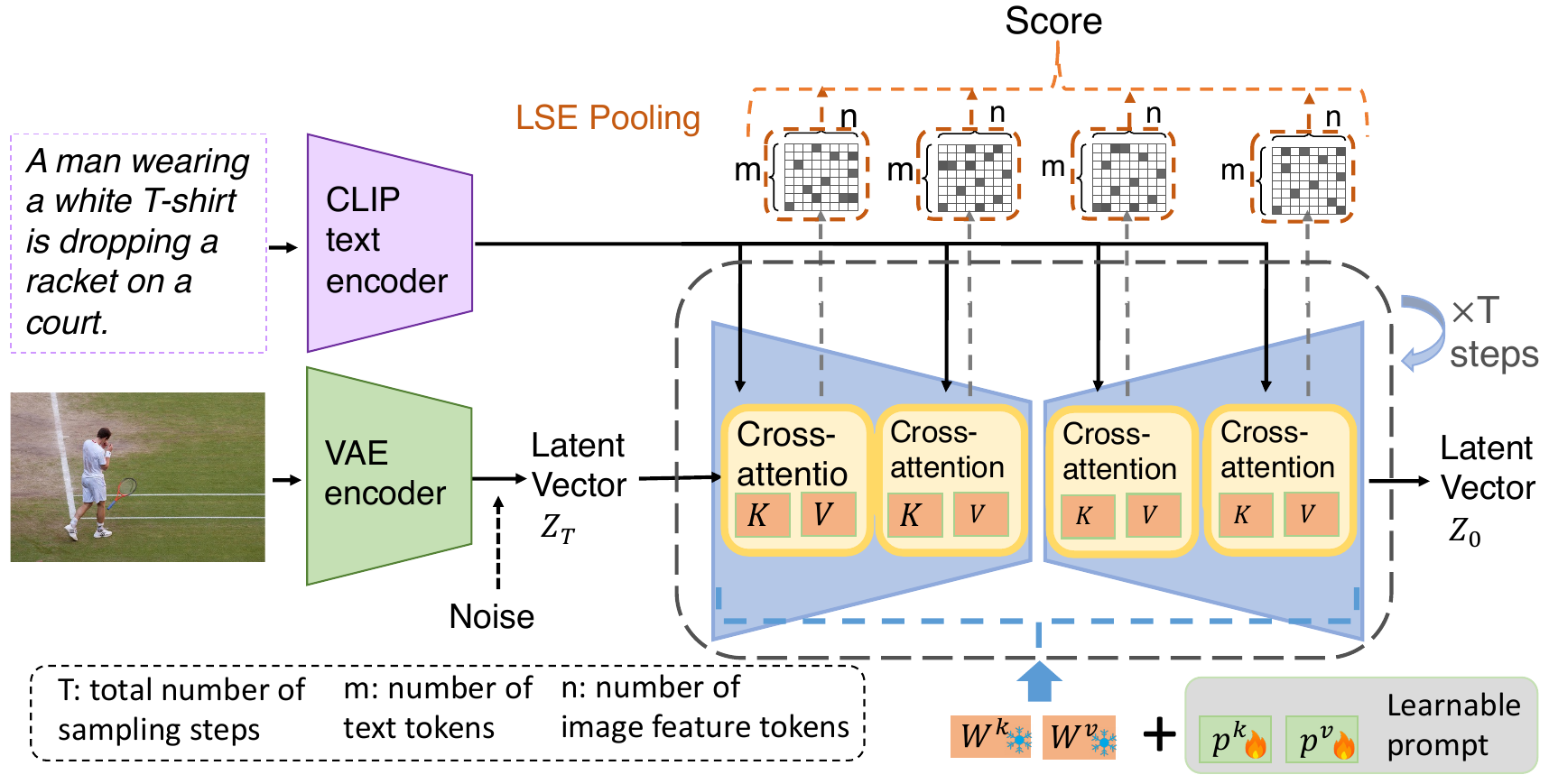}
\caption{The overview of our Discriminative Stable Diffusion framework, which measures how much the given images and texts matched use the cross-attention mechanism in the Stable Diffusion. Discriminative Stable Diffusion added learnable prompt over attention matrices (red boxes). The learnable prompt will receive gradients during training and updated, while the pretrained weights are fixed. The layer index in $m$ and $n$ is dropped for simplicity.
}
\label{fig:overview_dsd}
\end{figure*}

\subsection{Method Overview}
To learn the function $f$, the main idea is to leverage the powerful representations learned by a pre-trained Stable Diffusion model to perform image-text matching. There are three key modules in Discffusion, \textit{cross-attention score computation}, \textit{LogSumExp pooling}, and \textit{attention-based prompt learning}, as shown in Figure~\ref{fig:overview_dsd}. The cross-attention score computation module extracts the mutual influence between visual and textual information by computing the attention scores from the cross-attention matrix in U-Nets of the Stable Diffusion model. The LogSumExp pooling module pools these attention scores over all tokens in the text description to obtain a single matching score. Finally, the attention-based prompt learning module fine-tunes the model by updating the key and value mappings from text to latent features in the cross-attention layers under a few-shot setting. This allows the model to learn new image-text concepts while retaining the ability to capture complex and nuanced relationships between images and text. The model outputs a score that measures the alignment between the image and text, which can be used to adapt the model from a text-to-image generation task to an image-text matching task. Note that Discffusion is a general pipeline that can adopt other faster sampling strategies as well.

\subsection{Cross-attention Score Computation}
Cross-attention scores can be a measure of the relevance of an image and a text to each other~\cite{uniter, visualbert}.  Prior research~\cite{prompttoprompt} shows that the cross-attention within diffusion models governs the layout of generated images and the scores in cross-attention maps represent the amount of information flows from a text token to a latent pixel. They are calculated by taking the dot product of the representations of the image and text in a latent space, and normalizing by the product of their norms. We propose to adapt cross-attention scores as a way to better capture the complex relationships between images and text in the image-text matching task. In the sequel, we elaborate on our strategy in depth.

Stable Diffusion~\cite{stable_diffusion} is trained to generate images from text prompts, and as such, it has learned strong compressed representations of both text and images. This enables us to use these representations to learn the function $f$ for image-text matching.

More specifically, a text prompt $y$ is first encoded into an intermediate representation $r_y = \tau_\theta(y) \in \mathbb{R}^{m \times d_\tau}$ using a domain-specific encoder $\tau_\theta$, where $m$ represents the number of text tokens. We then encode each image $x \in \mathcal{X}$ where $x \in \mathbb{R}^{H \times W \times 3}$ in RGB space into a latent image representation $z=\mathcal{E}(x)$, where $\mathcal{E}(x)$ is the image encoder. The noisy version of the input $z$ is obtained by: \begin{equation}
z_t=\sqrt{\bar{\alpha}_t} z+\sqrt{1-\bar{\alpha}_t} \epsilon 
\text {   \ \ \       for    \ \ \   } \epsilon \sim \mathcal{N}(\mathbf{0}, \mathbf{I}),
\label{noisy_latents}
\end{equation} where constants $\bar{\alpha}_t$ are hyper-parameters inherited from~\cite{ddim}, which will control the level of noises applied to the latent representation. The encoder $\epsilon_\theta$ in the U-Net~\cite{unet} of the pre-trained text-to-image generation model then encode $z_t$ into $r_x = \varphi_i\left(z_t\right)$, where $\varphi_i\left(z_t\right) \in \mathbb{R}^{n \times d_\epsilon^i}$, $n$ and $i$ denote the number of image feature tokens and index of layers respectively  (we dropped the superscript $i$ for $n$ for notation simplicity). This forms a (flattened) intermediate depiction within the U-Net that uses $\epsilon_\theta$,  which are subsequently integrated into intermediate layers of the U-Net via a cross-attention mechanism defined as Attention$(Q, K, V )=\operatorname{softmax}\left(\frac{Q K^T}{\sqrt{d}}\right)\cdot V$, with
$
Q=r_x \cdot W^{q^{(i)}},  K=r_y \cdot W^{k^{(i)}},  V=r_y \cdot W^{v^{(i)}}.
$
Here, $W^{q^{(i)}} \in \mathbb{R}^{d_\epsilon^i \times d}, W^{k^{(i)}} \in \mathbb{R}^{d_\tau \times d}, W^{v^{(i)}} \in \mathbb{R}^{d_\tau \times d}$ are learnable projection matrices~\cite{perceiver, attention}. These matrices map the inputs to query, key, and value matrices, respectively, where 
$d$ is the output dimension of the projection operations in the attention computation.

\subsection{LogSumExp Pooling (LSE)}
To compute the function $g$ and quantitatively evaluate the degree of semantic alignment between an
image and a text prompt, we leverage LogSumExp (LSE) pooling~\cite{logsumexp} as a means of aggregating the attention maps generated by the cross-attention mechanism in our model. By using
LSE pooling, we are able to take into account the relative importance of different image and text
tokens in the attention map, rather than simply averaging or summing all elements in the map. This
has several benefits. 
Firstly, LSE pooling is able to handle large values and outliers in the attention
map more robustly than other pooling methods, such as average or sum pooling. Secondly, LSE
pooling has high numerical stability during training. Thirdly, LSE pooling is able to better preserve
the ordering of values in the attention map, allowing for more interpretable and accurate matching
scores.

For notation simplicity, we drop the batch and attention head dimension, the attention map matrix is denoted as \(A \in \mathbb{R}^{n \times m}\), where \(n\) represents the number of image tokens (height × width) in the latent
space and \(m\) represents the number of text tokens. To compute the image-text pair matching score, we utilize the LogSumExp (LSE) pooling operator, which is applied directly to each row of the attention matrix \(A\), resulting in a vector of length $m$, and then averaged to produce the final score:
\begin{equation}
f(A) = \frac{1}{n} \sum_{i=1}^{n} \frac{1}{\lambda} \log \left( \sum_{j=1}^{m} \exp(\lambda A_{ij}) \right)
\label{combined_score_computation}
\end{equation}
Here, \(A_{i,:}\) represents the \(i\)-th row of the matrix \(A\), Ave is the average operator, and \(\lambda\) is a scaling factor that magnifies the importance of the most relevant pairs of image region features and attended text sentence vectors. By default, we set \(\lambda = 1\). This operation computes the log-sum-exponential over each row, summarizing the contribution of each text token to the image features, and then averages these values across different timesteps to produce a single scalar score for the image-text pair \((y, x)\). We further enhance the model's capability for image-text matching through attention-based prompt learning, which we will introduce in the next section.

\subsection{Attention-based Prompt Learning for Stable Diffusion}
We aim to adapt the latent diffusion probabilistic model to the image-text matching task leveraging only a few examples, that is, under the few-shot setting. The task of fine-tuning aims at updating the mapping from the given text to the aligned image distribution, and the text features are only input to $W^k$ and $W^v$ projection matrix in the cross-attention block. Inspired by~\cite{ptuningv2,prompt_tuning,liPrefixTuningOptimizingContinuous2021a}, we propose the use of learnable prompts, which are added to the attention matrices in our model. Specifically, as shown in Figure~\ref{fig:overview_dsd}, we introduce learnable prompt embedding matrices, which are added element-wise to the key and value attention matrices at the identified layer of the Stable Diffusion model. We keep the original $W^k$ and $W^v$ frozen, and only update the learnable prompt weights added to them. As our addition operation applies to all layers and sampled timesteps, we will omit superscripts {$t$} and layer $l$ for notational clarity and obtains:
\begin{equation}
W^{k^{\prime}}=W^k+p^k, W^{v^{\prime}}=W^v+p^v.
\label{prompt}
\end{equation}
 Both $W^k$ and $W^v$ are frozen and $p^k$ and $p^v$ receive gradients and are updated during training, which is inspired by LoRA~\cite{huLoRALowRankAdaptation2021} and can further improve the training efficiency.
 This allows the model to adapt to new instances by attending to relevant information in the intermediate representation of the text inputs, $\tau_\theta(y)$. With the learned prompt embeddings in the few-shot scenario, we can effectively adapt the Stable Diffusion to image-text matching performance. For optimization, we use the margin-based triplet loss function between the predicted match score and the true match score. Let $L$ be the loss, we have:
\begin{equation}  
 L = \sum_{i} \max\left(0, f(x_{\text{neg}}, y) - f(x_{\text{pos}}, y) + M\right),
 \label{loss}
\end{equation}
where
\( f(\cdot, \cdot) \)  represents the score function defined in Eq.~(\ref{combined_score_computation}) computed from the cross-attention map $A$ with image $x$ and text $y$, $i$ denotes the sample index, 
\( {x_{\text{pos}}} \) is the groundtruth image corresponding to the \( i \)-th text \( y \),
\( {x_{\text{neg}}} \) is the negative image, and
\( M \) is a predefined margin where we use $0.2$ in our experiments.

\paragraph{Dynamic Attention Head Weighting}
We propose a method for adjusting the weights of different attention heads in the cross-attention of our model. 
We compute the gradient of the output of each attention head with respect to the input, and use this gradient to weight the contribution of each head to the final output of the model as follows:
\begin{equation}
\operatorname{Attr}_h(A)=A_h \odot \sum_{k=1}^H \frac{\partial {f}(A)}{\partial A_h},
\label{attributed_attention}
\end{equation} where $\odot$ is element-wise multiplication, $A_h \in \mathbb{R}^{n \times m}$ denotes the $h$-th head's attention weight matrix, and $\frac{\partial f(A)}{\partial A_h}$ computes the gradient along $A_h$. The $(i, j)$-th element of $\operatorname{Attr}_h(A)$ computes the interaction in terms of the $h$-th attention head. The gradient weights are detached and utilized solely as scalar multipliers during training. During training, we then update the attention map with $\operatorname{Attr}_h(A)$. By adjusting these weights, we are able to estimate the gradient flow, keep the most essential token alignment, guide the matching process between text and image and achieve better performance in our task.
The inference time of Discffusion is introduced in the next section.

\subsection{Inference}
During inference, Discffusion leverages the latent representations of both images and text to compute a matching score that determines how well the image corresponds to the text prompt. Specifically, for each pair of image $x$ and text $y$ in the batch, Discffusion first obtains the noisy latent representation $z_t$. Then, the text prompt $y$ is encoded into its latent representation $r_y$ using the domain-specific encoder $\tau$. The U-Net is used to obtain the intermediate image representation $r_x$ from the noisy latent representation $z_t$. The attention maps $A$ are computed using the latent representations $r_y$ and $r_x$. Finally, the matching score $f(A)$ is computed from the attention maps from different sampling timesteps, where Discffusion can adopt various sampling strategies designed for diffusion-based generative models. This process integrates both high-level and low-level features from the image and text during the diffusion process, effectively leveraging the strong representation of the diffusion model to perform image-text matching.

\section{Accelerating Diffusion Models for Zero-Shot Classification}

 Given an image $\boldsymbol{x}$, the goal is to predict the most probable class assignment
\begin{equation}
\begin{aligned}
\hat{y} & =\underset{\mathrm{y}_i}{\arg \max }p\left(y=\mathrm{y}_i \mid \boldsymbol{x}\right) 
\\ & =\underset{\mathrm{y}_i}{\arg \max } p\left(\boldsymbol{x} \mid y=\mathrm{y}_i\right) \cdot p\left(y=\mathrm{y}_i\right) \\
& =\underset{\mathrm{y}_i}{\arg \max } \log p\left(\boldsymbol{x}\mid y=\mathrm{y}_i\right),
\label{eq:3}
\end{aligned}
\end{equation}
where we assume a uniform prior $p\left(y=\mathrm{y}_i\right)=\frac{1}{k}$ that can be dropped from the arg max.

Convert the label $\mathrm{y}_i$ from each class name into text prompts using a dataset-specific template (e.g. $\mathrm{y}_i \rightarrow$ $\mathbf{c}_i$: A photo of a $\mathrm{y}_i$). Then we can convert eq.~\ref{eq:3} to be solved via VLB~\cite{kingma2013auto} by:
\begin{equation}
\begin{aligned}
\hat{y} & =\underset{\mathrm{y}_i}{\arg \max } \log p_\theta\left(\boldsymbol{x} \mid y=\mathrm{y}_i\right) \\
& \approx \underset{\mathrm{y}_i}{\arg \min }  \mathcal{L}_{\text {Diffusion }}\left(\boldsymbol{x}, \mathrm{y}_i\right) \\
& =\underset{\mathrm{y}_i \in\left[\mathrm{y}_i\right]}{\arg \min } 
 \mathbb{E}_{t, \epsilon}\left[\boldsymbol{w}_t\left\|\epsilon-\epsilon_\theta\left(\mathbf{x}_t, \mathbf{c}\right)\right\|^2\right],
\end{aligned}
\label{vlb}
\end{equation}
and $\boldsymbol{w}_t$ is a weight assigned to the timestep $t$.

\section{Accelerated Sampling}

In this section, we introduce an improved, hierarchical sampling strategy that enhances the efficiency of the sampling process for using pretrained diffusion models as classifier and optimizes the process of class prediction.

\paragraph{Class Scoring and Prediction}
Upon obtaining a noisy image, we apply Stable Diffusion to denoise and predict $\boldsymbol{x}$ from $\boldsymbol{x}_t$, yielding $\hat{\epsilon}=\epsilon{\theta}\left(\boldsymbol{x}_t, \mathbf{c}, t\right)$. We designate the squared error of the prediction, $||\epsilon-\hat{\epsilon}||_2^2$, as the score for $\left(\boldsymbol{x}, \mathrm{y}_i\right)$. We compute this score for each class $N$ times. The final step involves weighting the scores based on the corresponding $\boldsymbol{w}_t$ and averaging them across all sampled timesteps to generate a prediction score for each class.

\paragraph{Hierarchical Sampling Strategy}
In contrast to the conventional sampling strategy suggested by~\cite{diffusion_classifier}, which allocates equal sample numbers to each class at every timestep, our approach places emphasis on classes with higher prediction probabilities.

Our strategy maintains a beam of classes, initially sized to $C/b$, where $C$ is the total number of classes in the dataset and $b$ is the \texttt{BeamFactor} hyper-parameter which determines the number of classes to retain during the sampling process. Given an input text prompt $\mathbf{c}$, the process begins by sampling $N$ instances from the starting timestep $t_0$. However, unlike previous methods, this process is not repeated for every class. Instead, we retain only the top $C/b$ classes that demonstrate the highest performance after each timestep. This selective approach continues until a single class consistently achieves the highest probability across $t_s$ additional samplings or each timestep has been sampled for $N$ times.

This future work offers novel insights into the capabilities of stable diffusion. We posit that text-to-image diffusion models can learn powerful representations and serve as an efficient classifier. The hierarchical sampling strategy we introduce serves as a stepping stone towards making these models more accessible and practical for a wider range of applications, thereby unlocking new potential for their deployment in real-world scenarios.

\section{Experiments}
\subsection{Datasets}
We use the Compositional Visual Genome (ComVG)~\cite{visualgenome} and RefCOCOg~\cite{refcocog} datasets to do image-text matching, which requires model's ability to understand fine-grained details, spatial relationships, and compositionality of image and text pairs. Additionally, we include the VQAv2 dataset, which we adapted for image-text matching by concatenating questions with their corresponding answers to form unique text prompts for matching with images, to demonstrate the versatility and robustness of Discffusion across different vision and language tasks. Apart from these, Winoground~\cite{thrush2022winoground} and VL-checklist~\cite{vlchecklist} are also included for evaluating the understanding ability of compositionality.

\paragraph{Compositional Visual Genome (ComVG)~\cite{visualgenome}} is a reconstructed dataset of the Visual Genome~\cite{visualgenome} dataset, which contains 108,007 images annotated with 2.3 million relationships. These relationships are represented as subject-predicate-object triplets and include both action and spatial relationships. ComVG was created by selecting a subset of 542 images from Visual Genome that contain clear relationships, resulting in a total of 5400 data points. It mainly contains three subcategories: subject, predicate, and object. We perform evaluation on the entire dataset as well as on the `Subject', `Predicate', and `Object' subcategories, respectively.

\paragraph{RefCOCOg~\cite{refcocog}} is a reconstructed dataset of the MS-COCO~\cite{coco} dataset, including 85,474 referring expressions for 54,822 objects in 26,711 images, with a focus on images containing between 2 and 4 objects of the same category. 

\paragraph{VQAv2~\cite{goyal}}
The VQAv2 dataset~\cite{goyal} contains questions such as `Binary', `Other', `Numbers', it is commonly converted to a classification task with 3,129 answer classes with frequency large than 9. In our setting, we modify the candidate text to be the concatenation of question and answer pair for each question and perform matching with images. We perform evaluation on all the modified dataset as well as on the `Binary' and `Other' subcategories, respectively.

\begin{table*}[t]
\centering
  \caption{{Comparison of accuracy (\%) under the fine-tuning setting with 5\% training data  and zero-shot setting (Average of three runs). Discffusion can beat CLIP and Diffusion Classifier by a large margin consistently across all these datasets under the fine-tuning setting, demonstrating the superiority of our approach compared with traditional vision and language models pre-trained for discriminative tasks.  Note that OFA's pre-training datasets include Visual Genome (VG), RefCOCOg, and VQAv2, and BLIP2's pre-training datasets also include Visual Genome (VG) and COCO~\cite{coco} (whose images are the sources images for RefCOCOg and VQAv2). Therefore, for a fair comparison, CLIP should be the main baseline to evaluate the effectiveness of Discffusion.}}
\resizebox{\columnwidth}{!}{%
\begin{tabular}{llcccccccccc}
    \toprule
    & \multirow{2}*{Method} & \multicolumn{4}{c}{Compositional Visual Genome} & \multicolumn{2}{c}{RefCOCOg} & \multicolumn{3}{c}{VQAv2} \\
    \cmidrule(lr){3-6} \cmidrule(lr){7-8} \cmidrule(lr){9-11}
    & & {Subjects} & {Objects} & {Predicate} & {All} & {Top-1 Acc.} & {Top-5 Acc.} & {Binary} & {Other} & {All} \\
    \midrule 
   \multirow{5}{*}{{Zero-shot}}  & {CLIP } & 78.87 & 80.59 & 58.60 &74.20 & 68.10  & 82.79 & 66.31 & 4.90 & 17.55\\
    & {BLIP2*} & 80.96 & 84.07 & 64.14 &80.65 &75.28 & 89.20 & 67.73 & 6.73 & 20.49\\
    & {OFA*} & 80.79 & 82.49 & 60.89 &76.28 &72.05 & 88.52 & 66.01 & 6.37 & 19.57\\
    & {Diffusion Classifier} & 79.85 & 82.41 & 63.28 &77.05 & 73.02 & 88.60& 67.10 & 5.95 & 19.64\\
    & {Discffusion } & 80.62 & 84.74 & 63.27 & 78.11 & 73.07 & 89.16 & 67.51 & 6.82 & 20.43 \\
     \midrule
  \multirow{7}{*}{{Fine-tuned}}   & {CLIP (Fine-tuning)} & 80.77 & 82.49 & 60.50 & 76.10 & 69.88 & 84.57 & 66.94 & 5.10 & 17.86 \\
    & {CLIP (Prompt Learning)} & 78.88 & 79.51 & 60.41 & 74.24 & 69.40 & 84.48 & 67.32 & 5.16 & 18.03 \\
    & {BLIP2*} & \textbf{81.12} & 84.23 & \underline{64.30} & \underline{80.81} & \underline{76.43} & \underline{90.35} & \textbf{71.23} & \underline{7.23} & \textbf{20.99} \\
    & {OFA*} & 80.80 & 82.50 & 60.90 & 76.29 & 73.31 & 89.78 & 66.45 & 6.81 & 20.01 \\
    & {Diffusion Classifier} & 80.48 & 82.52 & 63.42 & 77.50 & 73.40 & 89.83 & 68.94 &6.78 & 19.88 \\
    & {Discffusion} & 80.78 & \underline{84.90} & 63.43 & 78.27 & 75.87 & 89.96 & 70.60 & 6.91 & 20.52 \\
    & {Discffusion (w/ pre-training)} & \underline{80.98} & \textbf{84.97} & \textbf{65.17} & \textbf{80.89} & \textbf{76.91} & \textbf{90.42} & \underline{71.04} & \textbf{7.25} & \underline{20.83} \\
    \bottomrule
  \end{tabular}
  }
  \label{coco}
\end{table*}


\subsection{Experimental Setup} 
We use Stable Diffusion v2.1-base with the xFormer~\cite{xFormers2022} and flash attention~\cite{flashattention} implementation, which utilizes the LAION~\cite{laion} dataset for pre-training. On the RefCOCOg dataset, we sample 10 text prompts from the pool each time, and the model is asked to do the correct matching given the image and the 10 sampled text prompts.  We first evaluate our method under the zero-shot setting and select the variant with the best performance (using attention maps with dynamic attention head weighting, averaged across all U-Net layers, and using the LogSumExp, see the ablation studies section for details). We then test Discffusion under the setting where we train the model with only 5\% of the dataset~\cite{yoo2021gpt3mix}, demonstrating its adaptation capability using limited data.

\subsection{Baselines}
 We mainly compare Discffusion with several baseline models, including CLIP, BLIP2, and Diffusion Classifier. For CLIP, we use the CLIP-ViT/H-14 model with the OpenCLIP variant as the backbone for the fair comparison.

\begin{itemize}
    \item CLIP (Fine-tuning)~\cite{clip}: We fine-tune the CLIP model, adjusting only the last layer.
    \item CLIP (Prompt Learning)~\cite{cocoop}: Inspired by prompt learning strategies, we incorporate learnable prompts to textual inputs conditioned on individual images.
    \item BLIP2~\cite{blip2}: A vision-language pre-training model that bridges the modality gap using a lightweight Querying Transformer.
    \item Diffusion Classifier~\cite{diffusion_classifier}: Extracts standard classifiers from class-conditional diffusion models by choosing the conditioning that best predicts the noise added to the input image.
\end{itemize}

\subsection{Experiments: Comparison under the Fine-tuned Setting}
In order to facilitate a fair comparison, we adapt the use of the Discffusion model to two distinct settings on ComVG, given the different resolutions of Stable Diffusion and CLIP. The first setting involves resizing all images in the dataset to a resolution of 224x224 first, and then upsampling and resizing to 512x512 for the use of Discffusion, so that Discffusion will not take advantage of its higher input resolution. The results are shown in Table~\ref{coco}. The other setting involves directly resizing all images to 224x224 and 512x512 base resolution, which is employed by CLIP and Stable Diffusion, with the results shown in the Appendix. 

We present a comprehensive comparison with other state-of-the-art methods: BLIP2~\cite{blip2} and OFA (base)~\cite{wang2022ofa}, specifically under the few-shot scenario using only 5\% of the data in Table~\ref{coco}. Our results show that our method is competitive, trailing BLIP2 closely, which is the state-of-the-art across numerous vision and language tasks. It's worth noting that OFA's pre-training datasets include Visual Genome (VG), RefCOCOg, and VQAv2, and BLIP2's pre-training datasets also include Visual Genome (VG) and COCO~\cite{coco} (whose images are the sources images for RefCOCOg and VQAv2). Therefore, for a fair comparison, CLIP should be the baseline to evaluate the effectiveness of our approach for few-shot learning. As shown in Table~\ref{coco}, Discffusion can beat CLIP consistently across all these datasets. {In addition, with pre-training on the corresponding datasets, Discffusion outperforms all other methods, including the state-of-the-art BLIP2 on Compositional Visual Genome, RefCOCOg, and the `Other' category on VQAv2.}


\subsection{Experiments: Zero-shot Transfer from Training on the MS-COCO Dataset}
We assess the zero-shot transfer capabilities of the Discffusion model and its comparison with notable baselines such as CLIP, BLIP2, Diffusion Classifier~\cite{diffusion_classifier}. {We fine-tuned all these models on the MS-COCO dataset, and then performed the evaluation on downstream datasets}. Given that BLIP2's pre-training datasets include Visual Genome (VG) and COCO~\cite{coco}, and considering the overlap of MS-COCO images with those in ComVG and RefCOCOg, we selected two commonly used evaluation benchmarks: Winoground~\cite{thrush2022winoground} and VL-checklist~\cite{vlchecklist}. 
The evaluation results are depicted in Table~\ref{tab:winoground_vlchecklist}. Discffusion outperforms all others on Winoground and VL-checklist, showcasing its superior ability to generalize the learned visual-linguistic representations to datasets beyond the ones used during fine-tuning.

\subsection{{Experiments: Zero-shot Evaluation on the Winoground and VL-checklist Dataset}}
{We conducted additional experiments to evaluate the potential of Discffusion for zero-shot inference with 50 sampling steps on the Winoground~\cite{thrush2022winoground} and VL-checklist datasets, which require compositional understanding ability. The top of Table~\ref{tab:winoground_vlchecklist} shows a comparison of Discffusion with other zero-shot approaches. Discffusion outperforms CLIP and is competitive with Diffusion Classifier under the zero-shot setting on both the two datasets, even though Discffusion is primarily designed as a fine-tuned method.} The detailed performance across categories \textless relation, object, both\textgreater\ in the format (text score, image score, group score) is shown in Table~\ref{tab:winoground} in the Appendix. As shown, Discffusion excels notably in the `relation' category, demonstrating proficiency in discerning shuffled elements like verbs, adjectives, prepositions, and adverbs. This suggests that Discffusion, with its generative-task-oriented pre-training and our adaptation strategy, offers enhanced capability for handling fine-grained nuances, spatial interrelationships, and intricate compositionalities compared to conventional discriminative vision and language models.

\begin{table}[t]
  \caption{{Comparison of accuracy (\%) on Winoground and VL-checklist across three runs for CLIP, BLIP2, and Discffusion, both zero-shot and fine-tuned on MS-COCO. For BLIP2's evaluation on Winoground, it is approached as an image-text matching task to ensure a fair comparison, rather than prompting the model to answer the rewritten question from captions~\cite{BLIP_winoground}. 
 BLIP2 gains little from fine-tuning on MS-COCO because BLIP2's pre-training datasets already include MS-COCO~\cite{coco}, and Winoground is quite different from MS-COCO as it requires compositional understanding ability. Similarly, Diffusion Classifier, designed as a zero-shot approach, does not benefit much from fine-tuning. As can be seen, under the fine-tuned setting, Discffusion outperforms all other approaches by a large margin across both datasets in each category. The detailed results on Winoground with 95\% confidence intervals across different categories are presented in Table~\ref{tab:winoground}.}}
  \centering
  \resizebox{0.8\columnwidth}{!}{
    \begin{tabular}{lllllllll}
      \toprule
     & \multirow{2}*{Method}   &\multicolumn{3}{c}{Winoground} & \multicolumn{4}{c}{VL-checklist} \\
      \cmidrule(lr){3-5} \cmidrule(lr){6-9}
     & & {Text} & {Image} & {Group} & {Attribute} & {Object} & {Relation} & {Average} \\
      \midrule
    \multirow{4}{*}{{Zero-shot}} & CLIP & 30.68 & 11.91 & 8.36 & 67.82 & 75.67 & 67.11 & 70.18 \\
     & BLIP2 & 31.78 & 12.37 & 9.02 & \textbf{79.00} & \textbf{84.05} & \textbf{73.55} & \textbf{78.86} \\
     & Diffusion Classifier & \textbf{37.75} & 12.25 & 9.00 & 64.19 & 75.98 & 66.99 & 68.23 \\
     & Discffusion (Ours) & 34.11 & \textbf{13.38} & \textbf{11.04} &73.12 &79.58 &69.48&74.06\\
     \midrule
      & CLIP & 31.50 & 12.00 & 9.25 & 67.90 & 75.75 & 68.15 & 70.73 \\
    {MS-COCO}  & BLIP2 & 28.25 & 13.00 & 9.09 & 79.64 & 84.70 & 73.95 & 79.97 \\
   {Fine-tuned} &  Diffusion Classifier & 26.76 & 10.00 & 8.75 & 65.90 & 76.81 & 67.14 & 68.94 \\
     & Discffusion (Ours) & \textbf{35.75} & \textbf{14.50} & \textbf{12.75} &\textbf{79.80} &\textbf{85.90} &\textbf{74.58}&\textbf{80.52}\\
      \bottomrule
    \end{tabular}
  }
  \label{tab:winoground_vlchecklist}
\end{table}

\begin{table}[t]
\centering
  \caption{Ablation study on the ComVG dataset across three runs using cosine similarity, maximum value from each column of the attention map, and the smoothed maximum (LogSumExp pooling); and the amount of noise added during the diffusion process: using consistent noise levels of $0.4$, $0.8$, and using ensembling. The accuracy numbers are in (\%).}
\begin{tabular}{lccccc}
    \toprule
    & Method/Noise Level & {Subjects} & {Objects} & {Predicate} & {All}\\
    \midrule
    \multirow{4}{*}{Method} & Cosine similarity &50.7 & 42.5& 30.5&43.0 \\
    & Maximum (column-wise) & 58.8&59.5 &40.4 & 57.7 \\
    & Maximum (row-wise) & 57.8&59.0 &40.1 & 57.0 \\
    & LogSumExp & 78.9& 80.1& 61.2& 75.4 \\
    \midrule
    \multirow{3}{*}{Noise Level} & Noise level 0.4 & 75.5& 77.5& 58.4& 72.5 \\
    & Noise level 0.8 & 75.7& 77.7& 58.5& 72.7 \\
    & Ensembling & 76.1& 78.0& 59.2& 73.1 \\
    \bottomrule
  \end{tabular}
  \label{merged_ablation_study}
\end{table}

\section{Related Work}
\paragraph{Diffusion Probabilistic Models (DPMs)}
Diffusion probabilistic models (DPMs) have been widely used as generative models for images in recent years. These models, which include diffusion~\cite{sohl2015thermodynamics} and score-based generative models~\cite{song2019generative}, have been shown to outperform generative adversarial networks (GANs)~\cite{goodfellow2014generative} in many cases~\cite{label_efficient_semantic_segmentation}. In the past two years, significant progress has been made in the development of DPMs, with a focus on improving sampling techniques such as classifier-free guidance~\cite{ho2021classifier}. DPMs are typically implemented using convolutional U-Net architectures~\cite{ronneberger2015u} which contain cross-attention layers. ~\cite{prompt_to_prompt} finds that replacing attention maps in the cross-attention module of text-to-image generation diffusion models can edit image attributes. Just scaling the attention maps of the respective word can adjust the effect of a particular word in the prompt. ~\cite{structured_diffusion} demonstrates that one can retain the compositional semantics in the generated image by manipulating the cross-attention. ~\cite{multi_diffusion} proposes to fine-tune the key and value mapping from text to latent features in the cross-attention layers of text-to-image diffusion model to compose multiple new concepts in the image. In the context of image-text matching, the attention scores between the text and image representations in the DPMs can reflect the degree of alignment between them.

\paragraph{Few-shot Learning for Vision and Language Tasks}
Vision and Language discriminative models pre-trained on large-scale image-text pairs have demonstrated great potential in multimodal representation  learning~\cite{align,flip,florence,clip,llava}. Among them, CLIP~\cite{clip} benefits from 400M curated data and defines various prompt templates to carry out zero-shot image classification. Like CLIP, several different few-shot learners were proposed. GPT~\cite{gpt3}, as a strong few-shot learner, is capable of performing a new language task by learning from only a few training instances.  Frozen~\cite{frozen} is developed based on GPT and made into a multimodal few-shot learner by expanding the soft prompting to include a collection of images and text. The concept of prompt learning~\cite{first_prompt} has been widely explored in natural language processing (NLP) and computer vision. It allows pre-trained models to adapt to various downstream tasks with minimal number of data by introducing a small prompt layer~\cite{first_prompt, ptuningv2}. 
In the context of image-text matching, prompt learning has been used to fine-tune pre-trained models for the task~\cite{he2022cpl}. In our work, instead of adding learnable prompts over the inputs or between transformer layers~\cite{visualprompttuning}, we introduce learnable prompts over the attention layers. In our paper, our primary research question is the adaptation of pre-trained generative diffusion models into discriminative models for specific tasks. This focus is driven by the challenges and opportunities presented by utilizing diffusion-based processes in a discriminative setting, specifically for the image-text matching task, which has distinct characteristics compared to the modeling approaches mentioned above.

\paragraph{Generative Models for Discriminative Tasks}
There has been a significant amount of research on using generative models for discriminative tasks in the past decades. ~\cite{discriminative_classifier} compare the discriminative classifier with generative classifier. ~\cite{generative_recognition} apply deep generative models to the recognition task. 
For diffusion models, recently, ~\cite{diffusion_classifier,diffusion_classifier2} propose to use pre-trained diffusion models for zero-shot classification. ~\cite{diffusion_mae} formulate diffusion models as masked autoencoders and achieves state-of-the-art classification accuracy on video tasks.  Different from these works, we are the first to explore the use of cross-attention maps in pre-trained diffusion models for discriminative tasks, specifically the image-text matching task. Another line of works use diffusion models as data source and then training a discriminative model on the synthetic data generated from it~\cite{synthetic_data_for_generative1,synthetic_data_for_generative2,synthetic_data_for_generative3}. Differs from these works, our approach emphasizes the direct adaptation of generative diffusion models, leveraging their pre-existing structures and knowledge without the need to generate synthetic data.

\chapter{Structured Multimodal Reasoning}
\label{char:mgtmqa}

\section{Introduction}
\begin{figure}[t]
    \centering
    \includegraphics[width=0.8\columnwidth]{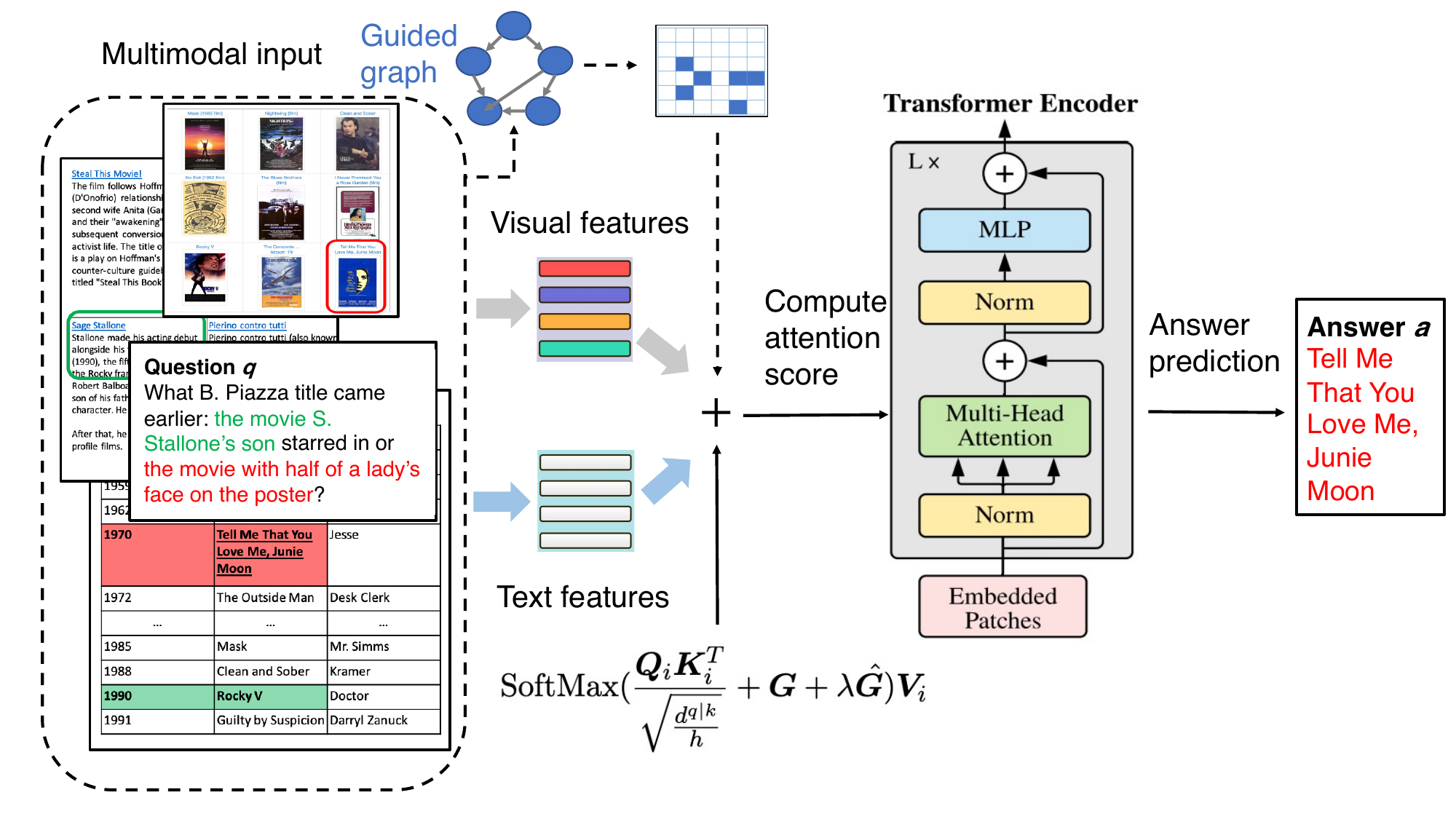}
    \caption{Overview of the Multimodal Graph Transformer. It takes visual features, text features, and their corresponding generated graphs as inputs. The generated graph is first converted to an adjacency matrix to induce the mask matrix $\bm{G}$. The modified quasi-attention score in the Transformer is computed to infer the answer. In the formula, $\bm{G}$ is the graph-induced matrix constructed by concatenating adjacency matrices from both the vision and language ends. $\hat{\bm{G}}$ is the trainable bias. The input features from different modalities are fused along with graph information to perform downstream reasoning.}
    \label{fig:teaser_mqa}
\end{figure}

A myriad of complex real-world tasks require both prior knowledge and reasoning
intelligence~\cite{reason1,reason2}. These days, vision-and-language reasoning tasks such as as vision question answering (VQA)~\cite{vqa} and multimodal question answering (MultiModalQA)~\cite{multimodalqa} post further needs for integrating structured info from different input modalities and thus perform reasoning. Towards this, two questions yield: What is the best way to integrate prior knowledge and reasoning components from multiple modalities in a single model? How would such an integration lead to
accurate models, while being more computationally efficient and allowing for significantly more interpretability? Such questions are important
to address when scaling reasoning systems to real-world use cases.

These years, there are a spectrum of methods in the literature exploring different ways of integrating structured prior information. Graph neural networks (GNNs)~\cite{graph_neural_networks}, 
have been widely used in representation learning on graphs.
Some experts tried to investigate the embedding of the structured information by resorting to them. However, GNNs are inefficient~\cite{graph_neural_networks} and they can barely compete with Transformer models. Besides, most GNNs are designed to learn node
representations on fixed and homogeneous graphs. 
Thereby, it is suboptimal to operate GNNs on vision-and-language tasks such as visual question answering (VQA), where graphs encountered in these problems (e.g. scene graphs) can be more complex;
Alternatively, knowledge graphs (KGs), such as Freebase~\cite{knowledge_graph}, represent world-level factoid information of entities and their relations in
a graph-based format, surfaced these years. They have been successfully used
in vision and language applications including VQA~\cite{okvqa}.  However, they have not been dedicated to be applied to our scenario, more concretely, we aim at filling the gap of capturing prior knowledge in Transformer models.

To mitigate deficiencies of the existing methods, this paper proposes a novel plug-and-play graph-involved Transformer-based method for multimodal question answering tasks. Our method is {\em Multimodal Graph Transformer} in the sense that it is built upon the well-established Transformer backbone, albeit with several key fundamental differences. 
First, we introduce a systematic scheme to convert text graphs, dense region graphs, and semantic graphs from vision and language tasks to adjacency matrices to use in our method. 
Second, instead of directly computing the attention score, we learn the newly proposed quasi-attention score with graph-induced adjacency matrices live at its heart, to signify the importance of learning relative importance as a highly effective inductive bias for computing the quasi-attention score.  
Third, different from previous Transformer methods, where self-attention
are fully learned from data, we switch gears to introduce the graph-structured information in the self-attention computation to guide the training of Transformers as shown in Figure~\ref{fig:teaser_mqa}.

\begin{figure}[htbp]
	\begin{center}
 	\includegraphics[width = \columnwidth]{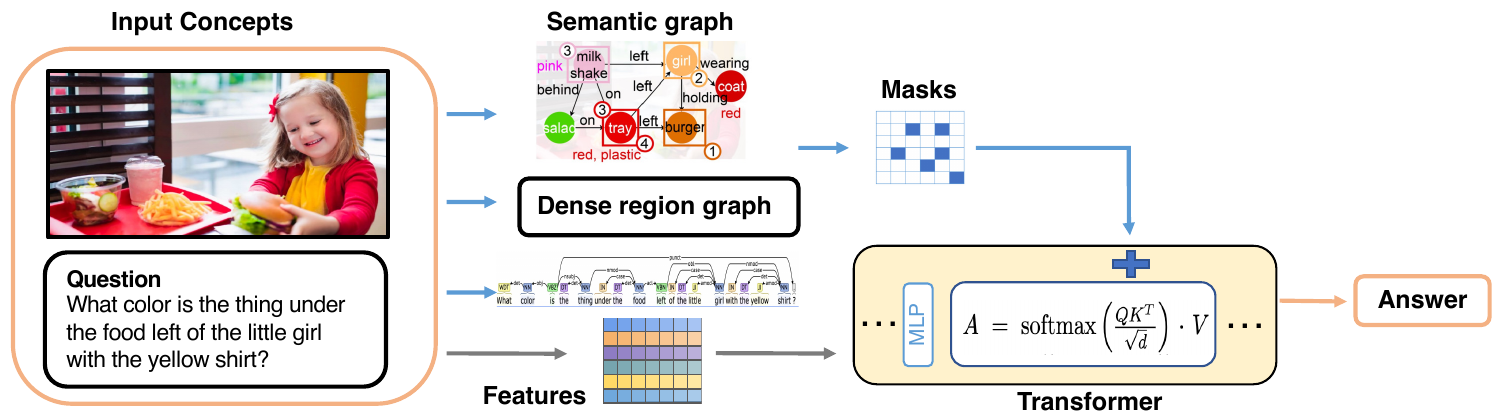}
 	\caption{The figure illustrates the overall framework of our Multimodal Graph Transformer. The input from different modalities are processed and transformed into corresponding graphs, which are then converted into masks and combined with their features to be fed into Transformers for downstream reasoning. In detail, semantic graphs are created through scene graph generation methods, dense region graphs are extracted as densely connected graphs, and text graphs are generated through parsing. 
	}\label{fig:overview_mmqa}
	\end{center}
 \end{figure}

\section{Integrating Structural Graph Representations into Multimodal Reasoning}

\subsection{Background on Transformers}
The Transformer layer~\cite{vaswani2017attention} consists of two modules: a multi-head attention and a feed-forward network (FFN). Specifically, each head is represented by four main matrices: the query matrix $\boldsymbol{W}_{i}^{q} \in$ $\mathbb{R}^{d^{m} \times d^{q} / h}$, the key matrix $\boldsymbol{W}_{i}^{k} \in \mathbb{R}^{d^{m} \times \frac{d^{k}}{h}}$, the value matrix $\boldsymbol{W}_{i}^{v} \in \mathbb{R}^{d^{m} \times \frac{d^{v}}{h}}$, and the output matrix $\boldsymbol{W}_{i}^{o} \in \mathbb{R}^{\frac{d^{v}}{h} \times d^{o}}$, and takes the hidden states $\boldsymbol{H} \in \mathbb{R}^{l \times d^{m}}$ of the previous layer as input, where $d$ denotes the dimension of the model, $h$ represents the number of head, and $i$ denotes the index of layer number. The output of attention is given by:\\
\begin{equation}
\boldsymbol{Q}_{i}, \boldsymbol{K}_{i}, \boldsymbol{V}_{i}=\boldsymbol{H} \boldsymbol{W}_{i}^{q}, \boldsymbol{H} \boldsymbol{W}_{i}^{k}, \boldsymbol{H} \boldsymbol{W}_{i}^{v}
\end{equation}
\begin{equation}
\operatorname{Attention}\left(\boldsymbol{Q}_{i}, \boldsymbol{K}_{i}, \boldsymbol{V}_{i}\right)=\operatorname{SoftMax}\left(\frac{\boldsymbol{Q}_{i} \boldsymbol{K}_{i}^{T}}{\sqrt{\frac{d^{q, k}}{h}}}\right) \boldsymbol{V}_{i} \; 
\end{equation}
\begin{equation}\boldsymbol{H}_{i}=\operatorname{Attention}\left(\boldsymbol{Q}_{i}, \boldsymbol{K}_{i}, \boldsymbol{V}_{i}\right) \boldsymbol{W}_{i}^{o}
\end{equation}
where $\boldsymbol{Q}_{i} \in \mathbb{R}^{l \times \frac{d^{q}}{h}}, \boldsymbol{K}_{i} \in \mathbb{R}^{l \times \frac{d^{k}}{h}}, \boldsymbol{V}_{i} \in$ $\mathbb{R}^{l \times \frac{d^{v}}{h}}$ are obtained by the linear transformations of $\boldsymbol{W}_{i}^{q}, \boldsymbol{W}_{i}^{k}, \boldsymbol{W}_{i}^{v}$ respectively. $\operatorname{Attention}(\cdot)$ is the scaled dot-product attention operation. Then output of each head is transformed to $\boldsymbol{H}_{i} \in \mathbb{R}^{l \times d^{o}}$ by $\boldsymbol{W}_{i}^{o} .$

\subsection{Framework overview} 
The entire framework of the proposed Multimodal Graph Transformer method is depicted in Figure~\ref{fig:overview_mmqa}.  Without loss of generality, we assume the end task is VQA in the following discussion while noting that our framework can be applied to other vision-language tasks, such as multimodal question answering.

Given the input images and questions, the framework first constructs three graphs, including the semantic graph, dense region graph, and text graph, which will be described in more detail in the following sections. The graph $G=(\mathcal{V}, \mathcal{E})$, where $\mathcal{V}$ represents the set of nodes in the graph and $\mathcal{E}$ represents the edges connecting them, is fed into Transformers to guide the training process.

\subsection{Multimodal graph construction}
We build three types of graphs and feed them into Transformers: \emph{text graph}, \emph{semantic graph}, and \emph{dense region graph}. We now introduce them in detail.
\paragraph{Text graph} 
\begin{figure}[tbbp]
	\begin{center}
 	\includegraphics[width = 0.8\columnwidth]{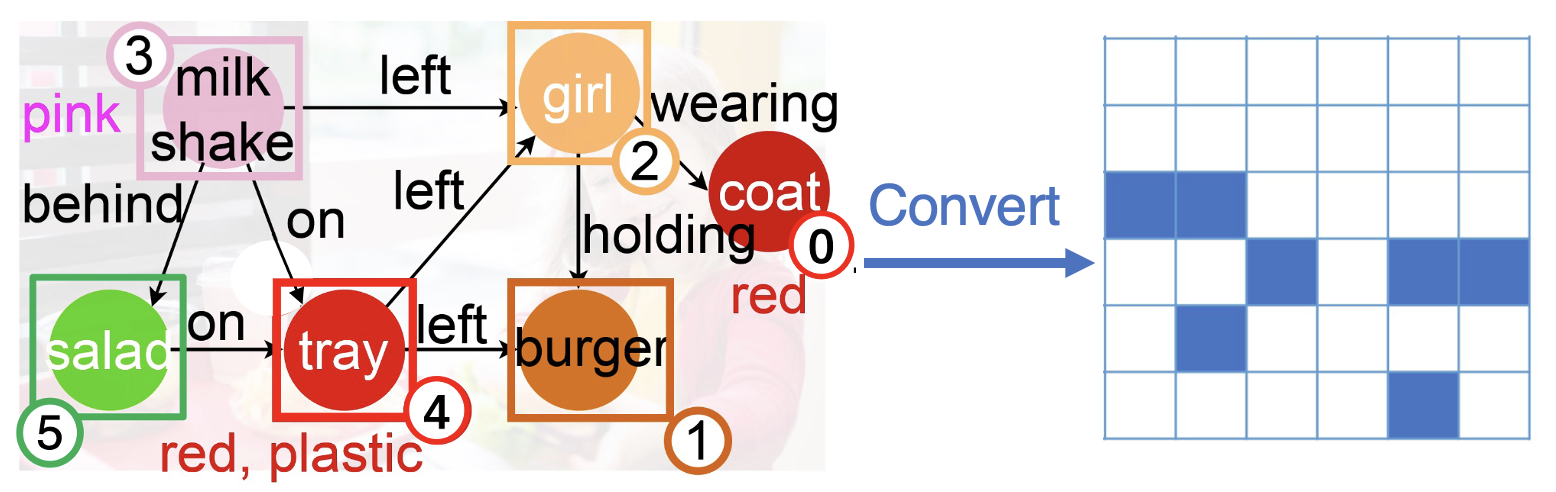}
 	\caption{ The naive demonstration of converting a semantic graph into an adjacency matrix. Cells in blue means `0's for that element in the graph matrix, while white ones means `-inf's. We employ the matrix as the mask when computing the quasi-attention. 
	}\label{fig:convert}
	\end{center}
 \end{figure}
The task of Visual Question Answering involves a combination of an image, a question, and its corresponding answer. To process the question, we extract the entities and create a text graph representation. We then build the graph $G=(\mathcal{V}, \mathcal{E})$ as shown in the left of Figure~\ref{fig:overview_mmqa}. The set of nodes, $\mathcal{V}$, represents the entities and the set of edges, $\mathcal{E}$, represents the relationships between the pairs of entities. This results in:

\begin{itemize}
\item A set of $N$ entities, each represented by a vector of token embeddings, that constitute the nodes of the graph.
\item A set of pairwise relations between entities, forming the edges of the text graph. The relationship between entities $i$ and $j$ is represented by a vector $e_{i j}$ which encodes the relative relationships.
\end{itemize}

\paragraph{Semantic graph}
In tasks such as multimodal question answering, there might be additional inputs in the form of tables or lengthy paragraph sentences. To handle these inputs, a linear representation of the table can be created and a semantic graph can be constructed using a similar approach. They are processed using the scene graph parser~\cite{zhong2021learning}, which transforms the text sentence into a graph of entities and relations, as depicted in Figure~\ref{fig:convert}. The output of the scene graph parser includes:

\begin{itemize}
\item A set of $N$ words that constitute the nodes of the semantic graph, where $N$ is the number of parsed words in the texts.
\item A set of possible pairwise relations between words, such as "left" and "on" as shown in Figure~\ref{fig:convert}, which constitute the edges of our graph. An edge between words connecting $j$ to $i$ is represented by $e_{i j}$, namely, the connectivity is indicated as: $e_{i j}= \begin{cases}0, & i, j \text{ \ not connected} \\ 1, & i, j \text{ \ connected} \end{cases}$. 
\end{itemize}

\paragraph{Dense region graph}
The visual features are extracted by slicing the input images into patches and flattening them. A dense region graph $G=(\mathcal{V}, \mathcal{E})$ is then converted into masks, with $\mathcal{V}$ being the set of extracted visual features and $\mathcal{E}$ being the set of edges connecting each feature node, following the method described in~\cite{vilt}. This results in a graph that is nearly fully connected.

The resulting three graphs are then transformed into adjacency matrices, where the elements are either -$\infty$ or zero. The conversion process is depicted in Figure~\ref{fig:convert} using the semantic graph as an example. These adjacency matrices are used inside the scaled dot-product attention to control the flow of information, by masking out (setting to $-\infty$) the values.

\subsection{Graph-involved quasi-attention}
\begin{figure}[tbp]
	\begin{center}
 	\includegraphics[width = 0.9\columnwidth]{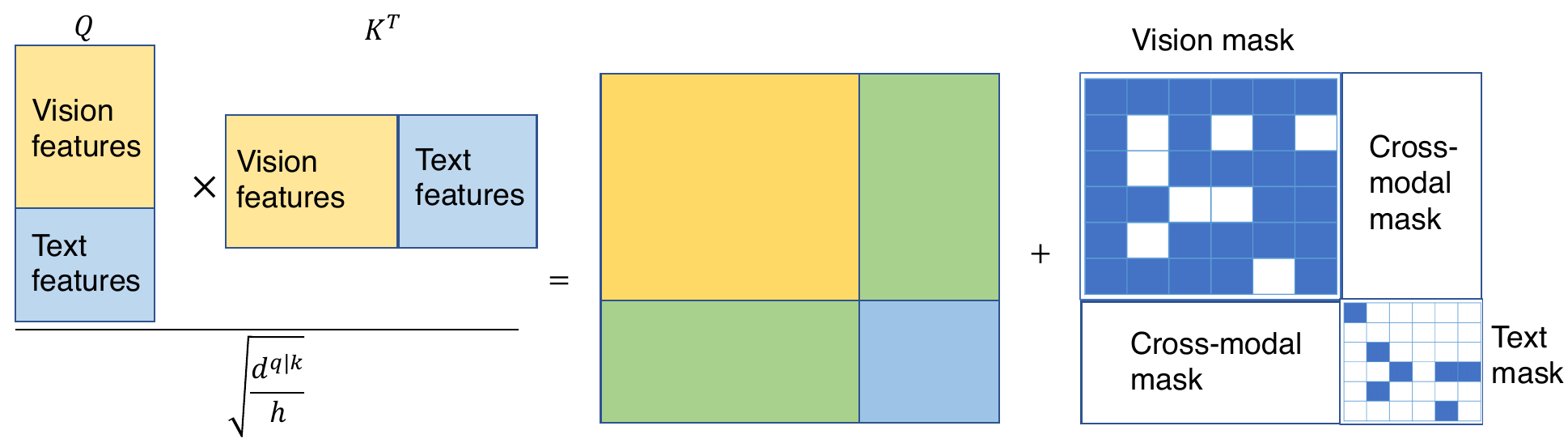}
 	\caption{ A naive demonstration of adding the graph-induced mask while computing the quasi-attention when the inputs are from two modalities. The visual mask is the mask converted from the dense region graph and the text mask is converted from the text graph. The cross-modal mask, which is always set as an all-zero matrix, is imposed to encourage the model to learn the cross-attention between the image features and text features, thus facilitating the alignment across them.
	}\label{fig:graph_mask}
	\end{center}
 \end{figure}

In order to effectively utilize structured graph knowledge in our self-attention computation, we incorporate the graph as an extra constraint in each attention head by converting it into an adjacency matrix. The graph matrix, denoted as $\bm{G}$, is constructed by combining various masks. An illustration of this process can be seen in Figure~\ref{fig:graph_mask}. The visual mask is generated from the dense region graph, while the text mask is derived from the text graph. Additionally, the cross-modal mask is set to an all-zero matrix to encourage the model to learn the cross-attention between visual and text features, thereby promoting alignment across the different modalities.

Within the context of adding graph information,
when vision graph mask and text graph mask are concatenated and aligned with image and text features, we believe that a more flexible masking-out mechanism is beneficial, rather than keeping a single constant mask matrix inside the Softmax operation. Drawing insights from ~\cite{liu2021swin}, where they include a relative position bias to each head in computing similarity, we also intuitively parameterize a trainable bias $\hat{\bm{G}}$ and involve it in the training process. Finally, we compute the quasi-attention as follows:
\begin{equation}
\begin{aligned}
\operatorname{Attention}=\operatorname{SoftMax}(\frac{\boldsymbol{Q}_{i}\boldsymbol{K}_{i}^{T}}{\sqrt{\frac{d^{q \mid k}}{h}}}+\bm{G}+\lambda \hat{\bm{G})  } \boldsymbol{V}_{i} ,
\end{aligned}
\label{eq_additive_term}
\end{equation}

 where $\lambda$ is the tradeoff hyper-parameter that controls the contribution of
$\bm{\hat{G}}$, and $\bm{G}$ is our graph-induced matrix constructed by concatenating a graph matrix both from the vision and the language end.
Here for clear clarification, we use $\bm{G}$ and $\hat{\bm{G}}$ to
distinguish the graph matrices fixed and trainable, respectively. During training, $\bm{G}$ is frozen as before and does not receive gradient updates, while $\hat{\bm{G}}$ contains trainable parameters.  

We now introduce the motivation behind adding two types of graph matrices. We perform the masking process by adding $\bm{G}$ when computing the quasi-attention because 
it can be interpreted as a form of attentional pooling (learning to align), in
which each element of $\bm{G}$ pools all relevant information across all elements of the relative importance matrix computed by $\left(\frac{\boldsymbol{Q}_{i} \boldsymbol{K}_{i}^{T}}{\sqrt{\frac{d^{q \mid k}}{h}}}\right)$. Hence during fine-tuning, the model ignores redundant features and only focuses on useful information. The mask can also force the model to learn the cross attention between features from the images and questions and perform aligning across them. 
And the trainable bias $\hat{\bm{G}}$ captures information gained during the training process. Such information is valuable for fine-tuning, making the Transformer more robust and helping it gain numerical stability.

\subsection{Training}
The interdependence of output features from various modalities calls for a unified optimization approach for the Transformers in both the visual question answering and multimodal question answering tasks. To accomplish this, we implement a kind of end-to-end training, which ensures the optimality of the models. The final outcome of our models is a classification logit, which is generated by the VQA models that select the best answer from the available candidate answers. To evaluate the accuracy of the models, we compute the cross-entropy loss~\cite{cross_entropy} using the output logits produced by the Transformer. This measure helps us determine the difference between the predicted class probabilities and the actual class labels.

\begin{table*}[ht]
 \caption{Accuracy (\%) comparison of different methods on the VQA task. Ours has the second best performance and is comparable to state-of-the-art methods. After applying our proposed quasi-attention mechanism and exploiting the use of graphs, there is also a 2\% improvement of overall accuracy on the LXMERT baseline, suggesting the generalization ability of our method.}
  \centering
  \resizebox{\textwidth}{!}{
  \begin{tabular}{ccccc}
    \toprule
     Dataset &Method & Open questions & Binary questions & Overall accuracy\\
    \midrule
   \multirow{5}*{GQA}& LXMERT~\cite{lxmert} & -&-&60.0  \\   & LXMERT w/ Graph~\cite{lxmert} & -&-&61.4  \\
   & HANs~\cite{kim2020hypergraph} & -&-& 69.4 \\
   & NSM~\cite{hudson2019learning} & 49.3& 78.9& 63.2  \\
   & OSCAR~\cite{li2020oscar} & -&-&61.6 \\
   & VinVL~\cite{zhang2021vinvl} & - &-& 65.1 \\
   & Multimodal Graph Transformer (Ours) &  59.4& 80.5&68.7 \\
    \midrule
\multirow{6}*{VQA v2}& LXMERT~\cite{lxmert} & -&-&72.4  \\
& HANs~\cite{kim2020hypergraph} & -&-&65.1  \\
& NSM~\cite{hudson2019learning} & -&-&63.0  \\
& OSCAR~\cite{li2020oscar} & -&-&73.8 \\
& VinVL~\cite{zhang2021vinvl} & - &-&  76.6 \\
    &Multimodal Graph Transformer (Ours)&66.5&87.0&74.5 \\
    \bottomrule
  \end{tabular}}
 \label{tab:real_results}
\end{table*}

\section{Experiments}
\subsection{Datasets}

\paragraph{VQA v2}
The VQA v2 dataset~\cite{goyal} extends the VQA~\cite{vqa} dataset to better balance visual and textual information through the collection of complementary images. Each question in VQA v2 is associated with a pair of similar images with different answers, resulting in a total of 1.1 million QA pairs and 204,000 images.
The data split for VQA v2 includes a training set with 83,000 images and 444,000 questions, a validation set with 41,000 images and 214,000 questions, and a test set with 81,000 images and 448,000 questions.
The annotated answers are in natural language, but they are commonly converted to a classification task with 3,129 answer classes. As described by~\cite{bottom}, the model selects the answer to each question from a set of 3,129 most frequent answers. Following this convention, we fine-tune the multimodal graph transformer model on the VQAv2 training and validation sets, while reserving 1,000 validation images and related questions for internal validation.

\paragraph{GQA}
The GQA dataset contains 22M questions over 113K images. The questions in GQA are designed to require multi-hop reasoning to test the reasoning skills of VQA models. GQA greatly increases the complexity of the semantic
structure of questions, leading to a more diverse function
set. The real-world images in GQA also bring in a bigger
challenge in visual understanding. We treat the task as the classification task reffering to the VQA v2 setting.

\paragraph{MultiModalQA}
MultiModalQA (MMQA) contains 29, 918 questions. We split the dataset with reference to the public split. Around 60\% of the questions in MMQA are compositional. The answer for each question can be a single answer or a list of answers.

\subsection{Baselines}
We compare with
four state-of-the-art VQA models:
LXMERT~\cite{lxmert}, NSM~\cite{hudson2019learning}, OSCAR~\cite{li2020oscar}, and VinVL~\cite{zhang2021vinvl}.  

\begin{itemize}
\item LXMERT~\cite{lxmert} designs five pretraining tasks: masked language
modeling, feature regression, label classification,
cross-modal matching, and image question answering to
pretrain a large Transformer model. Towards this, a large-scale Transformer~\cite{vaswani2017attention} 
model is built that consists of three encoders: an object
relationship encoder, a language encoder,
and a cross-modal encoder. 
\item NSM~\cite{hudson2019learning} predicts a probabilistic graph
that represents its underlying semantics and performs sequential reasoning over the graph to traversing its
nodes to make the inference. 
\item OSCAR~\cite{li2020oscar} uses object tags detected in images as anchor points to significantly ease
the learning of alignments, improving previous methods and using self-attention to learn image-text semantic alignments.
\item VinVL~\cite{zhang2021vinvl} developed a new
object detection model to create better visual features of images than previous classical object detection models. 
\end{itemize}

We compare with four baselines introduced in the MultiModalQA paper~\cite{multimodalqa}: Question-only~\cite{questiononly}, Context-only~\cite{questiononly}, AutoRouting, ImplicitDecomp. 
\begin{itemize}
\item Question-only is a sequence-to-sequence
model that directly generates the answer given the question.
\item Context-only first predicts the question type using the classifier and then feed in the relevant context to predict the answer.
\item AutoRouting first determines the modality where the answer is expected to occur, and then runs the corresponding single-modality module. 
\item ImplicitDecomp is a 2-hop implicit decomposition
baseline and so far the state-of-the-art method on the MultiModalQA dataset.
\end{itemize}

\subsection{Implementation details}
The input texts undergo preprocessing using a scene graph parser which extracts entities and their relationships. The text features are obtained through a pre-trained BERT tokenizer, allowing us to extract text spans of individual entities and text spans containing two related entities. As for images, we employ the methods described in~\cite{vit, vilt} to extract visual features and create graph masks. This involves resizing the shorter edge of the input images while preserving the aspect ratio and limiting the longer edge, followed by patch projection and padding for batch training. The resulting patch embeddings are used as inputs along with constructed dense region graph that is densely connected. The Transformer backbone used in this setting is the pretrained VIT-B-32~\cite{vit} version, consisting of 12 layers with a hidden size of $H$ = 768, layer depth of $D$ = 12, patch size of $P$ = 32, a multi-layer perceptron size of 3072, and 12 attention heads. To test this setting, all inputs and graphs are merged and processed by the Transformer backbone, which learns from features from different modalities.

\begin{table}[ht]
\small
 \caption{EM (\%) and F1 (\%) of Multimodal Graph Transformer and its Transformer baseline on questions in MultiModalQA that require reasoning over multiple modalities. We also quote the results from the MultiModalQA~\cite{multimodalqa} paper. Incorporating graph information into the Multimodal Graph Transformer can boost about 2\% F1 and 4\% EM performance.} 
  \centering
  \scalebox{1}{
  \begin{tabular}{ccc}
    \toprule
   Method& EM  & F1
    \\
    \midrule
    Question-only& 16.9&19.5\\
    Context-only&6.6&8.5\\
    \midrule
    AutoRouting& 32.0&38.2\\
    ImplicitDecomp&46.5&51.7\\
     \midrule
     Human&84.8&90.1\\
    \midrule
     \midrule
     Multimodal Transformer w/o Graph & 50.1&56.4\\
     Multimodal Graph Transformer (Ours) &52.1&57.7\\
     
    \bottomrule
  \end{tabular}}
  \vskip 0.1in
 \label{tab:real_results2}
\end{table}

\begin{table*}[ht]
\small
 \caption{Ablation Studies on the GQA and VQA v2 datasets. The figure demonstrates the effectiveness of incorporating graph information into the Transformer architecture through ablation studies performed on the GQA and VQA. The results of these studies clearly indicate that including graph information can lead to an improvement in performance.}
  \centering
  \scalebox{0.8}{
  \begin{tabular}{cccccc}
  \toprule
  Dataset  & Method & Open questions & Binary questions & Overall accuracy\\
    \midrule
 \multirow{3}*{GQA} 
   &One-modality Transformer & 47.7 & 78.1&62.7 \\ & Multimodal Transformer w/o Graph &49.9 & 81.0& 65.4 \\
   & Ours & \textbf{60.1} & \textbf{90.2}& \textbf{72.4} \\
   \midrule 
 \multirow{3}*{VQA v2} & One-modality Transformer w/ one Transformer &60.5& 85.4&70.1 \\
   & Multimodal Transformer w/o Graph & 64.8&86.3&72.1  \\
    &Ours  &\textbf{66.7}& \textbf{87.2}&\textbf{74.6} \\
    \bottomrule
  \end{tabular}
  }
 \label{tab:real_results1}
\end{table*}
\subsubsection{MultiModalQA}
We further investigate the effectiveness of our proposed method on MultiModalQA~\cite{multimodalqa}, a recently introduced and demanding task that requires joint reasoning across various modalities such as texts, images, tables, etc. We employ a Multimodal Graph Transformer to tackle the task, using the same approach for extracting vision and text features as in VQA. Additional modalities, such as tables, are encoded by linearizing them and utilizing pre-trained models like RoBERTa-large~\cite{liu2019roberta}. After generating text graphs, semantic graphs, and dense region graphs from input questions, text, tables, and images, we feed them along with the extracted features into the Transformer.

\subsection{Results and analysis}
Table~\ref{tab:real_results} presents a comparison of the accuracy of our proposed method on the GQA dataset with previous state-of-the-art methods. Our proposed method ranks second in terms of accuracy and outperforms the third best method by a substantial margin, with an absolute improvement of over 3\% in overall accuracy. The performance of our method is comparable to the state-of-the-art method.

We also conducted experiments on the VQA v2 dataset, and the results are summarized in Table~\ref{tab:real_results} and Table~\ref{tab:real_results1}. As shown, there are significant improvements over methods without graphs, suggesting that incorporating graph information into the Transformer is effective.

Additionally, after incorporating our proposed graph method into LXMERT, we can observe a boost in overall accuracy on the GQA dataset, demonstrating the generalization ability of the proposed method in incorporating graph information into quasi-attention computation.

Table~\ref{tab:real_results2} compares the Exact Match (EM) and average F1 score of our proposed method on the MultiModalQA dataset with the baseline. The results show that our proposed method outperforms the baseline without the aid of graph information, demonstrating the generalization of our method to more complicated vision-and-language reasoning tasks.

\subsection{Ablation studies}
We perform ablation studies to verify the necessity of using two-stream inputs with the help of graphs to deal with input from different modalities, with
GQA dataset as our testing bed. For all experiments, we use the overall accuracy as the evaluation metric.

The results presented in Table~\ref{tab:real_results1} show the superiority of our proposed Multimodal Graph Transformer over the method where a single modality input is fed into a Transformer. Our method, which involves dividing the input streams into two separate parts and processing each part through a Transformer, outperforms the Multimodal Transformer without Graph. This demonstrates the beneficial effect of incorporating graph information into the processing of the input data and performing training. The use of different input features with the help of graphs allows for a better alignment of the information from different modalities, which is reflected in the improved performance of our proposed method.

\subsection{Qualitative results}
\begin{figure}[htbp]
	\begin{center}
 	\includegraphics[width = \columnwidth]{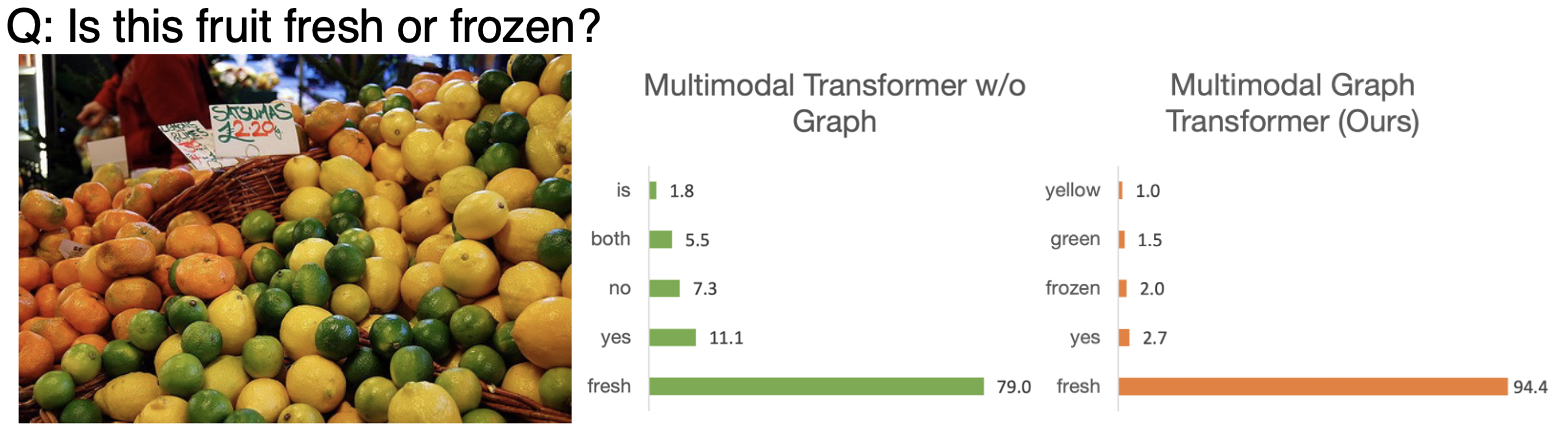}
 	\caption{A qualitative comparison from VQA v2. {\em fresh} is the ground truth. Predictions from the Multimodal Graph Transformer (ours) are more relevant to the contents of the input image and achieve a higher confidence score over the ground truth.}
	\label{fig:Qualitative_Results}
	\end{center}
 \end{figure}

One qualitative example is shown in Figure~\ref{fig:Qualitative_Results}. As can be seen, predictions from Multimodal Graph Transformer are more relevant to contents of the input image as the graph information improves the inferring ability of the Transformer, which further indicates the effectiveness of Multimodal Graph Transformer.

\section{Related Works}
\subsection{Multimodal question answering}
Visual Question Answering (VQA)\cite{vqa,he2021towards} has been a prominent topic in the field of multimodal question answering, garnering significant attention and advancing significantly since the introduction of the first large-scale VQA dataset by\cite{vqa}. To answer VQA questions, models typically leverage variants of attention to obtain a representation of the image that is relevant to the question~\cite{andreas2016neural, Yang2015StackedAN, xu2016ask, mcb, lu2016hierarchical,he2020meddialog}. A plethora of works~\cite{liang2021graghvqa, hudson2018compositional, yi2018neural, xiong2016dynamic, BAN, teney2017graph} have attempted to enhance the reasoning capability of VQA models, with~\cite{teney2017graph} proposing to improve VQA using structured representations of the scene contents and questions. They developed a deep neural network that leverages the structure in these representations and builds graphs over scene objects and question words. The recent release of MultiModalQA~\cite{multimodalqa}, a dataset that demands joint reasoning over texts, tables, and images, has received widespread attention. However, similar to VQA, existing MultiModalQA methods have not fully utilized structured information from the input concepts. To address this, we propose a combination of multimodal graph learning and Transformer models to improve question answering across inputs from multiple different modalities.

\subsection{Attention mechanisms}
The attention mechanism~\cite{show_attend_tell, attention, bert}, has dramatically advanced the field of representation learning in machine learning. The attention mechanism is introduced in~\cite{vaswani2017attention} and widely used in language tasks (i.e., abstract summarization~\cite{xu2020self}), machine translation~\cite{bahdanau2014neural}, reading comprehension~\cite{dai2020funnel}, question answering~\cite{min2019knowledge}, etc. ~\cite{sgnet} proposes using syntax to guide the text modeling by incorporating explicit syntactic constraints into attention mechanisms. Meanwhile, it has seen increasing application in multimodal tasks~\cite{li2020oscar, nam2017dual, lu2016hierarchical}, where it is usually used for learning of interactions between multiple inputs.  Following their success, Transformer models have also shown impressive
results on several vision-and-language tasks~\cite{chen2019uniter, hu2020iterative, he2022parameter, videobert}. ~\cite{graph_transformer} proposes Graph Transformer Networks (GTNs) that can generate new graph structures and learn effective node representation on the new graphs in an end-to-end fashion.  Different from these works, our work incorporates graph information from different modalities into the Transformer to improve the reasoning ability.

\subsection{Exploiting graphs in multimodal reasoning}
Considering that graph priors can
transfer commonalities and mitigate the gap between visual and language domains, researchers explore how to use
graphs~\cite{graph_vl1, graph_vl2} properly in both tasks. In recent years, many classes of GNNs have been developed for both tasks which are divided into two approaches: spectral~\cite{bruna2013spectral} and non-spectral methods~\cite{chen2018fastgcn}. Graphs can also be transferred into latent variables by GCN~\cite{gcn1, gcn2},
which can be directly utilized by models. However, the need for
aligning graph priors from different modalities to do reasoning limits the use of graph priors. Our work addresses this problem via the graph-involved quasi-attention mechanism.

\subsection{Pretraining}
Pretrained models in computer vision~\cite{VGG, resnet} and NLP~\cite{bert, yang2019xlnet, liu2019roberta},
have achieved state-of-the-art performances in many downstream tasks~\cite{thongtan-phienthrakul-2019-sentiment, white2017inference, comclip, image_text_matching1, image_text_matching2}.
Other pretrained models~\cite{lu2019vilbert, videobert} based on BERT~\cite{bert} and ViLT~\cite{vilt} also demonstrate their effectiveness on downstream vision-language tasks. Recent works on vision-language pretraining such as OSCAR~\cite{li2020oscar} perform cross-modal alignment in their visual-language pretraining models. Likewise, our proposed method includes cross-modality alignment, which is critical for reasoning.
Our proposed modular plug-and-play graph-involved quasi-attention mechanism is also model-agnostic and can be also applied to other pretrained Transformer-based vision and language models.

\chapter{Benchmarking and Evaluating Multimodal World Models}

\section{Introduction}

\begin{figure}[h]
  \centering
  \includegraphics[width=0.8\textwidth]{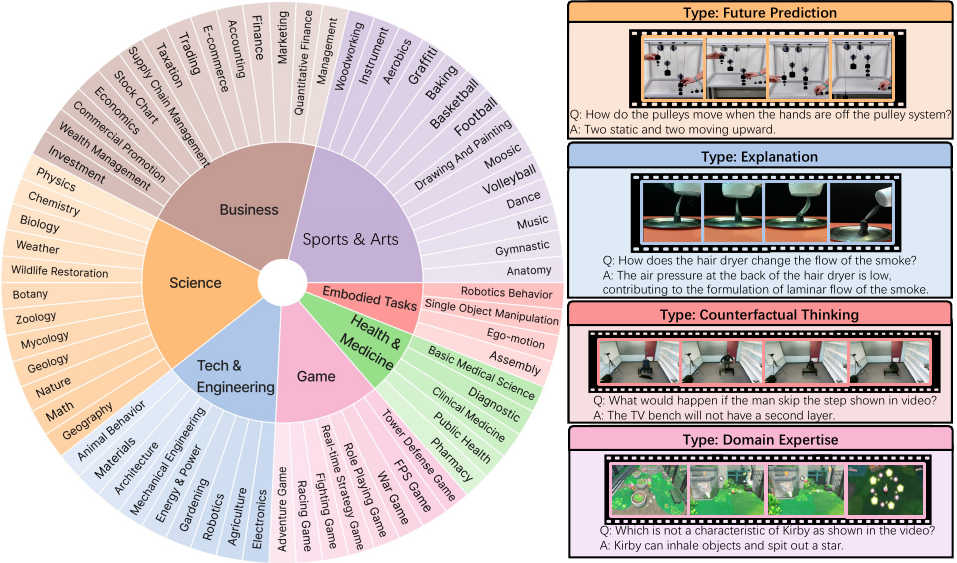}
  \caption{\benchmarkname covers seven broad disciplines and  69 subdisciplines, focusing on the evaluation of multi-faceted reasoning beyond perception (e.g., explanation, counterfactual thinking, future prediction, domain expertise). On the right are four video samples from the Science, Tech \& Engineering, Embodied Tasks, and Game disciplines.} 
  \label{fig:overview}
\end{figure}

Foundation models, such as Large Language Models (LLMs)~\cite{openai2023gpt4, touvron2023llama, jiang2023mistral,anil2023palm} and Multimodal LLMs (MLLMs)~\cite{team2023gemini,videollava,li2023videochat,Maaz2023VideoChatGPT,chen2023minigptv2}, have demonstrated remarkable abilities in text and image domains, igniting debates about their potential pathways to Artificial General Intelligence (AGI). This raises a critical question: how well do these models understand the dynamics of the real world? Are they equipped with an inherent World Model~\cite{lecun2022path,worldknowledge,worldmodels,pandora} that can understand and reason about the underlying principles and causalities of the dynamic, multimodal world?

Videos, with their rich, dynamic portrayal of the real world, are ideally suited for evaluating the "world modeling" capabilities of MLLMs.
Existing video understanding benchmarks~\cite{mvbench,videobench,perceptiontest,mvbench}, however, fall short in two key perspectives for such evaluations. 
First, as LeCun et al.~\cite{lecun2022path} discussed, the world model should be able to \emph{(1) estimate missing information about the state of the world not provided by perception, and (2) predict plausible future states of the world}. Evaluation of such capabilities requires \textbf{multi-faceted reasoning} beyond perception level, including explaining the video dynamics, counterfactual thinking of alternative consequences, and predicting future activities within videos.
Moreover, the \textbf{multi-discipline} nature of the multimodal world necessitates a grasp of diverse fundamental principles---ranging from physics and chemistry to engineering and business. 
Hence, domain expertise across a variety of disciplines is imperative for a thorough evaluation of a model’s world understanding towards AGI~\cite{morris2023agi,yue2023mmmu}.

Therefore, we introduce \benchmarkname, a multi-discipline multi-faceted multimodal video understanding benchmark to comprehensively evaluate MLLMs' abilities in reasoning and interpreting real-world dynamics.  \benchmarkname encompasses a wide range of disciplines and presents multi-faceted reasoning challenges that demand a combination of visual, auditory, and temporal understanding. 
It consists of 1,910 videos that span seven common disciplines, including \emph{Art \& Sports}, \emph{Business}, \emph{Science}, \emph{Health \& Medicine}, \emph{Embodied Tasks}, \emph{Tech \& Engineering}, and \emph{Games}, and 69 subdisciplines (see Figure~\ref{fig:overview}) such as Robotics, Chemistry, Trading, and Agriculture, thereby fulfilling the objective of breadth in discipline coverage. The dataset includes a total of 1,559 question-answer pairs and video captions annotated and reviewed by humans.
Meanwhile, for multi-faceted reasoning, \benchmarkname mainly contains seven kinds of questions focusing on \emph{explanation} (explaining the phenomenon in videos), \emph{counterfactual thinking} (answering what-if questions), \emph{future prediction} (predicting future events), \emph{domain expertise} (answering domain-specific inquiries), \emph{temporal understanding} (reasoning about temporal information), and etc. Four video examples with these questions from different disciplines are depicted in Figure~\ref{fig:overview}. To serve as a comprehensive benchmark, \benchmarkname comprises two datasets: a human-annotated dataset for evaluating MLLMs on the whole video and a synthetic dataset designed to analyze MLLMs' perception within single visual or audio modalities. We evaluate 15 MLLMs that can handle videos or image sequences on \benchmarkname, including both open-source (e.g., Video-LLaVA-7B~\cite{videollava}) and proprietary models (GPT-4o~\cite{gpt4o} and Gemini~\cite{team2023gemini}).

\begin{table}[t]
\centering
\caption{Comparison between \benchmarkname and previous benchmarks for real-world video understanding on a variety of criteria. Multi-faceted include Explanation (\texttt{Explain.}), Counterfactual Thinking (\texttt{Counter.}), Future Prediction (\texttt{Future.}) and Domain Expertise (\texttt{Domain.}) \benchmarkname is the first multi-discipline and multitask video understanding benchmark that covers wider reasoning questions, and also included first-party data annotations.  }
\label{tab:video-datasets}
\setlength{\tabcolsep}{3pt}
\resizebox{\linewidth}{!}{
\begin{tabular}{lccccccc}
\toprule
\multirow{2}{*}[0em]{\textbf{Benchmarks}} 
& \multirow{2}{*}[0em]{ $\begin{array}{l}
    \textbf{Multi-}\\
     \textbf{Discipline} \\
\end{array}$}
& \multirow{2}{*}[0em]{$\begin{array}{l}
    \textbf{Multi-}\\
     \textbf{Task} \\
\end{array}$}
& \multicolumn{4}{c}{\textbf{Multi-Faceted Reasoning}} 
& \multirow{2}{*}[0em]{$\begin{array}{l}
    \textbf{First-Party}\\
     \textbf{Annotation} \\
\end{array}$} \\
\cmidrule{4-7}
&&
& $\begin{array}{c}
    \texttt{Explain.}\\
\end{array}$      
& $\begin{array}{c}
    \texttt{Counter.}\\
\end{array}$
& $\begin{array}{c}
    \texttt{Future.}\\
\end{array}$
& $\begin{array}{c}
    \texttt{Domain.}\\
\end{array}$
&  \\
\midrule
MovieQA~\cite{tapaswi2016movieqa} &  &  &\cmark&&&  & \cmark 
\\TVQA~\cite{lei2018tvqa} &  &  &\cmark&&&  & \cmark \\
ActivityNet-QA~\cite{yu2019activitynet} &  &  & &&&  & \cmark  \\
MSVD-QA~\cite{xu2017video}~\cite{msr-vtt} &  &  &&&&  & \cmark  \\
MSRVTT-QA~\cite{msr-vtt} &  &  &&&&  & \cmark  \\
Sports-QA~\cite{sportsqa} &  &  & &\cmark&& \cmark & \cmark  \\
VaTeX~\cite{wang2019vatex} &  & \cmark &  &  &  &  & \cmark  \\
VALUE~\cite{li2021value} &  & \cmark &  &  &  &  &   \\
Video-Bench~\cite{ning2023video} &  & \cmark & &&\cmark& \cmark &   \\
MVBench~\cite{mvbench} &  & \cmark &&\cmark&\cmark&  &  \\
Perception Test~\cite{perceptiontest} &  & \cmark &\cmark&\cmark&\cmark&  &  \\
{VideoMME~\cite{videomme}} &  &  &&&\cmark&\cmark  & \cmark \\
{MMBench-Video~\cite{fang2024mmbench}} &  &  &&\cmark&\cmark& \cmark &  \cmark\\
{TempCompass~\cite{liu2024tempcompass}} && \cmark& & & \cmark& \cmark&\cmark\\
{ViLMA~\cite{kesen2023vilma}} & & \cmark& & & \cmark& \cmark&\cmark\\
{VITATECS~\cite{li2023vitatecs}} & & & & \cmark& \cmark& \cmark&\cmark\\
{NExT-QA~\cite{nextqa}} &&\cmark &\cmark & &\cmark & &\cmark\\
{CVRR~\cite{cvrr}} && &\cmark & &\cmark & &\cmark\\
{Causal-VidQA~\cite{causalvidqa}} & & &\cmark&\cmark&\cmark & &\cmark \\
\benchmarkname (Ours) & \cmark & \cmark & \cmark & \cmark & \cmark & \cmark & \cmark  \\
\bottomrule
\end{tabular}}
\end{table}

\section{Multi-discipline Multi-faceted World Model Evaluation}
We build the \benchmarkname benchmark on three key design principles: multi-discipline coverage, multi-faceted reasoning, and temporal reasoning. It spans various disciplines that require domain expertise and incorporates diverse reasoning skills such as explanation, counterfactual thinking, and future prediction. The benchmark consists of two parts: a human-annotated dataset and a synthetic dataset.\textbf{ The human-annotated dataset serves as the main testbed to evaluate MLLMs from multiple perspectives.} The synthetic dataset is divided into two subsets, each designed to assess MLLMs' perception behavior based on visual and audio inputs, respectively.

\subsection{Manual Data Collection}
\label{sec:data_collection}
We collect videos from YouTube with the Creative Licence in seven disciplines:
Art $\&$ Sports (18.5\%), Business (12.0\%), Science (20.4\%), Health $\&$ Medicine (12.0\%), Embodied Tasks (12.0\%), Tech $\&$ Engineering (12.9\%), and Game (12.2\%). For Art $\&$ Sports, 29 videos are collected from the SportsQA dataset~\cite{sportsqa}. And for Embodied Tasks, 24 videos are sourced from IKEA Assembly~\cite{ben2021ikea}, RT-1~\cite{rt-1}, and Ego4D~\cite{grauman2022ego4d} datasets to increase video diversity.

Our manual benchmark collection takes two
stages. In the first stage, we conduct a detailed examination of each of the seven primary disciplines to identify a comprehensive range of subdisciplines for inclusion in our benchmark. 

\begin{figure}[tb]
  \centering
    \includegraphics[width=0.8\textwidth]{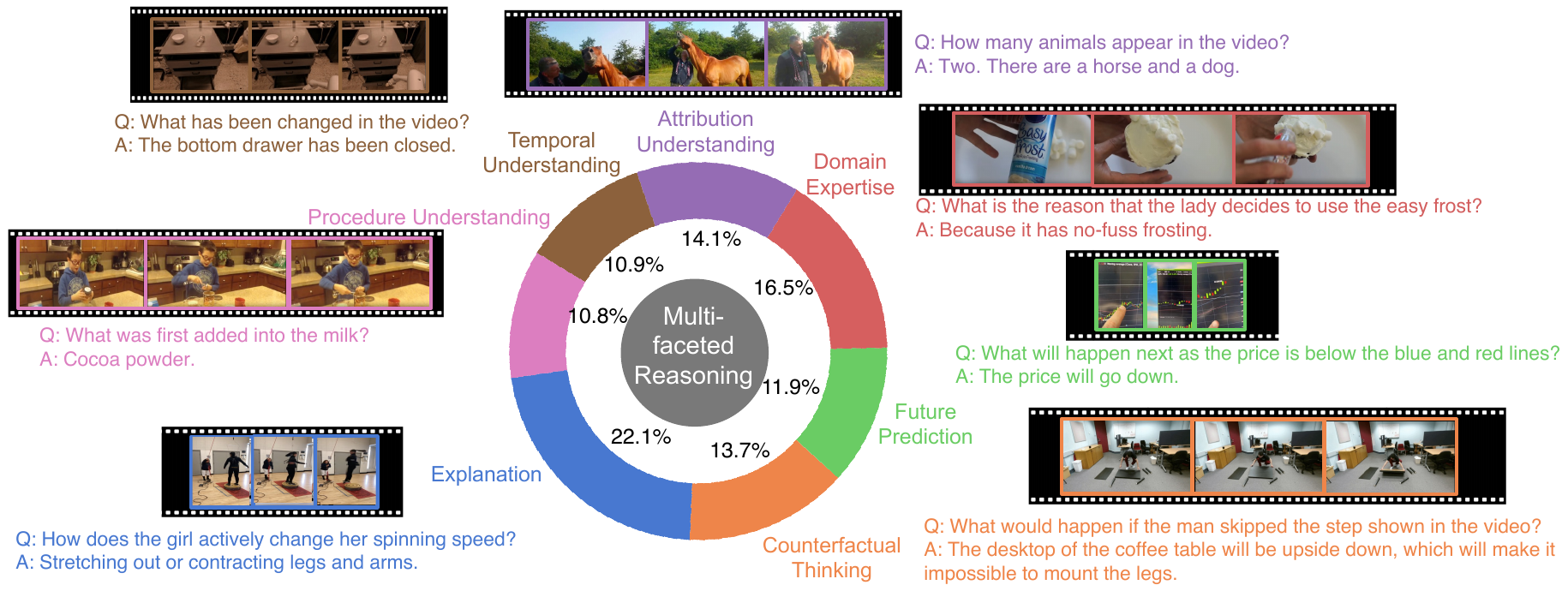}
    \caption{{The questions in \benchmarkname are designed to evaluate seven primary understanding and reasoning abilities of models. Each question is annotated with all relevant categories. The figure showcases one example question for each reasoning category, based on its main category.}}
  \label{fig:questions_per_type}
\end{figure}

During the second stage, our team began the task of annotating questions, answers, and options. All annotators were asked to carefully watch the collected videos and create questions with corresponding answers and options, ensuring that understanding the video content and applying temporal reasoning were necessary to determine the correct answers. We also ensured that the clarity, correctness, and grammatical accuracy of the questions and answers were verified using GPT-4o, and that the questions could not be correctly answered without video input. We craft questions that primarily test seven aspects of multimodal video understanding also from the perspective of \textbf{multi-faceted reasoning}: 1) Explanation: Questions ask the model to elucidate the underlying logic or purpose within the video;
2) Counterfactual Thinking: Tests the model's ability to hypothesize and consider alternative outcomes;
3) Future Prediction: Aims to predict future events based on the current scenario, challenging the model’s foresight;
4) Domain Expertise: Evaluates the model's depth of knowledge in specific fields, such as how to assemble a coffee table;
5) Temporal Understanding: Assesses the model's capability to reason about temporal sequences and dynamics;
6) Attribution Understanding: These questions focus on identifying cause-and-effect relationships within the video, including tasks like counting;
7) Procedure Understanding: Tests the model's ability to comprehend and explain procedural tasks shown in the video. The detailed distribution and examples are shown in Figure~\ref{fig:questions_per_type}. For quality control, we ensure each annotation is cross-checked by at least two professional researchers to ensure accuracy and prevent annotation errors.

\subsection{Automated Data Collection}
\label{automatic_data}
Understanding real-world dynamics requires models to process both audio and visual modalities. To evaluate MLLMs' perception abilities in these modalities, we designed an automated data collection pipeline. This pipeline collects targeted videos and generates QA pairs based on either audio or visual information, ensuring the model's capabilities are assessed independently for each modality. By using information from a single modality to generate QA pairs, our pipeline ensures that the synthetic data remains unbiased regarding input modality.

\begin{figure}[tb]
  \centering
  \includegraphics[width=0.8\textwidth]{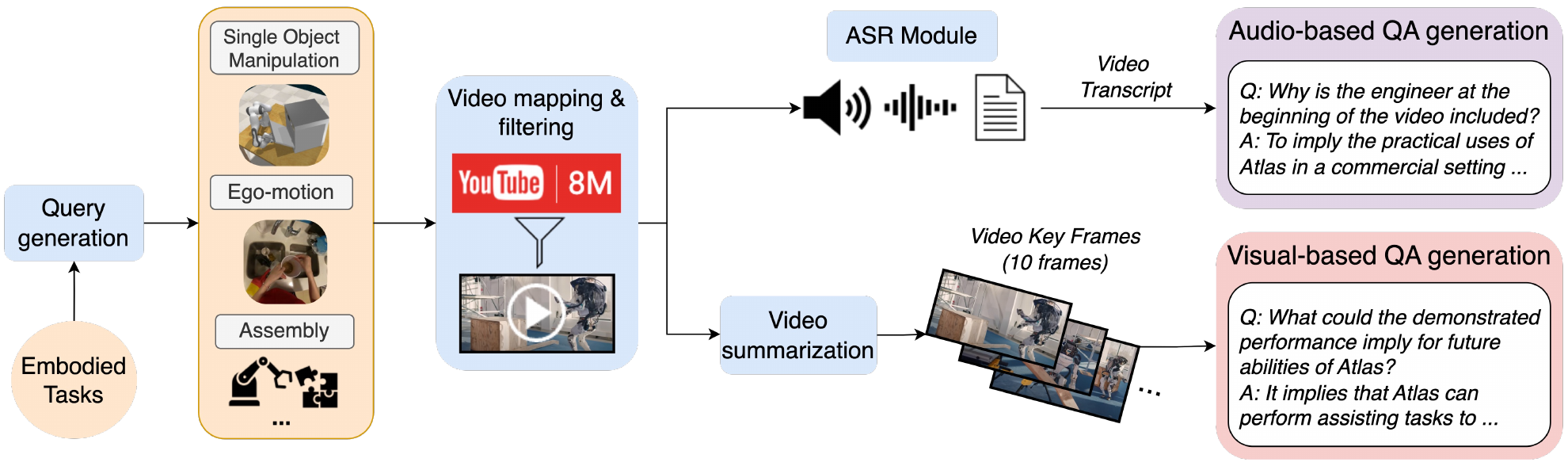}
  \caption{Schematic diagram of the synthetic data generation pipeline in \benchmarkname. It starts with generating subdiscipline-specific queries, followed by video retrieval from YouTube-8M~\cite{youtube8m} and YouTube. Keyframes are extracted for visual-based QA generation, and videos are transcribed using an ASR module for audio-based QA generation. 
  }
  \label{fig:example}
\end{figure}

The synthetic data generation pipeline is illustrated in Figure~\ref{fig:example}. We employ a systematic approach to gather videos with Creative Commons licenses from YouTube and the extensive YouTube-8M dataset~\cite{youtube8m}. This method ensures a diverse and comprehensive collection of video data, which is important for the robust evaluation of multimodal video understanding models.

\noindent \textbf{Video Collection and Processing~}
We start with the video~\textit{Query Generator}. We start with the same seven disciplines as the manually collected dataset. For each discipline, a set of subdisciplines is defined to encapsulate a wide spectrum of topics, ensuring a diverse and comprehensive dataset.
Once the queries are generated, the \textit{Video Mapping and Filtering} step is initiated. We perform mapping of videos to YouTube-8M and online videos, constrained by a strict time limit of two minutes per query, keeping only the most pertinent videos that satisfy the predefined criteria. Simultaneously, the works in conjunction with the video transcripts to extract key terms and concepts. This iterative process refines the search parameters and enhances the semantic richness of the dataset by identifying and encoding the salient themes present in the videos. The \textit{Video Summarization} module utilizes Query-focused video summarization techniques based on Katna\footnote{
\url{https://github.com/keplerlab/katna}} and UniVTG~\cite{univtg}. This module selects ten representative frames from each video, distilling the essence of the content while preserving the narrative context. This summarization facilitates efficient storage and quicker processing times, which are crucial for large-scale analysis.

\noindent \textbf{QA Generation~}
The final stage in our pipeline is the \textit{QA / Caption Generation} module, where we leverage the capabilities of GPT-4V to generate accurate and contextually relevant questions and answers, as well as captions, based on the video frames and transcripts. This step not only provides rich annotations for each video but also equips the dataset with a multimodal dimension that supports various downstream tasks such as video QA, captioning, and more.

\noindent \textbf{Quality of the Synthetic Dataset~}
Human evaluators were engaged to ascertain the reasonableness of automatically generated questions and answers, ensuring that the synthetic dataset maintains a high standard of quality and relevance. The findings from this human evaluation phase are detailed in Section D of the Appendix, offering insights into the dataset's efficacy and the realism of its constructed queries and responses.

\begin{table}[t]
\caption{Key Statistics of the \benchmarkname Benchmark. The main subset is the human-annotated subset. Synthetic Subset I contains generated QA pairs focused exclusively on the audio content, while Synthetic Subset II contains QA pairs focused exclusively on the visual content of the video.}
\centering
\resizebox{0.8\textwidth}{!}{
\begin{tabular}{lrrrr}
\toprule
\multicolumn{1}{l}{\textbf{Statistics}} & \multicolumn{1}{l}{\textbf{Main Subset}} & \multicolumn{1}{l}{\textbf{Synthetic I}} & \multicolumn{1}{l}{\textbf{Synthetic II}}  \\
\midrule
\#Discipline/\#Subdiscipline & 7/61 & 7/51 & 7/54  \\
\#Videos & 417 & 746 & 747 \\
\#QA pairs & 1,559 & 2,969 & 2,099\\
Avg Video Lengths (s) & 102.3 & 103.4 & 115.8 \\
\hdashline
Avg \#Questions per Video & 4.05 & 3.98 & 2.81  \\
Avg \#Options & 3.90 & 4.00 & 4.00  \\
Avg Question Length & 11.39 & 15.12 & 17.56 \\
Avg Option Length & 7.27 & 6.01 & 5.19  \\
Avg Answer Length & 6.42 & 6.71 & 5.67  \\
Avg Caption Length & 27.00 & 71.87 & 82.33  \\
{\# Unique Words in Questions} & {1,913} & {2,528} & {2,279} \\
{\# Unique Words in Answers} & {2,292} & {2,981} & {2,657} \\
\bottomrule
\end{tabular}
}
\label{tab:merged_benchmark_stats_total}
\end{table}
Finally, the statistics of automated curated data, which is used for the ablation study, are shown in Table~\ref{tab:merged_benchmark_stats_total}.
The taxonomy of our dataset is shown in Figure~\ref{fig:overview}. We note that only a portion of the subdisciplines are shown due to space concerns. Please refer to the Appendix for full information.

\section{Experiments}
\subsection{Experimental Settings}
\label{sec:implement}
In our study, we compare MLLM's performance on the \benchmarkname benchmark, including GPT-4o~\cite{gpt4o}, GPT-4V~\cite{gpt4-v}, Gemini Pro~\cite{team2023gemini}, Claude-3.5-Sonnet~\cite{claude}, Video-Chat~\cite{li2023videochat}, Video-ChatGPT~\cite{Maaz2023VideoChatGPT}, Video-LLaMA~\cite{zhang2023videollama}, Video-LLaVA~\cite{videollava}, ChatUnivi~\cite{jin2023chatunivi}, mPLUG-Owl~\cite{ye2023mplug}, Otter~\cite{li2023otter}, ImageBind-LLM~\cite{han2023imagebind}, PandaGPT~\cite{su2023pandagpt}, LWM~\cite{lwm}, 
and X-Instruct-BLIP~\cite{panagopoulou2023xinstructblip}. For proprietary model, we adhere to the default settings provided by their official APIs. They both take ten image frames extracted from the video content as the input. The Gemini Pro is set to process visual input and configured with safety settings to filter a range of harmful content. The configuration thresholds are set to `BLOCK\_NONE'. For PandaGPT, we set `top\_p' to 0.7 and `temperature' to 0.5. For VideoChat, we set `max\_frames' to 100. For X-Instruct-BLIP, the model is implemented using four image frames. We use GPT-4-32K as the judge for judging whether the model answer is correct when it can not mapped to the option letter using the rule-based method. For others, we all use the default setting. All inferences are run on a NVIDIA A6000 workstation. The detailed implementation is given in the Appendix.

\begin{table}[t]
\centering
\caption{MLLM accuracy across diverse disciplines (averaging over three runs). 
GPT-4V and Gemini Pro lead at most disciplines and achieve the best overall accuracy. The best open-source model Video-LLaVA-7B outperforms them on Embodied Tasks and perform similarly on Art \& Sports. {All data are annotated by humans.}}
\label{maineval}
\setlength{\tabcolsep}{3pt}
\resizebox{\linewidth}{!}{
\begin{tabular}{llllllllll}
\toprule
\multirow{2}{*}{\textbf{Model}}        
& \textbf{Art\& } & \multirow{2}{*}{\textbf{Business}} & \multirow{2}{*}{\textbf{Science}} & \textbf{Health\&} & \textbf{Embodied } & \textbf{Tech\& } & \multirow{2}{*}{\textbf{Game}} & \multirow{2}{*}{\textbf{Average}} \\
& \textbf{Sports}& & &  \textbf{Medicine}& \textbf{Tasks}& \textbf{Engineering}& \\
\midrule
Random Choice&25.03 & 25.09 & 26.44 & 25.00 & 26.48 & 30.92 & 25.23 & 26.31 \\ 
\midrule
\multicolumn{9}{c}{\textit{Proprietary MLLMs}} \\
\midrule
GPT-4o~\cite{gpt4o} & \underline{47.87} \tiny{$\pm$1.47} & \textbf{91.14} \tiny{$\pm$0.87} & \textbf{73.78} \tiny{$\pm$2.88} & \textbf{83.33} \tiny{$\pm$1.47} & \underline{62.94} \tiny{$\pm$3.47} & \textbf{75.53} \tiny{$\pm$2.61} & \textbf{80.32} \tiny{$\pm$2.05} & \textbf{62.54} \tiny{$\pm$0.79} \\
Claude-3.5-Sonnet~\cite{claude} & \textbf{54.58} \tiny{$\pm$0.45} & 63.87 \tiny{$\pm$0.40} & 59.85 \tiny{$\pm$1.28} & 54.51 \tiny{$\pm$1.28} & 30.99 \tiny{$\pm$0.40} & 58.87 \tiny{$\pm$0.61} & 59.44 \tiny{$\pm$0.68} & \underline{54.54} \tiny{$\pm$0.29} \\
GPT-4V~\cite{gpt4-v}         & 36.17  \tiny{$\pm$0.58  }& \underline{81.59}  \tiny{$\pm$1.74  }& \underline{66.52}  \tiny{$\pm$1.86  }& 73.61  \tiny{$\pm$0.49  }& 55.48  \tiny{$\pm$2.70  }& 61.35  \tiny{$\pm$1.00  }& \underline{73.49}  \tiny{$\pm$1.97  }& 52.30 \tiny{$\pm$0.49     }        \\
Gemini Pro~\cite{team2023gemini}          & 37.12  \tiny{$\pm$2.68  }& 76.69  \tiny{$\pm$2.16  }& 62.81  \tiny{$\pm$1.83  }& \underline{76.74}  \tiny{$\pm$1.30  }& 43.59  \tiny{$\pm$0.33  }& \underline{69.86}  \tiny{$\pm$2.01  }& 66.27  \tiny{$\pm$2.60  }& 51.02  \tiny{$\pm$1.35 } \\
\midrule 
\multicolumn{9}{c}{\textit{Open-source MLLMs}}  \\
\midrule
Video-LLaVA-7B~\cite{videollava}      & 35.91  \tiny{$\pm$0.96  }& 51.28  \tiny{$\pm$0.87  }& 56.30  \tiny{$\pm$0.76  }& 32.64  \tiny{$\pm$0.49  }& \textbf{63.17}  \tiny{$\pm$1.44  }& 58.16  \tiny{$\pm$1.00  }& 49.00  \tiny{$\pm$3.16  }& 44.60  \tiny{$\pm$0.58   }       \\
Video-Chat-7B~\cite{li2023videochat} & 39.53  \tiny{$\pm$0.06  }& 51.05  \tiny{$\pm$0.00  }& 30.81  \tiny{$\pm$0.21  }& 46.18  \tiny{$\pm$0.49  }& 40.56  \tiny{$\pm$0.57  }& 39.36  \tiny{$\pm$0.00  }& 44.98  \tiny{$\pm$0.57  }& 40.11  \tiny{$\pm$0.06 } \\
ChatUnivi-7B~\cite{jin2023chatunivi} & 24.47  \tiny{$\pm$0.49  }& 60.84  \tiny{$\pm$1.51  }& 52.00  \tiny{$\pm$0.73  }& 61.11  \tiny{$\pm$1.96  }& 46.15  \tiny{$\pm$2.06  }& 56.74  \tiny{$\pm$1.33  }& 52.61  \tiny{$\pm$2.84  }& 39.47  \tiny{$\pm$0.42 } \\
mPLUG-Owl-7B ~\cite{ye2023mplug} & 29.16  \tiny{$\pm$1.62  }& 64.10  \tiny{$\pm$1.84  }& 47.41  \tiny{$\pm$3.29  }& 60.07  \tiny{$\pm$1.30  }& 23.78  \tiny{$\pm$3.47  }& 41.84  \tiny{$\pm$5.09  }& 62.25  \tiny{$\pm$3.16  }& 38.94  \tiny{$\pm$1.52  }
\\
Video-ChatGPT-7B ~\cite{Maaz2023VideoChatGPT} &26.84  \tiny{$\pm$0.69  }& 39.16  \tiny{$\pm$3.02  }& 36.45  \tiny{$\pm$1.31  }& 53.12  \tiny{$\pm$0.00  }& 36.60  \tiny{$\pm$3.25  }& 41.49  \tiny{$\pm$1.74  }& 36.55  \tiny{$\pm$2.27  }& 33.27  \tiny{$\pm$0.97  } \\
PandaGPT-7B~\cite{su2023pandagpt} & 25.33  \tiny{$\pm$0.54  }& 42.66  \tiny{$\pm$3.02  }& 39.41  \tiny{$\pm$2.67  }& 38.54  \tiny{$\pm$3.07  }& 35.43  \tiny{$\pm$0.87  }& 41.84  \tiny{$\pm$2.79  }& 40.16  \tiny{$\pm$4.65  }& 32.48  \tiny{$\pm$0.45 } \\
ImageBind-LLM-7B~\cite{han2023imagebind} & 24.82  \tiny{$\pm$0.16  }& 42.66  \tiny{$\pm$0.99  }& 32.15  \tiny{$\pm$1.11  }& 30.21  \tiny{$\pm$1.47  }& 46.85  \tiny{$\pm$1.14  }& 41.49  \tiny{$\pm$1.50  }& 41.37  \tiny{$\pm$0.57  }& 31.75  \tiny{$\pm$0.14 } \\
X-Instruct-BLIP-7B~\cite{panagopoulou2023xinstructblip} & 21.08  \tiny{$\pm$0.27  }& 15.85  \tiny{$\pm$0.87  }& 22.52  \tiny{$\pm$1.11  }& 28.47  \tiny{$\pm$0.49  }& 18.41  \tiny{$\pm$1.44  }& 22.34  \tiny{$\pm$0.87  }& 26.10  \tiny{$\pm$0.57  }& 21.36  \tiny{$\pm$0.18 } \\
LWM-1M-JAX~\cite{lwm} & 12.04  \tiny{$\pm$0.53  }& 17.48  \tiny{$\pm$0.57  }& 15.41  \tiny{$\pm$0.91  }& 20.49  \tiny{$\pm$0.98  }& 25.87  \tiny{$\pm$1.98  }& 21.99  \tiny{$\pm$2.19  }& 11.65  \tiny{$\pm$3.01  }& 15.39  \tiny{$\pm$0.32 } \\
Otter-7B~\cite{li2023otter} & 17.12  \tiny{$\pm$1.17  }& 18.65  \tiny{$\pm$0.87  }& ~~9.33  \tiny{$\pm$0.36  }& ~~6.94  \tiny{$\pm$0.98  }& 13.29  \tiny{$\pm$1.51  }& 15.96  \tiny{$\pm$1.74  }& 15.26  \tiny{$\pm$0.57  }& 14.99  \tiny{$\pm$0.77}  \\
Video-LLaMA-2-13B~\cite{zhang2023videollama} & ~~6.15  \tiny{$\pm$0.44  }& 21.21  \tiny{$\pm$0.66  }& 22.22  \tiny{$\pm$1.45  }& 31.25  \tiny{$\pm$1.70  }& 15.38  \tiny{$\pm$1.14  }& 19.15  \tiny{$\pm$1.74  }& 24.90  \tiny{$\pm$5.93  }& 14.03  \tiny{$\pm$0.29 } \\
 \bottomrule
\end{tabular}}
\end{table}

\subsection{Evaluation Strategy}
Our dataset contains multiple-choice questions and captions corresponding to each video, supporting tasks such as video question answering and video captioning. In our evaluation setup, we focus on video question answering by measuring a model’s accuracy in selecting the correct answer from the provided options. This method is straightforward to quantify and provides objective assessment. However, one challenge is reliably mapping the model’s predictions to one of the predefined choices.

To address this, we employ two mapping strategies. We employ two mapping strategies. The first method employs automated scripts to parse the models' predictions and compare the parsed results with the ground truth, similar to the approach used in~\cite{yue2023mmmu}; The second method involves models freely generating answers, which are then evaluated by GPT-4. Given the question, correct answer, and model's prediction, GPT-4 returns a True or False judgment. This approach is based on recent works in model evaluation~\cite{Maaz2023VideoChatGPT, hsu2023gpt,hackl2023gpt,liu2023gpteval}.

We validated the second GPT-4-based evaluation approach with human evaluators, showing an error rate of only 4.76\% across 189 examples, demonstrating its reliability as an evaluator. Detailed results for human evaluation and both evaluation strategies are provided in Appendix. All results presented in the main paper are based on the second evaluation approach.

\subsection{Main Evaluation Results on Human-annotated Data}
\label{sec:results}
We show in Table~\ref{maineval} the main evaluation results of different MLLMs. Among these, GPT-4o emerges as the top performer, followed by Claude-3.5-Sonnet. Video-LLaVA also demonstrates strong results, primarily due to the extensive training data which consists of 558K LAION-CCSBU image-text pairs and 702K video-text pairs from WebVid~\cite{webvid}. Its superior performance may also be attributed to the adoption of CLIP ViT-L/14 trained in LanguageBind~\cite{videollava} as its vision model and the inclusion of a large volume of image-video-text pairings within the
training data. On the other hand, models like Otter and LWM perform poorly across most disciplines, possibly due to their weaker backbone and architecture used. Otter uses the LLaMA-7B language encoder and a CLIP ViT-L/14 vision encoder, both of which are frozen, with only the Perceiver resampler~\cite{awadalla2023openflamingo} module fine-tuned, which may lead to the lower performance.
Additionally, four MLLMs perform even worse than random, highlighting the challenging nature of \benchmarkname.

\begin{table}[t]
\centering
\caption{{Results of different MLLMs on multi-faceted reasoning. All data are annotated by humans.}}
\label{multifacetedreasoning}
\resizebox{\linewidth}{!}{
\begin{tabular}{lcccccc}
\toprule
\multirow{2}{*}{\textbf{Model}} & \multirow{2}{*}{\textbf{Explanation}} & \textbf{Counterfactual} & \textbf{Future} & \textbf{Domain} & \textbf{Attribution} & \textbf{Temporal} \\
 &  & \textbf{Thinking} & \textbf{Prediction} & \textbf{Expertise} & \textbf{Understanding} & \textbf{Understanding} \\
\midrule
\multicolumn{7}{c}{\textit{Proprietary MLLMs}} \\
\midrule
GPT-4o~\cite{gpt4o} & \textbf{56.68} \tiny{$\pm$0.72} & \textbf{75.88} \tiny{$\pm$1.47} & \textbf{82.48} \tiny{$\pm$0.69} & \textbf{69.05} \tiny{$\pm$0.49} & \textbf{65.10} \tiny{$\pm$1.15} & \textbf{40.90} \tiny{$\pm$2.42} \\
GPT-4V~\cite{gpt4-v} & 44.90 \tiny{$\pm$0.07} & 64.90 \tiny{$\pm$0.58} & 78.59 \tiny{$\pm$1.55} & 61.07 \tiny{$\pm$0.17} & 59.61 \tiny{$\pm$0.85} & 27.17 \tiny{$\pm$1.00} \\
Claude-3.5-Sonnet~\cite{claude} & 51.94 \tiny{$\pm$0.23} & 62.75 \tiny{$\pm$0.16} & 71.78 \tiny{$\pm$0.40} & 66.79 \tiny{$\pm$0.45} & 40.00 \tiny{$\pm$0.55} & 25.77 \tiny{$\pm$0.46} \\
Gemini Pro~\cite{team2023gemini} & 48.58 \tiny{$\pm$1.07} & 65.49 \tiny{$\pm$0.42} & 65.45 \tiny{$\pm$1.05} & 53.87 \tiny{$\pm$1.31} & 43.92 \tiny{$\pm$1.40} & 24.65 \tiny{$\pm$1.00} \\
\midrule 
\multicolumn{7}{c}{\textit{Open-source MLLMs}}  \\
\midrule
Video-LLaVA~\cite{videollava} & 42.46 \tiny{$\pm$0.61} & 42.55 \tiny{$\pm$0.85} & 64.96 \tiny{$\pm$0.69} & 47.86 \tiny{$\pm$0.58} & 36.86 \tiny{$\pm$1.95} & 34.45 \tiny{$\pm$1.19} \\
Video-Chat-7B~\cite{li2023videochat} & 41.66 \tiny{$\pm$0.06} & 43.73 \tiny{$\pm$0.32} & 45.74 \tiny{$\pm$0.20} & 40.95 \tiny{$\pm$0.10} & 30.59 \tiny{$\pm$0.00} & 25.77 \tiny{$\pm$0.23} \\
Video-ChatGPT-7B~\cite{Maaz2023VideoChatGPT}  & 32.13 \tiny{$\pm$0.38} & 39.02 \tiny{$\pm$1.12} & 47.45 \tiny{$\pm$2.09} & 33.69 \tiny{$\pm$1.08} & 21.18 \tiny{$\pm$2.00} & 23.53 \tiny{$\pm$0.76}  \\
ImageBind-LLM-7B~\cite{han2023imagebind} & 29.51 \tiny{$\pm$0.27} & 26.86 \tiny{$\pm$0.58} & 50.61 \tiny{$\pm$0.20} & 33.93 \tiny{$\pm$0.17} & 34.90 \tiny{$\pm$1.40} & 19.89 \tiny{$\pm$0.91} \\
PandaGPT-7B~\cite{su2023pandagpt} & 29.55 \tiny{$\pm$0.41} & 37.45 \tiny{$\pm$1.80} & 46.47 \tiny{$\pm$1.05} & 33.93 \tiny{$\pm$0.45} & 26.27 \tiny{$\pm$2.24} & 28.01 \tiny{$\pm$0.82} \\
ChatUnivi-7B~\cite{jin2023chatunivi} & 33.91 \tiny{$\pm$0.31} & 48.82 \tiny{$\pm$0.48} & 61.80 \tiny{$\pm$0.53} & 45.95 \tiny{$\pm$0.68} & 33.33 \tiny{$\pm$0.64} & 22.97 \tiny{$\pm$0.91} \\
Video-LLaMA-2-13B~\cite{zhang2023videollama} & 10.55 \tiny{$\pm$0.29} & 23.92 \tiny{$\pm$0.97} & 25.30 \tiny{$\pm$1.11} & 16.31 \tiny{$\pm$1.03} & 8.63 \tiny{$\pm$0.85} & 6.16 \tiny{$\pm$1.00} \\
X-Instruct-BLIP-7B~\cite{panagopoulou2023xinstructblip} & 23.05 \tiny{$\pm$0.24} & 15.29 \tiny{$\pm$0.28} & 27.25 \tiny{$\pm$0.53} & 21.07 \tiny{$\pm$0.51} & 24.31 \tiny{$\pm$0.64} & 11.20 \tiny{$\pm$0.82} \\
LWM-1M-JAX~\cite{lwm} & 11.62 \tiny{$\pm$0.39} & 18.82 \tiny{$\pm$0.55} & 30.66 \tiny{$\pm$0.34} & 17.98 \tiny{$\pm$0.26} & 21.57 \tiny{$\pm$0.85} & 7.00 \tiny{$\pm$0.46} \\
Otter-7B~\cite{li2023otter} & 16.91 \tiny{$\pm$0.54} & 10.98 \tiny{$\pm$0.42} & 15.82 \tiny{$\pm$0.20} & 13.10 \tiny{$\pm$0.68} & 17.65 \tiny{$\pm$0.00} & 9.52 \tiny{$\pm$1.00} \\
mPLUG-Owl-7B~\cite{ye2023mplug} & 35.20 \tiny{$\pm$1.17} & 49.61 \tiny{$\pm$1.31} & 55.47 \tiny{$\pm$1.58} & 47.74 \tiny{$\pm$1.07} & 24.71 \tiny{$\pm$2.00} & 20.17 \tiny{$\pm$0.69} \\
\bottomrule
\end{tabular}}
\end{table}

\noindent \textbf{Study on Multi-faceted Reasoning}
~{Table~\ref{multifacetedreasoning} illustrates the multi-faceted reasoning performance of each MLLM.} GPT-4o emerges as the strongest model across all facets. Notably, in temporal understanding, the open-sourced Video-LLaVA outperforms all other models except GPT-4o, likely due to its extensive training on high temporal resolution video data, enhancing its Spatiotemporal reasoning abilities. This is further reflected in its high scores on Embodied Tasks (the best) and Art \& Sports, both of which involve dense Spatiotemporal information, as shown in Table~\ref{maineval}.

\noindent \textbf{Study on MLLM Performance at Different Difficulty Levels for Average Humans}

\begin{figure}[tb]
  \centering
  \includegraphics[width=0.8\textwidth]{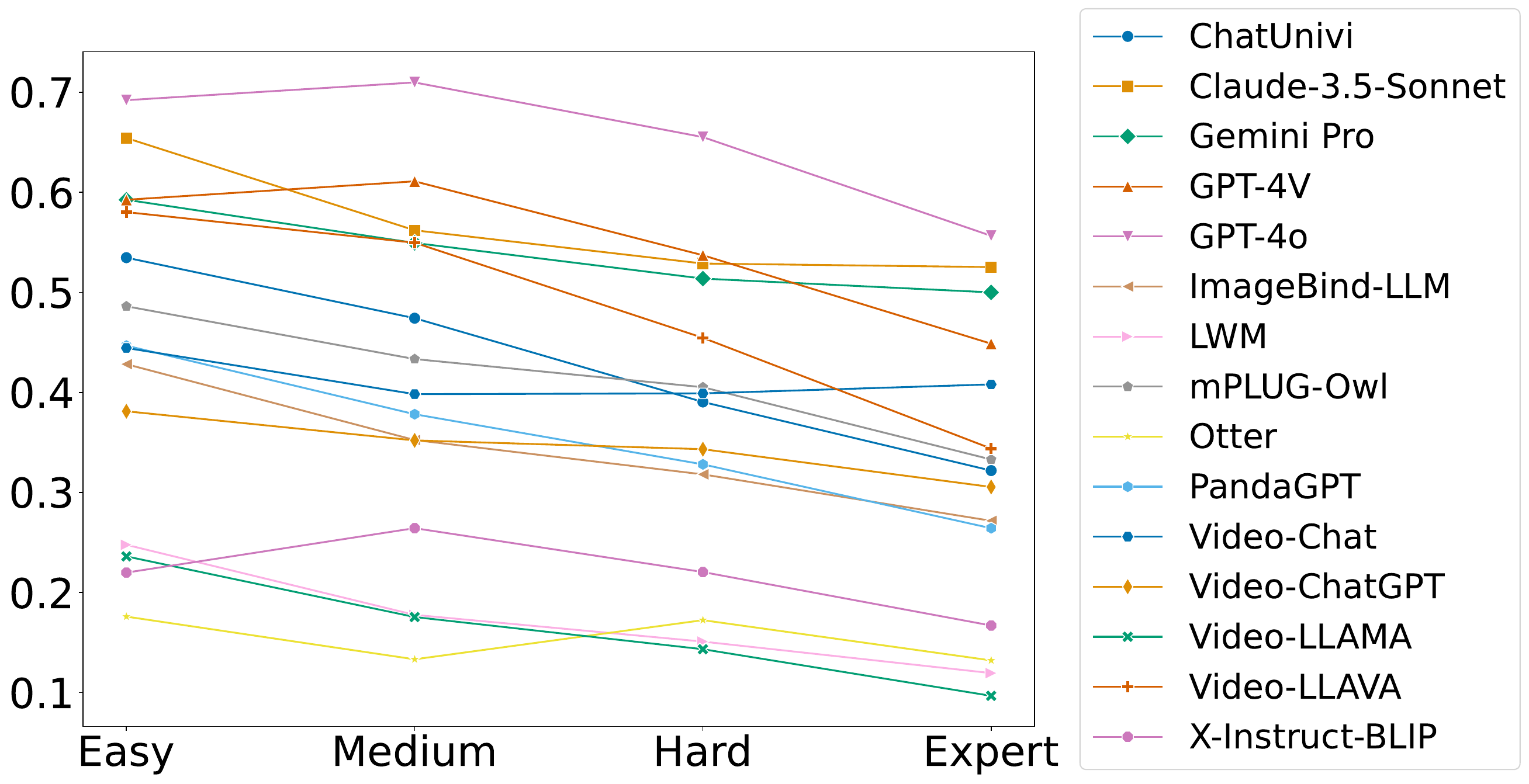}
  \caption{Accuracy of MLLMs at difficulty levels for average humans. Average human difficulty levels are defined by 3 turkers' performance per question: Easy (3/3 correct answers), medium (2/3 correct), hard (1/3 correct), and expert (0/3 correct).}
  \label{fig:subfig1}
\end{figure}

\begin{figure}[tb]
  \centering
  \includegraphics[width=0.6\textwidth]{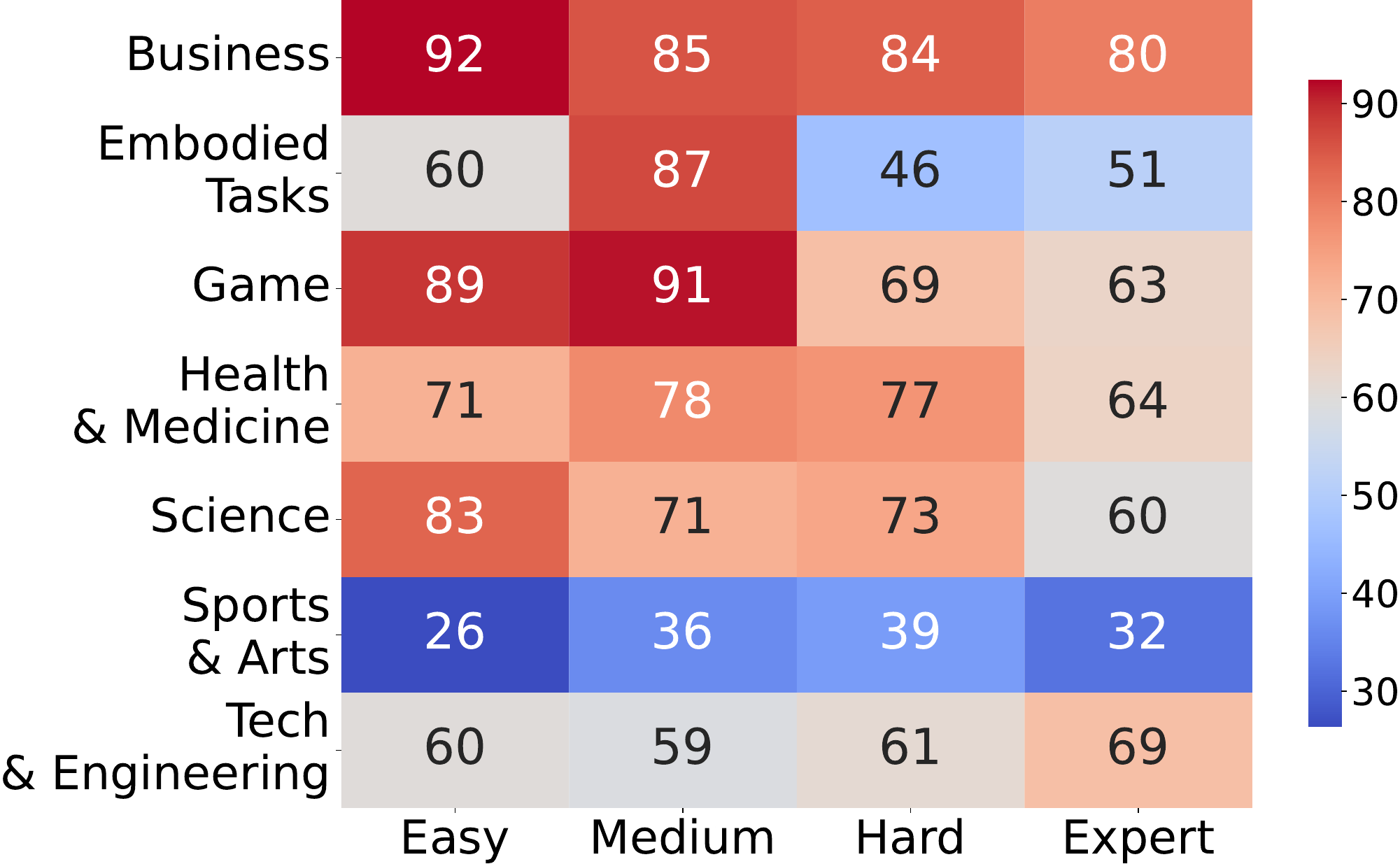}
  \caption{GPT-4V results by disciplines at different difficulty levels for average humans. Average human difficulty levels are defined by 3 turkers' performance per question: Easy (3/3 correct answers), medium (2/3 correct), hard (1/3 correct), and expert (0/3 correct).}
  \label{fig:subfig2}
\end{figure}

Figure~\ref{fig:subfig1} indicate some correlation between the difficulty levels as perceived by humans and the performance of MLLMs.  The difficulty levels are defined based on the \textbf{average human performance}. MLLMs generally follow a trend where accuracy decreases as the difficulty level increases, which aligns with human performance patterns. However, the correlation is not perfect, suggesting that while models and humans share some common ground in understanding question difficulty, there are also notable differences in their capabilities. The data reveals that MLLMs exhibit different skill sets compared to humans. As highlighted in Figure~\ref{fig:subfig2}, models like GPT-4V can correctly answer expert-level questions that humans often get wrong, particularly in disciplines such as Business and Health \& Medicine, where humans often struggle, yet they sometimes falter on easier questions, likely due to the lack of contextual understanding. Notably, discrepancies in disciplines like Art \& Sports and Tech \& Engineering highlight areas where MLLMs’ performance does not align with human results, suggesting different perception, cognition, and reasoning abilities in handling abstract concepts. These differences suggest that MLLMs can complement human capabilities, offering potential for enhanced task performance by combining the data-driven insights of models with human intuition and contextual knowledge.

\noindent \textbf{Error Analysis}
{To gain deeper insights into the limitations of current open-sourced MLLMs and provide guidance for developing next-generation models, we prompted the models to explain their reasoning, particularly when errors occurred. We grouped and identified common error patterns into seven distinct categories. We conducted a comparative test by posing the error-inducing questions for GPT-4V to other MLLMs, as GPT-4V was used as a representative model due to its strong performance and its ability to highlight errors common across MLLMs.} 

{Our analysis revealed that Video-LLaVA exhibited the lowest error frequencies among open-source MLLMs Its superior performance, particularly in reducing Visual Perception Errors (PE), Hallucination Errors (HE), and Reasoning Errors (RE), can also be linked to its use of the CLIP ViT-L/14 model in LanguageBind~\cite{zhu2023languagebind}. In contrast, mPLUG-Owl showed higher rates of Visual Perception Errors, possibly due to its reliance on weaker video embedder architectures. Furthermore, VideoChat outperformed Video-LLaMA due to its GMHRA~\cite{li2023videochat} module for temporal aggregation, demonstrating the importance of effective temporal aggregation in reducing errors. Common trends across all models included frequent hallucination errors and a lack of domain-specific knowledge, highlighting the need for accurate, noise-free training data and suggesting that techniques like Reinforcement Learning from Human Feedback (RLHF)~\cite{rlhf} could help mitigate these issues. While current MLLMs demonstrate strong multi-disciplinary world knowledge, they could benefit from enhanced domain-specific expertise, potentially through retrieval-based methods. Detailed qualitative examples and further analysis are provided in the Appendix.}

\begin{figure}[!t]
    \centering
    \includegraphics[width=\textwidth]{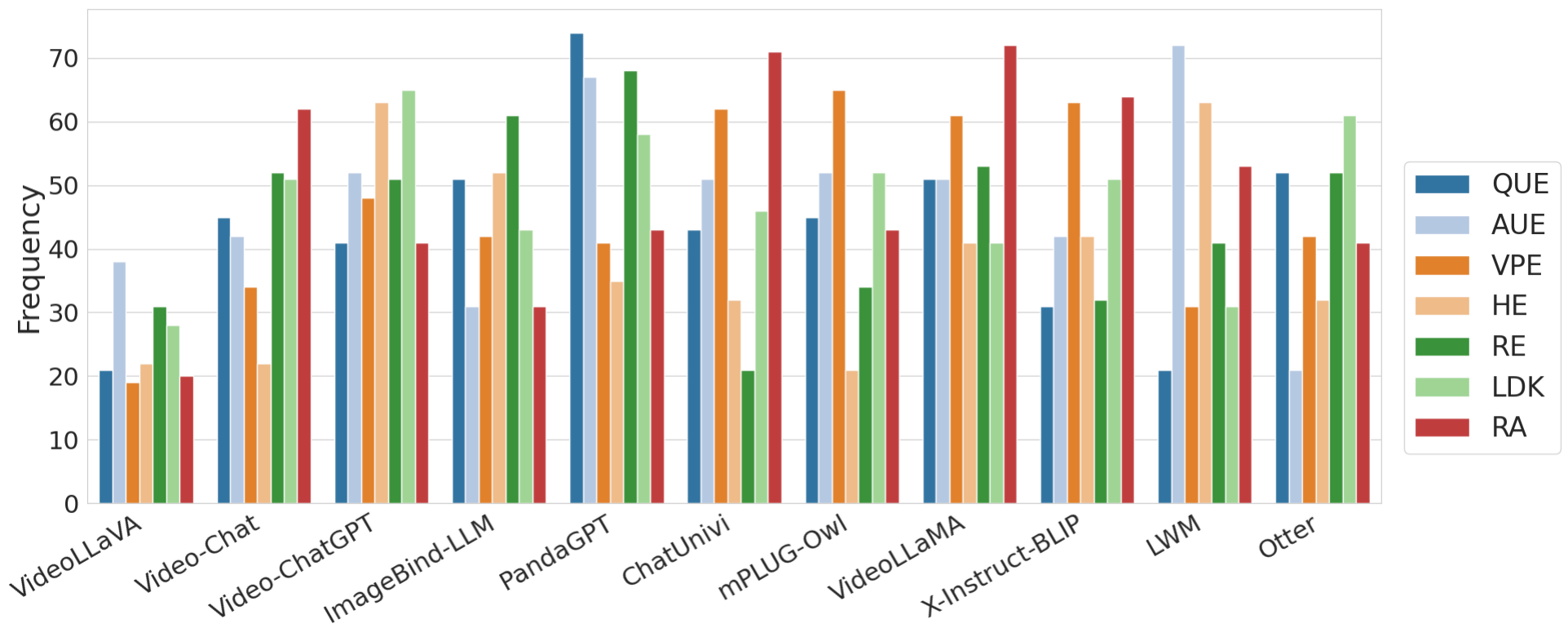}
    \caption{ {The frequency of different error types across various MLLMs. For each error type, 100 examples were evaluated. Error types are abbreviated as follows: QUE (Question Understanding Error), AUE (Audio Understanding Error), VPE (Visual Perception Error), HE (Hallucination Error), RE (Reasoning Error), LDK (Lack of Domain Knowledge), and RA (Reject to Answer).}}
    \label{fig:error_analysis}
\end{figure}

\subsection{Study on Modality of Perception on Synthetic Data}
We conducted ablation studies to evaluate how well MLLMs can perceive the world when limited to a single modality (audio or visual) using the synthetic dataset of \benchmarkname. In these experiments, we isolated scenarios where only one modality—either audio or visual—was available. Table~\ref{audio_only} presents the results, which assess the models' ability to interpret spoken language, background noises, and other audio elements without visual context, as well as their visual perception without any audio input. For the visual perception test, Gemini Pro performed the best, demonstrating its strong ability to process visual information. Interestingly, Video-Chat exhibited better audio perception than ChatUnivi, despite its poorer visual perception. This may be attributed to its use of the Whisper~\cite{whisper} speech recognition model. It also explains that in Table~\ref{maineval}, Video-Chat outperforms ChatUnivi in the Art \& Sports discipline, which requires a greater understanding of music, voice, and background audio. However, in other disciplines such as Science and Health \& Medicine, Video-Chat's performance is significantly worse.

\begin{table}[t]
\centering
\caption{Performance on Synthetic Subset I (Audio) and II (Visual). Synthetic Subset I contains QAs based solely on the audio content, while Synthetic Subset II focuses exclusively on the visual content of the video. We evaluated four MLLMs processing both audio and visual inputs along with Gemini Pro (for the audio setting, only providing the question).}
\label{audio_only}
\resizebox{\textwidth}{!}{
\begin{tabular}{lccccccccccccccccc}
\toprule
\multirow{2}{*}{Model}        
& \multicolumn{2}{c}{Art\&Sports} & \multicolumn{2}{c}{Business} & \multicolumn{2}{c}{Science} & \multicolumn{2}{c}{Health\&Medicine} & \multicolumn{2}{c}{Embodied Tasks} & \multicolumn{2}{c}{Tech\&Engineering} & \multicolumn{2}{c}{Game}& \multicolumn{2}{c}{Average} \\
& Audio & Visual & Audio & Visual & Audio & Visual & Audio & Visual & Audio & Visual & Audio & Visual & Audio & Visual & Audio & Visual\\
\midrule
Random Choice & 31.59 & 30.14 & 31.18 & 26.58 & 36.98 & 32.89 & 38.74 & 32.64 & 32.81 & 31.25 & 27.23 & 32.60 & 32.01 & 30.78 & 32.44 & 30.91 \\   
\hdashline
Video-Chat~\cite{li2023videochat} & \textbf{33.98} & 32.48 & \textbf{46.47} & 41.46 & \textbf{41.86} & 39.15 & \textbf{45.95} & 36.81 & 32.81 & 46.88 & \textbf{37.48} & 35.91 & \textbf{32.98}&46.70&\textbf{38.82} & 39.07 \\
ChatUnivi~\cite{jin2023chatunivi} & 30.03 & 43.22 & 30.19 & 52.85 & 38.75 & 54.59 & 34.76 & 50.69 & 20.14 & 40.63 & 24.17 & 46.41 & 29.98& 45.44 & 31.82 & 48.44 \\
Video-LLaMA~\cite{zhang2023videollama} & 30.15 & 30.23 & 36.18 & 33.17 & 31.33 & 31.34 & 30.90 & 32.78 & \textbf{33.13} & 30.05 & 31.18 & 30.55 & 20.49&27.20&29.08 & 30.47 \\
Otter~\cite{li2023otter} & 14.22 & 16.82 & 16.77 & 14.24 & 16.12 & 17.00 & 19.82 & 13.19 & 10.94 & 12.50 & 15.63 & 12.43 & 6.65 &10.44&12.83 & 13.41 \\
Gemini Pro~\cite{team2023gemini} & 20.88 & \textbf{61.38} & 29.43 & \textbf{77.35} & 30.62 & \textbf{74.26} & 30.14 & \textbf{81.53} & 22.57 & \textbf{70.31} & 18.83 & \textbf{66.22} & 29.96 & \textbf{65.01} & 24.45 &\textbf{69.97} \\
\bottomrule
\end{tabular}}
\end{table}

\section{Related Work}
\label{sec:related work}
\subsection{Multimodal Large Language Models (MLLMs)}
\paragraph{Emerging MLLMs~}
Recent advancements in Large Language Models (LLMs)~\cite{gpt4,google2023bard,touvron2023llama,vicuna2023,touvron2023llama2,bai2023qwen} have paved the way for several multimodal counterparts in the vision-and-language domain~\cite{dai2023instructblip,liu2023visual,liu2023improved,li2023otter,zhu2023minigpt,zheng2023minigpt,bai2023qwenvl}, and recently released GPT-4V~\cite{gpt4-v}, followed by Gemini Vision family~\cite{team2023gemini}. 
As LLMs have been applied to world modeling and simulation~\cite{llm_world_model}, MLLMs now extend their capabilities beyond text and image inputs. Pretrained on large-scale, diverse datasets, these models are equipped with commonsense, domain-specific knowledge, and broad generalizability.

VideoChat~\cite{li2023videochat} leverages the QFormer~\cite{blip2} to map visual representations to LLM~\cite{vicuna2023}, and performs a multi-stage training pipeline. Otter~\cite{li2023otter} proposes to conduct instruction finetuning based on Openflamingo~\cite{awadalla2023openflamingo}. PandaGPT~\cite{su2023pandagpt} employs the ImageBind~\cite{han2023imagebind} as the backbone and finetunes it. The mPLUG-Owl~\cite{ye2023mplug} introduces an abstractor module to perform visual and language alignment. VideoLLaMA~\cite{zhang2023videollama} introduces a frame embedding layer and also leverages ImageBind to inject temporal and audio information into the LLM backend. Chat-UniVi~\cite{jin2023chatunivi} uses clustering to do feature fusion. LWM~\cite{lwm} collects a large video and language dataset from public books and video datasets and trains a world model that is capable of processing more than millions of tokens.  

These MLLMs demonstrate emerging abilities in multi-disciplinary world knowledge and excel at multi-faceted reasoning tasks, such as inverse dynamic prediction—predicting intermediate steps between previous and next states, a crucial auxiliary task for next-state prediction~\cite{devlin2018bert,lu2019vilbert,paster2020planning} in real-world scenarios. In response to the emerging capabilities of MLLMs, we propose~\benchmarkname to evaluate their ability to understand real-world dynamics, underlying principles, and causalities, with the ultimate goal of achieving world modeling.

\paragraph{Benchmarking MLLMs~}
To evaluate MLLMs, there is a flourishing of analysis \cite{liu2023mitigating,zhang2023gpt4vision,comclip,yujie2024wildvisionarena,fan2024muffin,cui2023holistic, guan2023hallusionbench,yu2023mmvet,fu2023mme} and the establishment of innovative benchmarks such as VisIB-Bench~\cite{bitton2023visit} which evaluates models with real-world instruction-following ability given image inputs, MMMU~\cite{yue2023mmmu} designed to access models on college-level image-question pairs that span among different disciplines, and VIM~\cite{lu2023vim} which challenges the model's visual instruction following capability.

However, these recent analyses and benchmarks only cover the image input. Recently, video benchmarks such as Perception Test~\cite{perceptiontest} is proposed to focus on perception and skills like memory and abstraction. However, it uses scenarios with a few objects manipulated by a person, which limits the variety of contexts. In contrast, MMWorld operates in an open-domain scenario with diverse scenes; {MVBench~\cite{mvbench}, TempCompass~\cite{liu2024tempcompass} centers on temporal understanding, while \benchmarkname not only includes temporal reasoning but also evaluates other multi-faceted reasoning abilities such as counterfactual thinking and domain-specific expertise; }EgoSchema~\cite{Egoschema} focuses on natural human activity and behavior, but it does not cover the broad range of disciplines that MMWorld does. MLLMs that can perfectly solve MMWorld would unlock the ability to perform multifaceted, multidisciplinary reasoning and the potential to serve as a world model.

\subsection{Video Understanding Benchmarks}
Previous video benchmarks, as shown in Table~\ref{tab:video-datasets}, focus on video understanding tasks, including activity-focused on web videos~\cite{activityQA}, description-based question answering~\cite{VideoQA}, video completion~\cite{fu2023tvc}, and video infilling~\cite{himakunthala2023lets}. Recently, Video-Bench~\cite{videobench} introduces a benchmark by collecting videos and annotations from multiple existing datasets. Mementos~\cite{wang2024mementos} builds a benchmark for MLLM reasoning for input image sequences. STAR~\cite{star} builds a benchmark for situated reasoning in real-world videos. CLEVER~\cite{clevrer} builds a benchmark containing videos focusing on objects with simple visual appearance. DreamerV3~\cite{dreamerv3} learns a model of the environment and improves its behavior by imagining future scenarios that outperforms specialized methods across over 150 diverse tasks, with a single configuration. None of these benchmarks match the multi-discipline coverage that MMWorld provides. MMWorld, in contrast, presents a new benchmark designed to encompass interdisciplinary coverage, task diversity, and multifaceted reasoning capabilities—including future prediction, counterfactual thinking, and more—underpinned by original human annotations and integrated domain knowledge.

\chapter{Improving Spatiotemporal Awareness in Multimodal World Models}
\section{Introduction}
Humans posses an innate ability to perceive, track and interpret motion, spatial and temporal changes, ~\cite{DeFreitas2016Tracking, Marr1981Directional} enable rich interpretations of complex dynamic events from both egocentric and allocentric perspectives \cite{Burgess2006Spatial}. When observing an object move, we can inherently process any changes such as lateral shifts, rotational directions and periodic or repeated actions unfolding along a specific trajectory \cite{Burgess2006Spatial}. These sophisticated perceptual abilities are a product of our spatiotemporal cognition~\cite{Freyd1984Representational}, and form an essential foundation that allows us to comprehend and reason about physical phenomena, object interactions and causal relationships within our environment. ~\cite{Leslie1984Spatiotemporal, Johansson1973Visual} Vision-language models (VLMs), which can also potentially perceive the motions and spatialtemporal changes in videos, constitute a prominent class of methods designed to emulate or surpass human capabilities in integrated visual and linguistic reasoning~\cite{LeCun1989Backpropagation, Dosovitskiy2021An}. While prior work has focused on static visual understanding from mass training corpuses of language and visual data~\cite{Radford2021Learning} or understanding video such as captioning~\cite{Lu2019ViLBERTPT} and scene understanding~\cite{Chen_2022_CVPR}, we find that the exceptional performance in the prior mentioned tasks does not innately carry over to spatiotemporal capabilities. This limitation is notable given that contemporary state-of-the-art VLMs are typically trained on data sets of the order of hundreds of billions of tokens~\cite{liu2023world}. In contrast, human infants naturally develop robust spatiotemporal cognition within the first few months of life~\cite{Spelke2007Core}. Another key challenge that inhibits VLM performance in spatiotemporal tasks is the necessity to implicitly or explicitly reconstruct a four-dimensional (4D) representation of dynamic scenes and subsequently reason over such reconstruction~\cite{wang2024compositional}. As illustrated in~\cref{fig:teaser}, the car is advancing forwards and turning to the left in its own frame of reference. 

However, from the camera's perspective, its motion appears as a combination of heading to the right and receding into the distance despite the car being in the center of the frame due to camera view rotation. Human observers can seamlessly disentangle these complex dynamics, accurately interpreting trajectories by synthesizing diverse visual cues including camera rotation compensation, stationary scene landmarks, prior knowledge of 3D and 4D environmental structures, and perspective projections~\cite{Marr1981Directional,Burgess2006Spatial,Leslie1984Spatiotemporal,Freyd1984Representational}. The inability of current VLMs to similarly integrate these cues underscores an important gap. Furthermore, bridging this gap will require VLMs to develop more sophisticated mechanisms for reconstructing and reasoning over dynamic scenes, potentially drawing on  insights from cognitive science and neuroscience on how humans process and integrate spatial and temporal information.

\begin{figure}[t]
    \centering
    \includegraphics[width=0.45\textwidth]{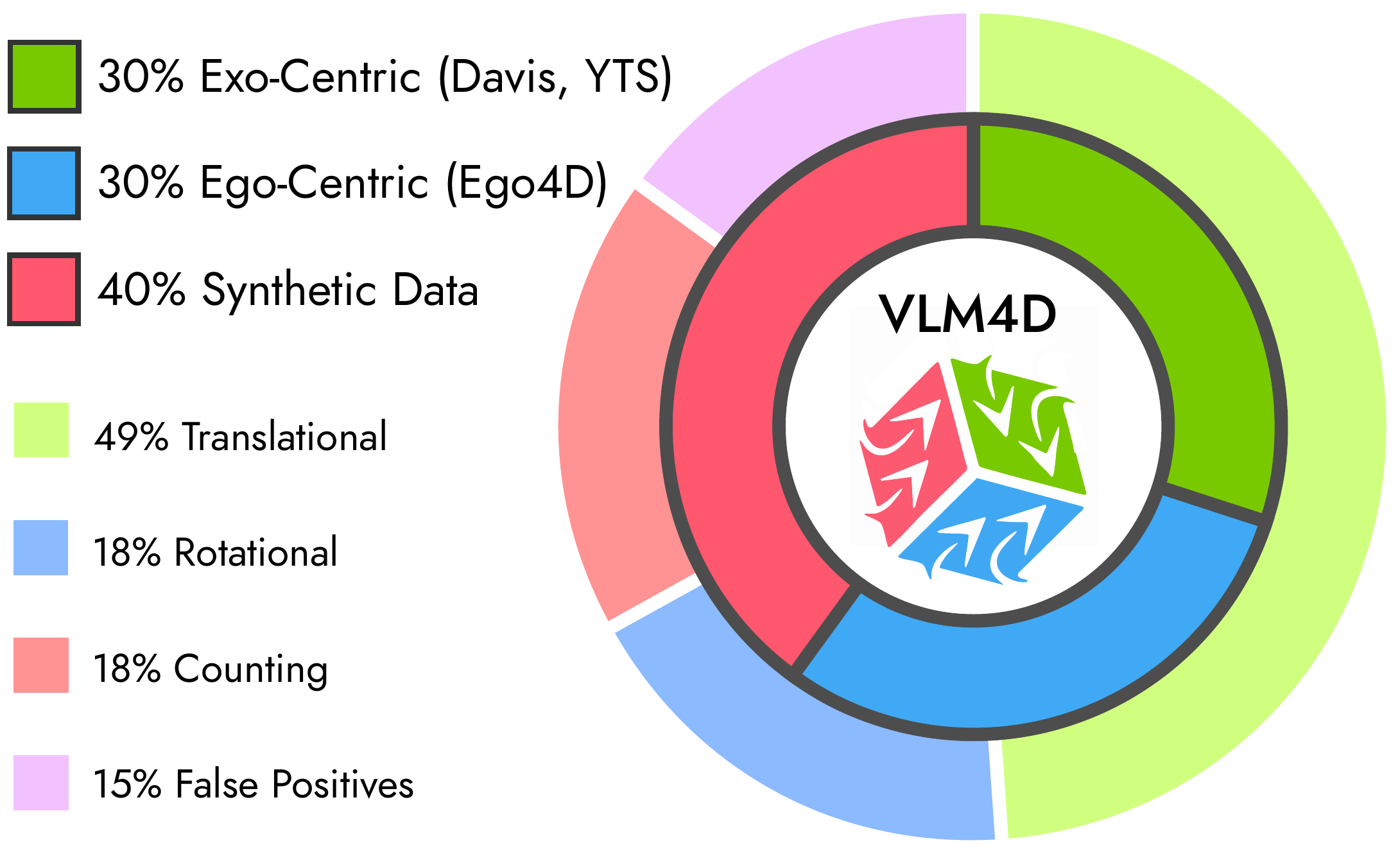}
    \caption{\textbf{Distribution of Dataset Sources and Annotations.} Breakdown of our dataset illustrating the proportions of data sourced from third-person (Davis, YouTube), first-person (Ego4D), and synthetic data, categorized by annotation types: translational, rotational, action, counting, and false positives.}
    \label{fig:data_piechart}
    \vspace{-0.3cm}
\end{figure}

\begin{figure*}[ht]
    \centering
    \includegraphics[width=1\textwidth]{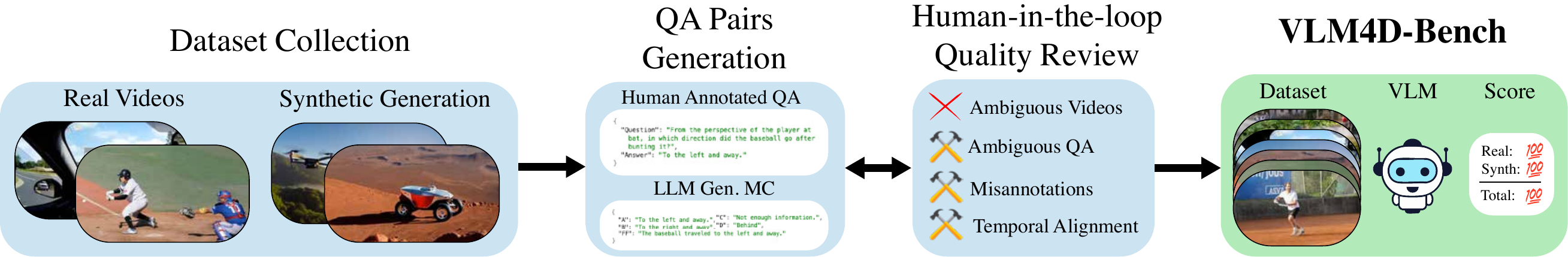}
    \caption{\textbf{Dataset Generation and Annotation Pipeline.} Our dataset was constructed by collecting real videos and generating synthetic data, followed by human-in-the-loop quality reviews to address ambiguous videos and annotations. After temporal alignment and quality assurance, human-annotated questions and answers were created, complemented by multiple-choice questions generated by large language models (LLMs). The final dataset includes real-world and synthetic video data with comprehensive VLM scoring metrics.}
    \label{fig:dataset_pipeline}
    \vspace{-0.4cm}
\end{figure*}

In order to effectively characterize and challenge the existing spatiotemporal reasoning abilities of VLMs, we directly evaluate their capacity to track complex directional movements and perspective transformations over time. We introduce \texttt{VLM4D}, a rigorous benchmark specifically designed to probe the spatiotemporal grounding capabilities of current vision-language models. Through this contribution, we aim to catalyze research that addresses the critical gap in spatiotemporal understanding and reasoning within VLMs and provide a foundational analysis highlighting key deficiencies in existing models.

We introduce \texttt{VLM4D}, the first benchmark specifically designed to test the spatiotemporal reasoning abilities of VLMs. \texttt{VLM4D} consists of 1,000 videos paired with over 2,000 question-answer pairs, each carefully designed to assess both spatial and temporal understanding jointly. The majority of these videos are sourced from datasets with rich spatiotemporal characteristics, thus ensuring a diverse range of motion-related scenarios. We augment the dataset with synthetic videos generated by a world-foundation model, Cosmos~\cite{agarwal2025cosmos}, that has been modified using techniques introduced in~\cite{he2024mojito} to obtain more accurate correspondence between motion-oriented prompts and the resulting generated video. Figure~\ref{fig:data_piechart} illustrates the composition of our dataset.

\subsection{Benchmark Construction}
Unlike prior work that often relies heavily on LLMs and VLMs to generate first iterations of benchmarks and datasets~\cite{chen2024sharegpt4video} followed by human quality control - we found that existing VLMs and automated methods showed significant limitations in terms of realiability and quality. This shortcoming necessitated direct human annotations that were then followed by augmentation by LLMs to ensure a high-quality benchmark. An overview of the benchmark curation pipeline is shown in \cref{fig:dataset_pipeline}. 

\paragraph{Real Video Data Collection} Real-world videos were sourced from datasets with rich spatiotemporal characteristics that ensured diverse motion and perspective variations. For egocentric data, we relied mainly on the Ego4D dataset~\cite{grauman2022ego4d}, while most exocentric data points were collected from the Davis~\cite{davis2017} and YouTube-VOS~\cite{xu2018youtube} datasets. To minimize confounders and to focus attention of VLM abilities to only spatiotemporal reasoning, we preprocessed the videos by temporally segmenting and centering them around the most relevant action, thus resulting in videos with an average duration of $5$-$15$ seconds. This ensures that the key event described in the question is clear and reduces ambiguities or confounders that would reduce VLM accuracy. 

\paragraph{Synthetic Video Generation}  
For synthetic video generation, we use Cosmos~\cite{agarwal2025cosmos} as our video generation backbone. To ensure that the generated videos align with the intended object moving directions, we incorporate input bounding boxes as additional spatial guidance. Specifically, we follow the approach introduced in~\cite{he2024mojito} modifying the diffusion forward steps to enforce object localization constraints at each timestep, ensuring consistency between the generated object direction and the user-specified trajectory. The average duration of generated synthetic videos is $5$ seconds. To maintain high-quality outputs, we perform a manual verification step after generation, filtering out low-quality videos and retaining only those that accurately match the specified directions. Once a video is generated, we use an LLM (GPT-4o) to generate two types of questions for evaluation: Direct questions, which are derived directly from the textual prompt used to generate the video; Counterfactual questions, which involve querying about non-existent objects in the generated scene. Both question types follow the format: ``What direction is the $\langle$Object Name$\rangle$ moving?", where the model must select one of four possible answers: ``left", ``right", ``not moving", or ``no $\langle$Object Name$\rangle$ there."

\begin{figure*}[t]
    \centering
    \includegraphics[width=0.8\textwidth]{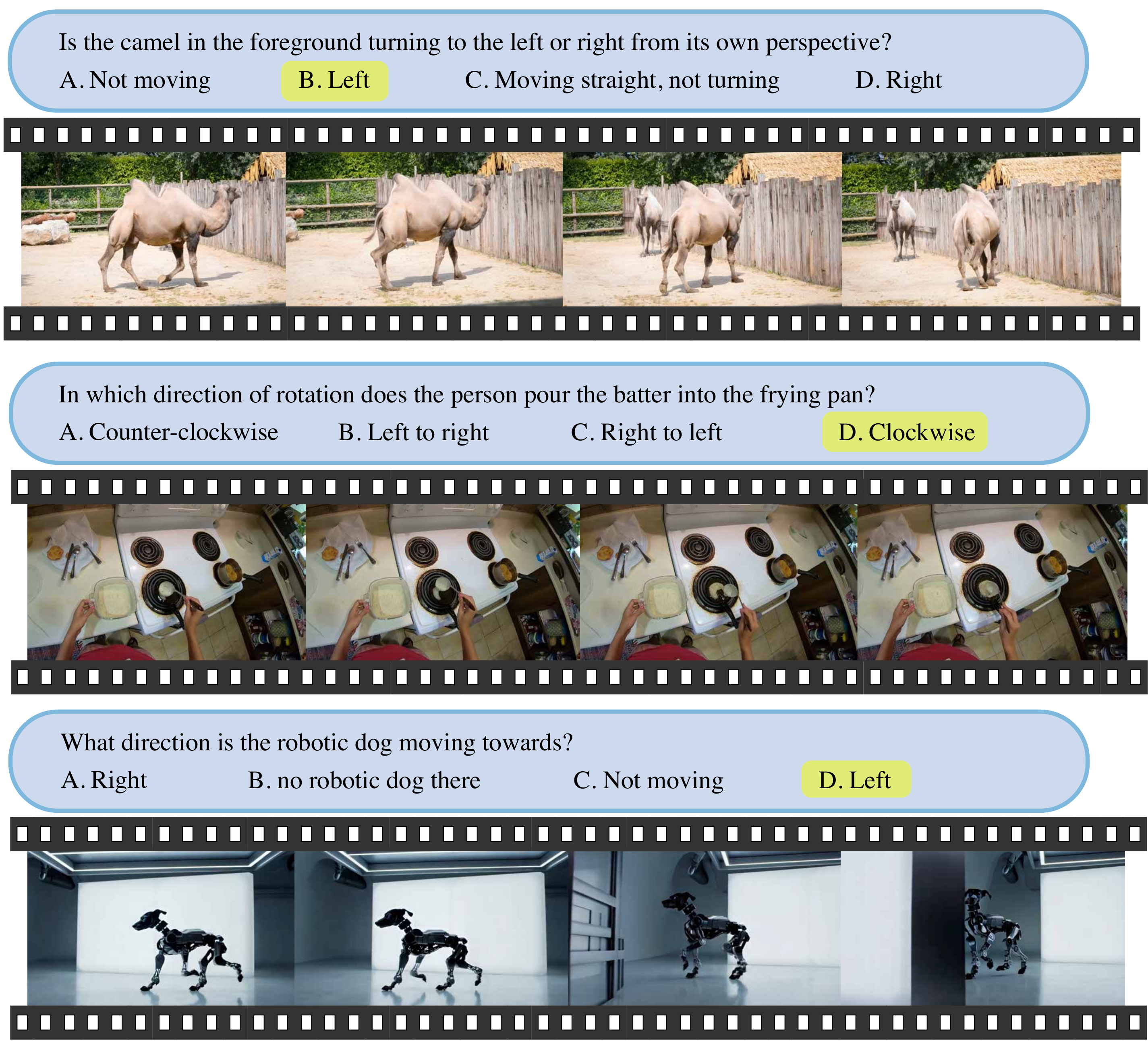}
    \caption{\textbf{Qualitative Examples of Dataset Annotations.} (Top) A third-person video with translational annotations (``camel turning left from its perspective"). (Middle) A first-person video with a rotational question (``clockwise rotation of ladle"). (Bottom) A synthetic scene with action recognition ``robotic dog moving left"). }
    \label{fig:qualitative_examples}
    \vspace{-0.5cm}
\end{figure*}

\paragraph{QA Generation and Quality Control} Question-answer pairs are primarily constructed through human annotations. The question answer pairs are then supplemented with alternative answers by an LLM (GPT-4o) for multiple choice (MC) questions. To ensure high-quality annotations, a rigorous human verification process was applied where ambiguous videos were filtered out and vague, misleading, or incorrect QA pairs were refined to allow for spatial and temporal alignment between the language and visual content. Figure~\ref{fig:qualitative_examples} showcases some qualitative examples of annotations for different types of videos.

\paragraph{Assessing Human Performance}
To establish a human performance baseline on our benchmark, we conducted an evaluation in which participants independently answered 100 randomly sampled questions from the dataset. The accuracy of human responses was then aggregated to approximate the performance of human spatiotemporal reasoning on the dataset. 

\subsection{Categorizing Spatiotemporal Performance}
To systematically evaluate spatiotemporal reasoning capabilities, we first categorize videos into two primary groups: egocentric (first-person) videos and exocentric (third-person) videos. Egocentric videos are sourced from the Ego-4D~\cite{grauman2022ego4d} dataset where scenes are captured from a head-mounted camera, thus offering dynamic video data that is inherently coupled with the individual's actions. Exocentric videos encompass a diverse range of recorded scenes, from sports footage to everyday scenes. Beyond this categorization, we also evaluate spatiotemporal performance across four dimensions: translational movement (TM), rotational movement (RM), spatiotemporal counting (STM), and false positives (FP). Translational movement assesses a model's ability to track linear motion within scenes, while rotational movement assesses the understanding of changes in orientation and perspective shifts over time. Spatiotemporal counting extends these core motion-based tasks by requiring a more complex reasoning strategy to determine the number objects performing a translation or rotational movement. Lastly, the false positives category measures the model's reliability in recognizing whether any motion took place. By structuring the benchmark along these axes, we aim for a comprehensive framework for assessing spatiotemporal reasoning (Figure~\ref{fig:radar_plot}).

\begin{table*}[h]
    \centering
    \resizebox{1.0\linewidth}{!}{
    \renewcommand{\arraystretch}{1.2}
    \arrayrulecolor{black} 
    \begin{tabular}{l l l c c c c c c c}
    \toprule
    \textbf{Organization} & \textbf{Model} & \textbf{Release} & \multicolumn{3}{c}{\textbf{Real}} & \multicolumn{3}{c}{\textbf{Synthetic}} & \textbf{Overall} \\
    \cmidrule(lr){4-6}
    \cmidrule(lr){7-9}
    & & & \textbf{Ego-centric} & \textbf{Exo-centric} & \textbf{Average} & \textbf{Directional} & \textbf{FP} & \textbf{Average} \\
    \midrule
         User Study & Human Performance &  & 99.6 & 99.7 & 99.7 & 91.8 & 100 & 95.9  & 98.3     \\
         Random & Random Selection  &  & 24.4 & 23.2 & 23.6 & 25.5 & 24.7 & 25.1 & 24.2    \\

        \midrule
        \rowcolor[HTML]{e1f0f5}\multicolumn{10}{l}{\textbf{Latest Proprietary VLMs}} \\
        \midrule
        OpenAI & GPT-4o & 2024-11 & \cellcolor[HTML]{FFCC99}{54.3} & \cellcolor[HTML]{FFCC99}{61.2} & \cellcolor[HTML]{FFCC99}{58.9} & 47.8 & 47.5 & 47.7 & \cellcolor[HTML]{FFCC99}{56.2} \\
        \arrayrulecolor{lightgray} \hdashline
        Google & Gemini-2.5-Pro & 2025-3 & \cellcolor[HTML]{FFB366}{\textbf{ 68.2}} & \cellcolor[HTML]{FFB366}{\textbf{70.5}} & \cellcolor[HTML]{FFB366}{\textbf{69.7}} & \cellcolor[HTML]{FFB366}{\textbf{71.3}} & \cellcolor[HTML]{FFCC99}{75.0} & \cellcolor[HTML]{FFB366}{\textbf{71.6}} & \cellcolor[HTML]{FFB366}{\textbf{70.2}} \\
        & Gemini-2.0-Pro & 2025-2 & 44.8 & 50.5 & 48.7 & 42.8 & 52.5 & 43.6 & 47.4     \\
        \hdashline
        xAI & Grok-2-Vision & 2024-12 & 44.1 & 48.8 & 47.3 & 49.0 & \cellcolor[HTML]{FFCC99}{75.0} & 51.4 & 48.3 \\
        \arrayrulecolor{black} \midrule
        \rowcolor[HTML]{e1f0f5}\multicolumn{10}{l}{\textbf{Open-source Image VLMs}} \\
        \midrule
        Meta 
        & Llama-4-Maverick-17B & 2025-4 & 48.8 & 52.9 & \cellcolor[HTML]{FFE5CC}{51.6} & \cellcolor[HTML]{FFCC99}{56.0} & 50.0 & \cellcolor[HTML]{FFE5CC}{55.5} & \cellcolor[HTML]{FFE5CC}{52.5} \\
        & Llama-4-Scout-17B & 2025-4 & 46.6 & 51.3 & 49.7 & 46.8 & 45.0 & 46.6 & 49.0 \\
        & Llama-3.2-90B-Vision & 2024-9 & 37.4 & 42.4 & 40.8 & 28.0 & \cellcolor[HTML]{FFB366}{\textbf{85.0}} & 33.2 & 38.9 \\
        & Llama-3.2-11B-Vision & 2024-9 & 35.2 & 36.1 & 35.8 & 38.3 & 62.5 & 40.5 & 36.9 \\
        \arrayrulecolor{lightgray} \hdashline
        Microsoft & Phi-4-Multimodal & 2025-3 & 39.9  & 36.0 & 37.3 & 34.8 & 2.5 & 31.8 & 36.0 \\
        & Phi-3.5-Vision & 2024-7 &  36.3 & 39.1 & 38.2 & 26.5 & 42.5 & 28.0 & 35.7 \\
        \arrayrulecolor{lightgray} \hdashline
        DeepSeek &  DeepSeek-VL2-Tiny & 2024-12 &  31.4 & 32.5 & 32.2 & 42.8 & 15.0 & 40.2 & 34.1 \\
        \arrayrulecolor{lightgray} \hdashline
        Shanghai AI Lab & InternVL2.5-38B & 2024-11 & 42.8  & \cellcolor[HTML]{FFE5CC}{53.2} & 49.7 & 37.5 & 62.5 & 39.8 & 47.3 \\
        & InternVL2.5-8B & 2024-11 & 40.8 & 41.1 & 41.0 & 40.8 & 55.0 & 42.1 & 41.3 \\
        & InternVL2-8B & 2024-6 & 33.2 & 38.2 & 36.5 & 34.8 & \cellcolor[HTML]{FFE5CC}{72.5} & 38.2 & 36.9 \\
        \arrayrulecolor{lightgray} \hdashline
        Mistral AI & Pixtral-12B & 2024-9 & 36.3 & 32.9 & 34.0 & 41.0 & 17.5 & 38.9 & 35.2  \\
        \arrayrulecolor{lightgray} \hdashline
        Rhymes & Aria & 2024-11 &  42.3 & 44.0 & 43.5 & 35.3 & 57.5 & 37.3 & 42.0  \\
        \arrayrulecolor{lightgray} \hdashline
        HuggingFaceM4 & Idefics3-8B & 2024-8 & 34.3 & 36.2 & 35.6 & 33.5 & 42.5 & 34.3 & 35.3 \\
        \arrayrulecolor{lightgray} \hdashline
        H2O & H2OVL-Mississippi-2B & 2024-10 &  37.0 & 33.3 & 34.5 & 27.3 & 27.5 & 27.3 & 32.7 \\
        \arrayrulecolor{black} \midrule
        \rowcolor[HTML]{e1f0f5} \multicolumn{10}{l}{\textbf{Open-source Video VLMs}} \\
        \midrule
        Alibaba & Qwen2.5-VL-7B & 2025-1 & 42.3 & 45.0 & 44.1 & 39.3 & 55.0 & 40.7 & 43.3 \\
        & Qwen2.5-VL-72B-AWQ & 2025-1 & 49.9 & 48.7 & 49.1 & \cellcolor[HTML]{FFE5CC}{54.3} & \cellcolor[HTML]{FFCC99}{75.0} & \cellcolor[HTML]{FFCC99}{56.1} & 50.8 \\
        & Qwen2-VL-7B & 2024-8 & 36.1 & 38.2 & 37.5 & 38.5 & 37.5 & 38.4 & 37.7 \\
        & Qwen2-VL-72B-AWQ & 2024-9 & 43.0 & 46.2 & 45.2 & 43.8 & \cellcolor[HTML]{FFCC99}{75.0} & 46.6 & 45.5 \\
        \hdashline
        DAMO & VideoLLama3-2B & 2025-1 & 48.6 & 43.7 & 45.3 & 29.0 & 60.0 & 31.8 & 42.2 \\
        & VideoLLama3-7B & 2025-1 & 47.4 & 45.0 & 45.8 & 39.5 & 60.0 & 41.4 & 44.7 \\
        & VideoLLama2.1-7B & 2024-10 & 43.0 & 36.0 & 38.2 & 31.5 & 40.0 & 32.3 & 36.8 \\
        & VideoLLama2-7B & 2024-6 & 36.3 & 16.5 & 23.0 & 25.8 & 37.5 & 26.8 & 23.9 \\
        \hdashline
        OpenGVLab & InternVideo2.5-8B & 2025-1 &  \cellcolor[HTML]{FFE5CC}{52.8} & 50.1 & 51.0 & 45.3 & 32.5 & 44.1 & 49.3 \\
        & InternVideo2-8B & 2024-8 & 37.2 & 37.9 & 37.6 & 40.5 & 0.0 & 36.8 & 37.4 \\
        \hdashline
        LLaVA & LLaVA-One-Vision-7B & 2024-9 & 32.5 & 33.1 & 32.9 & 32.8 & 45.0 & 33.9 & 33.1  \\
        & LLaVA-NeXT-Video-7B & 2024-6 & 30.3 & 30.9 & 30.7 & 24.5 & 25.0 & 24.6 & 29.2 \\
        & LLaVA-NeXT-Video-34B & 2024-6 & 37.2 & 34.9 & 35.7 & 31.5 & 60.0 & 34.1 & 35.3  \\
        \arrayrulecolor{black} \bottomrule
    \end{tabular}
    }
    \caption{\textbf{Evaluation on \texttt{VLM4D} Benchmark} across various proprietary and open-source VLMs. Top three performers in each category are highlighted from \colorbox[HTML]{FFB366}{dark} (highest) to \colorbox[HTML]{FFE5CC}{light} (third highest).}
\label{tab:result}
\end{table*}

\section{Evaluation of Spatiotemporal Awareness}

\subsection{Evaluation Setup}
\paragraph{Benchmark Models}
We evaluate over 10 of the most recently released VLMs thus covering a wide range of model sizes, architectures, and training methodologies. For open-source models, we include Llama-3.2-Vision~\cite{grattafiori2024llama}, DeepSeek-VL~\cite{lu2024deepseek}, InternVL2.5~\cite{chen2024expanding}, Pixtral~\cite{agrawal2024pixtral}, Aria~\cite{li2024aria}, Idefics~\cite{laurenccon2023obelics}, H2OVL~\cite{galib2024h2ovl}, Qwen2-VL~\cite{wang2024qwen2}, Qwen2.5-VL~\cite{yang2024qwen2}, VideoLLama2~\cite{cheng2024videollama}, VideoLLama3~\cite{zhang2025videollama}, Llava-One-Vision~\cite{li2024llava}, Llava-NeXT-Video~\cite{zhang2024llavanextvideo}, InternVideo2~\cite{wang2024internvideo2}, and InternVideo 2.5~\cite{wang2025internvideo2}. When available, we evaluate different sizes for each model, resulting in models ranging from 2 to 72 billion parameters. For closed-source VLMs, we evaluate GPT-4o~\cite{gpt4o}, Gemini 2.0 Pro~\cite{team2024gemini}, and Grok-2-Vision. 

\paragraph{Evaluation Settings}
The evaluations were performed in a zero-shot setting with video or a set of sampled frames of video followed by the prompt forming the input. For each model, we evaluate on two different inference settings. In the first setting, the model is directed to produce the answer immediately without any reasoning (DO), and in the second evaluation setting, the model is directed to create intermediate reasoning steps, Chain of Thought (CoT)~\cite{wei2022chain}, before inferring the final answer. Additional details about the evaluation setup and prompts are provided in the appendix.

\paragraph{Metrics} Following prior work~\cite{yang2024thinking} and given the nature of our target task, we use multiple-choice questions for evaluation. The primary metric is accuracy on the multiple choice questions (MCQ). Given the two inference settings mentioned previously, we employ LLM-as-Judge following~\cite{zhao2025mmvu} to grade the VLMs' outputs. LLM-as-Judge was utilized instead of performing string or template matching as we found that especially during CoT, various VLMs may output all possible answers during the reasoning process in varying frequencies and with slight modifications to the format of the possible answer choices in MCQ. Each MCQ contains four possible answers.

\subsection{Benchmark Results}
\begin{figure}
    \centering
    \includegraphics[width=0.5\linewidth]{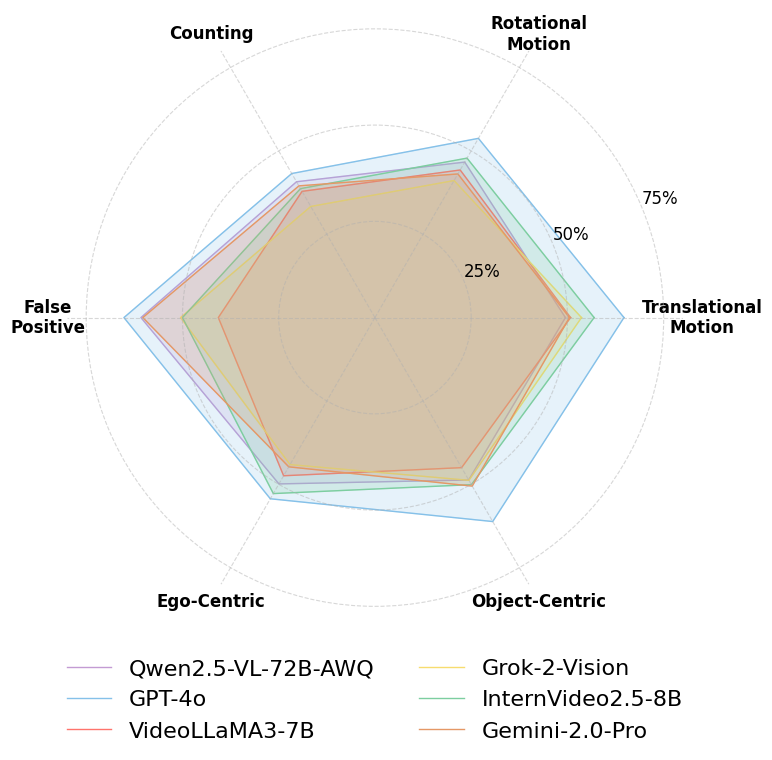}
    \caption{\textbf{Comparison of accuracy across types of spatiotemporal questions.} Model accuracy is shown only for the six top-performing VLMs.}
    \label{fig:radar_plot}
    \vspace{-0.3cm}
\end{figure}

\paragraph{VLMs Performance} The evaluation results in~\cref{tab:result} reveal several critical insights regarding the spatiotemporal reasoning capabilities of contemporary VLMs on the \texttt{VLM4D} benchmark. First, proprietary VLMs, particularly OpenAI’s GPT-4o, consistently outperform open-source models across nearly all real-world categories, highlighting the performance gap between closed-source and publicly available VLMs. Among open-source models, InternVideo2.5-8B and Qwen2.5-VL-72B-AWQ emerge as notable contenders, with Qwen2.5-VL-72B-AWQ achieving exceptional results on synthetic data, surpassing even GPT-4o. However, all models significantly trail behind human-level performance, emphasizing substantial room for improvement, especially in nuanced spatiotemporal reasoning. These findings underscore a critical gap in current VLM architectures, reinforcing the need for further research into structured 4D scene representations and improved spatiotemporal grounding strategies. We additionally show in ~\cref{fig:radar_plot} for the top-performing models their strengths and weaknesses in the fine-grained categories mentioned in the previous section. As expected, translational motion performs best, followed by rotational motion and spatiotemporal counting. 

\paragraph{Human Level Performance}
We use \textbf{Prolific}, an online platform designed to connect academic researchers with user research participants for human-level performance evaluation. The participants are English-speaking random users verified by this platform without prior knowledge of computer vision. We asked 51 candidates to answer the spatial awareness questions in our benchmark. Each question has four choices, and the user may select only one correct answer. We collect their answers and report the average precision in Table.~\ref{tab:result}

\section{Analysis: Why Existing Multimodal Foundation Models Don't Work Well?}

\subsection{Limited Spatiotemporal Cognition}
\begin{figure}
    \centering
    \includegraphics[width=0.5\linewidth]{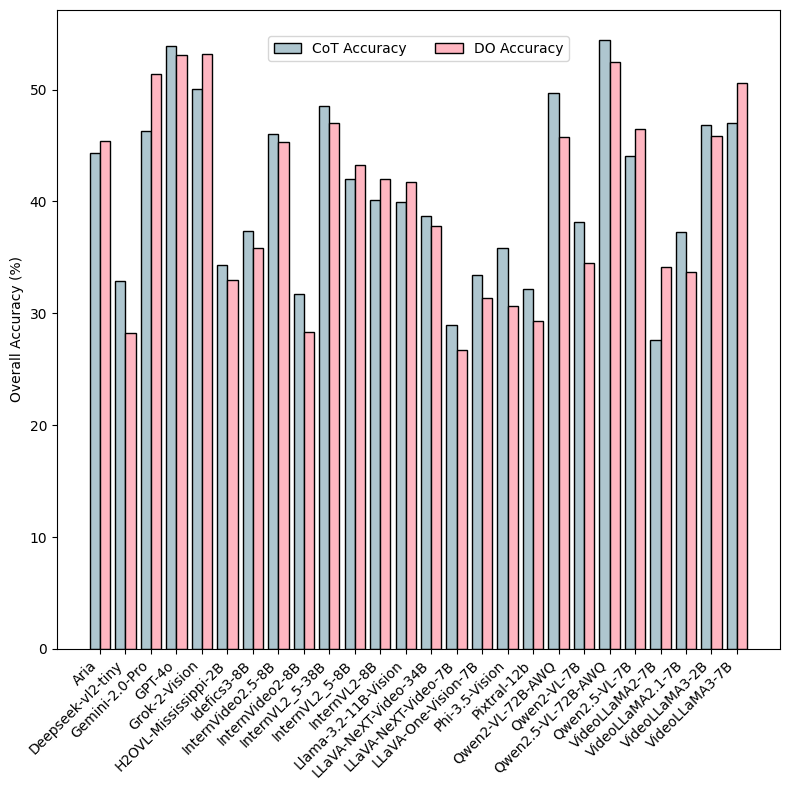}
    \caption{\textbf{Comparison of CoT and DO Accuracy Across Models.} Accuracy comparison between Chain-of-Thought (CoT) and Direct Output (DO) prompting across VLMs.}
    \label{fig:cot_vs_do}
    \vspace{-0.5cm}
\end{figure}

Despite significant advances in VLMs, their ability to understand and reason about motion, spatial relationships, and temporal coherence remains fundamentally underdeveloped ~\cite{chen2024we, zohar2024apollo}. Chain of Thought (CoT)~\cite{wei2022chain} is widely employed as a method to improve accuracy through step-by-step reasoning. We showcase a comparison between CoT and DO in ~\cref{fig:cot_vs_do}. Overall, there is no indication of a large advantage of CoT over all evaluated models. Upon deeper exploration of the CoT reasoning of some models, we observe that the reasoning process was primarily flawed in the following ways: irrelevant information and arriving at conclusions that are inconsistent with the reasoning process. Larger models exhibited strategies that would be similar to how a human processes spatiotemporal information, but the resulting execution falls short of human performance. This demonstrates a disconnect between its visual and linguistic knowledge. We provide examples of this behavior in the appendix. 

\subsection{Deficiencies in Spatiotemporal Labeling}
\begin{figure}[b]
    \centering
    \includegraphics[width=0.5\textwidth]{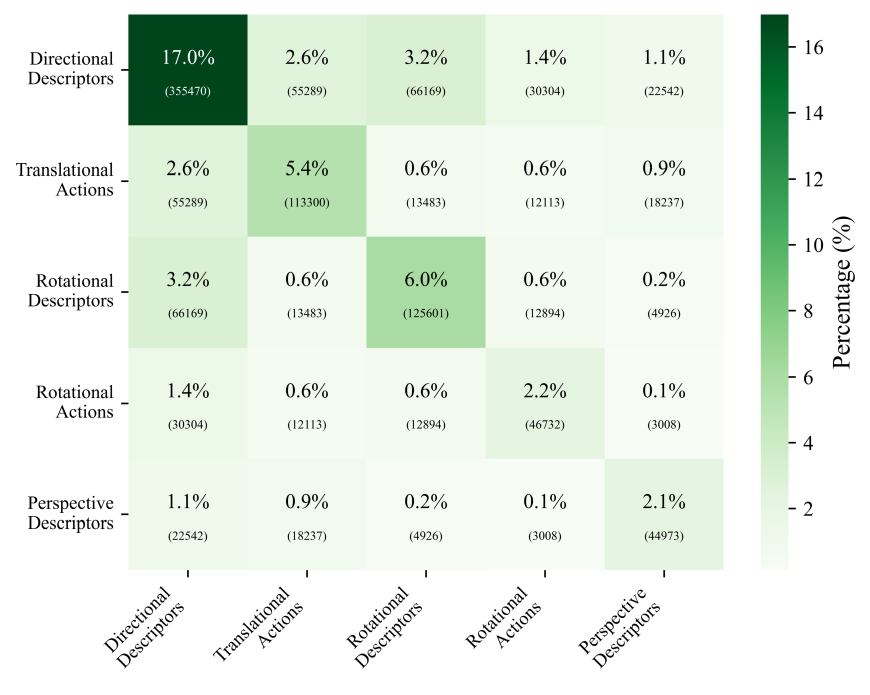}
    \caption{\textbf{Heatmap of Occurances of Spatial-Temporal Terms in popular video SFT datasets.}}
    \label{fig:sft_ds}
\end{figure}

Another avenue of exploration we undertook is to understand the richness of spatiotemporal labels in popular SFT VLM datasets. Typically, video captioning occurs at the `scene' level, lacking fine-grained temporal, spatial, and object-level details. We performed an extensive analysis, encompassing over 2 million samples~\cite{chen2024sharegpt4video, cui2025comprehensive, li2023videochat, li2023mvbench, zhang2024llavanext-video}. We performed this analysis through string-matching of spatiotemporal descriptors related to directionality, translational motion, rotation, and perspective shifts and provide the overall results in ~\cref{fig:sft_ds}. We then performed a manual finegrained evaluation of the ShareGPT4Video dataset~\cite{chen2024sharegpt4video} which we found had the highest density of spatiotemporal datasets. We found that from a sample of 100 labels that were detected as spatiotemporal, less than 10\% of them were judged as accurate upon human evaluation. This result underscores the inadequacy of current dense captioning approaches, which frequently generate spatiotemporal descriptors without capturing precise motion dynamics. We provide more detailed analysis and explanations in the supplement.

\section{Probing Future Solutions}
To probe promising future solutions for enhancing spatiotemporal video understanding, we propose two approaches that address some of the shortcomings of current state-of-the-art VLMs: fine-tuning a VLM on data-rich in spatiotemporal actions and the other leveraging 4D reconstruction and feature fields jointly with a VLM. SFT refines the model’s abilities by training on datasets that contain temporally and spatially rich actions and interactions. By integrating structured visual representations and targeted fine-tuning, these approaches enhance video-language models’ ability to interpret motion. The second method lifts the feature space of VLMs into a temporally coherent 4D feature field, providing structured scene representations that improve motion and spatial reasoning in the stage of decoding and inference.

\paragraph{Spatial-Temporal SFT}
We evaluate on a subset split of the real dataset by splitting the real-world dataset into a training and testing split (80\% / 20\%) and we try settings using synthetic/real/both for training. We conducted the experiments using Qwen 2VL (7B) and Qwen 2.5VL (7B) through LLama-Factory~\cite{zheng2024llamafactory}, and compared the performance before and after supervised fine-tuning in~\cref{tab:sft}. The results demonstrated an improvement in accuracy in spatiotemporal reasoning, suggesting that performance gains can be obtained through targeted training. However, the addition of synthetic data does not necessarily increase performance over using real data alone, suggesting the importance of synthetic data quality.

\paragraph{4D Feature Fields Reconstruction} Recent advances in 3D/4D reconstruction methods, such as Feature4X~\cite{zhou2025feature4x}, have significantly enhanced Vision-Language Model (VLM) performance on visual question answering (VQA) tasks by integrating structured 4D scene representations into the model’s inference stage. Inspired by these promising results, we investigate incorporating spatiotemporal awareness into the InternVideo2-8B model~\cite{wang2024internvideo2}, employing the 4D feature lifting strategy proposed by Feature4X. To assess this approach, we evaluate performance on a subset of the \texttt{VLM4D} benchmark, specifically leveraging all 50 videos from the DAVIS 2016 dataset~\cite{perazzi2016benchmark}.
Our experimental evaluation compares the inference results across three distinct input modalities: original 2D videos, reconstructed global-view RGB videos (4D), and reconstructed global semantic feature fields. As demonstrated in Table~\ref{tab:4D_recon_results}, the highest accuracy consistently results from the reconstructed semantic feature fields, highlighting the clear advantages of structured 4D representations. These findings confirm that global 4D feature field reconstruction enhances contextual understanding and mitigates artifacts associated with RGB rendering during reconstruction. However, the current approach requires per-scene optimization as a post-processing step, limiting its generalizability and making it computationally intensive.

\begin{table}[t]
    \centering
    \renewcommand{\arraystretch}{1.2}
    \begin{tabular}{l|cc}
        \toprule
        \textbf{Model} & \textbf{FF} & \textbf{MC} \\
        \midrule
        \rowcolor[HTML]{e1f0f5} \textit{Original Model} & \phantom{-} & \phantom{-} \\
        Qwen 2VL (7B) & 31.9 & 38.3 \\
        Qwen 2.5VL (7B) & 31.6 & 43.4 \\
        \midrule
        \rowcolor[HTML]{e1f0f5} \textit{Finetuned Model} & & \\
        Qwen 2VL (7B) (R) & \textbf{50.7} & 53.5 \\
        Qwen 2VL (7B) (S) & 38.9 & 41.0 \\
        Qwen 2VL (7B) (R+S) & 49.7 & 52.8 \\
        Qwen 2.5VL (7B) (R) & 48.9 & \textbf{56.3} \\
        Qwen 2.5VL (7B) (S) & 35.4 & 42.0 \\
        Qwen 2.5VL (7B) (R+S) & 39.2 & 48.3 \\
        \bottomrule
    \end{tabular}
    \caption{\textbf{SFT on Spatial-Temporal Datasets.} MC and FF refer to multiple-choice and freeform accuracy, while R, S, and R+S denote real, synthetic, and both real+synthetic usage of data.}
    \label{tab:sft}
\end{table}

\begin{table}[t]
    \centering
    \renewcommand{\arraystretch}{1.2}
    \begin{tabular}{l|cc}
        \toprule
        \textbf{Input Modality} & \textbf{Accuracy} \\
        \midrule
        \rowcolor[HTML]{e1f0f5} \textit{Chain of Thought Response} & \phantom{-} & \phantom{-} \\
        Original 2D Video & 36.0 \\
        Global View Video & 32.7 \\
        Global Feature Field & \textbf{37.4} \\
        \midrule
        \rowcolor[HTML]{e1f0f5} \textit{Direct Output Response} & & \\
        Original 2D Video & 24.3 \\
        Global View Video & 23.8 \\
        Global Feature Field & \textbf{29.0} \\
        \bottomrule
    \end{tabular}
    \caption{\textbf{InternVideo2 Accuracy with 4D Reconstruction.} Comparison of InternVideo2 accuracy given different input modalities from the same dataset.}
    \label{tab:4D_recon_results}
    \vspace{-0.3cm}
\end{table}

\section{Related Work}
\label{sec:related_work}

\paragraph{Spatiotemporal Understanding in Vision Language Models}%
Vision Language Models (VLMs) have evolved rapidly by fully leveraging the significant achievements of Large Language Models (LLMs)~\cite{brown2020language, devlin2019bert, wei2021finetuned, bai2023qwen, touvron2023llama, radford2018improving} and large-scale visual instruction tuning datasets~\cite{liu2023visual, zhu2023minigpt, dai2023instructblip}. While VLMs~\cite{gong2023multimodal, liu2023visual, zhu2023minigpt, abouelenin2025phi4minitechnicalreportcompact, hurst2024gpt, li2024llava, team2024gemini, wang2024qwen2} exhibit transformative potential for applications such as embodied AI~\cite{suglia2024alanavlm, driess2023palm, kim2024openvla}, robotics~\cite{wang2024vlm, patel2025real}, and world modeling~\cite{liu2024world, zhang2024combo}, most existing methods remain constrained to static images, focusing narrowly on spatial understanding while overlooking the dynamic temporal dimension inherent in real-world interactions. To bridge this gap, emerging research~\cite{li2023videochat, zhang2023video, cheng2024videollama, maaz2023video, zhang2025videollama} has begun exploring video modality integration, aiming to equip VLMs with spatial-temporal awareness critical for tasks like video comprehension, where both contextual details and motion dynamics are essential. For example, VideoLLM-MoD~\cite{wu2024videollm} proposes to address the efficiency issue when processing long-term video by mixture-of-depths. ~\cite{yuan2024videorefer} introduces VideoRefer to enhance the finer-level (like object-level) spatial-temporal video understanding of VLMs. Grounded-VideoLLM~\cite{wang2024grounded} also targets for fine-grained video understanding through incorporating an additional temporal stream. In this work, we aim to rigorously evaluate the 4D spatial-temporal reasoning capabilities of state-of-the-art VLMs, probing how and to what extent these models internalize spatial intelligence and temporal dependencies.

\paragraph{VLM Benchmarks}%
Following the development trends of VLMs, benchmarking VLMs shares the similar trajectory by first evaluating vision QA on static images~\cite{li2024seed, liu2024mmbench, he2024mmworld,yue2024mmmu}, to align with models’ early focus on 2D understanding. As VLMs evolved to tackle dynamic scenarios, benchmarks expanded to evaluate general-purpose video comprehension tasks that probe temporal coherence and event understanding~\cite{ning2023video, khattak2024good, li2024videoeval, fu2024video, li2024mvbench}. Notably, MMVU~\cite{zhao2025mmvu} further proposes a knowledge-intensive benchmark to assess the expert-level reasoning ability of current video-based large models. However, while these works assess perception and semantic understanding, they largely overlook the explicit evaluation of spatial-temporal awareness, a core capability for real-world applications requiring 4D (3D space + time) reasoning. Recent efforts like~\cite{yang2024thinking} pioneer benchmarks for 3D visual-spatial intelligence but restrict evaluation to static 3D scene, neglecting the interplay of object motion and temporal dynamics intrinsic to videos. In this work, we introduce \texttt{VLM4D}, the first benchmark designed to holistically evaluate the 4D intelligence in VLMs, unifying spatial understanding, temporal continuity, and motion reasoning. By curating tasks that demand precise analysis of dynamic interactions (e.g., direction prediction, perspective anticipation, and motion reasoning), \texttt{VLM4D} exposes critical gaps in current models' ability to internalize spatiotemporal relationships. Our work not only advances the granularity of VLM evaluation but also shares insights and potential solutions to improve the model performance.

\part{Generative World Modeling}

\begingroup
\let\clearpage\relax

\chapter{Multimodal Control for Text-to-Image Generation}
\section{Introduction}
In the realm of text-to-image (T2I) generation, diffusion models exhibit exceptional performance in transforming textual descriptions into visually accurate images. Such models exhibit extraordinary potential across a plethora of applications, spanning from content creation~\cite{stable_diffusion,saharia2022photorealistic,nichol2021glide,ramesh2021zero,yu2022scaling,chang2023muse}, image editing~\cite{balaji2022ediffi,kawar2023imagic,couairon2022diffedit,zhang2023text,valevski2022unitune,nichol2021glide,hertz2022prompt,brooks2023instructpix2pix,mokady2023null}, and also fashion design. We propose a new unified method that can tackle two problems in text-to-image generation: improve the training efficiency of T2I models concerning memory usage, computational requirements, and a thirst for extensive datasets~\cite{imagen2022, stable_diffusion, dalle2}; and improve their controllability especially when dealing with multimodal conditioning, e.g. multiple edge maps and at the same time follow the guidance of text prompts.  

Controllable text-to-image generation models~\cite{t2iadapter} often come at a significant training computational cost, with linear growth in cost and size when training with different conditions. Our approach can improve the training efficiency of existing text-to-image diffusion models and unify and flexibly handle different structural input conditions all together. We take cues from the efficient parameterization strategies prevalent in the NLP domain~\cite{pham2018efficient,huLoRALowRankAdaptation2021,zakenBitFitSimpleParameterefficient2021,houlsbyParameterEfficientTransferLearning2019} and computer vision literature~\cite{he2022parameter}.  The key idea is to learn shared decomposed weights for varied input conditions, ensuring their intrinsic characteristics are conserved.  Our method has several benefits: It not only achieves greater compactness~\cite{stable_diffusion}, but also retains the full representation capacity to handle various input conditions of various modalities; Sharing weights across different conditions contributes to the data efficiency; The streamlined parameter space aids in mitigating overfitting to singular conditions, thereby reinforcing the flexible control aspect of our model.

Meanwhile, generating images from multiple homogeneous conditional inputs, especially when they present conflicting conditions or need to align with specific text prompts, is challenging. To further augment our model’s capability to handle multiple inputs from either the same or diverse modalities, during training, we introduce a new training strategy with two new loss functions introduced to strengthen the guidance of corresponding conditions. This approach, combined with our compact parameter optimization space, empowers the model to learn and manage multiple controls efficiently, even within the same category (e.g., handling two distinct segmentation maps and two separate edge maps).

The contributions of this future work are summarized below: First, we propose a novel text-to-image generation model for efficient controllable image generation that substantially reduces training memory overhead and model parameters through decomposition of weights shared across different conditions. Second, we introduce a new training strategy to improve the flexible controllability. Compared with previous works, we can generate new images conditioning on multiple inputs from diverse compositions of multiple modalities.
Third,  we will show on-par performance with Uni-ControlNet~\cite{unicontrolnet} on controllable text-to-image generation withless trainable parameters and 30\% less training memory. Furthermore, we will exhibit enhanced data efficiency, effectively doubling the performance achieved with only half amount of training data.

While text-to-image diffusion models have shown promise results, their potential in downstream discriminative applications is largely uncharted. In this paper, we delve into the capabilities of these diffusion models and improve the efficiency of using them as zero-shot classifiers. Towards this, we introduce a novel hierarchical sampling strategy that significantly optimizes the computational demands of these zero-shot classifiers, making them faster and more feasible for real-world applications. Our work showcases the potential of text-to-image diffusion models as powerful tools for zero-shot classification.

The proliferation of diffusion models like Imagen~\cite{imagen}, Dalle-2~\cite{dalle2}, and Stable Diffusion~\cite{stable_diffusion} have shown promise in creating realistic, high-resolution images from text prompts. However, there's a gap in exploring their transferability to discriminative tasks and comparative performance against other models. This work aims to bridge this gap by leveraging Stable Diffusion for discriminative tasks, introducing an innovative acceleration technique for text-to-image diffusion models as zero-shot image classifiers. We propose a novel sampling technique that boosts sample efficiency by up to 2 times, showing substantial improvements in inference speeds across three benchmark datasets while maintaining comparable classification accuracy. This endeavor could significantly enhance computational efficiency, paving the way for broader deployment of these models in discriminative tasks.

\section{Method}
\begin{figure}[!t]
     \centering
     \includegraphics[width=0.88\textwidth]{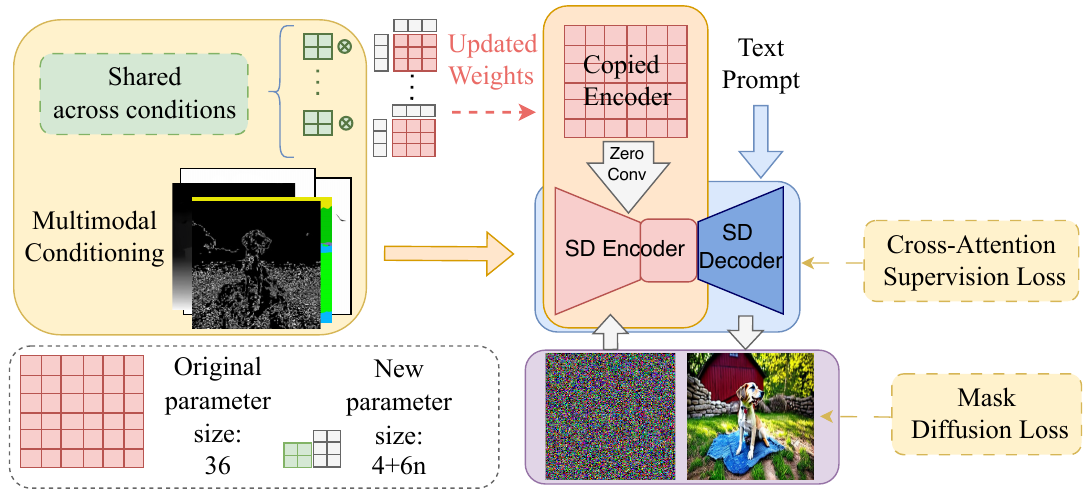}
     \caption{{\textbf{Overview of FlexEControl}: a decomposed green matrix is shared across \textbf{different input conditions}, significantly enhancing the model's efficiency and preserving the image content. During training, we integrate two specialized loss functions to enable flexible control and to adeptly manage conflicting conditions. In the example depicted here, the new parameter size is efficiently condensed to $4+6n$, where $n$ denotes the number of decomposed matrix pairs. }}
     \label{fig:overview_method}
\end{figure}
The overview of our method is shown in Figure~\ref{fig:overview_method}. In general, we use the {copied Stable Diffusion encoder (Stable Diffusion encoder block and Stable Diffusion middle block)} which accepts structural conditional input and then perform efficient training via parameter reduction using Kronecker Decomposition first~\cite{phm} and then low-rank decomposition over the updated weights of the copied Stable Diffusion encoder. To enhance the control from language and different input conditions, we propose a new training strategy with two newly designed loss functions. The details are shown in the sequel.

\subsection{Preliminary}
We use Stable Diffusion 1.5~\cite{stable_diffusion} in our experiments. This model falls under the category of Latent Diffusion Models (LDM) that encode input images \( x \) into a latent representation \( z \) via an encoder \( \mathcal{E} \), such that \( z = \mathcal{E}(x) \), and subsequently carry out the denoising process within the latent space \( \mathcal{Z} \). An LDM is trained with a denoising objective as follows:

\begin{equation}
\mathcal{L}_{\text{ldm}} = \mathbb{E}_{z,c,e,t}\left[\left\lVert \hat{\epsilon}_\theta(z_t \mid c,t) - \epsilon \right\rVert^2\right]
\end{equation}
where $(z, c)$ constitute data-conditioning pairs (comprising image latents and text embeddings), $\epsilon \sim \mathcal{N}(0, I)$ , $t \sim \text{Uniform}(1, T)$, and $\theta$ denotes the model parameters.

\subsection{Efficient Training for Controllable Text-to-Image (T2I) Generation}
 Our approach is motivated by empirical evidence that Kronecker Decomposition~\cite{phm} effectively preserves critical weight information. We employ this technique to encapsulate the shared relational structures among different input conditions. Our hypothesis posits that by amalgamating diverse conditions with a common set of weights, data utilization can be optimized and training efficiency can be improved. We focus on decomposing and fine-tuning only the cross-attention weight matrices within the U-Net~\cite{unet} of the diffusion model. As depicted in Figure~\ref{fig:overview_method}, the copied encoder from the Stable Diffusion will accept conditional input from different modalities. During training, we posit that these modalities, being transformations of the same underlying image, share common information. Consequently, we hypothesize that the updated weights of this copied encoder, $\Delta\boldsymbol{W}$, can be efficiently adapted within a shared decomposed low-rank subspace.  This leads to:
\begin{equation}
    \Delta\boldsymbol{W}=\sum_{i=1}^{n} \boldsymbol{H}_{\boldsymbol{i}} \otimes\left({u}_{{i}} {v}_{{i}}^{\top}\right)
\end{equation}
with $n$ is the number of decomposed matrices, ${u}_{{i}} \in \mathbb{R}^{\frac{k}{n} \times r}$ and ${v}_{{i}} \in \mathbb{R}^{r \times \frac{d}{n}}$, where $r$ is the rank of the matrix which is a small number, $\boldsymbol{H}_{\boldsymbol{i}}$ are the decomposed learnable matrices shared across different conditions, and $\otimes$ is the Kronecker product operation. The low-rank decomposition ensures a consistent low-rank representation strategy. This approach substantially saves trainable parameters, allowing efficient fine-tuning over the downstream text-to-image generation tasks.

The intuition for why Kronecker decomposition works for finetuning partially is partly rooted in the findings of~\cite{phm,mahabadiCompacterEfficientLowRank2021,he2022parameter}. These studies highlight how the model weights can be broken down into a series of matrix products and thereby save parameter space. As shown in Figure~\ref{fig:overview_method}, the original weights is 6x6, then decomposed into a series of matrix products. When adapting the training approach based on the decomposition to controllable T2I, the key lies in the shared weights, which, while being common across various conditions, retain most semantic information. {The Kronecker Decomposition is known for its multiplicative rank property and content-preserving qualities. For instance, the shared “slow”
weights~\cite{fast_weights} of an image, combined with another set of  “fast” low-rank weights, can preserve the original image's distribution without a loss in semantic integrity. This observation implies that updating the slow weights is crucial for adapting to diverse conditions. Following this insight, it becomes logical to learn a set of condition-shared decomposed weights in each layer, ensuring that these weights remain consistent across different scenarios. The data utilization and parameter efficiency is also improved.}

\subsection{Enhanced Training for Conditional Inputs}
\label{sec:enhanced_training}
We then discuss how to improve the control under multiple input conditions of varying modalities with the efficient training approach.

\paragraph{Dataset Augmentation with Text Parsing and Segmentation}
To optimize the model for scenarios involving multiple homogeneous (same-type) conditional inputs, we initially augment our dataset. We utilize a large language model (\texttt{gpt-3.5-turbo}) to parse texts in prompts containing multiple object entities. The parsing query is structured as: \texttt{Given a sentence, analyze the objects in this sentence, give me the objects if there are multiple.} Following this, we apply CLIPSeg~\cite{clipseg} to segment corresponding regions in the images, allowing us to divide structural conditions into separate sub-feature maps tailored to the parsed objects. These segmentation masks are selectively used to augment the dataset, specifically when there is a clear, single mask for each identified object. This selective approach helps maintain the robustness of the dataset and enhances training performance.

\paragraph{Cross-Attention Supervision Loss}
For each identified segment, we calculate a unified attention map, \(\boldsymbol{A}_j\), by averaging attention across layers and the relevant \(N\) text tokens:
\begin{equation}
\boldsymbol{A}_j = \frac{1}{L} \sum_{l=1}^L \sum_{i=1}^N \llbracket T_i \in \mathcal{T}_j \rrbracket \, \mathbf{CA}_i^l,
\end{equation}
where \(\llbracket \cdot \rrbracket\) denotes the Iverson bracket, \(\mathbf{CA}_i^l\) is the cross-attention map for token \(i\) in layer \(l\), and \(\mathcal{T}_j\) denotes the set of tokens associated with the \(j\)-th segment.

The model is trained to predict noise for image-text pairs concatenated based on the parsed and segmented results. An additional loss term, designed to ensure focused reconstruction in areas relevant to each text-derived concept, is introduced. This loss is calculated as the Mean Squared Error (MSE) deviation from predefined masks corresponding to the segmented regions:
\begin{equation}
\mathcal{L}_{\text{ca}} = \mathbb{E}_{z, t}\left[\left\|\boldsymbol{A}_i(v_i, z_t) - M_{i}\right\|_2^2\right],
\end{equation}
where \(\boldsymbol{A}_i(v_i, z_t)\) is the cross-attention map between token \(v_i\) and noisy latent \(z_t\), and \(M_{i}\) represents the mask for the \(i\)-th segment, which is derived from the segmented regions in our augmented dataset and appropriately resized to match the dimensions of the cross-attention maps.

\paragraph{Masked Diffusion Loss}
To ensure fidelity to the specified conditions, we apply a condition-selective diffusion loss that concentrates the denoising effort on conceptually significant regions.  This focused loss function is applied solely to pixels within the regions delineated by the concept masks, which are derived from the non-zero features of the input structural conditions. Specifically, the masks are binary where non-zero feature areas are assigned a value of one, and areas lacking features are set to zero. Because of the sparsity of pose features for this condition, we use the all-ones mask. These masks serve to underscore the regions referenced in the corresponding text prompts:

\begin{equation}
\mathcal{L}_{\text{mask}} = \mathbb{E}_{z, \epsilon, t}\left[\left\|(\epsilon - \epsilon_\theta(z_t, t)) \odot M\right\|_2^2\right],
\end{equation}

where $M$ represents the union of binary mask obtained from input conditions,
$z_t$ denotes the noisy latent at timestep $t$, $\epsilon$ the injected noise, and $\epsilon_\theta$ 

the estimated noise from the denoising network (U-Net).

The total loss function employed is:
\begin{equation}
\mathcal{L}_{\text{total}} = \mathcal{L}_{\text{ldm}}  + \lambda_{\text{ca}} \mathcal{L}_{\text{ca}} + \lambda_{\text{mask}} \mathcal{L}_{\text{mask}},
\end{equation}
with $\lambda_{\text{rec}}$ and $\lambda_{\text{attn}}$ set to 0.01. The integration of $\mathcal{L}_{\text{ca}}$ and $\mathcal{L}_{\text{mask}}$ ensure the model will focus at reconstructing the conditional region and attend to guided regions during generation.

\section{Experiments}


\subsection{Datasets}~\label{dataset}
In pursuit of our objective of achieving controlled Text-to-Image (T2I) generation, we employed the \textbf{LAION improved\_aesthetics\_6plus} \cite{laion} dataset for our model training. Specifically, we meticulously curated a subset comprising 5,082,236 instances, undertaking the elimination of duplicates and applying filters based on criteria such as resolution and NSFW score. Given the targeted nature of our controlled generation tasks, the assembly of training data involved considerations of additional input conditions, specifically edge maps, sketch maps, depth maps, segmentation maps, and pose maps. The extraction of features from these maps adhered to the methodology expounded in\cite{controlnet}.



\subsection{Experimental Setup}
\label{Experimental Setup}

\paragraph{Structural Input Condition Extraction}
{We start from the processing of various local conditions used in our experiments. To facilitate a comprehensive evaluation, we have incorporated a diverse range of structural conditions. These conditions include edge maps~\cite{canny, hed, mlsd}, sketch maps~\cite{sketchmap}, pose information~\cite{pose}, depth maps~\cite{depth}, and segmentation maps~\cite{segmentation}, each extracted using specialized techniques. These conditions are crucial for guiding the text-to-image generation process, enabling FlexEControl to produce images that are both visually appealing and semantically aligned with the text prompts and structural inputs. The additional details for extracting those conditions are given in the Appendix.}

\paragraph{Evaluation Metrics}
We employ a comprehensive benchmark suite of metrics including mIoU, SSIM, mAP, MSE, FID~\cite{fid}, and CLIP Score~\cite{hessel2021clipscore,clip}~\footnote{https://github.com/jmhessel/clipscore}.

{\paragraph{Baselines}
In our comparative evaluation, we assess T2I-Adapter~\cite{t2iadapter}, PHM~\cite{phm}, Uni-ControlNet~\cite{unicontrolnet}, and LoRA~\cite{huLoRALowRankAdaptation2021}. The implementation details are given in the Appendix.}

\paragraph{Implementation Details}
In accordance with the configuration employed in Uni-ControlNet, we utilized Stable Diffusion 1.5~\footnote{https://huggingface.co/runwayml/stable-diffusion-v1-5} as the foundational model. Our model underwent training for a singular epoch, employing the AdamW optimizer~\cite{adam} with a learning rate set at $10^{-5}$. Throughout all experimental iterations, we standardized the dimensions of input and conditional images to $512\times512$. The fine-tuning process was executed on P3 AWS EC2 instances equipped with 64 NVIDIA V100 GPUs.

{For baseline implementations, we compare~FlexEControl with T2I-Adapter~\cite{t2iadapter}, PHM~\cite{phm}, Uni-ControlNet~\cite{unicontrolnet}, and LoRA~\cite{huLoRALowRankAdaptation2021} where we implement LoRA and PHM layers over the trainable modules in Uni-ControlNet interms of generated image quality and controllability. The rank of LoRA is set to 4. For PHM~\cite{phm}, we implement it by performing Kronecker decomposition and share weights across different layer, with the number of decomposed matrix being 4.}

For quantitative assessment, a subset comprising 10,000 high-quality images from the \textbf{LAION improved\_aesthetics\_6.5plus} dataset was utilized. The resizing of input conditions to $512\times512$ was conducted during the inference process.

\subsection{Quantitative Results}
\label{Results}

\begin{table}[!t]
\centering
\caption{\textbf{Text-to-image generation efficiency comparison}: FlexEControl shows substantial reductions in memory cost, trainable parameters, and training time, highlighting its improved training efficiency with the same model architecture. Training times are averaged over three runs up to 400 iterations for consistency.}

\label{tab:efficency-table}
\begin{tabular}{lcccc}
\toprule
Models & Memory Cost $\downarrow$ & \# Params. $\downarrow$ & Training Time $\downarrow$ \\

\midrule
  Uni-ControlNet~\cite{unicontrolnet} & 20.47GB & 1271M & 5.69 $\pm$ 1.33s/it \\
  LoRA~\cite{huLoRALowRankAdaptation2021} & 17.84GB & 1074M  & 3.97 $\pm$ 1.27 s/it \\
  PHM~\cite{phm} & 15.08GB & 819M  & 3.90 $\pm$ 2.01 s/it \\
FlexEControl (\textbf{ours}) & \textbf{14.33GB} & \textbf{750M} & \textbf{2.15 $\pm$ 1.42 s/it} \\
\bottomrule
\end{tabular}
\end{table}

\begin{table*}[t]
\centering
\caption{\textbf{Quantitative evaluation of controllability and image quality} for single structural conditional inputs. FlexEControl performs overall better while maintaining much improved efficiency. }
\label{tab:table_performance}
\resizebox{\textwidth}{!}{
\begin{tabular}{@{}lccccccccccc@{}}
\toprule
\multirow{2}{*}{Models} & Canny & MLSD & HED & Sketch & Depth & Segmentation & Poses & \multirow{2}{*}{FID$\downarrow$} & \multirow{2}{*}{CLIP Score$\uparrow$} \\
 & (SSIM)$\uparrow$ & (SSIM)$\uparrow$ & (SSIM)$\uparrow$ & (SSIM)$\uparrow$ & (MSE)$\downarrow$ & (mIoU)$\uparrow$ & (mAP)$\uparrow$ & \\
\midrule
T2IAdapter~\cite{t2iadapter} & 0.4480 & - &-&0.5241&90.01&0.6983&\textbf{0.3156}&27.80&0.4957\\
{ControlNet}~\cite{controlnet} &  0.4989& 0.6172&0.4990&\textbf{0.6013}&  \textbf{89.08} & 0.7481& 0.2024 &27.62 &0.4931\\
Uni-Control~\cite{unicontrol} &  0.4977& 0.6374&0.4885&0.5509&  90.04 & 0.7143 & 0.2083 &27.80 &0.4899\\
Uni-ControlNet~\cite{unicontrolnet} &  0.4910& 0.6083&0.4715&0.5901&  90.17 & 0.7084 & 0.2125 &27.74 &0.4890\\
PHM~\cite{phm} & 0.4365 & 0.5712 & 0.4633 & 0.4878  & 91.38 & 0.5534 & 0.1664 &27.91 &0.4961\\
LoRA~\cite{huLoRALowRankAdaptation2021} & 0.4497 & 0.6381 & \textbf{0.5043} & 0.5097  & 89.09 & 0.5480 & 0.1538 &27.99 &0.4832\\
FlexEControl(\textbf{ours}) &\textbf{0.4990} &\textbf{0.6385} &0.5041 &0.5518 & 90.93 &\textbf{0.7496}&0.2093 &\textbf{27.55}&\textbf{0.4963}\\

\bottomrule

\end{tabular}}

\end{table*}

\begin{table*}[t]
\centering
\caption{\textbf{Quantitative evaluation of controllability and image quality} on~FlexEControl along with its variants and Uni-ControlNet. For Uni-ControlNet, we implement multiple conditioning by adding two homogeneous conditional images after passing them through the feature extractor.}
\label{tab:small_data}
\resizebox{\textwidth}{!}{
\begin{tabular}{@{}lcccccccccccc@{}}
\toprule
& \multirow{2}{*}{Models} & Canny & MLSD & HED & Sketch & Depth & Segmentation & Poses & \multirow{2}{*}{FID$\downarrow$} & \multirow{2}{*}{CLIP Score$\uparrow$} \\
&& (SSIM)$\uparrow$ & (SSIM)$\uparrow$ & (SSIM)$\uparrow$ & (SSIM)$\uparrow$ & (MSE)$\downarrow$ & (mIoU)$\uparrow$ & (mAP)$\uparrow$ & & \\
\midrule
\multirow{4}{*}{Single Conditioning} & Uni-ControlNet & 0.3268&0.4097&0.3177&0.4096&98.80&0.4075&\textbf{0.1433}&29.43 &0.4844  \\
& FlexEControl (w/o $L_{ca}$)&0.3698 &0.4905 &0.3870 &0.4855 & 94.90 &0.4449&0.1432 &28.03 &0.4874\\
& FlexEControl (w/o $L_{mask}$) & 0.3701 & 0.4894 &0.3805&\textbf{0.4879}&\textbf{94.30} & 0.4418 &0.1432&28.19&0.4570\\
& FlexEControl &\textbf{0.3711} &\textbf{0.4920} &\textbf{0.3871} &0.4869 & 94.83 &\textbf{0.4479}&0.1432 &\textbf{28.03} &\textbf{0.4877}\\
\hdashline
\multirow{4}{*}{Multiple Conditioning} 
& Uni-ControlNet & 0.3078 & 0.3962 &0.3054&0.3871&98.84 & 0.3981 &0.1393&28.75&0.4828\\
& FlexEControl (w/o $L_{ca}$) & 0.3642 & 0.4901 &0.3704&0.4815&94.95 & 0.4368 &0.1405&28.50&0.4870\\
& FlexEControl (w/o $L_{mask}$) & 0.3666 & 0.4834 &0.3712&0.4831&94.89 & 0.4400 &0.1406&28.68&0.4542\\
& FlexEControl &\textbf{0.3690} &\textbf{0.4915} &\textbf{0.3784} &\textbf{0.4849} & \textbf{92.90} &\textbf{0.4429}&\textbf{0.1411} &\textbf{28.24} &\textbf{0.4873}\\
\bottomrule
\end{tabular}}
\end{table*}

Table~\ref{tab:efficency-table} highlights FlexEControl's superior efficiency compared to Uni-ControlNet. It achieves a 30\% reduction in memory cost, lowers trainable parameters by 41\% (from 1271M to 750M), and significantly reduces training time per iteration from 5.69s to 2.15s.


Table~\ref{tab:table_performance} provides a comprehensive comparison of FlexEControl's performance against Uni-ControlNet and T2IAdapter across diverse input conditions.  After training on a dataset of 5M text-image pairs, FlexEControl demonstrates better, if not superior, performance metrics compared to Uni-ControlNet and T2IAdapter. Note that Uni-ControlNet is trained on a much larger dataset (10M text-image pairs from the LAION dataset). Although there is a marginal decrease in SSIM scores for sketch maps and mAP scores for poses, FlexEControl excels in other metrics, notably surpassing Uni-ControlNet and T2IAdapter. This underscores our method's proficiency in enhancing efficiency and elevating overall quality and accuracy in controllable text-to-image generation tasks.

To validate FlexEControl's effectiveness in handling multiple structural conditions, we compared it with Uni-ControlNet through human evaluations. Two scenarios were considered: multiple homogeneous input conditions (300 images, each generated with 2 canny edge maps) and multiple heterogeneous input conditions (500 images, each generated with 2 randomly selected conditions). Results, summarized in Table~\ref{tab:humaneval}, reveal that FlexEControl was preferred by 64.00\% of annotators, significantly outperforming Uni-ControlNet (23.67\%). This underscores FlexEControl's proficiency with complex, homogeneous inputs. Additionally, FlexEControl demonstrated superior alignment with input conditions (67.33\%) compared to Uni-ControlNet (23.00\%). In scenarios with random heterogeneous conditions, FlexEControl was preferred for overall quality and alignment over Uni-ControlNet.

In addition to our primary comparisons, we conducted an additional quantitative evaluation of~FlexEControl and Uni-ControlNet. This evaluation focused on assessing image quality under scenarios involving multiple conditions from both the homogeneous and heterogeneous modalities. The findings of this evaluation are summarized in Table~\ref{tab:quantitative_results_diverse_modalities}. FlexEControl consistently outperforms Uni-ControlNet in both categories, demonstrating lower FID scores for better image quality and higher CLIP scores for improved alignment with text prompts.

\subsubsection{Ablation Studies}
\label{ablation2}
To substantiate the efficacy of FlexEControl in enhancing training efficiency while upholding commendable model performance, and to ensure a fair comparison, an ablation study was conducted by training models on an identical dataset. {We trained~FlexEControl along its variants and Uni-ControlNet on a subset of 100,000 training samples from \textbf{LAION improved\_aesthetics\_6plus}. When trained with the identical data, FlexEControl performs better than Uni-ControlNet. The outcomes are presented in Table~\ref{tab:small_data}. Evidently, FlexEControl exhibits substantial improvements over Uni-ControlNet when trained on the same dataset. This underscores the effectiveness of our approach in optimizing data utilization, concurrently diminishing computational costs, and enhancing efficiency in the text-to-image generation process.}

\begin{table}[!t]
\centering
\caption{Human evaluation of FlexEControl and Uni-ControlNet under homogenous and heterogeneous structural conditions, assessing both human preference and condition alignment. "Win" indicates FlexEControl's preference, "Tie" denotes equivalence, and "Lose" indicates Uni-ControlNet's preference. Results indicate that under homogeneous conditions, FlexEControl outperforms Uni-ControlNet in both human preference and condition alignment.}
\label{tab:humaneval}
\medskip
\begin{tabular}{lccccc}
\toprule
Condition Type & Metric & Win & Tie & Lose \\
\midrule
\multirow{2}{*}{Homogeneous} & Human Preference (\%)& \textbf{64.00} & 12.33 & 23.67 \\
& Condition Alignment (\%)& \textbf{67.33} & 9.67 & 23.00 \\
\midrule
\multirow{2}{*}{Heterogeneous } & Human Preference (\%)&\textbf{9.80}&87.40&2.80 \\
& Condition Alignment (\%)&\textbf{6.60} & 89.49 &4.00 \\
\bottomrule
\end{tabular}
\end{table}

\begin{table}[!t]
\centering
\caption{Quantitative evaluation of controllability and image quality in scenarios with multiple conditions from heterogeneous and homogeneous modalities for FlexEControl and Uni-ControlNet. The 'heterogeneous' category averages the performance across one Canny condition combined with six other different modalities. The 'homogeneous' category represents the average performance across seven identical modalities (three inputs). }
\label{tab:quantitative_results_diverse_modalities}
\medskip
\begin{tabular}{lcccc}
\toprule
Condition Type & Baseline & FID$\downarrow$ & CLIP Score$\uparrow$ \\
\midrule
\multirow{2}{*}{Heterogeneous} & Uni-ControlNet  & 27.81 & 0.4869 \\
& FlexEControl& \textbf{27.47} & \textbf{0.4981} \\
\hdashline
\multirow{2}{*}{Homogeneous} & Uni-ControlNet & 28.98 & 0.4858 \\
& FlexEControl& \textbf{27.65} & \textbf{0.4932} \\
\bottomrule
\end{tabular}
\end{table}

\subsubsection{Additional Results on Stable Diffusion 2}
In our efforts to explore the versatility and adaptability of FlexEControl, we conducted additional experiments using the Stable Diffusion 2.1 model, available at \href{https://huggingface.co/stabilityai/stable-diffusion-2-1}{Hugging Face's Model Hub}. The results from these experiments are depicted in Table~\ref{tab:sd2_small}. FlexEControl can leverage the advancements in Stable Diffusion 2.1 to achieve even better performance in text-to-image generation tasks. For the sake of a fair comparison in the main paper, we conduct experiments using Stable Diffusion 1.5 model.

\begin{table*}[t]
\centering
\caption{Quantitative evaluation of controllability and image quality trained on a subset of 100,000 samples. Human poses are evaluated solely within portrait images.}
\label{tab:sd2_small}
\resizebox{\textwidth}{!}{
\begin{tabular}{@{}lccccccccccc@{}}
\toprule
\multirow{2}{*}{Models} & Canny & MLSD & HED & Sketch & Depth & Segmentation & Poses & \multirow{2}{*}{FID$\downarrow$} & \multirow{2}{*}{CLIP Score$\uparrow$} \\

 & (SSIM)$\uparrow$ & (SSIM)$\uparrow$ & (SSIM)$\uparrow$ & (SSIM)$\uparrow$ & (MSE)$\downarrow$ & (mIoU)$\uparrow$ & (mAP)$\uparrow$ & & \\
\midrule
FlexEControl &0.3711 &0.4920 &0.3871 &0.4869 & 94.83 &0.4479&0.1432 &28.03 &0.4877\\
FlexEControl-SD 2.1 &\textbf{0.3891} &\textbf{0.5273} &\textbf{0.4077} &\textbf{0.4960} & \textbf{93.58} &\textbf{0.4490}&\textbf{0.1562} &\textbf{25.08} &\textbf{0.5833}\\
\bottomrule
\end{tabular}}
\end{table*}

\section{Related Work}
FlexEControl is an instance of efficient training and controllable text-to-image generation. Here, we overview modeling efforts in the subset of efficient training towards reducing parameters and memory cost and controllable T2I.

\subsection{Efficient Training} Prior work has proposed efficient training methodologies both for pretraining and fine-tuning. These methods have established their efficacy across an array of language and vision tasks~\cite{he2020sample,yang2020covid}. One of these explored strategies is Prompt Tuning~\cite{prompt_tuning}, where trainable prompt tokens are appended to pretrained models~\cite{first_prompt, promptingvl, visualprompttuning}. These tokens can be added exclusively to input embeddings or to all intermediate layers~\cite{liPrefixTuningOptimizingContinuous2021}, allowing for nuanced model control and performance optimization. Low-Rank Adaptation (LoRA)~\cite{huLoRALowRankAdaptation2021} is another innovative approach that introduces trainable rank decomposition matrices for the parameters of each layer. LoRA has exhibited promising fine-tuning ability on large generative models, indicating its potential for broader application. Furthermore, the use of Adapters inserts lightweight adaptation modules into each layer of a pretrained transformer~\cite{houlsbyParameterEfficientTransferLearning2019, ruckleAdapterDropEfficiencyAdapters2021}. This method has been successfully extended across various setups~\cite{zhangTipAdapterTrainingfreeCLIPAdapter2021,clip_adapter,t2iadapter}, demonstrating its adaptability and practicality. Other approaches including post-training model compression~\cite{structural_prunning} facilitate the transition from a fully optimized model to a compressed version -- either sparse~\cite{sparsegpt,tarm,ltarm}, quantized~\cite{Q-diffusion,vector_quantized_diffusion}, or both. This methodology was particularly helpful for parameter quantization~\cite{dettmers2023qlora}. {Different from these methodologies, FlexEControl puts forth a new unified strategy that aims to enhance the efficient training of text-to-image diffusion models through the leverage of low-rank structure. FlexEControl is also a general approach that can be applied to UniControl~\cite{unicontrol} or other backbones.}

\subsection{Controllable Text-to-Image Generation}
Recent developments in the text-to-image generation domain strives for more control over image generation, enabling more targeted, stable, and accurate visual outputs, several models like T2I-Adapter~\cite{t2iadapter} and Composer~\cite{composer} have emerged to enhance image generations following the semantic guidance of text prompts and multiple different structural conditional control. However, existing methods are struggling at dealing with multiple conditions from the same modalities, especially when they have conflicts, e.g. multiple segmentation maps and at the same time follow the guidance of text prompts; Recent studies also highlight challenges in controllable text-to-image generation (T2I), showing that current models are struggling at handling controls from different conditions. Towards these, the Attend-and-Excite method \cite{Chefer2023AttendandExciteAS} refines attention regions to ensure distinct attention across separate image regions. ReCo \cite{yang2022reco}, GLIGEN \cite{Li2023GLIGENOG}, and Layout-Guidance \cite{chen2023trainingfree} allow for image generation informed by bounding boxes and regional descriptions. {~\cite{mo2024freecontrol} offers a training-free approach to multimodal control.
 FlexEControl improves the model's controllability by proposing a new training strategy, distinguishing itself by targeting the flexibility and efficiency of multimodal control, especially in scenarios with conflicting conditions from the same or different modalities (e.g., multiple segmentation maps combined with text prompts). }

\chapter{Dynamic Control for Text-to-Video Generation}

\section{Introduction}
\label{sec:intro}

\begin{figure*}[tp]
  \centering
    \includegraphics[width=0.8\textwidth]{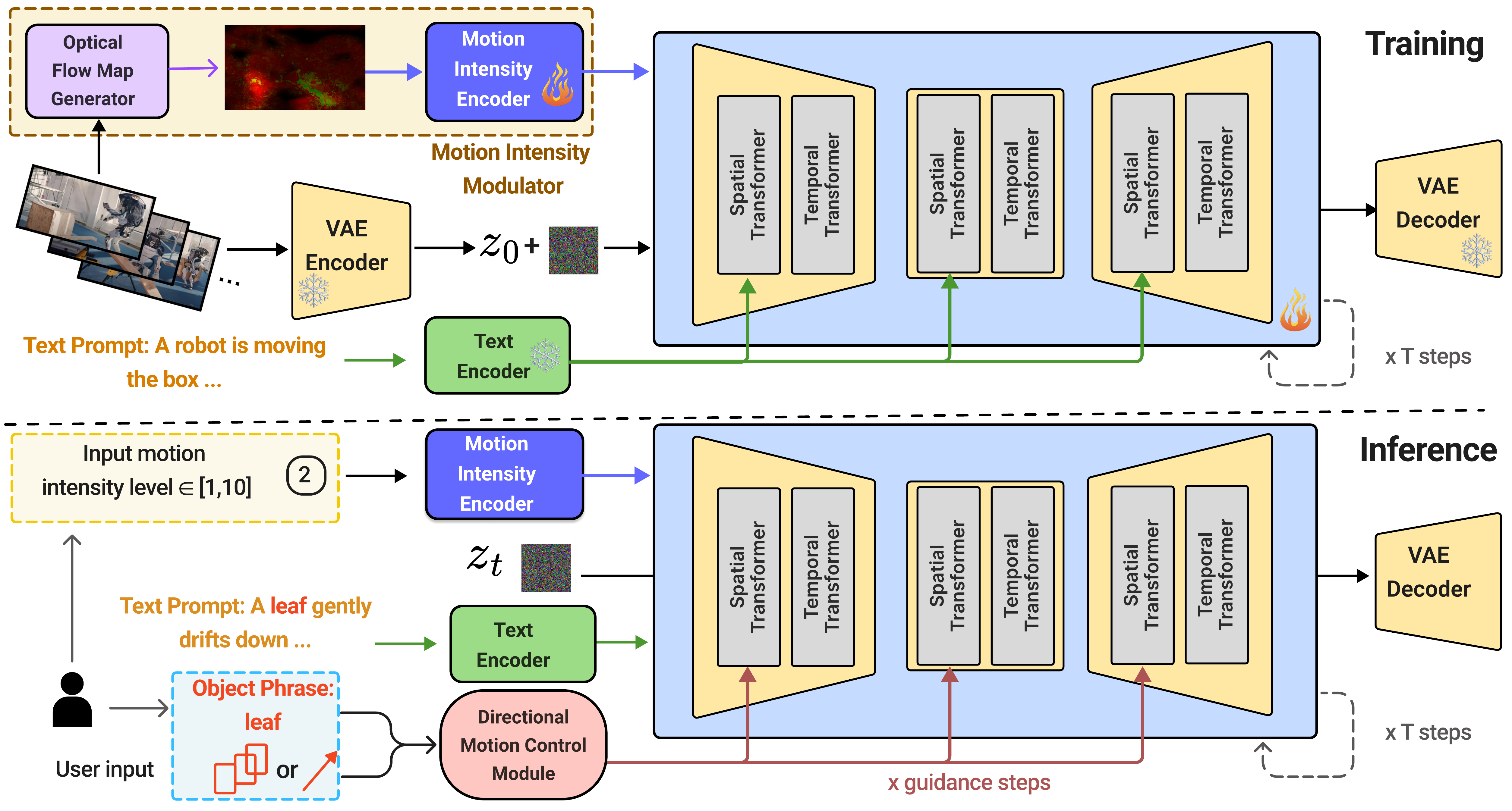}
    \caption{Overview of the Mojito framework. In the training pipeline (top), ~Mojito uses a VAE Encoder to transform input frames into latent features, processed by Spatial and Temporal Transformers within the U-Net. Motion intensity control is introduced through the \textit{Motion Intensity Modulator}, consisting of the Optical Flow Map Generator and the Motion Intensity Encoder. The~\textit{Directional Motion Control} module interprets object phrases within the prompt to align attention with specified trajectories. During inference (bottom), Mojito generates videos following user-defined motion intensity and directional guidance. }
    \label{Fig:overview_mojito}
\end{figure*}

Text-to-video (T2V) generation~\cite{he2024mojito,latentvideodiffusion,opensora,opensoraplan,easyanimate2024xu,yang2024cogvideox,vchitect2024} aims to produce diverse, high-quality videos with given text prompts. Unlike image generation~\cite{stable_diffusion,saharia2022photorealistic,nichol2021glide,ramesh2021zero,yu2022scaling,chang2023muse}, which produces a single static frame, video generation can extend beyond visual synthesis and has the potential to serve as a world modeling tool~\cite{nvidia2025cosmos,xiang2024pandora,wang2024world,kang2024far} and a physical AI engine~\cite{Genesis} for simulating real-world dynamics. As world modeling expects video generation models to control and generate the motion of physical objects, enabling realistic interactions with the environment, existing T2V models—despite their creative and powerful generative capabilities—struggle to produce large motion magnitudes~\cite{opensora,opensoraplan,pika2024,modelscope2023wang} or adaptable motion intensities (i.e., motion speed and magnitude) that align with user intent~\cite{nvidia2025cosmos,chen2024videocrafter2}, limiting their applicability.

Towards this, in this work, we aim to address a new research question:~\textit{Can text-to-video models be trained efficiently to control both motion direction and intensity in alignment with user intent?} The primary limitations of current models stem from several key challenges. \underline{First}, existing diffusion-based T2V models—whether U-Net or DiT-based~\cite{dit}—rely on diffusion processes in either pixel space or latent visual feature space during training, but lack dedicated motion modeling mechanisms. Consequently, they inherently struggle to adjust object motion in generated videos solely through text prompts and often fail to maintain temporal consistency in moving objects, resulting in flickering, trajectory inconsistencies, and visual artifacts across frames; \underline{Second}, capturing relative motion data is inherently complex in real-world videos due to the simultaneous movements of both cameras and objects, leading to lack of video data in training. There is also a lack of large-scale, annotated datasets specifically for motion direction and intensity. Existing video datasets rarely include detailed labels for motion dynamics, and obtaining human annotations is labor-intensive; \underline{Third}, training T2V models with such detailed annotations would also bring substantial computational resources.

To tackle these challenges, we present Mojito, the first diffusion model for text-to-video generation that simultaneously integrates text-based prompts with flexible motion controls, allowing for the precise modulation of both motion direction and intensity. 
To achieve this, Mojito introduces two novel modules apart of our trained video generation backbone: (1) the \textit{Motion Intensity Modulator} (MIM), which encodes input motion intensities as features and integrates them seamlessly into the diffusion framework during training.  (2) the \textit{Directional Motion Control} (DMC) module, which adjusts object motion direction through cross-attention guidance \textbf{without training}, aligning object trajectories with specified paths.

\section{Preliminaries}

\paragraph{Latent Video Diffusion.} 
Latent Diffusion Models (LDMs), such as Stable Diffusion~\cite{stable_diffusion}, consist of two main components: an autoencoder and a denoising network. The autoencoder encodes images into a latent space, while the denoising network operates within this latent space, progressively transforming random noise into a coherent image representation. Building on this, Latent Video Diffusion Models (LVDMs)~\cite{latentvideodiffusion} extend LDMs to video generation by incorporating temporal dynamics in the latent space. They are trained using a noise-prediction objective, minimizing the difference between predicted and actual noise added to the data:
\begin{equation}
    \mathcal{L}_{LDM} = \mathbb{E}_{i} \left[ \| \epsilon - \epsilon_\theta(z_t, t, c) \|_2^2\right],
\end{equation}
where \( \epsilon_\theta(\cdot) \) is the noise prediction function of the U-Net, and the condition \( c \) is provided to the U-Net through cross-attention, enabling control based on input text or other conditioning signals. The latent state \( z_t \) follows a Markov process, with Gaussian noise progressively added at each step to the initial latent state \( z_0 \):
\begin{equation}
    z_t = \sqrt{\bar{\alpha}_t} z_0 + \sqrt{1 - \bar{\alpha}_t} \epsilon, \quad \epsilon \sim \mathcal{N}(0, I),
\end{equation}
where \( \bar{\alpha}_t = \prod_{i=1}^t (1 - \beta_i) \) and $\beta_t$ is a step-dependent coefficient that controls noise strength at time $t$.

\paragraph{Motions in Video Generation.}\quad 
Motion in videos generally consists of two main components: \textit{direction} and \textit{intensity}. The direction of motion is derived from the trajectory or bounding boxes that an object follows across frames, while the intensity (or strength) reflects the speed and amplitude of movement along this path.

We define a motion trajectory as a sequence of spatial positions:
\begin{equation}
\mathcal{T} = \left\{(x_{1}, y_{1}), (x_{2}, y_{2}), \ldots, (x_{L}, y_{L})\right\},
\end{equation}
where each $(x_{i}, y_{i})$ represents the object’s position in frame $i$, with $x \in [0, W)$ and $y \in [0, H)$, where $L$ is the total number of frames, and $W$ and $H$ are the width and height of the video frame, respectively.

\section{The~Mojito Video Generation Framework}
\label{sec:motion_guided_video_generation}
\subsection{Methodology Overview}
Figure~\ref{Fig:overview_mojito} illustrates the architecture of~Mojito, which comprises two core modules for motion control: the \textit{Directional Motion Control} (DMC) module and the \textit{Motion Intensity Modulator} (MIM). These two modules work together to allow controlled motion in terms of direction and intensity, as well as to coordinate interaction with the generated video content. We first introduce the test-time, training-free ~\textit{Directional Motion Control}. Then we will introduce the~\textit{Motion Intensity Modulator} and finally the~Mojito backbone.

\subsection{Training-Free Directional Motion Control}
Existing text-to-video generation models often struggle to achieve user-defined motion direction, such as following a specific trajectory for an object. We address this limitation by introducing a method that dynamically guides motion direction during inference-time sampling, generating samples from a conditional distribution $p(z | y, \mathcal{T}, i)$ with additional directional controls. Here, $z$ represents the generated latent video, $y$ is the input text, and $\mathcal{T}$ specifies a user-defined input trajectory associated with certain tokens $y_n$. This directional control is achieved by modifying attention responses within cross-attention layers in the spatial transformer block of~Mojito, as shown in Figure~\ref{Fig:dmc}.

Cross-attention layers are essential for managing spatial layouts in generated content~\cite{prompt_to_prompt, training-free-guidance, training-layout-control}. The attention score $A_{u,n}^{(\gamma)}$ between spatial location $u$, text token $y_n$ in cross-attention layer $\gamma$ determines their association, with $\sum_{n=1}^N A_{u,n}^{(\gamma)} = 1$. This competitive interaction among text tokens helps guide object motion by biasing attention maps. By aligning attention focus along the trajectory $\mathcal{T}$ with the token $y_n$, we gain fine-grained frame-by-frame control over object placement without requiring additional training.

Specifically, for a given input trajectory $\mathcal{T} = \{(x_{i}, y_{i})\}_{i=1}^{L}$, where $L$ is the number of points in the trajectory, we expand each point $(x_{i}, y_{i})$ to create a bounding box $B_{i}$ for frame $i$. This bounding box is defined as:
\begin{equation}
B_{i} = \{ (x, y) : |x - x_{i}| \leq \Delta_x, |y - y_{i}| \leq \Delta_y \},
\end{equation}
where $\Delta_x$ and $\Delta_y$ are tolerance values that define the size of the bounding box around each trajectory point. These bounding boxes $B_{i}$ are allocated for each video frame $i$, ensuring alignment with the trajectory.

To align the attention maps with the bounding boxes over the course of the video, we define a frame-specific energy function $E_{i}$ at each video timestep $i$:
\begin{equation}
\label{energy_function}
E_{i}\left(A_{i}^{(\gamma)}, B_{i}, n\right) = \left(1 - \frac{\sum_{u \in B_{i}} A_{i,u,n}^{(\gamma)}}{\sum_u A_{i,u,n}^{(\gamma)}}\right)^2,
\end{equation}
where $A_{i,u,n}^{(\gamma)}$ denotes the attention score at video timestep $i$ in layer $\gamma$, spatial location $u$, and text token $y_n$ for the input object phrase. This function encourages attention to concentrate within the bounding box $B_{i}$ at each timestep $i$, thereby achieving effective alignment of the attention with the trajectory throughout the video sequence.

\paragraph{Temporal Smoothness Function.}\quad 
To ensure temporal coherence and avoid abrupt changes in motion, we introduce a temporal smoothness function across frames, defined as the expectation of squared differences in attention maps over consecutive frames:
\begin{equation}
\label{temporal_loss}
\mathcal{T}_s = \mathbb{E}_{i} \left[ \| A_{i,:,:}^{(\gamma)} - A_{i-1,:,:}^{(\gamma)} \|^2 \right],
\end{equation}
where $A_{i,:,:}^{(\gamma)}$ and $A_{i-1,:,:}^{(\gamma)}$ are attention maps for consecutive frames $i$ and $i-1$. Optimizing this function penalizes large frame-to-frame attention changes, promoting smooth motion across frames.

\paragraph{Gradient Update.}\quad 
The latent variable $\boldsymbol{z}^{(t)}$ at diffusion timestep $t$ is iteratively updated using the gradients of the combined energy functions:
\begin{equation}
\label{gradient_update}
\boldsymbol{z}^{(t)} \leftarrow \boldsymbol{z}^{(t)} - \sigma_t^2 \eta \nabla_{\boldsymbol{z}^{(t)}} \sum_{i=1}^{L} \sum_{\gamma \in \Gamma} \left( E_{i}\left(A_{i}^{(\gamma)}, B_{i}, n\right) + \lambda \mathcal{T}_s \right),
\end{equation}
where $\eta > 0$ controls the guidance strength, $\lambda$ controls the scale of the temporal smoothness function, and $\sigma_t = \sqrt{\frac{1 - \alpha_t}{\alpha_t}}$ adjusts for the noise level at diffusion timestep $t$. Scaling by $\sigma_t^2$ ensures that early diffusion timesteps with high noise receive weaker guidance, which intensifies as noise decreases in later timesteps, refining latent representations near the end of the diffusion process.

To generate a video, we alternate between gradient updates and denoising steps, iteratively refining the latent variable $\boldsymbol{z}^{(t)}$ to align cross-attention maps with the intended trajectory. The combination of iterative guidance and smoothness enforcement allows the model to produce coherent video sequences with precise spatial control.

\subsection{Motion Intensity Modulator}

The~\textit{Motion Intensity Modulator} (MIM) enables controlled adjustment of motion intensity in generated videos. During training, we compute the optical flow~\cite{optical_flow} strength between consecutive frames to quantify motion intensity, using it as a conditioning input for the diffusion model. This conditioning allows the model to capture varying levels of motion intensity, guiding the generation process based on the specified motion dynamics.

\paragraph{Optical Flow Map Generator.}\quad 
To quantify motion intensity in videos, we compute optical flow maps that capture the temporal dynamics between consecutive frames. These maps encode pixel-wise motion direction and magnitude, providing a structured representation of movement. For each video, frames are processed sequentially and converted to grayscale to facilitate efficient computation using the Farneback method~\cite{farneback2003two}. This approach estimates motion by analyzing pixel displacements between adjacent frames. The resulting magnitude values serve as a measure of motion intensity, which we normalize and discretize into integer levels ranging from $1$ to $10$. 

\paragraph{Motion Intensity Encoder.}\quad 
To incorporate the motion intensity into our model, we convert the optical flow strength of videos into embeddings. Similar to~\cite{kandala2024pix2gif}, rather than feeding the motion intensity as a raw numerical value, we transform it into a word form during training. This aligns with the CLIP model’s preference for text-based representations, as CLIP serves as our text encoder.

The motion intensity is embedded through a learned embedding layer \( \mathcal{M} \), which maps the motion intensity to an embedding vector \( c_M \). This vector is duplicated and concatenated with itself to ensure that its influence is strong enough when combined with the text embeddings.
\paragraph{Training and Loss Function.}\quad 
The model is trained by minimizing the following loss function:
\begin{equation}
\mathcal{L}_{LDM}^{\prime} = \mathbb{E}\left[\left\|\epsilon - \epsilon_\theta\left(z_t, t, c_T, c_M\right)\right\|_2^2\right].   
\end{equation} Here, \( c_M \) represents the motion intensity condition and \( c_T \) the text condition, which together provide a comprehensive conditioning signal for guiding the generation of temporally consistent, motion-controlled videos.

\paragraph{Inference.}\quad 
During inference, for motion intensity guidance, Mojito takes two inputs: a text prompt, and the motion intensity. These inputs are processed in two pathways: one directly through the diffusion model and the other through the MIM module. The text embedding $c_T$ is obtained by processing the text prompt via the CLIP text encoder, while the motion embedding $c_M$ is generated through the motion intensity embedding layer. The final conditioning input $c_L = c_T + c_M$ combines the text and motion embeddings, allowing the generation process to be guided with both semantic and motion-based conditioning.

\subsection{Model Backbone}
The backbone of~Mojito is based on a Conv-Spatial-Temporal Transformer architecture. The model structure consists of input blocks, a mid-block, and output blocks, each containing convolutional layers, temporal transformers, and spatial transformers. This design enables handling motion control in both the temporal and spatial dimensions. The details of model backbone are given in the~\textsc{Appendix}.

The~\textit{Temporal Transformer} processes video data along the time axis, capturing dependencies between frames by reshaping the input into temporal sequences. This block includes relative position encoding and optional causal masking for unidirectional attention, making it effective for capturing motion patterns across frames; The~\textit{Spatial Transformer} captures spatial relationships within each frame. Using linear or convolutional projections, this block reshapes and processes spatial information to learn localized patterns, which is essential for high-resolution video generation. Cross-attention layers enable alignment with external context when available, further refining spatial coherence in the generated content. The~\textit{Directional Motion Control} (DMC) module operates within Spatial Transformer to enforce trajectory-based motion guidance.

\section{Experiments}
\subsection{Datasets}
We use Panda70M~\cite{panda70m}, InternVID~\cite{internvid}, and Mira~\cite{ju2024miradata} to train both the~Mojito backbone and the Motion Intensity Modulator module. From Panda-70M, we extract a high-quality 10M subset for training. Additionally, we process 1,000 videos from Panda-70M for evaluation of motion intensity modulation. For motion direction guidance evaluation, we utilize the LaSOT dataset~\cite{fan2021lasot}, which provides paired text descriptions, bounding box annotations, and corresponding video sequences. For InternVID and Mira, we process and use the entire dataset for training. Further details on dataset preprocessing and usage can be found in the~\textsc{Appendix}.

\begin{table}[tbp]
\centering
\caption{Quantitative evaluation of different methods.}
\label{tab:quantitative_results}
\resizebox{0.7\textwidth}{!}{
\begin{tabular}{lcccc}
\toprule
\multirow{2}{*}{Method} & \multicolumn{3}{c}{Direction} & \multicolumn{1}{c}{Intensity} \\ 
\cmidrule(lr){2-4} \cmidrule(lr){5-5}
 & mIoU $\uparrow$ & AP50 $\uparrow$ & CD $\downarrow$ & Motion Alignment $\downarrow$ \\ 
\midrule
\midrule
OpenSora~\cite{opensora} & 8.4 & 0.9 & 0.31 &  0.281\\
OpenSora Plan~\cite{opensora} & 8.1 & 1.0 & 0.33 &  0.377\\
VideoCrafter2~\cite{chen2024videocrafter2} & 11.7 & 3.2 & 0.28 & 0.209 \\
\midrule
\cellcolor[gray]{0.9} Mojito & \cellcolor[gray]{0.9} \textbf{26.0} & \cellcolor[gray]{0.9} \textbf{17.1} &  \cellcolor[gray]{0.9} \textbf{0.18}& \cellcolor[gray]{0.9}\textbf{0.089} \\
\bottomrule
\end{tabular}}
\end{table}

\begin{figure}[tb]
  \centering
\includegraphics[width=\linewidth]{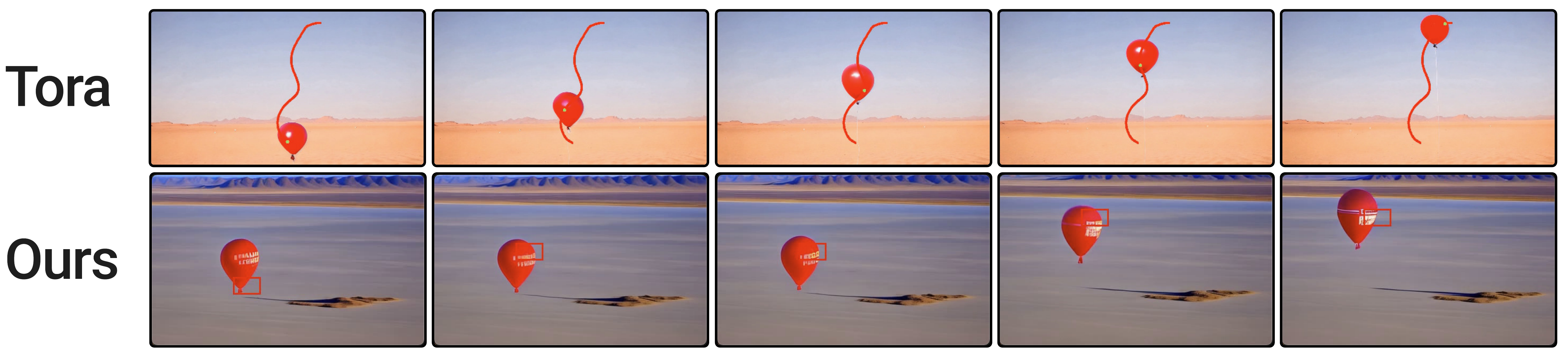}
\vspace{-4ex}
   \caption{Qualitative comparison of directional control with Tora. ~Mojito achieves motion control comparable to Tora while offering additional capabilities to specify objects and precise locations \textbf{without training}. The red bounding boxes, serving as inputs to~Mojito, guide the balloon to follow the specified trajectory.}
    \label{Fig:comparison}
\end{figure}

\subsection{Evaluation Metrics}
For quantitative evaluation, we employ commonly used metrics: Fréchet Video Distance (FVD)~\cite{unterthiner2019fvd} and CLIPSIM~\cite{dynamicrafter}, which calculates the average similarity across all generated frames in relation to the input caption. Specifically, we compute the FVD for a set of 300 videos, each consisting of 16 frames. 

To evaluate the effectiveness of directional guidance, we utilize the off-the-shelf OWL-ViT-large open-vocabulary object detector~\cite{oject_detector} to extract bounding box annotations from the generated videos. We then compute the Intersection-over-Union (IoU)~\cite{iou} between the detected and target bounding boxes to measure spatial alignment. We also compute the Centroid Distance (CD) as the distance between the centroid of generated objects and input bboxes, normalized to 1.
For motion intensity control, we compute the difference between the detected average optical flow in the generated video and the target motion intensity.

\subsection{Implementation Details}
\noindent \textbf{Model Backbone.}\quad  For the Variational Autoencoder (VAE) framework, the encoder processes input frames at a resolution of 320$\times$512, progressively downsampling through a series of residual blocks and convolutional layers with channel multipliers of $(1, 2, 4, 4)$. This results in a latent representation with a spatial resolution of $40\times64$. We use CLIP (ViT-H-14)~\cite{clip} as the text encoder, allowing for high-quality text-to-video alignment. For video processing, spatial and temporal downsampling factors of 8 are applied in both dimensions. The attention matrices (such as $q$ and $kv$) within the spatial transformer are structured as $(t, b, c)$, where $t$ represents the temporal length. Trajectory adjustments are applied only within the mid-block and the initial layers of the upsampling branch in the denoising U-Net~\cite{unet}, which provides a balance between maintaining video quality and achieving precise motion control. Additional implementation details and parameter settings are provided in the \textsc{Appendix}.

\noindent \textbf{Training.}\quad We use the DeepSpeed~\cite{deepspeed} framework to train our model, utilizing memory and computation optimizations for distributed training. The model training is carried out using the FusedLamb~\cite{fusedlamb} optimizer combined with the LambdaLRScheduler. We use a batch size of 4 and a base learning rate of $1.0 \times 10^{-5}$. More details can be found in the \textsc{Appendix}.

\subsection{Performance Evaluation}
\noindent \textbf{Quantitative Comparison.}\quad
We evaluate~Mojito against leading video generation models based on metrics for directional alignment and motion intensity alignment. For the baseline models, motion direction and intensity conditions are provided by rewriting the text prompts, as detailed in the \textsc{Appendix}. The results, summarized in Table~\ref{tab:quantitative_results}, demonstrate that~Mojito consistently outperforms all other approaches across both direction-related and intensity-related metrics, showcasing its superior ability to control motion direction and intensity.

\begin{table*}[t]
\centering
\caption{Human evaluation between Mojito, OpenSora~\cite{opensora}, OpenSora plan~\cite{opensoraplan}, and VideoCrafter2~\cite{chen2024videocrafter2}. It includes 2400 comparisons on 400 video pairs. Each video pair contains 5 ratings from different human annotators on Amazon Turk. The detailed setup is given in the \textsc{Supplementary Materials}.}
\label{HumanEvaluation}
\resizebox{\textwidth}{!}{
\begin{tabular}{@{}l|ccc|ccc|ccc@{}}
\toprule 
& Mojito & Tie & OpenSora   & Mojito & Tie & VideoCrafter2   & Mojito & Tie & OpenSora plan   \\
\midrule
\midrule
Direction Alignment &  \textbf{84.8\%} & 3.3\% & 12.0\% & \textbf{88.0\%} & 5.0\% & 7.0\% & \textbf{86.8\%} & 3.0\% & 10.3\% \\
Intensity Alignment & \textbf{46.8\%} & 22.3\% & 31.0\% & \textbf{40.0\%} & 28.5\% & 31.5\% & \textbf{43.5\%} & 23.8\% & 32.8\% \\
\bottomrule
\end{tabular}}
\end{table*}

\begin{figure}[tbp]
  \centering
\includegraphics[width=\linewidth]{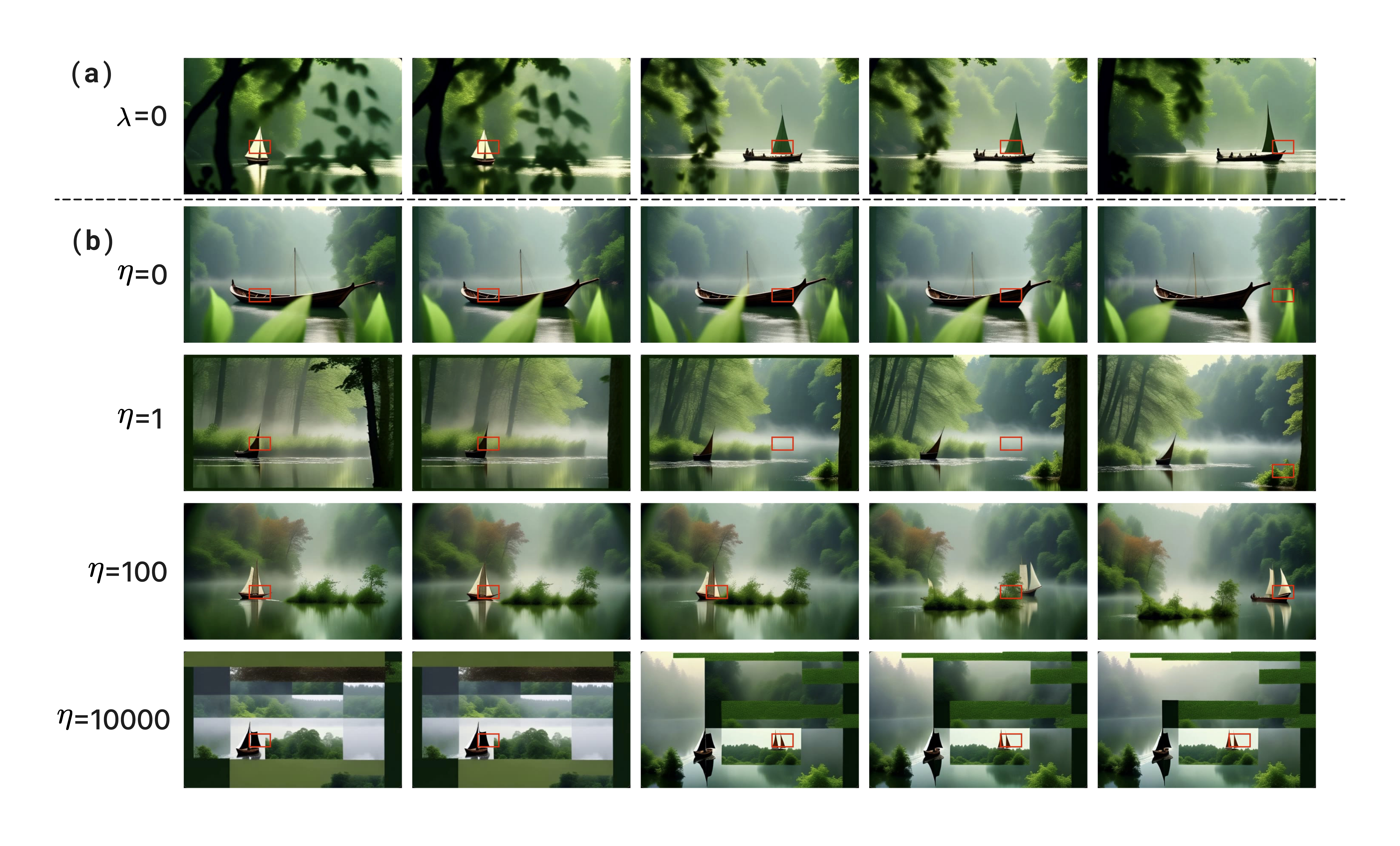}
    \caption{ (a) \textbf{Ablation Study on Temporal Smoothness Loss}: Without temporal smoothness loss, the generated sailboat exhibits inconsistencies across frames. (b) \textbf{Ablation Study on Guidance Strength}: Adjusting the guidance strength demonstrates a trade-off between video quality and trajectory alignment.}
    \label{Fig:direction}
\end{figure}

\begin{figure}[tbp]
  \centering
\includegraphics[width=\linewidth]{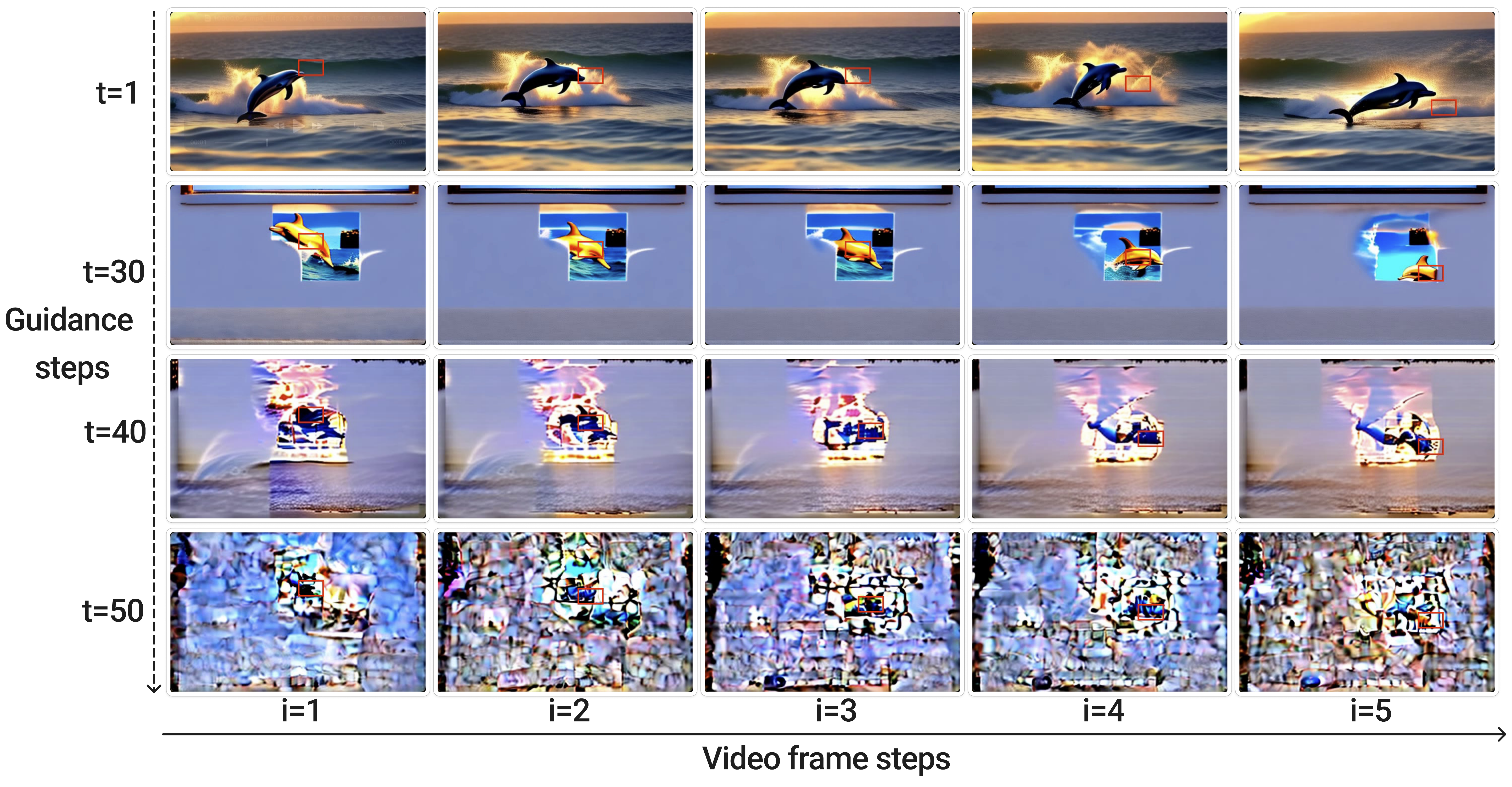}
\vspace{-4ex}
    \caption{ Ablation study on the effect of guidance steps. While increasing the number of guidance steps improves trajectory alignment, excessive guidance steps can degrade overall video quality.}
    \label{Fig:guidance_steps}
\end{figure}

\begin{table}[tbp]
\centering
\caption{Quantitative evaluation of trajectory alignment and video quality with different total numbers of guidance steps. Setting higher guidance steps generally enhances trajectory alignment while maintaining video quality.}
\label{tab:ablation_direction}
\resizebox{0.77\textwidth}{!}{
\begin{tabular}{lccccc}
\toprule
\multirow{2}{*}{Method} & \multirow{2}{*}{FVD $\downarrow$} & \multirow{2}{*}{CLIPSIM $\uparrow$} & \multicolumn{3}{c}{Direction} \\
\cmidrule(lr){4-6}
 & & & mIoU $\uparrow$ & AP50 $\uparrow$ & CD $\downarrow$ \\
\midrule
\midrule
$t=1$ & 423 & 0.2520 & 18.0 & 11.2 & 0.34  \\
$t=5$ & \textbf{422} & 0.2511 & 24.5 & 14.4 & 0.32  \\
\rowcolor[gray]{0.9} 
$t=10$ & \textbf{422} & \textbf{0.3700} & 26.0 & 17.1 & \textbf{0.28} \\
$t=30$ & 479 & 0.1880 & \textbf{28.3} & \textbf{18.9} & \textbf{0.28} \\
\bottomrule
\end{tabular}}
\end{table}

\begin{table}[tbp]
\centering
\caption{Quantitative evaluation of different designs for MIM. Combining motion intensity embeddings with text embeddings achieves the best overall performance, balancing video quality, semantic similarity, and motion alignment. Treating motion intensity embeddings as a global conditional input shows good motion alignment but significantly degrades video quality.}
\label{tab:ablation_intensity}
\resizebox{\linewidth}{!}{
\begin{tabular}{l|ccc}
\toprule
Method & FVD $\downarrow$ & CLIPSIM $\uparrow$ & Motion Alignment $\downarrow$ \\
\midrule
\midrule
w/o MIM & 421 & 0.250 & 0.307 \\
Global Conditional Input & 438 & 0.250 & 0.097 \\ 
Direct Text Input & \textbf{422} & 0.256 & 0.174\\
\midrule
\rowcolor[gray]{0.9}Combined with Text Embedding& 423 & \textbf{0.273} & \textbf{0.089} \\
\bottomrule
\end{tabular}}
\end{table}

\noindent \textbf{Qualitative Evaluations on Directional Control.}
We present a qualitative comparison with Tora~\cite{zhang2024tora} in Figure~\ref{Fig:comparison}. The text prompt used is ``{A \textcolor{red}{red helium balloon} floats slowly upward into the sky over a barren desert and expansive Gobi landscape.}" The benefits of Mojito are threefold. \underline{First}, Mojito features a \textit{training-free capability}, enabling precise directional control at test time without requiring additional training data or fine-tuning. Unlike training-based approaches like Tora, Mojito can dynamically adjust object trajectories during sampling at inference time, ensuring robust control without added computational cost in training.
\underline{Second}, Mojito provides \textit{fine-grained directional control} by allowing users to specify motion paths for named objects. This level of control contrasts with baseline models like Tora, which lack the capability to follow object-specific directional prompts and typically follow arbitrary trajectories. In Figure~\ref{Fig:comparison}, we showcase scenarios where Mojito is given specific object names (e.g., ``red helium balloon") and precise locations, enabling it to guide each object’s motion along user-defined paths. \underline{Third}, the DMC module in Mojito is model-agnostic and easily pluggable into different architectures, including DiT-based models.

\noindent \textbf{Human Evaluation.}\quad
We conduct human evaluations to assess the effectiveness of~Mojito in both motion intensity control and directional control. For evaluating the Motion Intensity Modulator, we compare~Mojito with baselines by rewriting the text prompts for baselines to explicitly specify the desired intensity level. The results of this evaluation are presented in Table~\ref{HumanEvaluation}.

\subsection{Ablation Study}

\subsubsection{Designs in Directional Motion Control}
In this section, we analyze the effects of various configurations for directional control, specifically examining different values of guidance strength ($\eta$), temporal smoothness function ($\lambda$), and guidance steps. The results are presented in Figure~\ref{Fig:direction}. The text prompt used is ``{A vintage wooden {sailboat} glides steadily down a mist-covered river.}" and ``A {dolphin} leaps through the waves of the ocean at sunrise."

\noindent \textbf{Temporal Smoothness Function.}\quad 
When $\lambda = 0$ (no temporal smoothness function), the generated video exhibits noticeable changes in views and inconsistent sailboat appearances. 

\noindent \textbf{Guidance Strength.}\quad 
With $\eta = 0$ (no guidance strength), the generated ``sailboat" does not follow specified bounding boxes at all. As $\eta$ increases, the sailboat begins to follow bounding boxes more closely. The best results reaches with $\eta = 100$, where the sailboat’s trajectory aligns well with the bounding boxes while maintaining good visual quality. However, setting $\eta$ to an excessively high value (e.g., $\eta = 10000$) results in severe degradation of video quality, with visible artifacts such as square patterns.

\noindent \textbf{Guidance Steps.}\quad
We perform an additional experiment to evaluate the effect of varying the number of guidance steps on video quality and trajectory alignment. The results, shown in Table~\ref{tab:ablation_direction}, indicate that increasing the number of guidance steps from $t=1$ to $t=10$ generally enhances trajectory alignment without significantly impacting video quality. However, when the number of guidance steps becomes too large (e.g., $t \geq 30$), the video quality deteriorates substantially. Qualitative results illustrating this effect are provided in Figure~\ref{Fig:guidance_steps}.

\subsubsection{Alternative Choices for Motion Intensity Modulator}

In this section, we explore various designs for the~\textit{Motion Intensity Modulator} (MIM). The primary approach in this paper involves combining a motion intensity embedding with text prompt embeddings and feeding them into~Mojito. Additionally, we experimented with several alternatives:

\noindent \textbf{Without MIM.}\quad Removing MIM during training and testing. In this setup, we perform inference by rewriting the text prompts to include the intensity level. 

\noindent \textbf{Global Conditional Input.}\quad Combining the intensity embedding with FPS and time embeddings, using this global embedding as a conditioning input for convolution layers. 

\noindent \textbf{Direct Text Input with Fine-tuning.}\quad In this approach, we rewrite the text prompts to include the motion intensity level and then fine-tune the model using this new data. Both training and inference are guided by the modified text input that explicitly specifies the motion intensity.

To compare each design, we evaluated the alignment of optical flow between the generated and input videos, as well as video quality and semantic similarity metrics. The quantitative results for each design are shown in Table~\ref{tab:ablation_intensity}, indicate that combining motion intensity embeddings with text embeddings yields the best performance. This demonstrates that treating motion intensity as an additional conditioning input and incorporating it into the cross-attention alongside text embeddings is a highly effective strategy.

\section{Related Works}
\begin{figure*}[tb]
  \centering
\includegraphics[width=0.95\textwidth]{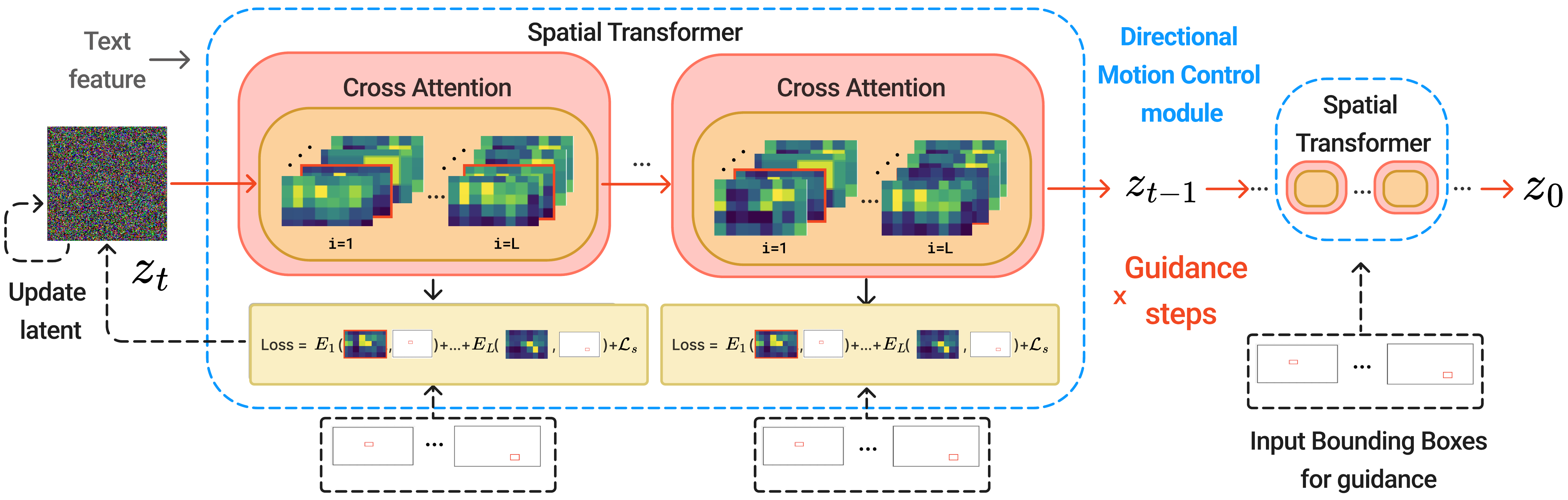}
    \caption{Overview of the~\textit{Directional Motion Control} module. The cross-attention map for the chosen word token in the given guidance step is marked with a red border. We compute the energy function and perform backpropagation during inference time to update latents. }
    \label{Fig:dmc}
\end{figure*}

\subsection{Diffusion-based Text to Video Generation}
With the rapid development of generative models, particularly diffusion models, numerous breakthroughs have been achieved in fields such as image generation~\cite{stable_diffusion} and video generation~\cite{latentvideodiffusion,easyanimate2024xu,opensora,opensoraplan,chen2024videocrafter2,yang2024cogvideox,vchitect2024}.  Regarding text-to-video (T2V) models, Make-A-Video~\cite{singer2022makeavideo}, Imagen Video~\cite{imagen2022} are cascaded models, while most
other works, such as LVDM~\cite{latentvideodiffusion}, ModelScope~\cite{wang2023modelscope}, and Align your Latents~\cite{align_latent}, are Stable Diffusion-based
models. They extend the SD framework to videos by incorporating temporal layers to ensure temporal consistency among frames and the spatial parameters are inherited from the pretrained SD UNet. ~Mojito follows a similar architectural approach, using spatial transformers for spatial information and accepting directional control, and incorporating temporal transformers to handle temporal coherence.

\subsection{Controllable Text-to-Visual Generation}

Recent advancements in text-to-visual generation have enabled more precise and stable control over generated outputs. In image generation, several models~\cite{t2iadapter, composer, unicontrolnet, unicontrol} have been developed to enhance control by leveraging semantic guidance from text prompts alongside various structural conditions. Some approaches utilize attention maps and bounding boxes to manage image layouts, enabling generation based on regional specifications~\cite{Chefer2023AttendandExciteAS, yang2022reco, Li2023GLIGENOG, chen2023trainingfree,training-free-guidance}. ~\cite{mo2024freecontrol} presents a training-free method for controllable image generation; In video generation, which involves additional complexity due to motion, recent works have explored methods for motion control. Training-based methods, such as those in~\cite{zhang2024tora, motionctrl}, allow for trajectory-based motion control, while~\cite{jain2024peekaboo} achieves video control through designed masking in the diffusion process. Unlike these approaches, our method leverages attention maps in a training-free manner for directional motion control. Our model can control the movement of specific objects within the generated video, and uniquely, it enables motion intensity control—providing a new level of flexibility in video generation not addressed in previous works.

 \chapter{Generating the 4D World}

\begin{figure}[!t]
    \centering
\includegraphics[width=\textwidth]{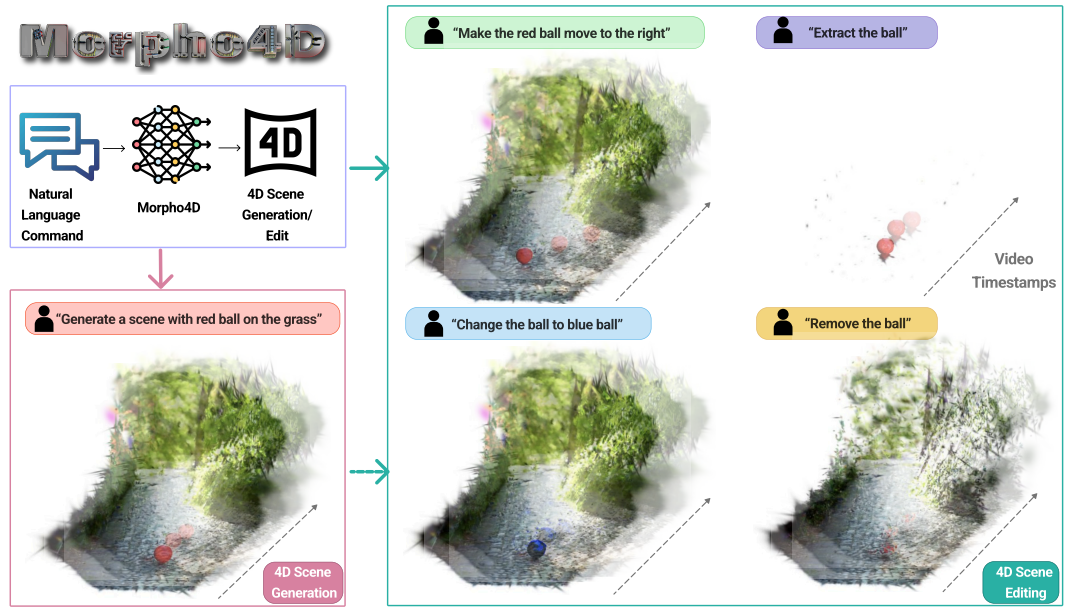} 
    \caption{Morph4D is a fully natural language-driven 4D scene generation engine that enables generation and editing of 4D scenes based on Language commands. Given a natural language input, Morph4D constructs a 4D scene and provides a unified framework for multiple tasks, including high-quality scene generation, interactive modification of object motion and appearance, and object extraction or removal.}
    \label{fig:teaser}
\end{figure}

\section{Introduction}
Recent advancements in video generation foundation models~\cite{kling2023,nvidia2025cosmos,chen2024videocrafter2,wan2.1, kong2024hunyuanvideo} have demonstrated remarkable capabilities by scaling models and data to unprecedented levels. As noted in Sora's report~\cite{sora}, ``\textit{Scaling video generation models is a promising path towards building general purpose simulators of the physical world}", state-of-the-art video generation models not only produces high-fidelity and surreal videos but has also ignited interest in using text-to-video generation models as world simulators~\cite{wang2024world,vidgen_worldmodel,wang2023world}. 

However, existing approaches remain limited to 2D generation, primarily relying on diffusion models or autoregressive models~\cite{xiang2024pandora,wang2023world,tian2024visual}, restricting them to single-view observations and non-interactive simulations. These limitations prevent such models from accurately capturing the true complexity of dynamic, multi-view environments.   

This naturally raises two fundamental questions:
\begin{itemize}
    \item \emph{Can the generative capability of text-to-video models be fully extended to 4D (spatial-temporal) scenarios?}  
    \item \emph{Can the generated 4D scenes be made interactive, controllable, and editable?}
\end{itemize}

Imagine an environment where a user can provide a natural language command, and the model generates a dynamic 4D scene—one that can be observed from multiple viewpoints and evolves over time. Beyond generation, these scenes should be editable, enabling users to modify object moving directions, change colors, extract or remove objects, as illustrated in Figure~\ref{fig:teaser}. Such a system would have wide-ranging applications, including video games, virtual reality (VR), and synthetic data generation for training vision-language models (VLMs)~\cite{cabon2020virtual,deitke2022,puig2018virtualhome,clevr}.

However, achieving this vision presents several key challenges. First of all, the 4D scene representation under this situation is not well defined. Second, the text-to-video generation models are designed for 2D video generation and lack the capability to model dynamic multi-view content. Third, the existing models do not support control over generated object motion directions, nor do they allow interactive scene modifications such as color changes, object extraction, or deletion.  

In this paper, we propose a new text-to-4D framework. Our model, namely Morpho4D, takes a language command as input, and parsing and generate an editable 4D scene that can be viewed from different viewpoints. Specifically, we introduce the dynamic control submodule, a new attention-based module that can understand and integrate natural language instructions from humans and edit the objection dynamic motion trajectories, and the static edit submodule, which can change the color appearance of the scene, extract the object, and delete the object. Overall, our main contributions can be summarized as follows:
\begin{itemize}
    \item We propose a new framework, named Morpho4D,  consisting of the \textit{command parameterizer}, \textit{scene generating module}, and \textit{scene editing module}, capable of producing editable 4D scenes giving natural language instructions and can edit scenes using natural language commands. 
    \item Within the scene editing module, we introduce the Dynamic Control Submodule, enabling precise motion direction control of objects in the generated 4D scenes through language-driven commands. We also introduce the Static Edit Submodule, allowing users to modify object appearance (color), extract objects, and remove objects interactively.  
    \item Our experimental results demonstrate that Morpho4D produces high-fidelity 4D content comparable to real-world scenes while maintaining superior controllability and editability compared to existing approaches.  
\end{itemize}

\begin{figure*}[t]
  \centering
   \includegraphics[width=\linewidth]{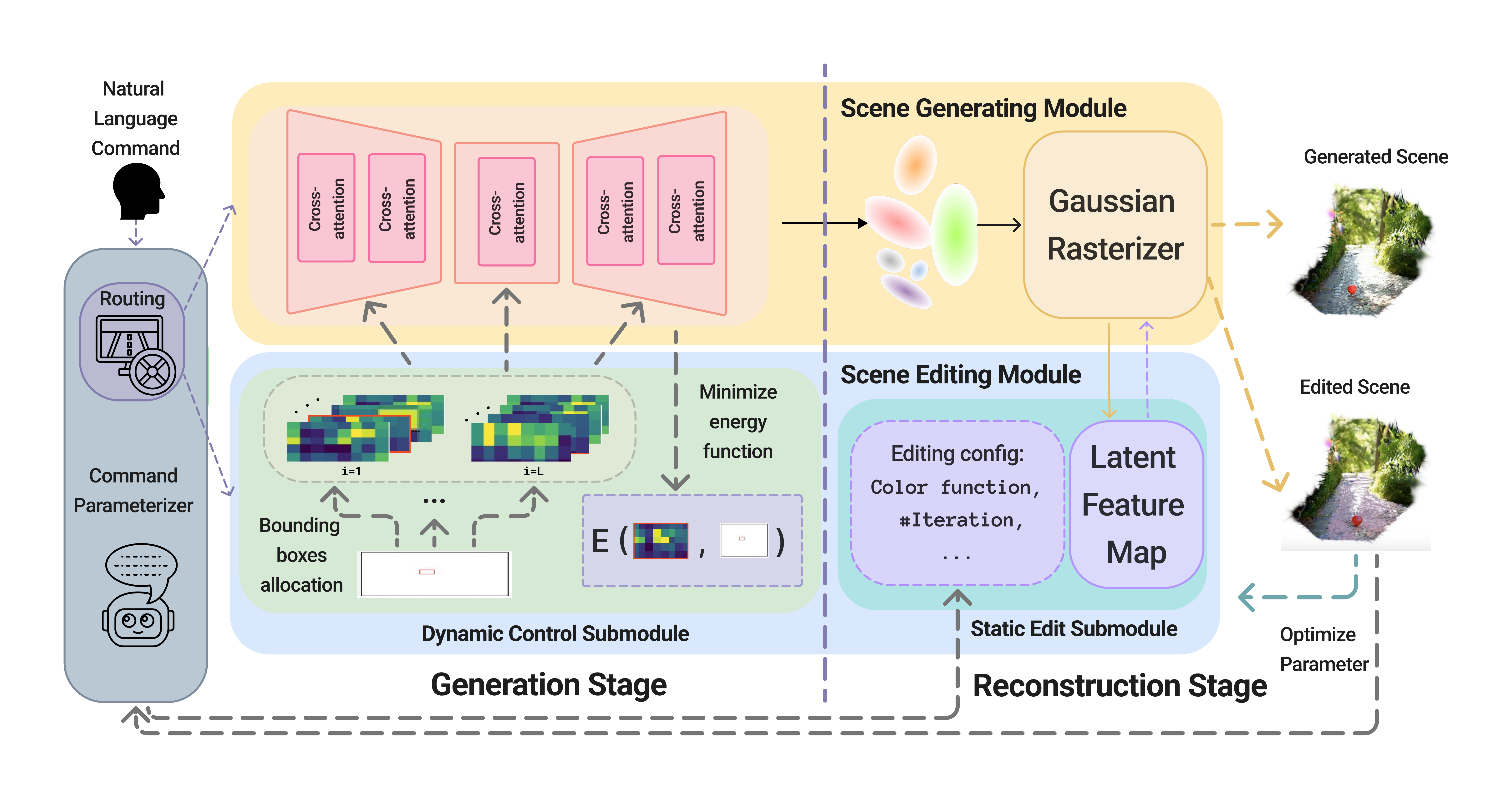}

   \caption{\textbf{Overview of the Morpho4D pipeline}. It consists of a \emph{command parameterizer} for natural language comprehension, a controllable \emph{scene generating module} which supports generation of 4D scenes following dynamic objects motion guidance, and an interactive \emph{scene editing module} for executing edits. }
   \label{fig:overall}
\end{figure*}

\section{Related Works}
\subsection{Diffusion Models for Visual Generation} 

Recent works in diffusion models have revolutionized visual generation across 2D, 3D, and 4D domains. In 2D, frameworks such as Stable Diffusion~\cite{stable_diffusion} and Imagen~\cite{imagen2022} enable high-fidelity text-to-image synthesis, while cascaded and latent diffusion approaches extend these capabilities to text-to-video generation~\cite{latentvideodiffusion,easyanimate2024xu,opensora,opensoraplan,chen2024videocrafter2,yang2024cogvideox,vchitect2024}. Building upon these successes, text-to-3D methods utilize Score Distillation Sampling to optimize neural scene representations by leveraging 2D diffusion priors~\cite{poole2022dreamfusion,tang2023dreamgaussian,li2024instant3d,lin2023magic3d,liu2023zero,sun2023dreamcraft3d}, or directly generate 3D models by 3D diffusion-based techniques~\cite{liu2024sherpa3d,xiang2024structured,chen20243dtopia,wu2024direct3d}. Recent efforts further extend diffusion-based methods to dynamic 4D scene generation by integrating video priors with 3D representations~\cite{singer2023text,yu20254real,bahmani2024tc4d,zhao2023animate124,bahmani20244d,ling2024align,ren2023dreamgaussian4d,xu2024comp4d,yin20234dgen}. While these approaches mostly focus on visual generation, our work introduces an interactive text-to-4D framework that not only generates high-fidelity spatiotemporal scenes but also enables editing and control through natural language instructions.

\subsection{4D Scene Reconstruction and Generation}

The advent of 3D Gaussian Splatting (3DGS)~\cite{kerbl20233d} marked a groundbreaking shift in efficient, high-fidelity 3D scene representation, excelling in novel view synthesis and reconstruction. Recent extensions to dynamic 4D settings~\cite{wu20234d,duan20244d,yang2023real,huang2024sc,liang2023gaufre,wang2024shape,lei2024mosca,mihajlovic2025splatfields,stearns2024dynamic} have further augmented its utility by integrating temporal deformation fields, enabling the modeling of temporal-consistent geometries and appearances. Building on this, numerous methods now tackle 4D scene reconstruction directly from monocular videos. Compared with NeRF-based counterparts~\cite{xian2021space,tretschk2021non,pumarola2020dnerf,li2020nsff,park2021nerfies,gao2021dynamic,park2021hypernerf,wu2022d}, the recent works including Shape-of-Motion~\cite{wang2024shape} and MoSca~\cite{lei2024mosca} have enhanced the reconstruction stability and generalizability from arbitrary video inputs by using Gaussian Splatting~\cite{kerbl20233d,lu2024scaffold,luiten2023dynamic}. Beyond reconstruction, the efficiency of 4DGS has catalyzed its adoption in 4D content generation~\cite{yin20234dgen,ren2024l4gm,ling2023align}, with recent works~\cite{yin20234dgen,li20244k4dgen,ren2023dreamgaussian4d,ren2024l4gm,sun2024eg4d} achieving remarkable quality under constrained settings—such as single-view, fixed-camera captures—by leveraging diffusion priors or score distillation.  Despite these advancements, existing approaches primarily focus on either 4D scene reconstruction or generation, without explicitly addressing the interactivity, controllability, and editability of those 4D scenes within a unified pipeline. Morpho4D bridges this gap, introducing a language-driven, multi-functional framework that seamlessly integrates high-quality generation, reconstruction, and intuitive editing.

\subsection{Language-guided Scene Editing} 
Recent advances in language-guided scene editing have shown promising results in static 3D settings by leveraging neural implicit representations and 2D diffusion priors for tasks such as appearance modification~\cite{chen2024gaussianeditor,wang2024gaussianeditor,zhuang2023dreameditor,khalid2024latenteditor,sella2023vox,xu2024gg,fang2024chat,wu2024gaussctrl}, object replacement~\cite{bartrum2024replaceanything3d,zheng2025editroom,zhuang2024tip,gordon2023blended}, and removal~\cite{zhuang2023dreameditor,gaussian_grouping,qiu2024feature}. In the realm of 4D editing, methods like 4D-Editor~\cite{jiang20234d}, CTRL-D~\cite{he2024ctrl}, Control4D~\cite{shao2024control4d}, and Instruct 4D-to-4D~\cite{mou2024instruct} extend these techniques to dynamic scenes by ensuring temporally consistent appearance edits across frames. However, these approaches primarily modify visual attributes while preserving the original motion dynamics. In contrast, our work advances this line of research by introducing motion control through natural language instructions, thereby enabling not only appearance modifications but also explicit edits to the underlying motion patterns in 4D scenes.

\section{Morpho4D: An Interactive, Controllable, and Editable Text-to-4D World Simulator}
Morpho4D is a unified framework that integrates multiple functionalities for generating and editing 4D scenes from natural language input. As illustrated in Figure~\ref{fig:overall}, Morpho4D consists of three core modules: the \textit{Command Parameterizer} Module, the 4D \textit{Scene Generation Module}, and the 4D \textit{Scene Editing Module}. The Command Parameterizer Module serves as the interface between natural language input and system execution. It interprets user instructions, routes them to the appropriate module, and converts them into structured, executable commands. The 4D Scene Generation Module is responsible for generating dynamic 4D scenes based on textual descriptions, capturing spatial and temporal instructions. The 4D Scene Editing Module enables interactive modifications, allowing users to adjust motion trajectories, alter object appearances (color), and manipulate scene elements (delete or extract) through natural language instructions. In the following sections, we will introduce these modules in detail.

\subsection{Preliminaries}
\paragraph{Video Diffusion Models}
\label{sec:preliminaries}

Video Diffusion Models (VDMs)~\cite{latentvideodiffusion,stable_video_diffusion} extend diffusion models to video generation by formulating a fixed forward diffusion process that progressively corrupts a 4D video sample $\mathbf{x}_0$ with noise in the latent space. This enables the model to learn a reverse denoising process to recover the original video. 

The forward diffusion process consists of $T$ timesteps, where noise is gradually added to the clean data $\mathbf{x}_0$ through a Markovian parameterization:
\begin{equation}
q(\mathbf{x}_t \mid \mathbf{x}_{t-1}) = \mathcal{N}(\mathbf{x}_t; \sqrt{1 - \beta_t} \mathbf{x}_{t-1}, \beta_t \mathbf{I}),
\end{equation}
\begin{equation}
q(\mathbf{x}_t \mid \mathbf{x}_0) = \mathcal{N}(\mathbf{x}_t; \sqrt{\bar{\alpha}_t} \mathbf{x}_0, (1 - \bar{\alpha}_t) \mathbf{I}),
\end{equation}
where $\beta_t$ is a predefined variance schedule during diffusion sampling, $\alpha_t = 1 - \beta_t$, and $\bar{\alpha}_t = \prod_{i=1}^{t} \alpha_i$. 
The reverse process then attempts to reconstruct $\mathbf{x}_{t-1}$ from $\mathbf{x}_t$ by learning a denoising distribution:
\begin{equation}
p_{\theta} (\mathbf{x}_{t-1} \mid \mathbf{x}_t) = \mathcal{N} (\mathbf{x}_{t-1}; \boldsymbol{\mu}_{\theta} (\mathbf{x}_t, t), \boldsymbol{\Sigma}_{\theta} (\mathbf{x}_t, t)).
\end{equation}
Here, the mean $\boldsymbol{\mu}_{\theta}$ and variance $\boldsymbol{\Sigma}_{\theta}$ are the estimated Gaussian
mean and variance predicted by the denoising network. The update step is typically computed as:
\begin{equation}
\mathbf{x}_{t-1} = \frac{1}{\sqrt{\alpha_t}} \left( \mathbf{x}_t - \sqrt{1 - \alpha_t} \epsilon_\theta(\mathbf{x}_t, t, \mathbf{c}) \right) + \sigma_t \mathbf{z},
\end{equation}
where $\alpha_t$ is the noise schedule coefficient, $\sigma_t$ is the stochastic noise factor, and $\mathbf{z} \sim \mathcal{N}(0, I)$ represents Gaussian noise injected at each step for improved sample diversity. The latent feature update can be modified to control the generation direction.

\paragraph{3D Gaussian Splatting}
In dynamic scene reconstruction approaches~\cite{wang2024shape, lei2024mosca, stearns2024dynamic}, the scenes was represented with dynamic 3D Gaussians~\cite{luiten2023dynamic}—a set of persistent 3D Gaussians~\cite{kerbl20233d} that deform over time to model motion. These representations cann efficiently capture spatiotemporal variations in monocular video reconstructions. To created 4D scenes, we build upon MoSca~\cite{lei2024mosca}, which resconstruct 4D scenes with single-view partial observations by leveraging priors from 2D foundation models~\cite{unidepth,zoedepth,teed2020raft,cotracker,harley2022particle} and by imposing regularization constraints on Gaussian motion trajectories. The key in the techniques is the use of 4D Motion Scaffold, a structured graph $(\mathcal{V}, \mathcal{E})$ that governs the deformation of individual 3D Gaussians $\mathcal{G} = \{G_j\}_{j=1}^n$.

\subsection{Language-Guided 4D Scene Generation}

\subsubsection{LLM as a Command Parameterizer}
Given an input natural language instruction $\mathcal{L}$, Morpho4D employs a large language model (LLM) agent $\mathcal{A}$ to interpret the command, extract semantic attributes, and dynamically route the request to the appropriate execution module. The agent $\mathcal{A}$ formalizes this routing process by mapping the input $\mathcal{L}$ to an execution plan $\mathcal{P}$:

\begin{equation}
\mathcal{A}: \mathcal{L} \to \mathcal{P} = (\mathcal{M}, \mathcal{Q}),
\end{equation}

where $\mathcal{M} \in \{\text{GEN}, \text{EDIT}\}$ represents the routing decision, selecting either the scene generating module ($\mathcal{G}$) or the scene editing module ($\mathcal{E}$), and $\mathcal{Q}$ represents a set of structured queries extracted from the input.

\subsubsection{Scene Generating Module}
If the LLM agent $\mathcal{A}$ determines $\mathcal{M} = \text{GEN}$, it routes the request to the scene generating module $\mathcal{G}$. 

To construct the initial scene representation, we leverage state-of-the-art text-to-video generation models. However, existing models such as Cosmos~\cite{nvidia2025cosmos} and Hunyuan~\cite{kong2024hunyuanvideo} often struggle to accurately follow input motion directions, as illustrated in Figure~\ref{fig:failure}. To mitigate this issue, we introduce an inference-time guidance mechanism that dynamically adjusts motion trajectories while sampling from the conditional distribution: $p(z | y, \mathcal{T}, i) $, where $z$ represents the generated latent features, $y$ is the input text, and $\mathcal{T}$ specifies a predefined trajectory associated with motion-related tokens $y_n$. This adjustment ensures that generated objects move according to user-specified directions without requiring additional training. The LLM agent first parses the motion direction description from the natural language input into a structured trajectory representation and assigns corresponding bounding boxes. These bounding boxes serve as guidance inputs for the scene generator, ensuring that objects move in the specified direction and at the indicated speed within the generated 4D scene.

\paragraph{Bounding Box Definition.} 
For a given translated trajectory $\mathcal{T} = \{(x_{i}, y_{i}, t_{i})\}_{i=1}^{L}$, where $L$ is the number of key points, $(x_{i}, y_{i})$ represents the spatial location at time step $t_{i}$, we define the bounding box $B_{i}$ for frame $i$ as:
$B_{i} = \{ (x, y) : |x - x_{i}| \leq \Delta_x, |y - y_{i}| \leq \Delta_y \}$,
where $\Delta_x$ and $\Delta_y$ define the spatial tolerance, determining the size of the bounding box. These values are influenced by both object size and the motion attributes extracted from the language.

\paragraph{Frame-Wise Bounding Box Allocation.} 
To account for motion speed and direction, we introduce a velocity-dependent expansion factor. If the language prompt describes fast movement (e.g., "The car moves quickly to the right"), the bounding boxes are spaced farther apart between frames to reflect rapid displacement. We define the displacement vector between consecutive points as:
$v_{i} = \frac{\| (x_{i+1} - x_{i}, y_{i+1} - y_{i}) \|}{t_{i+1} - t_{i}}$,
where $v_{i}$ represents the velocity magnitude. The bounding box displacement between frames is then scaled by a velocity factor $\lambda(v_i)$:
$x_{i+1} = x_i + \lambda v_i \cdot (x_{i+1} - x_{i}), y_{i+1} = y_i + \lambda v_i \cdot (y_{i+1} - y_{i})$.
Here, $\lambda$ is a scaling hyper-parameter that increases with velocity, ensuring that high-speed moving objects specified in the text prompts receive more widely spaced bounding boxes across frames. This adaptive allocation ensures that motion dynamics described in natural language are accurately reflected in the generated scene.

\begin{figure}[t]
  \centering
  \includegraphics[width=\linewidth]{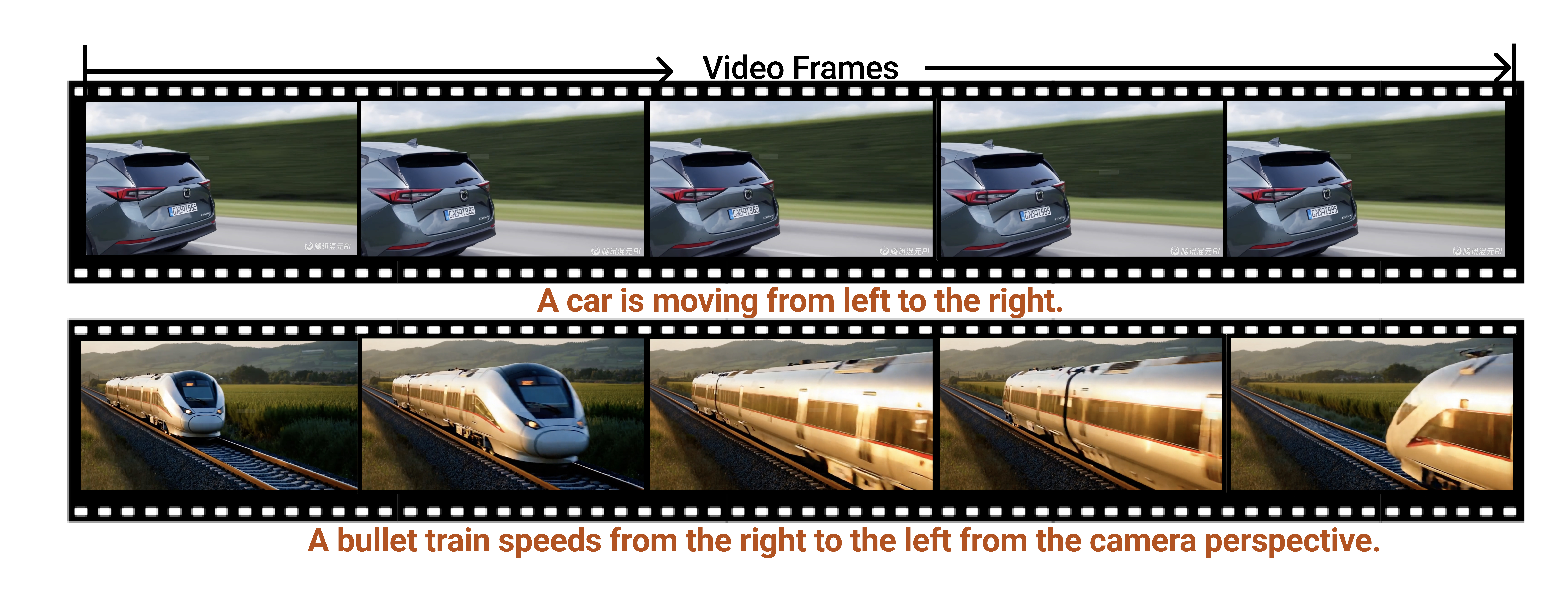}
  \caption{\textbf{Failure cases of state-of-the-art video generation models in adhering to spatial instructions from text prompts}. The generated object motions move in the opposite direction of the specified text prompt. The first row presents videos generated by Hunyuan~\cite{kong2024hunyuanvideo}, while the second row shows results from Cosmos~\cite{nvidia2025cosmos}.}
  \label{fig:failure}
\end{figure}

\paragraph{Dynamic Control Submodule.}  
To enforce directional control, we introduce the Dynamic Control Submodule with guidance bounding boxes, where we modify the cross-attention layers by biasing attention scores toward locations along $\mathcal{T}$. Specifically, at each sampling step $i$, we adjust the attention response for layers attending to motion-related text tokens. This is implemented using a trajectory-aligned attention weighting mechanism inspired by~\cite{he2024mojito}, which directs attention toward spatial regions prescribed by the trajectory. Figure~\ref{fig:overall} provides an overview of this modification. By dynamically altering attention weight distributions during generation, we effectively steer object placement and movement to align with input descriptions.

Controlling spatial layouts in generative models via cross-attention has been explored in 2D scenarios~\cite{prompt_to_prompt, training-free-guidance, training-layout-control}. We extend this approach by guiding attention maps to follow a predefined trajectory $\mathcal{T}$ over time. The cross-attention score $A_{u,n}$ measures the association between spatial location $u$ and text token $y_n$, with the sum over tokens constrained to one:

\begin{equation}
\sum_{n=1}^N A_{u,n} = 1.
\end{equation}

To enforce alignment with $\mathcal{T}$, we bias the attention maps to concentrate within the trajectory-defined bounding boxes $B_i$ at each timestep $i$. This is achieved through a frame-specific energy function:

\begin{equation}
E_{i}\left(A_{i}, B_{i}, n\right) = \left(1 - \frac{\sum_{u \in B_{i}} A_{i,u,n}}{\sum_u A_{i,u,n}}\right)^2,
\end{equation}
where $A_{i,u,n}$ represents the attention score at video timestamp $i$, spatial location $u$, and text token $y_n$. Minimizing this energy function encourages the attention distribution to remain within $B_i$, effectively guiding object placement frame-by-frame. During generation, we iteratively adjust attention maps at each denoising step to minimize $E_i$ and update latent scene features like~\cite{he2024mojito}, ensuring that the model remains aligned with $\mathcal{T}$ across successive time steps.

For models adopting the DiT~\cite{dit} architecture (e.g.,~\cite{nvidia2025cosmos}), input data is represented as a latent tensor of shape $T \times C \times H \times W$, where $T$ denotes the temporal dimension. The input video data undergoes 3D patchification via a linear projection layer, which extracts non-overlapping patches of size $(p_t, p_h, p_w)$, where $p_t$ is the patch size for the temporal dimension, $p_h$ is the height, and $p_w$ is the width, and is mapped into token embeddings for the denoiser network. This transforms the latent representation into a sequence of tokens of length $\frac{T H W}{p_t p_h p_w}$, ensuring compatibility with the model's spatiotemporal processing pipeline. To integrate directional guidance into the latent feature updates, we would modify the cross-attention map based on global visual features. Since the input undergoes 3D patch embedding in subsequent processing, we first reshape the latent feature to restore its original spatial-temporal ratio before patchification. This ensures that attention modifications remain spatially coherent and correctly aligned with the scene structure before tokenization.

This dynamic control submodule provides an effective mechanism for ensuring that generated objects follow the intended motion direction while maintaining structural consistency across frames. By integrating trajectory-aware cross-attention modifications within the model's latent space, we achieve improved control over object movement without requiring modifications to the underlying model architecture.

\paragraph{Scene Reconstruction.}
Starting from an initial generated scene representation, following~\cite{zhou2025feature4x, lei2024mosca}, we build a dynamic 3D representation that can effectively support the scene editing tasks. Given the monocular representation per frame from the 2D world generator $\mathcal{I}=\left\{I_1, \ldots, I_t\right\}$, we reconstruct the underlying dynamic 3D scene with a set of
dynamic 3D Gaussians, augmented with a unified latent feature embedding that jointly distills various 2D
foundation features useful for editing. We leverage dynamic 3D reconstruction to fuse multi-view and multi-frame 2D features into a unified 3D representation. 

To achieve this, we augment existing 3D Gaussian attributes with a latent feature $\mathcal{F}$. We learn $\mathcal{F}$ along with lightweight task-specific decoders $\{\mathcal{D}^1, \ldots, \mathcal{D}^{S}\}$, where the decoder maps the latent feature $\mathcal{F} \in \mathbb{R}^D$ to the editing feature space $\mathcal{F}^{s} \in \mathbb{R}^{D_s}$. 

During optimization, we attach a feature vector $f_j \in \mathbb{R}^{D}$ to each 3D Gaussian $G_j \in \mathcal{G}$, warp $G_j$ to the target timestep $\tau$ following~\cite{lei2024mosca}, and rasterize $f_j$ using the same approach as Gaussian color $c_j$ in~\cite{zhou2024feature}. The RGB and feature reconstruction at viewpoint $v$ and timestep $\tau$ are calculated via:
\begin{equation}
\begin{aligned}
\hat{I}_{\tau}^{v} &= \operatorname{rasterize}(v, \{\operatorname{warp}(G_j, \tau), c_j)\}_{G_j \in \mathcal{G}}), 
\end{aligned}
\end{equation}
\begin{equation}
\begin{aligned}
\hat{F}_{\tau}^{v} &= \operatorname{rasterize}(v, \{\operatorname{warp}(G_j, \tau), f_j)\}_{G_j \in \mathcal{G}}).
\end{aligned}
\end{equation}
The reconstructed feature map $\hat{F}_{\tau}$ (with viewpoint $v$ omitted for brevity) is passed through the corresponding decoder $\mathcal{D}^s$ to obtain the task-specific feature representation $\hat{F^{s}_{\tau}}$, which is supervised against the ground truth feature map obtained from the 2D encoder $\mathcal{E}^{s}$. The feature loss $L_{\text{feat}}$ is optimized with the original MoSca~\cite{lei2024mosca} loss terms:
\begin{align}
L_{\text{feat}} = \sum_{s=1}^{S} \operatorname{MSE}(\hat{F^s_{\tau}}, F^s_{\tau}), \\
\hat{F^{s}_{\tau}} = \mathcal{D}^s(\hat{F_{\tau}}), 
\end{align}
\begin{align}
\quad \quad F^s_{\tau} = \mathcal{E}^s(I_{\tau}).
\end{align}
Similar to~\cite{zhou2025feature4x}, 
We use an MLP-based decoder trained on 2D rendered feature maps but applied directly to 3D Gaussian features during inference.

\begin{figure}[tbp]
  \centering
\includegraphics[width=\linewidth]{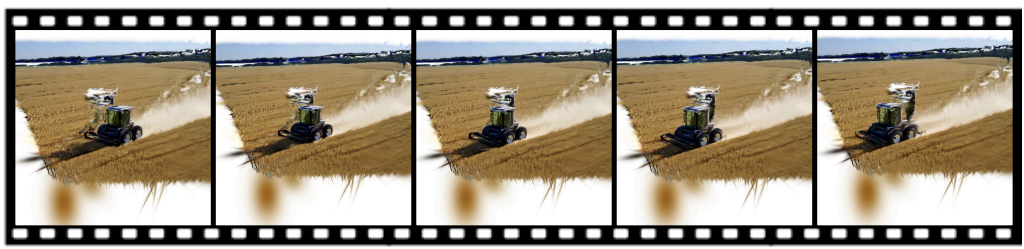}
  \caption{\textbf{Qualitative examples of 4D scene editing in Morpho4D for object motion control during the generation stage.} Morpho4D allows specifying different object motion directions in natural language forms and subsequently changes the scene to ensure objects move according to the given instructions.}
   \label{fig:motion_example}
\end{figure}

\begin{figure*}[htbp]
  \centering
\includegraphics[width=\linewidth]{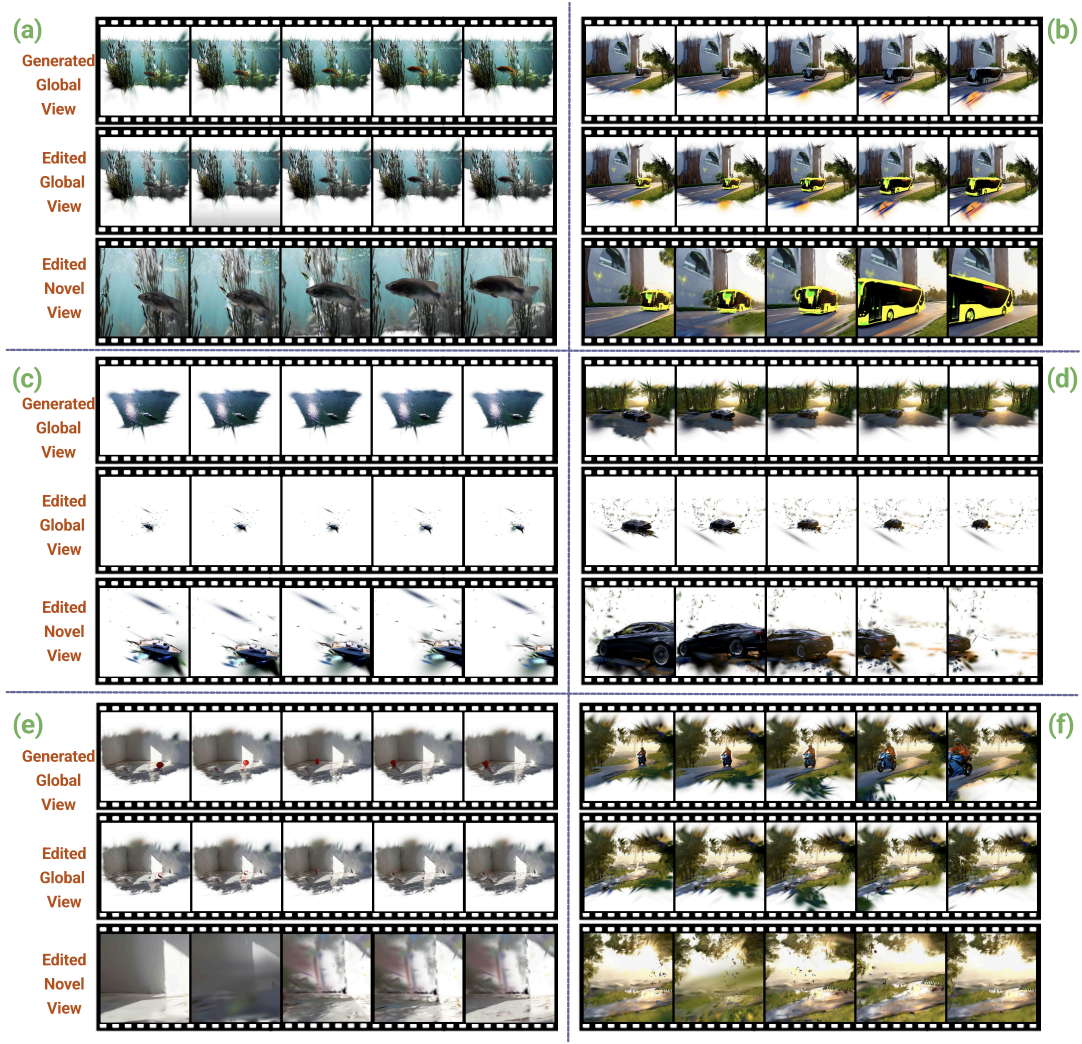}
  \caption{\textbf{Qualitative examples of 4D scene editing in Morpho4D during the reconstruction stage}.  
  (a) and (b) demonstrate \textit{color editing}, (c) and (d) show \textit{object extraction}, while (e) and (f) illustrate \textit{object removal}. In each subfigure: The first row shows the generated global view from the text prompt; The second row presents the global view after scene editing; The third row displays the novel view after editing.  The language commands for each example are as follows: \underline{(a)} "\textit{The fish swims through the crystal-clear waters from right to left}" to generate the scene, followed by "\textit{Make the color of the fish and seaweed black.}" \underline{(b)} "\textit{The bus is moving from right to left}" to generate the scene, followed by "\textit{Make the bus yellow.}" \underline{(c)} "\textit{A serene boat glides gracefully through tranquil waters from left to right}" to generate the scene, followed by "\textit{Extract the boat.}" \underline{(d)} "\textit{A car is moving from right to left through a serene sunlit landscape}" to generate the scene, followed by "\textit{Extract the car.}" \underline{(e)} "\textit{A small, vibrant red rubber ball is bouncing from right to left}" to generate the scene, followed by "\textit{Delete the ball.}" \underline{(f)} "\textit{A sleek black motorcycle is gliding effortlessly from right to left}" to generate the scene, followed by "\textit{Delete the motorcycle.}"}
   \label{fig:editing_example}
\end{figure*}

\begin{table*}[t]
\centering
\caption{Quantitative comparison between real-world scenes (Davis) and Morpho4D generated 4D scenes with two different base models.}
\label{tab:reconstruction_quality}
\resizebox{\textwidth}{!}{
\begin{tabular}{lccccc}
\toprule
\toprule
\textbf{Method} & \textbf{BRISQUE~\cite{brisque} $\downarrow$} & \textbf{NIQE~\cite{niqe} $\downarrow$} & \textbf{CLIP Similarity~\cite{clip} $\uparrow$} & \textbf{QAlign Quality~\cite{wu2023qalign} $\uparrow$} & \textbf{QAlign Aesthetic~\cite{wu2023qalign} $\uparrow$}\\
\midrule
\rowcolor[gray]{0.9}  \multicolumn{6}{l}{\textit{Overall Average}}\\
Davis (Real) & 31.639 & 3.551 & 0.250 & \textbf{3.432} & \textbf{2.254}\\
Backbone I & \textbf{18.380} & \textbf{3.286} & \textbf{0.263} & 3.350 & 2.114\\
Backbone II & 23.411 & 3.392 & 0.261 & 3.309 & 2.074\\
\midrule
\multicolumn{6}{l}{\textit{Example Scene Comparisons:}} \\
\midrule
\rowcolor[gray]{0.9}  \multicolumn{6}{l}{\textit{Scene: sheep}} \\
Davis (Real) & 18.090 & \textbf{2.173} & 0.266 & \textbf{4.371} & \textbf{2.891}\\
Backbone I & \textbf{9.638} & 3.454 & \textbf{0.300} & 3.623 & 1.918\\
Backbone II & 14.476 & 3.951 & 0.290 & 3.734 & 2.071\\
\midrule
\rowcolor[gray]{0.9} \multicolumn{6}{l}{\textit{Scene: snowboard}} \\
Davis (Real) & 34.904 & 3.719 & 0.269 & 2.715 & 1.991\\
Backbone I & \textbf{18.163} & \textbf{2.836} & \textbf{0.313} & \textbf{4.011} & \textbf{2.306}\\
Backbone II & 29.504 & 3.412 & 0.279 & 2.451 & 1.970\\
\midrule
\rowcolor[gray]{0.9}  \multicolumn{6}{l}{\textit{Scene: elephant}} \\
Davis (Real) & 16.815 & \textbf{2.317} & 0.284 & \textbf{4.048} & \textbf{2.764}\\
Backbone I & \textbf{15.140} & 3.171 & \textbf{0.308} & 3.634 & 2.511\\
Backbone II & 16.877 & 3.982 & 0.301 & 3.765 & 2.590\\
\bottomrule
\bottomrule
\end{tabular}}
\end{table*}

\subsection{Scene Editing Module}
If the LLM agent $\mathcal{A}$ determines $\mathcal{M} = \text{EDIT}$, it routes the request to the scene editing module $\mathcal{E}$, which modifies an existing 4D scene $\mathcal{S}$ based on the structured queries $\mathcal{Q}$. The editing operations include:
\begin{itemize}
    \item Appearance Editing: Changing the object color.
    \item Object Manipulation: Removing or extracting objects from the scene.
\end{itemize}
We further utilize the LLM agent to parse the input languages into executable commands and then perform follow-up executions, such as Color Editing, Object Removal, and Object Extraction. The modules can also be extended to support more executions.

The LLM agent is then used to optimize configuration parameters based on natural language prompts, perform precise queries, and iteratively refine results, enabling intelligent 4D scene manipulation. For instance, given a user instruction such as "Delete the ball" or "Change the ball's color to blue," the agent first parses the prompt and generates a set of configuration options with varying parameters relevant to the task. Specifically, it computes the probability $\mathbf{p}(\tau\mid j)$ of a 3D Gaussian being associated with a prompt $\tau$:
\begin{equation}
\mathbf{p}(\tau \mid j) = \frac{\exp (s)}{\sum_{s_i \in \mathcal{T}} \exp (s_i)},
\end{equation}
where $s$ is the cosine similarity between the semantic feature $f_j$ of the 3D Gaussian and the query feature $q(\tau)$:

\begin{equation}
s = \frac{f_j \cdot q(\tau)}{\|f_j\|\|q(\tau)\|}.
\end{equation}

The LLM agent then iterates over different threshold values to filter Gaussians with low probability scores and generate sample images using the 4D feature field, evaluating which configuration best aligns with the intended edit.

Once the optimal configuration is determined, the LLM agent $\mathcal{A}$ applies the selected parameters consistently across all frames in the video sequence, ensuring coherence in dynamic 4D scene editing. For instance, when modifying an object's color, the LLM agent iteratively adjusts the threshold to isolate and modify only the target object while preserving the rest of the scene. This process continues until either the input maximum iteration count parameter is reached or the threshold falls below a specified limit, ensuring precise and controlled edits. The details of this process are given in the Supplements.

\section{Experiments}  
\subsection{Datasets}  
We evaluate Morpho4D by comparing its generated 4D scenes against real-world videos from the DAVIS dataset~\cite{davis2017}. Specifically, we construct textual prompts from DAVIS annotations to generate corresponding 4D scenes and compare with reconstructed 4D scenes from DAVIS videos, assessing their quality relative to real-world counterparts.

\subsection{Quantitative Results on Generated Scene Quality}
We evaluate the quality of the reconstructed 4D scenes across 25 scenes using four metrics: 
\begin{itemize}  
    \item BRISQUE (Blind/Referenceless Image Spatial Quality Evaluator)~\cite{brisque} is a no-reference image quality assessment metric that measures perceptual distortions based on natural scene statistics. Lower BRISQUE scores indicate higher perceptual quality.  
    \item NIQE (Natural Image Quality Evaluator)~\cite{niqe} is another no-reference quality metric that evaluates the deviation of an image from learned natural scene statistics. Lower NIQE scores indicate better alignment.
    \item CLIP Similarity~\cite{clip} quantifies the semantic consistency between generated and real-world scenes by computing the cosine similarity between image embeddings extracted from the CLIP model.  
    \item QAlign~\cite{wu2023qalign} is the current state-of-the-art method for image quality assessment, leveraging a large multimodal model fine-tuned on publicly available image quality assessment datasets.

\end{itemize}
The results are summarized in Table~\ref{tab:reconstruction_quality}, where we compare our method on two video generation backbones with real-world scenes (Davis).  Backbone I is modified from CogVideoX~\cite{hong2022cogvideo}, and Backbone II is modified from Cosmos~\cite{nvidia2025cosmos} to achieve spatial control and 4D scene generation. The results demonstrate that our generated 4D scenes achieve comparable or better quality than real-world scenes using both two backbones, with Backbone I showing significantly better BRISQUE scores, slightly better NIQE scores, and improved CLIP similarity. This improvement can be attributed to the quality of Morpho4D generated 4D scenes, as well as the fact that Morpho4D generated camera views remain fixed throughout, resulting in higher metric scores, whereas real-world video sequences typically involve dynamic camera movement, which can introduce additional variance in evaluation.

\subsection{Qualitative Results}
We present qualitative results demonstrating the capabilities of our method in 4D scene generation and editing. As shown in Figure~\ref{fig:editing_example}, our approach generates 4D scenes with realism comparable to real-world environments. Additionally, it enables dynamic object motion editing, allowing objects to be manipulated to move in different directions based on user instructions (Figure~\ref{fig:motion_example}). Beyond motion control, our method supports appearance modifications, such as color editing (Figure~\ref{fig:editing_example}), ensuring consistent alterations across frames. Moreover, it facilitates structural modifications, including object extraction, which isolates objects without disrupting the surrounding scene, and object removal, which eliminates target objects while preserving background coherence. These results highlight the versatility of our approach in controlling, modifying, and refining 4D scene generation.

\part{Conclusion and Future Work}
\chapter{Conclusion}

This thesis has primarily focused on closing the loop between multimodal foundation models and world models. We explored the challenges and opportunities of perceiving, reasoning, and generating in the multimodal world, with a focus on integrating vision and language through foundation models and generative architectures. We began by investigating how large-scale vision-language models can be adapted effectively to real-world perceptual tasks using efficient training strategies, subspace-based adaptations, and compositional reasoning mechanisms.

We further introduced methods to enable counterfactual thinking in multimodal models, advancing their ability to go beyond pattern recognition and exhibit deeper reasoning. In addition, we proposed novel techniques for improving compositional generalization, aligning causal structures between vision and language, and facilitating better zero-shot generalization.

On the generation front, we introduced frameworks that bring graph-structured knowledge and control signals into the generative process, enabling fine-grained control over both text-to-image and text-to-video generation. We also proposed and benchmarked a unified evaluation framework for multimodal world models, culminating in the MMWorld benchmark, which provides a standardized setting for assessing multimodal perception, reasoning, and generation.

Altogether, this thesis contributes new insights, algorithms, and benchmarks for building more flexible, controllable, and cognitively capable multimodal systems. The contributions of this dissertation move us closer to the long-term goal of building multimodal world models that can not only perceive and generate, but also reason about, intervene on, and predict the complex dynamics of the world.

\chapter{Future Work}

While this thesis makes significant strides in advancing the perception, reasoning, and generation capabilities of multimodal foundation models, several open research directions remain. These represent promising avenues for developing more capable, robust, and generalizable multimodal world models.

\section{Scalable World Models with Long-Horizon Reasoning}

Despite recent progress in temporal modeling, current models often struggle to maintain coherence or perform reasoning over long sequences, especially in video and interactive tasks. Future research should explore scalable architectures that combine memory-based modeling with structured priors that promote long-term consistency. Moreover, integrating symbolic reasoning with neural representations could enable models to reason over higher-order temporal abstractions, such as plans, causes, and consequences. Applications include procedural video understanding, instructional generation, and simulating long-term outcomes in dynamic environments.

\section{Interactive Multimodal Agents}

Much of today's multimodal modeling focuses on passive perception and offline generation. A natural extension is to build interactive agents that perceive, act, and learn in real time. These agents should be capable of grounding language in action, formulating questions, interpreting feedback, and adapting to new tasks through interaction. Achieving this goal requires integrating components from reinforcement learning, visual navigation, continual learning, and dialog systems. This opens the door to practical applications such as robotic assistance, virtual tutors, and multimodal question-answering agents situated in simulation or real-world environments.

\section{Multimodal Causality and Explanation}

Understanding and modeling causality remains an underexplored yet crucial capability for multimodal systems. Future work should aim to develop models that not only predict associations but can also identify cause-effect relations, perform interventions, and generate explanations across modalities. For instance, given a change in an image or video, the model should be able to articulate the underlying cause or hypothesize plausible alternatives. Techniques such as causal representation learning, counterfactual data augmentation, and disentangled generative modeling could play key roles in this direction. Enhancing causal reasoning would not only improve generalization but also provide interpretability and robustness.

\section{Unified Control Across Modalities}

Controlling generative models has traditionally focused on specific modalities (e.g., text-to-image or text-to-video), but real-world scenarios often demand control across heterogeneous inputs and outputs. A promising future direction is the development of unified control frameworks that can seamlessly handle various control signals—such as textual prompts, motion trajectories, sketches, and semantic maps—across images, video, 3D scenes, and audio. Research here involves both architectural design (e.g., modular or compositional encoders) and training strategies that encourage generalization across modalities. Success in this area would lead to highly flexible and user-controllable generative systems.

\section{More Comprehensive Benchmarks}

While this thesis introduces the MMWorld benchmark to evaluate perception, reasoning, and generation in multimodal models, future work could expand its scope in several directions. These include incorporating real-world simulation environments (e.g., Habitat, AI2-THOR) and adding embodied interaction tasks. Evaluating models in closed-loop settings—where perception and action feed into each other—would better reflect real-world use cases. Additionally, incorporating more diverse evaluation metrics that assess causality, robustness, and controllability can provide a more holistic view of model capabilities.

\section{Incorporating More Modalities}

The current research primarily focuses on visual and linguistic modalities. Expanding the scope to include richer signals—such as depth maps, segmentation masks, scene graphs, SLAM outputs, and even tactile or audio data—can lead to more grounded and informed models. For instance, conditioning text-to-image or video generation on segmentation maps or depth could enhance spatial accuracy and realism. SLAM representations may enable scene-consistent generation across views or time. Future models should be designed to flexibly fuse diverse and structured inputs, supporting multimodal synthesis that reflects the richness and variety of the real world.

By addressing these directions, future research can further bridge the gap between current multimodal models and human-like world understanding. Ultimately, the goal is to develop general-purpose multimodal world models that not only perceive and generate, but also reason, interact, and adapt within complex and dynamic environments.

\endgroup


\bibliographystyle{plain}
\bibliography{main}

\end{document}

%% file: PEViT/fewshot.tex
\begin{table*}[t]
  \centering
  \resizebox{\columnwidth}{!}{
\begin{tabular}{lcccccccccccccccccccccccc}
\toprule
Method &
  \rotatebox[origin=c]{90}{Caltech101} & \rotatebox[origin=c]{90}{CIFAR10} & \rotatebox[origin=c]{90}{CIFAR100} & \rotatebox[origin=c]{90}{Country211} & \rotatebox[origin=c]{90}{DTD} & \rotatebox[origin=c]{90}{EuroSat} & \rotatebox[origin=c]{90}{FER2013} & \rotatebox[origin=c]{90}{FGVCAircraft} & \rotatebox[origin=c]{90}{Food101} & \rotatebox[origin=c]{90}{GTSRB} & \rotatebox[origin=c]{90}{HatefulMemes} & \rotatebox[origin=c]{90}{KittiDistance} & \rotatebox[origin=c]{90}{MNIST} & \rotatebox[origin=c]{90}{Flowers102} & \rotatebox[origin=c]{90}{OxfordPets} & \rotatebox[origin=c]{90}{PatchCamelyon} & \rotatebox[origin=c]{90}{SST2} & \rotatebox[origin=c]{90}{RESISC45} & \rotatebox[origin=c]{90}{StanfordCars} & \rotatebox[origin=c]{90}{VOC2007} &
 \rotatebox[origin=c]{90}{Ave Acc ($\uparrow$)} &
\rotatebox[origin=c]{90}{\#Params ($\downarrow$)} &
\rotatebox[origin=c]{90}{PE ($\uparrow$)} 
  \\ \midrule
 Fine-tuning &
  87.64 &
  \multicolumn{1}{r}{91.11} &
  \multicolumn{1}{r}{71.52} &
  \multicolumn{1}{r}{15.75} &
  \multicolumn{1}{r}{54.36} &
  \multicolumn{1}{r}{85.24} &
  \multicolumn{1}{r}{52.72} &
  \multicolumn{1}{r}{26.22} &
  \multicolumn{1}{r}{83.28} &
  \multicolumn{1}{r}{74.05} &
  \multicolumn{1}{r}{55.64} &
  \multicolumn{1}{r}{39.15} &
  \multicolumn{1}{r}{65.55} &
  \multicolumn{1}{r}{80.55} &
  \multicolumn{1}{r}{87.31} &
  \multicolumn{1}{r}{64.92} &
  \multicolumn{1}{r}{59.09} &
  \multicolumn{1}{r}{75.61} &
  \multicolumn{1}{r}{57.21} &
  \multicolumn{1}{r}{82.95} &
  \multicolumn{1}{r}{65.49} &
  87,878,739 &
  0.498 
  \\ 
Linear-probing &
  90.96 &
  \multicolumn{1}{r}{90.35} &
  \multicolumn{1}{r}{67.31} &
  \multicolumn{1}{r}{17.36} &
  \multicolumn{1}{r}{62.04} &
  \multicolumn{1}{r}{72.95} &
  \multicolumn{1}{r}{51.91} &
  \multicolumn{1}{r}{29.52} &
  \multicolumn{1}{r}{83.82} &
  \multicolumn{1}{r}{56.47} &
  \multicolumn{1}{r}{55.83} &
  \multicolumn{1}{r}{40.37} &
  \multicolumn{1}{r}{77.50} &
  \multicolumn{1}{r}{92.29} &
  \multicolumn{1}{r}{88.03} &
  \multicolumn{1}{r}{59.00} &
  \multicolumn{1}{r}{59.36} &
  \multicolumn{1}{r}{78.10} &
  \multicolumn{1}{r}{68.30} &
  \multicolumn{1}{r}{84.99} &
  \multicolumn{1}{r}{\textcolor{dblue}{66.32}} &
  \textbf{29,523} &
  \textcolor{dblue}{0.663} 
  \\
Adapter-tuning &
  90.18 &
  \multicolumn{1}{r}{90.14} &
  \multicolumn{1}{r}{73.57} &
  \multicolumn{1}{r}{16.83} &
  \multicolumn{1}{r}{57.13} &
  \multicolumn{1}{r}{67.97} &
  \multicolumn{1}{r}{41.76} &
  \multicolumn{1}{r}{30.52} &
  \multicolumn{1}{r}{83.58} &
  \multicolumn{1}{r}{58.50} &
  \multicolumn{1}{r}{48.91} &
  \multicolumn{1}{r}{37.18} &
  \multicolumn{1}{r}{80.34} &
  \multicolumn{1}{r}{90.78} &
  \multicolumn{1}{r}{86.52} &
  \multicolumn{1}{r}{59.92} &
  \multicolumn{1}{r}{58.70} &
  \multicolumn{1}{r}{79.22} &
  \multicolumn{1}{r}{67.68} &
  \multicolumn{1}{r}{82.22} &
  \multicolumn{1}{r}{65.08} &
 1,237,587  &
  0.647 
  \\
LoRA &
  87.64 &
  \multicolumn{1}{r}{90.52} &
  \multicolumn{1}{r}{69.69} &
  \multicolumn{1}{r}{17.12} &
  \multicolumn{1}{r}{50.16} &
  \multicolumn{1}{r}{74.03} &
  \multicolumn{1}{r}{51.04} &
  \multicolumn{1}{r}{20.01} &
  \multicolumn{1}{r}{83.76} &
  \multicolumn{1}{r}{42.96} &
  \multicolumn{1}{r}{55.88} &
  \multicolumn{1}{r}{48.05} &
  \multicolumn{1}{r}{61.36} &
  \multicolumn{1}{r}{74.28} &
  \multicolumn{1}{r}{85.49} &
  \multicolumn{1}{r}{63.20} &
  \multicolumn{1}{r}{57.04} &
  \multicolumn{1}{r}{62.09} &
  \multicolumn{1}{r}{54.89} &
  \multicolumn{1}{r}{80.33} &
  \multicolumn{1}{r}{61.48} &
 176,979 &
  0.614 
  \\
Compacter &
  89.02 &
  \multicolumn{1}{r}{79.96} &
  \multicolumn{1}{r}{44.33} &
  \multicolumn{1}{r}{28.22} &
  \multicolumn{1}{r}{52.93} &
  \multicolumn{1}{r}{50.48} &
  \multicolumn{1}{r}{35.46} &
  \multicolumn{1}{r}{41.13} &
  \multicolumn{1}{r}{78.28} &
  \multicolumn{1}{r}{66.90} &
  \multicolumn{1}{r}{47.60} &
  \multicolumn{1}{r}{57.72} &
  \multicolumn{1}{r}{85.82} &
  \multicolumn{1}{r}{88.29} &
  \multicolumn{1}{r}{79.23} &
  \multicolumn{1}{r}{61.83} &
  \multicolumn{1}{r}{64.22} &
  \multicolumn{1}{r}{63.76} &
  \multicolumn{1}{r}{64.79} &
  \multicolumn{1}{r}{75.84} &
  \multicolumn{1}{r}{62.79} &
 77,907&
  0.628 \\
  \hdashline
  KAdaptation &
  { 88.96} &
 \multicolumn{1}{r}{{ 90.03}} &
  \multicolumn{1}{r}{{ 73.92}} &
  \multicolumn{1}{r}{{ 17.53}} &
  \multicolumn{1}{r}{{ 63.97}} &
  \multicolumn{1}{r}{{ 76.25}} &
  \multicolumn{1}{r}{{ 47.45}} &
  \multicolumn{1}{r}{{ 30.04}} &
  \multicolumn{1}{r}{{ 84.38}} &
  \multicolumn{1}{r}{{ 80.71}} &
  \multicolumn{1}{r}{{ 55.86}} &
  \multicolumn{1}{r}{{ 42.29}} &
  \multicolumn{1}{r}{{ 85.20}} &
  \multicolumn{1}{r}{{ 93.19}} &
  \multicolumn{1}{r}{{ 89.05}} &
  \multicolumn{1}{r}{{ 63.39}} &
  \multicolumn{1}{r}{{ 59.18}} &
  \multicolumn{1}{r}{{ 79.96}} &
  \multicolumn{1}{r}{{ 70.21}} &
  \multicolumn{1}{r}{{ 84.49}} &
  \multicolumn{1}{r}{{ \textbf{68.92}}} &
  \textcolor{dblue}{79,699}&
  \textbf{0.689}
  \\
  \bottomrule
\end{tabular}
}
 \caption{ The averaged 5-shot experimental result comparison on 20 datasets from ELEVATER benchmark~\cite{vision_benchmark} in terms of accuracy (\%) and number of trainable parameters (\#Params) across random seeds of \{$0$, $1$, $2$\}. The vision transformer (ViT-B-224/32) via CLIP pretraining is evaluated.  Our method achieves the best tradeoff between accuracy and parameter efficiency: it obtains the best average accuracy among all efficient model adaptation methods, while updating only 0.09\% of the model parameters in CLIP. We color each accuracy value as the \textbf{best} and \textcolor{dblue}{second best}: the same hereinafter.} 
 \label{tab:fewshot}
\end{table*}